# Proposta di nuovi strumenti per comprendere come funziona la cognizione.

Devis Pantano[*]

Marzo 2014

## Abstract

English version

Novel tools to understand how cognition works

I think that the main reason why we do not understand the general principles of how knowledge works (and probably also the reason why we have not yet designed and built efficient machines capable of artificial intelligence), is not the excessive complexity of cognitive phenomena, but the lack of the conceptual and methodological tools to properly address the problem. It is like to build up Physics without the concept of number, or to understand the origin of species without including the mechanism of natural selection.
In this paper I propose some new conceptual and methodological tools, which seem to offer a real opportunity to understand the logic of cognitive processes.
I propose a new method to properly treat the concepts of structure and schema, and to perform on them operations of structural analysis. These operations allow to move straightforwardly from concrete to more abstract representations. With these tools I will suggest a definition for the concept of rule, of regularity and of emergent phenomena. From the analysis of some important aspects of the rules, I suggest to distinguish them in operational and associative rules. I propose that associative rules assume a dominant role in cognition. I also propose a definition for the concept of problem. At the end I will briefly illustrate a possible general model for cognitive systems.

---

[*] Per commenti ed informazioni mi si può contattare al seguente indirizzo:
devis.pantano@unipd.it

Versione italiana

Credo che la causa principale per cui non si conoscono i principi generali di come funziona la conoscenza, e probabilmente anche il motivo per il quale non si sono ancora progettate e costruite macchine efficienti capaci di intelligenza artificiale, non sia l'eccessiva complessità dei fenomeni cognitivi, ma il fatto che sono mancati gli strumenti concettuali e metodologici corretti per affrontare il problema. È un po' come se si cercasse di edificare la fisica senza il concetto di numero, o se si cercasse di comprendere l'origine delle specie senza aver compreso il meccanismo di selezione naturale.

In questo lavoro propongo alcuni nuovi strumenti metodologici e concettuali, che sembrano offrire la concreta possibilità di capire la logica dei processi cognitivi.

Propongo una nuova metodologia per trattare, in modo preciso, i concetti di struttura, di schema, e per eseguire delle speciali operazioni di analisi strutturale. Queste operazioni permettono di passare da rappresentazioni concrete ad altre più astratte in modo naturale, senza forzature. Munito di questi strumenti mi è possibile proporre una definizione per i concetti di regola, di regolarità e di proprietà emergente. Analizzando alcuni aspetti importanti delle regole propongo di distinguerle in operazionali e associative. Propongo che siano le regole associative ad assumere un ruolo dominante nell'attività cognitiva. Propongo inoltre una definizione per il concetto di problema. Alla fine illustro, in forma succinta, un possibile modello generale di sistema cognitivo.

# Sommario

# 1 Un'introduzione generale

Lo scopo di questo capitolo introduttivo è spiegare al lettore per quale motivo dovrebbe spendere il proprio tempo a leggere questo lavoro. Non mi è facile condensare in un riassunto l'intero contenuto. Questa introduzione è composta di due parti: una prima che contiene la sintesi delle idee principali, una seconda che riassume brevemente alcuni degli altri punti importanti.

## 1.1 Una breve sintesi delle idee principali

Chiedo al lettore di considerare l'ipotesi che le difficoltà che si incontrano nel comprendere la logica dei fenomeni cognitivi non siano dovute all'eccessiva complessità di questi, ma al fatto che sono mancati gli strumenti concettuali adeguati per affrontarli. È già capitato in passato che, una volta messi a punto gli strumenti idonei, sia stato possibile, in modo repentino, comprendere la logica di fenomeni che, fino a poco prima, sembravano molto complessi ed ingarbugliati. Si pensi, ad esempio, a quanto è accaduto con la comprensione del meccanismo di selezione naturale.
A volte si può essere indotti a ritenere di non avere bisogno di approfondire ulteriormente idee e concetti che ci sembra di saper gestire. Si può ritenere di avere compreso già in maniera sufficiente le cose per le quali qualcuno ci sta proponendo una rielaborazione. Personalmente, alcuni anni or sono, quando a dire il vero la presente teoria non era ancora matura, mi sono trovato in questa situazione con dei colleghi che si sono dimostrati restii nel comprendere l'importanza di acquisire gli strumenti per descrivere, con precisione ed accuratezza, concetti fondamentali quali quello di struttura e quelli di regola.
Proviamo ad immaginare quale sarebbe la reazione di un antico romano, se tornando indietro nel tempo, gli proponessimo di cambiare il suo sistema di numerazione per sostituirlo con il nostro. Questi potrebbe mostrarsi riluttante, in fondo egli già avrebbe uno strumento per descrivere le quantità che interessano per la vita pratica. "Perché", potrebbe chiedersi, "devo fare fatica ad imparare un nuovo sistema di numerazione se quello che uso già funziona?". Sospetto che non sarebbe banale spiegare le notevolissime possibilità offerte dal nuovo sistema di numerazione facendo solo discorsi teorici. In fondo anche al giorno d'oggi la maggior parte delle persone capiscono che la scienza è qualcosa di molto potente essenzialmente perché ne vedono i notevolissimi effetti pratici, e non perché ne comprendono i metodi!

In questo lavoro mostro che una volta identificati e messi appunto gli strumenti adeguati, il problema di capire la cognizione si semplifica notevolmente.
Per comprendere come funziona la conoscenza è sostanzialmente necessario trovare la maniera per definire con precisione, e trattare in maniera adeguata, alcuni concetti importanti per i quali però solitamente ci accontentiamo di una comprensione solo intuitiva.

I punti salienti della formulazione che propongo derivano, per buona parte, dal tentativo di precisare i concetti di **struttura** e di **regola**, e dalla ricerca degli strumenti per trattare, in modo non ingenuo, il **problema del confronto** tra le strutture delle cose e dei fenomeni, al fine di riuscire a identificare tutte le **regole emergenti** che possono essere utilmente sfruttate per la cognizione.

Una delle domande più importanti alla quale dovremmo cercare di rispondere è la seguente: com'è codificata l'informazione all'interno della mente? In cosa è scritto il "mentalese"? Si tratta di **simboli,** di **rappresentazioni pittoriche,** o di qualcosa d'altro?

Chi cerca di capire la logica della conoscenza s'imbatte in questo problema. In effetti, attualmente la nostra cultura scientifica mette a disposizione principalmente due concetti per descrivere il tipo di informazioni che sono contenute all'interno della mente: i **simboli** e, appunto, le cosiddette **rappresentazioni pittoriche**. Queste ultime sono spesso anche indicate con il termine: "rappresentazioni subsimboliche".

Da un lato sembra che, almeno in parte, il pensiero possa essere descritto e rappresentato mediante il linguaggio, quindi mediante dei "simboli". I successi della logica formale nel rendere automatizzabile una parte dei processi d'inferenza, hanno indotto molti a pensare che i simboli siano adeguati per rappresentare le informazioni all'interno della mente e quindi per rappresentare la conoscenza. Dall'altro lato, chi studia come il cervello processa le prime informazioni sensoriali, vale a dire "gli stimoli prossimali" che generano poi le percezioni, sa bene che esse hanno un aspetto molto lontano da **stringhe di simboli**. Queste informazioni sono molto più vicine a delle rappresentazioni "pittoriche" delle cose, sono vicine a oggetti quali i disegni e le fotografie. Come si conciliano questi due aspetti così diversi dell'informazione? Per cercare di superare il problema, molti indicano come "rappresentazioni subsimboliche" le rappresentazioni pittoriche che si trovano all'inizio dei processi di analisi sensoriale, e pensano che queste siano destinate a trasformarsi, con il procedere dei processi cognitivi, in rappresentazioni puramente simboliche.

In realtà credo di poter mostrare che, in queste forme, sia il concetto di simbolo sia quello di rappresentazione pittorica, o subsimbolica, sono inadeguati: non sono gli strumenti idonei per capire come funziona la conoscenza. È come se si cercasse di comprendere e descrivere le leggi fisiche privi di una parte degli strumenti matematici fondamentali: senza il concetto di derivata, senza saper scrivere equazioni e usando un sistema di numerazione del tutto inadeguato, come quello che usavano gli antichi romani.

Mancando gli strumenti di base, i concetti intuitivi di simbolo e di rappresentazione simbolica, da soli, non sono assolutamente idonei allo scopo.

Con gli strumenti giusti si può comprendere che, in realtà, le stringhe di simboli e le rappresentazioni pittoriche hanno degli aspetti fondamentali in comune che costituiscono il vero "substrato portante" della conoscenza. Sono questi aspetti comuni a fornire la possibilità di rappresentare l'informazione all'interno della mente e all'interno di ogni sistema capace di cognizione.

Credo sia possibile mostrare che la cognizione si basa sullo sfruttamento delle **corrispondenze strutturali** che sussistono tra oggetti e fenomeni distinti. Entità che sono fisicamente distinguibili per la composizione delle rispettive sostanze, possono esibire, nella loro forme e nei loro comportamenti, delle corrispondenze di struttura.

Nel 1902 Henri Poincaré pubblicò un trattato di epistemologia: "La scienza e l'ipotesi". In questo lavoro egli giunse a una conclusione che ritengo particolarmente importante. Secondo Poincaré *"la scienza può solo farci conoscere i rapporti tra le cose, al di là di questi rapporti non c'è alcuna realtà conoscibile!"*...

Sostanzialmente propongo che questi "rapporti tra le cose", che secondo il pensiero di Poincarè esauriscono il conoscibile, consistano proprio nelle "strutture" delle cose stesse. Propongo la congettura che all'interno di un sistema cognitivo siano rappresentabili, e quindi conoscibili, solo le strutture degli oggetti e dei fenomeni della realtà esterna.

Penso che l'idea di struttura sia quindi centrale ma, per edificare una teoria in grado di spiegare i principi generali dei processi cognitivi, la formulazione di questo concetto deve passare da qualcosa d'intuitivo a un oggetto matematicamente ben definito e nello stesso tempo facilmente maneggiabile.

Servono gli strumenti per far questo. Essi devono permettere di rappresentare le strutture delle cose, dei fenomeni, e delle situazioni, che appartengono alla nostra quotidianità. Servono quindi, nella sostanza, gli strumenti per comprendere come le informazioni strutturali sono codificate all'interno della nostra mente.

Purtroppo in questo campo esistono delle profonde lacune. La nostra cultura scientifica non fornisce, per quanto è a mia conoscenza, nulla di realmente adatto a questo scopo. Ho cercato di colmare questa mancanza elaborando una metodologia che presento nei primi capitoli di questo lavoro.

Muniti di questi strumenti, si può mostrare che in realtà sia le rappresentazioni pittoriche, sia le manipolazioni di stringhe di simboli formali, si basano sullo sfruttamento di corrispondenze strutturali: in ogni **schema di manipolazione simbolica,** utilizzato per compiere inferenze, ci sono sempre delle corrispondenze strutturali con il fenomeno che queste manipolazioni rappresentano. Queste corrispondenze strutturali possono essere difficili da scorgere poiché le manipolazioni "simulano" processi di "gestione delle rappresentazioni cognitive" (e per di più di loro aspetti astratti) che avvengono all'interno della mente, e che quindi non fanno parte dei fenomeni "direttamente osservabili".

Nelle rappresentazioni pittoriche, in modo inverso, è assai più semplice identificare le corrispondenze strutturali con l'oggetto rappresentato. In realtà si può mostrare che per costruire la rappresentazione stessa è necessario mettere assieme, nella maniera opportuna, una serie di oggetti che rappresentano le "parti componenti" della struttura rappresentata. Questi oggetti svolgono una funzione del tutto simile a quella assunta dai "simboli" che si usano nelle stringhe formali, qualora non ci si preoccupi, come imposto dall'assunto base del formalismo, del loro significato. A prima vista i simboli usati in un calcolo logico sembrano diversi dagli oggetti giustapposti per costruire le rappresentazioni pittoriche, ma in realtà hanno la stessa funzione!

Una nuova formulazione del concetto di struttura è dunque uno degli strumenti essenziali per riuscire a comprendere la logica della mente; ma da solo non è sufficiente, mancano ancora degli ingredienti.

Accanto a quello di "struttura" un altro ingrediente fondamentale è costituito dalla definizione del **concetto di regola**.

Le regole sono fondamentali, sono il "**motore di ogni processo d'inferenza**" e quindi sono il motore di ogni processo cognitivo.

Nonostante la loro fondamentale importanza, a tutt'oggi, per quanto è a mia conoscenza, non è disponibile una definizione esatta di cosa sia una regola! Come possiamo cercare di comprendere come funziona la cognizione se non siamo in grado di definire con precisione che cosa è una regola, vale è dire se non sappiamo in che cosa consistono, con precisione, i mattoni fondamentali di ogni processo d'inferenza?

Ho cercato di rimediare anche a questa lacuna e a tale scopo propongo la congettura che ogni regola (e ogni regolarità) sia sempre riconducibile all'esistenza di corrispondenze strutturali. Nella sostanza congetturo che sia possibile identificare una regolarità se, e solo se, è possibile trovare delle "coincidenze" tra due o più strutture[1].

Questa congettura (ovviamente se corretta) è molto importante perché ci dice che cosa dobbiamo andare a cercare per identificare ciò che fa funzionare la cognizione.

---

[1] Vedremo che per estendere questa congettura ad ogni possibile regola è necessario comprendere come si costruiscono le rappresentazioni astratte del mondo, da quelle più semplici a quelle più complesse

Studiando le varie tipologie di regole, e come queste sono utilizzate all'interno di un sistema cognitivo "naturale", mi sono reso conto che, all'atto pratico, è molto importante distinguerle in due categorie: quella delle regole che possiamo chiamare **operazionali** e quella delle regole che possiamo invece indicare come **associative**. Sono possibili anche altre classificazioni per le regole, ma questa è particolarmente importante per comprendere come funziona "l'intelligenza naturale".

Alla categoria delle regole di tipo operazionale appartengono quelle che richiedono l'uso di formule e l'esecuzione di una serie di calcoli (le leggi della fisica rientrano in questo caso).

Le regole associative hanno invece un aspetto molto diverso. Esse, pur potendo essere anche alquanto complesse, fondamentalmente si basano su un semplice meccanismo associativo tra l'identificazione di fatti che fanno da "cause" e altri che fanno da "effetti". Questi "fatti" devono essere descritti in termini strutturali.

Studiando queste due tipologie di regole, in particolare come sono utilizzate nella normale pratica cognitiva, si scopre che per l'intelligenza naturale le **regole associative sono predominanti**; senza di esse la cognizione stessa non sarebbe possibile.

Sono ovviamente importanti, e per certe abilità assolutamente necessarie, anche le regole operazionali, ma queste hanno spesso, in un certo senso, un ruolo subordinato. Sono spesso le regole associative a coordinare e a gestire l'applicazione delle regole operazionali. Nella sostanza le regole associative assumono un ruolo "dominante" nell'attività cognitiva.

Vedremo che nella pratica dell'intelligenza naturale le regole operazionali possono essere utilizzate solo per compiti che richiedono "un basso livello di astrazione", come per il coordinamento senso-motorio, dove è necessario eseguire una serie di operazioni di calcolo, spesso molto impegnative. Quando però si passa a compiti di "più alto livello", che interessano rappresentazioni appena più "astratte", sono le regole associative a prevalere nettamente.

Ho affermato che le regole consistono in "corrispondenze strutturali", che quelle predominanti sono di tipo associativo e che l'informazione all'interno di un sistema cognitivo consiste in rappresentazioni delle strutture delle cose. Questi punti implicano che, per scoprire e utilizzare le regole, è fondamentale "confrontare le strutture". In questo passaggio si manifesta un problema molto importante: come si confrontano delle strutture? Ingenuamente si potrebbe essere indotti a pensare che questo confronto debba avvenire cercando semplicemente di "far sovrapporre" le "rappresentazioni estese" delle cose. Ad esempio, nella geometria di Euclide, per eseguire il confronto si suggerisce di muovere delle figure senza deformale, fino a sovrapporle e "guardare se coincidono in tutte le loro parti".

Ebbene quest'approccio è **ingenuo**, non è sufficiente. Se ci limitassimo a far questo non saremmo in grado di individuare la grande maggioranza delle regole possibili. In realtà si può mostrare che, per confrontare in modo efficiente delle strutture, è necessario essere in grado di confrontare anche le loro astrazioni, e per far questo è necessario aggiungere, alle rappresentazioni delle strutture di base (quelle che si realizzano entro dei simulatori "tradizionali") altre rappresentazioni che si occupano sempre delle medesime informazioni, ma che sono nello stesso tempo più flessibili e molto più ricche delle prime.

Queste rappresentazioni si ottengono da quelle di base, che per intenderci sono cose molto vicine alle rappresentazioni delle geometrie tridimensionali degli oggetti, eseguendo su di esse delle opportune operazioni di analisi. Attraverso queste operazioni si passa a "**rappresentazioni esplicite**" del complesso del **contenuto informativo strutturale** delle prime. Con queste nuove rappresentazioni le operazioni di confronto diventano più semplici e diviene possibile, punto fondamentale, scoprire la presenza di regole associative che coinvolgono solo "**una parte**" del contenuto strutturale complessivo.

L'impostazione e gli strumenti che propongo nei primi capitoli di questo lavoro per trattare le strutture sono pensati proprio per consentire di eseguire in modo naturale queste operazioni di **derivazione** e di **esplicitazione separata** di tutti i "contenuti strutturali parziali".

La trattazione del concetto di struttura è concepita per eseguire in modo naturale delle "**operazioni di derivazione strutturale**", che permettono di passare dalle strutture di base ad altre che sono spesso astrazioni e generalizzazioni delle prime. Il tutto permette di massimizzare la capacità di identificare la presenza di regolarità e permette di cogliere in modo naturale **i fenomeni emergenti**. Le principali operazioni di derivazione strutturale sono pensate per identificare una parte importante di quelle che possiamo indicare come il complesso delle "**strutture emergenti**". Anche questo punto è cruciale perché la cognizione funziona grazie alla capacità di sfruttare non solo le regole basilari che gestiscono i fenomeni fisici, ma anche il **complesso delle regole emergenti**. Senza questa possibilità fondamentale, la capacità di generare previsioni sarebbe estremamente limitata. Nella sostanza siamo in grado di generare previsioni a lungo termine e di pianificare azioni utili alla sopravvivenza, grazie al fatto che dal substrato delle leggi fondamentali, che regolano i fenomeni più basilari, **emergono** moltissime regolarità e moltissime regole sfruttabili in modo molto più efficiente. Senza questo fenomeno della comparsa di **regole emergenti** la cognizione di fatto non sarebbe possibile. È interessante il fatto che la maggior parte delle regole emergenti sia di tipo associativo.

Un altro concetto importante è che per realizzare un "sistema cognitivo", vale a dire un sistema capace di cognizione, bisogna in primo luogo implementare la "conoscenza semantica di base", che non è costituita da elementi come i "principi della logica" e altre astrazioni, ma è invece fatta di abilità e conoscenze pratiche molto basilari, che sono sostanzialmente quelle che permettono di riconoscere, attraverso i sensi, le cose che osserviamo, di

"riconoscere le situazioni" e di utilizzare una serie di "regole di comportamento" per finalizzare le azioni verso l'ottenimento di "situazioni che sono valutate come positive".

Senza aver prima acquisito una buona base di questo tipo di conoscenze è molto difficile costruire quelle di livello più elevato.

## 1.2 Breve riassunto di altri punti salienti

Credo non sia sbagliato considerare, per alcuni dei loro aspetti salienti, i sistemi capaci di cognizione come degli speciali "simulatori della realtà".

Nel 1943 lo psicologo scozzese Kenneth Craik suggeriva che: *"Quando l'organismo porta dentro di sé un "modello in piccola scala" della realtà esterna e delle sue possibili azioni, diventa per esso possibile esplorare diverse alternative, optare per quella migliore, reagire a possibilità future prima ancora che si verifichino, utilizzare la conoscenza di eventi passati per affrontare il presente e il futuro, e in generale rispondere alle emergenze che gli si presentano in maniera molto più completa, sicura e competente."*

Oggi la tecnologia dei calcolatori mette a disposizione simulatori molto potenti, ma che funzionano in modo diverso rispetto ai "sistemi cognitivi naturali". Secondo alcune stime i maggiori supercomputer attuali si avvicinano, per capacità di calcolo, alla "potenza computazionale" del nostro cervello (e forse già le superano per alcuni aspetti). Ciò nonostante questi simulatori non sono dei sistemi intelligenti. Oggi le simulazioni al calcolatore sono molto usate in varie discipline, non solo per scopo scientifico o tecnologico, ma anche per scopi ludici, pensiamo ad esempio ai moderni videogiochi. Il funzionamento di queste simulazioni è basato sull'utilizzo di relativamente poche regole operazionali, la maggior parte delle quali utilizza un numero limitato di equazioni per calcolare come i processi simulati evolvono nel tempo.

Ebbene penso che anche il nostro cervello utilizzi delle simulazioni, ma lo faccia con un approccio ben diverso in alcuni aspetti importanti. Uno di questi è che il cervello, invece di usare poche regole operazionali, utilizza moltissime regole nelle quali prevale l'aspetto associativo. Penso inoltre che costruisca le rappresentazioni del mondo operando non a "unico livello", come si fa con i simulatori attuali, ma generando contemporaneamente molti "**strati di rappresentazioni**", e applicando a queste anche **diversi "strati di regole"**.

Questi strati, costituiti da rappresentazioni e regole, sono legati gli uni agli altri da precisi rapporti gerarchici. Il passaggio da quelli più basilari a quelli di livello più elevato avviene in modo naturale, seguendo sempre lo stesso "schema di base".

Come detto, molte delle regole utilizzate dal cervello sono diverse rispetto a quelle usate nelle attuali simulazioni al calcolatore. Per permettere al lettore di costruirsi una prima idea sull'aspetto di una parte di queste regole, può essere utile un esempio; invito comunque il lettore a tenere presente che, per illustrarlo, dovrò utilizzare un sistema di rappresentazione particolare: il nostro linguaggio.

Molti ricercatori concordano che le informazioni all'interno della mente sono codificate in modo diverso da quello del codice linguistico; non sapendo quale sia questo "codice", ma ritenendo che comunque esso esista, è stato coniato un termine per designarlo: "il mentalese" (Fodor). Il "mentalese" sarebbe, secondo questa concezione, il vero codice del pensiero. Il senso del concetto è che non sappiamo quale sia questo codice, ma crediamo che esista, e per questo lo indichiamo con questa buffa parola.

Nell'illustrazione dell'esempio non uso quindi il mentalese, ma uso il nostro linguaggio, anzi uso una sua forma un po' più vicina a un altro codice, che può essere usato nei calcolatori: quello delle "rappresentazioni logico formali".

Supponiamo, dunque, di essere in grado di costruire delle rappresentazioni che, espresse "quasi" nel nostro linguaggio, suonano del tipo:

Se è data la situazione che:
X è un oggetto, X è sospeso (nel senso che non poggia sulla superficie di qualcosa), e X è trattenuto da Y.

E se avviene l'evento che:
Y smette di trattenere X

Si può applicare la regola associativa:
Allora X cadrà verso il basso.

Questa regola è associativa, proprio perché si associano semplicemente la descrizione di alcune condizioni iniziali e il verificarsi di un evento particolare, un effetto ben prevedibile, che proponiamo quindi come conclusione della nostra inferenza. Essa dice che "se si smette di trattenere qualcosa che è sospeso, allora questo cadrà". Un fatto importante è che si tratta di una **regola molto affidabile**. Essa, infatti, funziona "quasi" sempre. In quei pochi casi particolari nei quali non funziona (ad esempio se X è un palloncino riempito di elio), il nostro cervello è comunque in grado di riconoscere, sempre attraverso rappresentazioni opportune, le condizioni che fanno sì che la regola non sia applicabile.

Di regole di questo tipo ne emergono moltissime non appena si acquisisce la capacità di passare dalle rappresentazioni concrete a quelle più astratte.

Il nostro cervello utilizza moltissime regole di questo tipo. Molte di esse sono assai più complesse di quella appena descritta, proprio perché in genere esse dipendono da molti fatti circostanziali. Anche la loro affidabilità dipende quindi da questi fatti circostanziali, ma il nostro cervello è in grado di riconoscerli e rappresentarli in modo molto efficiente. Credo lo faccia costruendo in realtà, una gerarchia di rappresentazioni dove sono "rese esplicite" moltissime singole

"informazioni strutturali". Se dovessi cercare di descrivere, usando il nostro linguaggio, solo una parte di tutti questi "fatti strutturali", che il nostro cervello è in grado di identificare e usare, avrei bisogno di migliaia di pagine, invece delle poche righe che ho usato nell'esempio proposto.

Chi si occupa di cercare di costruire sistemi di intelligenza artificiale usando gli strumenti della logica (o altri simili), potrebbe affermare (in parte con ragione) che in realtà la regola enunciata nell'esempio mostrato è proprio costituita da simboli, ed è esprimibile con i metodi della logica formale. Nel campo dell'intelligenza artificiale sono stati prodotti molti tentativi di implementare delle "ontologie formali", o altri sistemi simbolici di vario tipo, per rappresentare con forme linguistiche, e/o con gli strumenti della logica, i "fatti del mondo" e le loro regole. Una delle difficoltà maggiori è che quanto ne viene fuori è molto complesso, ma soprattutto non è chiaro quale metodo usare per provare a collegare questi linguaggi simbolici con la percezione del mondo reale.

Come detto all'inizio di questo capitolo, penso sia sbagliato contrapporre concetti come quelli di "rappresentazioni subsimboliche" a quelli di "rappresentazioni simboliche-linguistiche". Credo che il "codice" corretto per comprendere come funziona la cognizione debba essere diverso, e richieda la messa appunto di nuovi opportuni strumenti. Credo che, utilizzando questi strumenti, sia possibile mostrare che in realtà sia "le rappresentazioni subsimboliche", sia quelle più astratte, come lo sono anche le "stringhe di oggetti" che utilizziamo per le computazioni logiche, possano essere rappresentate con gli stessi metodi, con gli stessi "strumenti computazionali"; è anche possibile mostrare che il passaggio dalle prime alle altre (o meglio ad astrazioni che per certi aspetti sono simili alle "proposizioni" che si usano in logica) può avvenire in modo naturale, senza forzature, senza un netto cambiamento di "paradigma rappresentazionale".

Ad ogni modo, prima di illustrare alcune anticipazioni sul metodo che propongo, penso possa essere interessante affrontare un altro problema cruciale per l'attività cognitiva: quello del **confronto e riconoscimento di "entità complesse"**.

Per comprendere come si può realizzare un simulatore in grado di usare le regole associative emergenti può essere utile partire pensando a un possibile sistema ancora "non intelligente", ma che realizza al suo interno delle prime simulazioni della realtà circostante. Ragionando su cosa manca a questo sistema si possono comprendere alcuni passaggi importanti.

Pensiamo dunque a un sistema ipotetico, che sia dotato dell'equivalente di alcuni dei nostri organi di senso, e sia capace, per ipotesi, di ricostruire al proprio interno una rappresentazione tridimensionale dell'ambiente circostante. Questo sistema dovrebbe quindi essere dotato dell'equivalente della vista e dovrebbe essere capace di costruire una "proiezione virtuale interna" del "percepito".

È necessario premettere che in realtà non è per nulla semplice realizzare un sistema artificiale di questo genere: per tradurre le informazioni sensoriali visive grezze, in rappresentazioni 3D, serve già "molta intelligenza". Ad ogni modo, per questa introduzione, chiedo al lettore di partire dall'ipotesi di avere già a disposizione un primo sistema che sia in grado di ricostruire le geometrie degli oggetti direttamente osservati, vale a dire quelli che cadono sotto lo sguardo, il tutto in tempo (quasi) reale.

Supponiamo quindi di avere a disposizione un sistema che costruisce, al suo interno, una "realtà virtuale" in modo simile a quella generata in molti videogiochi moderni e che corrisponde proprio a quanto percepito in un certo momento. Un apparato di questo genere, in grado di fare solo questa cosa, non è sicuramente ancora un sistema intelligente. Un simulatore 3D non ha conoscenza delle cose e delle regole del mondo. Al massimo può essere usato per generare delle previsioni sulle traiettorie degli oggetti in movimento, ma che funzionano solo su scale temporali abbastanza brevi e che entrano subito in crisi non appena gli oggetti in movimento interagiscono con altri. La nostra realtà fisica non è prevedibile, se non su scale temporali limitate, usando solo delle semplici simulazioni tridimensionali.

A questo sistema manca la capacità di riconoscere le singole **cose** e le **situazioni** ricostruite al suo interno e la capacità di applicare a questi **"riconoscimenti"** le **regole associative** (emergenti) alle quali sono soggette le situazioni che si determinano. Si tratta di quelle regole che permettono di prevedere, quando possibile, come le situazioni possono evolvere, ma anche di quelle regole che permettono di pianificare le azioni in modo finalizzato! Se un sistema sa prevedere gli eventi può anticiparli. Se sa prevedere gli effetti delle proprie azioni può correggerle[2].

Al sistema manca dunque la capacità di **riconoscere** le cose concrete, le loro relazioni statiche e dinamiche, e con ciò la capacità di riconoscere **le situazioni** alle quali danno luogo.

---

[2] La "**finalizzazione dell'attività cognitiva**" e dei comportamenti può essere concepita, e credo accuratamente descritta, classificando una parte significativa delle situazioni possibili in "**desiderabili**" e "**indesiderabili**". Buona parte della nostra attività cognitiva (se non tutta) è finalizzata a trovare il modo per "pilotare gli eventi della realtà" per **ottenere situazioni desiderabili** e per **evitare quelle indesiderabili**.
Spesso "**risolvere un problema**" significa trovare entro lo "spazio delle azioni possibili", una percorso che porta dalla "situazione attuale" a quella desiderata

Per riconoscere un oggetto dobbiamo in qualche modo confrontare le nuove informazioni, ricavate dai sensi, con qualcosa che abbiamo memorizzato in passato. Per ipotesi, nel nostro sistema in esame, queste informazioni consistono in una ricostruzione tridimensionale delle loro geometrie. Come possiamo procedere per confrontare queste ricostruzioni?

Come accennato, il modo ingenuo, forse il primo che viene in mente, consiste nel prendere le rappresentazioni di due oggetti, ricostruiti nel nostro simulatore tridimensionale, e provare a "sovrapporle" con qualche movimento rigido che non le deformi. Ma un metodo di questo genere difficilmente può funzionare. Che efficienza può avere? È facile convenire che nella nostra esperienza quotidiana abbiamo a che fare molto spesso con oggetti che non mantengono una forma rigida, ma che siamo comunque in grado di riconoscere. Continuamente abbiamo a che fare con "categorie di oggetti" che si "assomigliano nella struttura", ma che non sono per nulla "perfettamente sovrapponibili" e che, nonostante ciò, siamo in grado di riconoscere come dello stesso tipo.

Se ragioniamo sulla questione, non è difficile convenire che il modo con il quale confrontiamo le nostre "ricostruzioni interne" delle cose deve essere un altro, non è pensabile che ci limitiamo a "cercare di sovrapporre" le ricostruzioni geometriche. Come possiamo procedere?

Supponiamo di essere in grado di aggiungere al nostro sistema, in grado di ricostruire le rappresentazioni 3D e di memorizzarle, un altro **apparato speciale**, in grado di identificare l'insieme di tutte le **proprietà strutturali** che sono presenti in una certa ricostruzione interna di un oggetto e di renderle esplicite producendo una loro elencazione digitalizzata, quindi di tipo binario.

Quest'apparato deve, per ipotesi, essere in grado di produrre dei "bit equivalenti" che indicano se è presente o no una specifica proprietà strutturale. Questo sistema dovrebbe quindi fornire in uscita moltissimi "bit equivalenti", che di volta in volta si accendono o restano spenti in funzione del fatto che sia o no presente la specifica proprietà strutturale alla quale sono individualmente associati.

Ad esempio, se l'oggetto osservato è una bottiglia, il nostro apparato dovrebbe essere in grado di fornire un elenco di bit accesi, associati a moltissime informazioni strutturali esplicite che appartengono alla struttura di quella specifica bottiglia. Molte di queste informazioni dovrebbero essere "particolari strutturali", "dettagli di forma", "categorie di relazioni tra le parti componenti" e varie altre, che possono essere difficili da descrivere linguisticamente. Per fare degli esempi posso provare a elencarne alcune, ma si tenga presente che in realtà devono essere molte di più di quelle che posso nominare, e che spesso si tratta di "dettagli strutturali" difficili da descrivere a parole. Nel caso di una bottiglia i tanti singoli "bit accesi" potrebbero indicare che: l'oggetto è di forma allungata, ha una forma cilindrica, la base del cilindro corrisponde a un cerchio, dalla parte opposta alla base il cilindro si restringe, da questa parte presenta un foro, il rapporto della base rispetto all'altezza è compreso entro un certo intervallo ….e molte altre.

Se ammettiamo di essere in grado di produrre questi elenchi, possiamo pensare di spostare le operazioni di confronto dalle rappresentazioni 3D "estese", all'elenco di bit. In effetti, in teoria, se l'analisi e l'esplicitazione delle proprietà strutturali è stata fatta correttamente, possiamo aspettarci che due forme identiche (e nella stessa posizione) produrranno esattamente gli stessi elenchi di bit a 1. Potremo anche pensare di memorizzare direttamente questi elenchi, invece delle rappresentazioni 3D.

Si tratterebbe quindi di spostare, almeno in parte, il problema della rappresentazione delle cose dalle rappresentazioni tridimensionali "estese" a questi elenchi di bit. Può funzionare? E, nel caso, quale sarebbe il vantaggio di quest'operazione?

Ebbene credo che quest'operazione possa comportare un vantaggio enorme per l'attività cognitiva, a patto di produrre questi elenchi nel modo opportuno. Questo vantaggio consiste nel fatto che, selezionando nel modo corretto particolari sottoinsiemi di questi grandi elenchi, è possibile identificare non una singola forma precisa, bensì un'intera **classe di forme,** accomunate da specifiche "**parentele strutturali**". Credo si tratti proprio di quelle parentele che ci permettono, specialmente nelle prime fasi di apprendimento, di riconoscere in modo flessibile sia singoli oggetti, sia categorie di questi. Anche se due oggetti non coincidono perfettamente per le loro forme tridimensionali, essi possono presentare comunque delle "**somiglianze strutturali**", che ne permettono il riconoscimento e la classificazione.

Possiamo inoltre pensare di utilizzare alcuni sottoinsiemi della grande lista di bit citata, per costruire delle "rappresentazioni invarianti" del medesimo oggetto. L'idea è che alcuni di questi sottoinsiemi si possano attivare (talvolta in modo parziale) in maniera "sufficientemente univoca" tale da permettere il riconoscimento di un particolare oggetto in modo indipendente dalla posizione specifica che esso assume, dalla scala nella quale è rappresentato, e anche quando è parzialmente occluso. In altre parole, l'idea è che sia possibile identificare particolari sottoinsiemi, del grande elenco di bit, che costituiscono una specie di "firma" di un oggetto specifico o di una certa classe di oggetti. L'idea generale, quindi, è di utilizzare queste "firme" per riconoscere le cose, e soprattutto, per riconoscere le "**regolarità strutturali**" presenti nelle rappresentazioni del mondo che ci circonda.

Quanto fin qui illustrato è solo una notevole semplificazione di come penso si debba realmente procedere; lo scopo di queste spiegazioni introduttive è solo quello di consentire al lettore di abbozzare alcune prime idee.

Si tenga anche presente che un metodo come quello sommariamente illustrato è in grado di identificare solo una certa classe delle "**proprietà cognitive**" che possono essere associate alle cose concrete: si tratta di quelle proprietà che riguardano esclusivamente quello che possiamo chiamare "**contenuto** **strutturale interno**" delle nostre rappresentazioni di partenza (che entro le ipotesi fatte sono appunto le rappresentazioni 3D). Si può mostrare che molto spesso identifichiamo proprietà che vanno oltre al "contenuto strutturale" delle cose; ad esempio spesso riconosciamo gli oggetti in base alle loro "proprietà funzionali". Spesso, osservando un oggetto, riusciamo a comprendere, per altro sempre esaminandone la forma, quindi la struttura, come lo possiamo usare, e in questo modo siamo in grado di classificarlo opportunamente.

Queste capacità di associare alle cose concrete delle proprietà funzionali possono essere apprese solo con molta esperienza, solo con "l'uso concreto" delle cose. Questa esperienza non può essere presente all'inizio dell'apprendimento, e ciò comporta che nelle fasi iniziali le cose possono essere riconosciute solo in funzione delle loro "**proprietà strutturali**".

Come vedremo nei capitoli che seguono, credo che in questa idea, seppur semplificata, di produrre degli elenchi di "bit equivalenti" associati al riconoscimento di specifiche proprietà strutturali, ci sia del vero[3]. Come accennato, penso che la sua utilità principale consista nel fatto che attraverso questi elenchi espliciti sia possibile identificare agevolmente molte regolarità che altrimenti non sarebbero "visibili".

Ad ogni modo da sola, questa descrizione di come si "dovrebbe procedere", all'interno di un sistema cognitivo, nelle attività di "esplicitazione" delle proprietà strutturali, al fine di "semplificare le operazioni di confronto", non aggiunge molto di nuovo, rispetto ai concetti di pattern recognition, e di estrazione delle features. Se il tutto si riducesse a questo, si potrebbe continuare a pensare che una volta riconosciute le varie "features"/"proprietà strutturali", si otterrebbe proprio il passaggio alle rappresentazioni simboliche: i vari bit che esplicitano i singoli riconoscimento potrebbero essere usati per costruire espressioni di logica formale. Inoltre quanto finora presentato non spiega come si fa concretamente a decodificare le "proprietà strutturali" corrette, e come si fa a passare da immagini bidimensionali alla ricostruzione tridimensionale dei singoli oggetti. Queste due passaggi sono in realtà delle attività molto complesse, che attualmente costituiscono delle sfide tra le più difficili tra i problemi affrontati nel campo dell'IA.

---

[3] Ragionando sulla logica che sta alla base delle "rappresentazioni strutturali", si può mostrare che in realtà non conviene agire semplicemente producendo alla rinfusa delle lunghe liste disordinate di "esplicitazioni" di proprietà strutturali, ma conviene bensì procedere per gradi, identificando anche una serie di altre "strutture derivate" che stanno le une rispetto alle altre in specifici ed importanti rapporti di derivazione. Vedremo che anche questi rapporti gerarchici reciproci fanno parte delle informazioni che è necessario rendere esplicite per essere in grado di "analizzare al meglio" il contenuto strutturale delle rappresentazioni di partenza.

Il punto è che una cosa è cercare di affrontare questi problemi senza "una visione chiara" della logica profonda dei vari passaggi, un'altra cosa è approcciarsi a essi muniti di strumenti che permettono di capire qual è il loro senso, e quindi con idee ben chiare di cosa si vuole ottenere.

Il valore aggiunto della teoria che propongo credo consista in primo luogo nella sua capacità di spiegare cosa sono le strutture, cosa sono le regole e cosa sono le proprietà emergenti. Ciò ci consente di capire come può essere codificata l'informazione all'interno dei sistemi cognitivi secondo una prospettiva nuova, che indica come passare in modo naturale dalle cosiddette "rappresentazioni subsimboliche" ad altre più astratte. Ciò ci consente anche di capire come possono essere scoperte e utilizzate le regole che, come già affermato, costituiscono "il motore di ogni processo cognitivo". Credo che questo a sua volta ci consenta di avvicinarci alla comprensione della logica generale dell'attività cognitiva, sia a livello globale, che a piccola scala. Credo che, una volta comprese in modo chiaro quali sono le finalità dei vari processi, diventi molto più semplice identificare le soluzioni concrete per la loro implementazione concreta.

### 1.3    Alcune anticipazioni sui metodi proposti

La metodologia che propongo nel capitolo 3 per rappresentare le strutture è concepita sia per le rappresentazioni più basilari sia per quelle astratte, e consente di passare dalle une alle altre in modo naturale.

Gli strumenti che servono per rappresentare quei "rapporti tra le cose", che, come spiega Poincarè, "esauriscono la realtà conoscibile", devono avere alcune importanti caratteristiche. Devono poter essere precisi, essere ben matematizzabili, essere idonei sia a descrivere le operazioni matematiche basilari, sia gli operatori più complessi (ma che siano concretamente definibili). Nello stesso tempo questi strumenti devono funzionare anche per descrivere le strutture degli oggetti e dei fenomeni della nostra quotidianità; devono essere in grado di descrivere le nostre percezioni del mondo esterno, sia a livello dei primi stimoli sensoriali, e sia a livello più astratto.

Per soddisfare a tutte queste richieste propongo un metodo per rappresentare ogni struttura "di prima specie" (avremo modo di vedere cosa significa di "prima specie") che si basa sulla specificazione di tre punti:
- un certo insieme di "**parti** componenti",
- almeno un metodo per "distinguerle" dal punto di vista delle loro "**proprietà interne**" (quando possibile),
- dei metodi per descrivere le "**relazioni esterne**" che intercorrono tra queste parti di una struttura.

Questa impostazione sfrutta l'accorgimento di focalizzare l'attenzione sulla contrapposizione tra "distinguibilità interna" e "distinguibilità esterna". Credo che questo stratagemma sia cruciale perché permette di eseguire agevolmente delle operazioni di "cambio di scala" nel modo di osservare le medesime informazioni strutturali. Chiamo queste operazioni indicandole come "**derivazioni strutturali di tipo quoziente**". Esse permettono di passare da una struttura di base a una struttura **quoziente** della prima che ha, per nuove parti componenti, oggetti che corrispondono a porzioni della struttura precedente, vale a dire a entità che sono composte, a loro volta, da più parti della struttura di base.

Il porre l'accento, con i metodi opportuni, su ciò che rende "distinguibili le parti", consente anche di eseguire un'altra operazione di derivazione strutturale molto importante, che indico con il termine "**morfismo**". Con le operazioni di morfismo si procede a rendere "meno forti" le distinguibilità tra gli elementi che costituiscono una struttura. In questo modo, attraverso un morfismo, si può passare da rappresentazioni strutturali che identificano singoli oggetti specifici, ad altre che identificano invece intere classi di oggetti. Inoltre si possono costruire rappresentazioni che restano invarianti per operazioni di spostamento, rotazione, cambio scala, parziale occultamento e altre ancora, e che quindi possono essere usate per riconoscere la medesima entità anche quando si presenta in modi diversi.

Chiaramente, perché il tutto funzioni, è necessario che tutte queste cose siano sempre ben definibili con metodi computazionali, quindi algoritmizzabili. Se affermo che una struttura è un'entità composta da:

- un certo insieme di parti,
- un sistema di distintinguibilità interna tra le parti e
- un complesso di "relazioni esterne";

allora devono esserci dei metodi concreti per descrivere tutto ciò. Vedremo come questo si può fare nel capitolo 3.

Il concetto di struttura che qui propongo è diverso da quello d'insieme. In un certo senso è più ricco e più concreto. Ad esempio, vedremo che le strutture quozienti non sono semplicemente una collezione di sottoinsiemi degli elementi della struttura di partenza. Questo perché nelle strutture devono essere rispettate le "relazioni esterne tra le parti": non tutte le partizioni (in senso insiemistico) di una struttura di partenza, generano delle strutture derivate valide.

In questo lavoro, per indicare i "costituenti base" di una struttura, al posto termine "elemento", preferisco usare la parola "parte". Ciò è legato proprio alla logica dell'operazione di quoziente strutturale. Quando la eseguiamo, cambiamo prospettiva di osservazione. Il concetto di parte è leggermente diverso da quello di "porzione" in modo sottile, ma non per questo banale. Quando, durante l'attività cognitiva, consideriamo una porzione di qualcosa come un "ente a sé stante", lo facciamo sempre per dei validi motivi, seguendo una logica, e questa "nuova entità" diventa spesso l'elemento costituente di una rappresentazione strutturale diversa rispetto a quella di partenza. Cambiamo appunto "la scala di osservazione".

Può essere utile un esempio. Supponiamo di avere un insieme di mattoncini per costruzioni, disposti sopra un tavolo. Se con questi costruisco un oggetto particolare, tipo una piccola automobile, cambierò la loro disposizione strutturale. In questo caso non cambierò le loro "distinguibilità interne", ma agirò sul complesso delle loro relazioni esterne. Se due mattoncini appaiono uguali, al punto tale che scambiandoli l'uno con l'altro ottengo esattamente la stessa automobilina, allora significa che questi due mattoncini non sono distinguibili dal punto di vista delle loro "proprietà interne".

Supponiamo ora di avere molti più mattoncini e di costruire non una sola, ma molte piccole automobili. Questi nuovi oggetti possono essere considerate le parti componenti di nuove strutture. Le loro distinguibilità interne saranno ora ciò che distingue i vari modelli di automobilina che costruisco, e se tra le tante ne costruisco alcune identiche, allora queste non saranno "distinguibili internamente".

Posso anche disporle, le une rispetto le altre, secondo diversi tipi di "relazioni esterne", ottenendo altri "oggetti composti" di scala superiore, che possono essere a loro volta tra loro ben distinguibili. Posso disporle in modo da simulare un autodromo, oppure allinearle per simulare un parcheggio, oppure disporle in modo da realizzare uno sfasciacarrozze…. Posso continuare ancora, e costruire altre cose, come delle case, delle strade, dei ponti e usare questa volta l'autodromo, il parcheggio, lo sfasciacarrozze, come parti componenti di un oggetto di scala ancora superiore: la simulazione di un'intera città.

La cosa importante da notare è che per rappresentare tutti questi diversi soggetti strutturali: singolo mattoncino, automobilina, autodromo, città… possiamo usare sempre lo stesso schema di base, possiamo sempre continuare a descrivere le nostre strutture come composte da:

- un certo insieme di parti,
- un sistema di distintinguibilità interna tra le parti e
- un complesso di "relazioni esterne".

L'esempio illustrato può essere utile per mostrare che quando cognitivamente spostiamo la nostra attenzione da un oggetto particolare, a un altro di scala maggiore, dove il primo è una "parte componente" del secondo, eseguiamo un operazione di quoziente. In questo caso possiamo continuare a utilizzare oggetti strutturali "concreti" nella loro interezza.

Per mostrare come avviene invece un'operazione di morfismo, dobbiamo già abbandonare gli oggetti concreti ed entrare entro il mondo delle "rappresentazioni interne". Un punto importante del metodo è che per ogni rappresentazione, diciamo "completa", di una "struttura intera", è importante procedere a "rendere esplicite" tutte le sue singole proprietà strutturali. Come illustrato più indietro, semplificando un po', possiamo pensare di far questo producendo degli "elenchi" delle proprietà strutturali che le varie strutture possono avere. Ogni struttura da esaminare potrà avere molte proprietà strutturali specifiche, ma accadrà spesso che strutture diverse possiedano alcune proprietà in comune.

Invece di produrre, ex novo, di volta in volta, delle nuove liste per ogni nuova struttura in esame, possiamo pensare di usare delle grandissime liste già pronte; quindi delle liste che contengono già tutte, o quasi, le proprietà, che si possono presentare. In questo modo queste grandissime liste dovranno solo essere "spuntate" per indicare quali sono le proprietà presenti nell'oggetto sotto esame. Come illustrato, il tutto si può fare utilizzando, al posto del segno di spunta, dei singoli bit. Anzi, per molte proprietà, ma non per tutte, invece di usare un sistema binario, può essere utile produrre un numero che in qualche modo esprime "quanto" la singola proprietà strutturale riconosciuta assomiglia a quella presente nell'elenco.

Come già accennato, penso che un sistema reale non debba limitarsi a produrre solo dei lunghi elenchi di queste proprietà, ma sia fondamentale rispettare i "rapporti gerarchici" che possono avere le varie strutture. Quindi credo che questi elenchi di proprietà debbano essere costruiti in modo da rispettare i corretti riferimenti, e che sia fondamentale saper indicare se una data esplicitazione si riferisce a "una proprietà interna" di una specifica parte di struttura, o a una "relazione esterna" che intercorre tra questa e un'altra, ecc…

Qualora le proprietà interne e le relazioni esterne siano rese in forma esplicita, con i rispettivi riferimenti, è possibile eseguire delle operazioni di morfismo, che, come detto, consistono nell'inibire ciò che rendere distinguibili, le une dalle altre, le parti componenti di una specifica struttura. Ripeto il concetto che queste distinguibilità possono essere dovute a "proprietà interne" e a "relazioni esterne".

Se nelle nostre elencazioni referenziate delle varie proprietà, procediamo a eliminarne alcune (spegnendo, o meglio, semplicemente ignorando dei "bit equivalenti" specifici), ciò che otterremo saranno appunto dei "morfismi" della struttura di partenza. Queste operazioni di morfismo permettono di "generalizzare" poiché le rappresentazioni ottenute con esse sono "più permissive" e possono quindi funzionare per intere classi di strutture.

Possiamo anche pensare di eseguire un'operazione inversa rispetto a quella di quoziente, e di passare da una struttura derivata, alla sua primitiva. Credo convenga assumere che possano esistere delle strutture "primitive" che non sono riducibili ad altre più basilari. Queste "strutture base" sono particolari, e possono non possedere un sistema di "distinguibilità interna" tra le loro parti componenti (si pensi ai punti di una varietà). Propongo di usare il concetto matematico di grafo per descrivere questi oggetti. In matematica si usa il concetto di varietà continua, ma per gli scopi di questo lavoro, giacché dobbiamo avere a che fare con strutture che sono "effettivamente costruibili", penso convenga preferire oggetti discreti, costituiti sempre da un numero finito di parti. Alla bisogna, possiamo usare lo stratagemma di approssimare ogni struttura continua d'interesse pratico con un grafo sufficientemente denso.

Un concetto importante è che in tutte le strutture quozienti i metodi per descrivere le relazioni esterne dipendono da quelle che esistevano nelle strutture più basilari. Quindi le nuove parti di una struttura quoziente in qualche modo "ereditano" le loro relazioni strutturali esterne da quelle che le loro costituenti avevano in quella di partenza. Se ammettiamo che la struttura base (per quelle di "prima specie") sia sempre riconducibile a un grafo, ne risulta che in quelle derivate le nuove "proprietà interne" e le nuove "relazioni esterne" avranno sempre una buona base di riferimento dalla quale ricavare, con metodi computazionali, le loro "caratteristiche strutturali" che dovranno essere rese esplicite.

Man mano che si sale lungo la gerarchia delle derivazioni alcune di queste "relazioni esterne" corrispondono a concetti che esprimono relazioni spaziali e temporali, come: stare sotto, sopra, vicino, distante, di lato, venire prima, dopo, essere contemporanei… Si possono però anche descrivere relazioni di tipo logico ed inferenziale, ad esempio essere in relazione di causa-effetto, l'essere una condizione necessaria per l'accadimento di qualcosa, l'essere una condizione sufficiente, costituire un impedimento e varie altre.

Vedremo che, a ridosso delle prime informazioni sensoriali, si ha a che fare con strutture più basilari mentre, man mano che si astrae, si procede con il formulare rappresentazioni che hanno un aspetto diverso, e le cui parti componenti sono costituite da "entità cognitive" che sono molto simili ai nostri concetti (e forse corrispondono proprio a questi). Per generalizzare proporrò, nel capitolo 5, che ad "alto livello di astrazione" le singole situazioni possano essere rappresentate come "insiemi strutturati di soggetti cognitivi". Il senso di questo terminologia potrà essere compreso più avanti. Per ora posso anticipare che anche questi oggetti appartengono al "paradigma delle rappresentazioni strutturali". Anche per questi si può applicare vantaggiosamente l'accorgimento di distinguere tra le "proprietà interne" e le "relazioni esterne" presenti tra i vari "soggetti cognitivi".

L'intera teoria si poggia su alcune ipotesi, che ritengo molto plausibili, ma che per rigore scientifico devo presentare come congetture. In particolare ve ne sono che considero di "riferimento" perché da esse dipende una parte importante della logica di quanto espongo.

La prima è che l'idea di Poincarè sui limiti del conoscibile possa essere reinterpretata in termini strutturali e che quindi, nella sostanza, della realtà esterna ad un sistema cognitivo siano conoscibili solo le strutture delle cose e quanto concerne le operazioni possibili su queste strutture.

La seconda afferma che ogni regola (e ogni regolarità), consiste sempre in "coincidenze strutturali".

Dopo la presente introduzione, che spero possa servire almeno ad incuriosire il lettore, propongo nel secondo capitolo alcune argomentazioni a sostegno della tesi che sono le "rappresentazioni strutturali" e non quelle "simboliche formali" a costituire il "materiale di base" della cognizione.

Nel terzo illustro la mia proposta per una metodologia per trattare le strutture. Si tratta di un capitolo abbastanza tecnico ma essenziale.

Nel successivo passo a illustrare cosa sono le regolarità e le regole, propongo una congettura per una loro definizione ed espongo alcuni primi punti per classificarle in modo opportuno.

Nel quinto capitolo espongo una trattazione per "le proprietà emergenti", proponendo, tra gli altri, i concetti di struttura emergente e quelli di regola e di logica emergente.

Nel sesto capitolo introduco il concetto di "soggetto cognitivo", argomentandone la necessità.

Nel settimo illustro alcuni importanti approfondimenti sulle regole.

Nell'ottavo propongo una possibile definizione per il concetto di problema.

Nel capitolo nove è esposta una presentazione veloce di come può essere costituito e di come può funzionare un sistema cognitivo.

Quanto esposto nei prossimi capitoli è applicabile a sistemi che funzionano secondo computazione classica. In prima istanza l'intero lavoro è concepito per sistemi computazionali deterministici, potenzialmente emulabili con calcolatori di tipo tradizionale. Non ho elementi per escludere a priori che la natura abbia trovato il modo per sfruttare forme di computazione diverse da quelle illustrate.

# 2 Esistono valide alternative alle rappresentazioni simboliche?

## 2.1 Introduzione

Il primo problema da affrontare consiste nel cercare di comprendere in cosa consistono le informazioni all'interno di un sistema cognitivo, qual è la loro "forma", e quali sono i principi basilari che determinano la loro codifica.

Credo che alcune prime importanti indicazioni su queste cose si possano già estrarre dall'analisi dei processi di comunicazione, qualora ci si ponga come obiettivo quello di capire quali sono le differenze tra le cosiddette rappresentazioni "subsimboliche" e i messaggi che invece usano simboli.

Sappiamo tutti che in molte circostanze si possono usare per comunicare anche delle rappresentazioni pittoriche. Spesso un disegno, o lo schizzo di uno schema, funziona molto meglio di mille parole quando si tratta di comunicare particolari "contenuti cognitivi". Nello stesso tempo è anche vero che i concetti più astratti non sono rappresentabili in modo pittorico.

Se oggi uno studente si chiede che cosa è l'informazione e conduce delle rapide ricerche, con buona probabilità gli sembrerà che la risposta debba essere contenuta in quella che è chiamata "teoria dell'informazione", se non altro per il nome di questa; ma è davvero così? Le idee contenute in questa teoria ci spiegano davvero che cosa è l'informazione, o si limitano a cogliere solo alcuni aspetti del fenomeno?

Credo che la "teoria dell'informazione" fornisca degli ottimi e preziosi strumenti; di alcuni di essi mi servirò anche in questo lavoro quando affronterò il tema del "contenuto informativo interno" delle rappresentazioni strutturali. Questa elegante teoria non è però in grado di farci afferrare, nella corretta prospettiva, alcuni aspetti che ritengo fondamentali dell'informazione.

## 2.2 Come possiamo comunicare?

Che cosa succede realmente quando comunichiamo? Pur riconoscendo che si tratta di concetti complessi, che ancora non possono essere inquadrati tramite delle definizioni univoche e precise, attualmente si tende a fare riferimento proprio alle idee che attingono dalla **teoria dell'informazione**. In un testo recente si afferma che "la comunicazione è lo scambio intenzionale d'informazioni effettuato attraverso la produzione e la percezione di segni presi da un sistema condiviso di segni convenzionali".

Ritengo di poter argomentare che questo modo di inquadrare il fenomeno non è completo poiché si possono individuare "oggetti di scambio", per il processo di comunicazione, che non rientrano entro la categoria dei "segni convenzionali". Inoltre penso che dall'analisi di questo fatto si possano estrarre interessanti indicazioni per comprendere la logica di come le informazioni dovrebbero essere codificate in un sistema intelligente.

Come accennato, il nome utilizzato per la teoria menzionata (appunto "teoria dell'informazione") probabilmente non costituisce la scelta migliore. Esso, infatti, sembra lasciare intendere che questa teoria fornisca un inquadramento teorico esaustivo del fenomeno. Questa teoria è indubbiamente utilissima e illustra un metodo che permette di identificare una misura che, in effetti, è direttamente associabile alla "**quantità di informazione**" presente in un messaggio. Tuttavia il fatto che esista questo metodo per quantificarla non comporta necessariamente che con esso si sia anche compreso che cosa è l'informazione. Banalmente, il fatto che in certi contesti sia possibile associare in maniera pertinente un numero ad una certa entità non significa assolutamente che tale numero ne costituisca anche la rappresentazione esaustiva di tutto il complesso dei suoi aspetti e delle sue proprietà (la misura della lunghezza di un tavolo, non è anche la descrizione di cosa è un tavolo!).

Nella teoria dell'informazione, per inquadrare il fenomeno della comunicazione si utilizza uno schema abbastanza semplice che individua tre elementi fondamentali: una sorgente dell'informazione, un canale di trasmissione e un destinatario. Si ha:

Sorgente -------> canale di trasmissione -------> Destinatario.

Si ammette che Sorgente e Destinatario debbano condividere un certo insieme di "**segni convenzionali**". Essi costituiscono, in tale visione, gli elementi fondamentali del linguaggio di comunicazione.

Devo segnalare che nei lavori più attenti si evita di parlare di "sistema condiviso di segni convenzionali", ma si preferisce parlare in termini di "eventi distinguibili", a1, a2, .... ak , che possono essere generati da una sorgente. Si ammette quindi che la sorgente possa spedire attraverso il canale di trasmissione un certo "messaggio" al destinatario, costituito da uno o più di tali "eventi distinguibili". La "misura dell'informazione" scambiata è calcolata in base alla probabilità che ha il destinatario di ricevere un dato messaggio ancor prima che esso sia trasmesso. Tale probabilità può essere assegnata con metodi statistici, in particolare calcolando la frequenza con cui il messaggio specifico è stato prodotto (in genere, ma non esclusivamente, considerando la storia passata del processo in esame). In questa sede non credo sia interessante dilungarsi sull'esposizione della metodologia matematica utilizzata per i calcoli. Mi limito a segnalare che se le probabilità associate agli eventi a1, a2.... ak sono p1,p2... pk, allora la "quantità di informazione" ricevuta con la rilevazione dell'evento a2 da parte del destinatario è calcolata tramite la formula: $\mathbf{i2 = \log 1/(p2)}$, quindi $i2=-\log(p2)$. Inoltre, diversamente dalla teoria della probabilità, l'informazione associata a due eventi a1 e a2 che si presentano insieme (o meglio in diretta successione), non è il prodotto delle singole probabilità di a1 e

a2, ma bensì la loro somma. Attraverso altre interessanti considerazioni è possibile fornire una misura della cosiddetta "entropia" associata ad una sorgente di informazione. Con essa si possono affrontare vari problemi teorici interessanti, come quello dell'ottimizzazione dei codici per la massimizzazione del rapporto segnale-rumore, per stimare quanto un messaggio può essere comprimibile, e vari altri.

Lo schema sopra illustrato per il processo di comunicazione, si può applicare quando sia la sorgente e sia il destinatario hanno specifiche proprietà. In particolare si richiede che essi, in qualche modo, condividano già a priori l'insieme {a1...ak} di eventi utilizzati nel processo, e che siano inoltre in grado di discriminare un evento da un altro.

Sorgente e destinatario possono in molti contesti essere dei sistemi artificiali, ad esempio dei calcolatori. Per questo genere di sistemi è davvero possibile una trasmissione diretta delle informazioni.

Quando però le "entità" che devono comunicare sono persone (o animali) la situazione è certamente più complessa e non pare si possa applicare con facilità il concetto di "trasmissione". Parlare di trasmissione diretta di "informazioni cognitive" da una mente all'altra è più vicino al concetto di telepatia che a un'analisi del processo di comunicazione interpersonale.

Per le persone la trasmissione diretta dei pensieri non è possibile e per comunicare è necessario passare attraverso l'ambiente e la mediazione dei sensi del destinatario. Questo passaggio richiede che la sorgente produca "qualche cosa" (oggetto permanente o fenomeno temporaneo) che sia in grado di stimolare i sensi del destinatario.

Possiamo quindi schematizzare il processo nel modo seguente:

Sorgente --> produzione di "qualche cosa" nell'ambiente --> Destinatario

Questa "mediazione" dell'ambiente e dei sensi è inevitabile. Ciò che per noi è interessante analizzare è il genere degli oggetti e dei fenomeni che possiamo utilizzare per questa mediazione. Possiamo utilizzare dei suoni, quindi produciamo una vibrazione nell'aria che stimola l'udito dei nostri interlocutori. Possiamo utilizzare dei disegni, o delle immagini, o degli scritti, quindi passiamo in questo caso attraverso la vista; con la scrittura braille si passa attraverso il tatto. I sistemi possibili sono dunque molteplici, e molteplici sono anche gli oggetti e i fenomeni che possiamo usare.

Possiamo ragionevolmente schematizzare pensando che la prima persona, quella che fa da sorgente, sia in grado di mettere a fuoco un certo "**contenuto cognitivo**" (qualunque cosa esso sia) che costituisce quanto egli desidera comunicare ad altri. Con ciò egli desidera, di fatto, che nella mente di altri si generi, in qualche maniera, un "contenuto cognitivo" che sia in qualche modo simile al proprio. Il punto veramente interessante è che tutti gli oggetti o i fenomeni prodotti dalla sorgente, per stimolare l'induzione di un certo contenuto cognitivo nella mente del destinatario, in funzione proprio del modo in cui essi svolgono questa funzione, possono essere classificati entro due categorie fondamentali. Una di queste è molto nota ed è quella dei segni convenzionali, che possiamo anche chiamare **simboli**; l'altra categoria invece non è solitamente considerata nella maniera opportuna.

### 2.3   Un'alternativa ai simboli

I suoni delle parole possono essere classificati come simboli. Così è anche per i segni d'inchiostro che formano uno scritto. Ma cose come un disegno o una scultura non possono essere classificate come segni convenzionali. Indubbiamente questi possiedono la proprietà di indurre nella mente dell'osservatore un certo "contenuto cognitivo" che può essere scelto da chi li produce, quindi sono a tutti gli effetti oggetti atti alla comunicazione. **Essi però non sono per nulla convenzionali**. Se in un disegno sono rappresentati in maniera chiara certi soggetti, questi potranno essere riconoscibili da chiunque sia in grado di vedere in modo indipendente da particolari convenzioni preaccordate.

Si considerino due fogli di carta nei quali in uno vi è un disegno ben fatto di un certo oggetto e nell'altro una descrizione particolareggiata scritta dello stesso. Chi conosce la lingua utilizzata nello scritto, sarà in grado, con quel foglio, di rappresentarsi l'oggetto, ovviamente questo non sarà possibile a chi invece non conosce tale lingua, mentre il disegno sarà in grado di assolvere il compito indipendentemente dalla lingua conosciuta da chi lo osserva.

Per quale motivo succede questo? Perché il disegno riesce a comunicare in maniera "pressoché universale" un certo "contenuto cognitivo" mentre ciò non avviene per lo scritto? In fondo si tratta, in tutte e due i casi, di segni d'inchiostro su un foglio di carta: cosa possiede allora il disegno che lo scritto non ha? In cosa consiste la differenza tra queste due tipologie di messaggi?

### 2.4   Le rappresentazioni costrutturate e le simulazioni

Torniamo alla domanda posta alla fine del paragrafo precedente: quale è la differenza importante tra un disegno e uno scritto che ritraggono e descrivono la medesima cosa?

La differenza consiste nel fatto che **il disegno possiede "delle corrispondenze strutturali" con il soggetto rappresentato**, cosa che invece non avviene per lo scritto. Questo è il punto saliente. Nel prossimo capitolo propongo un metodo che consente di attribuire un significato preciso e ben definito al concetto di "corrispondenza strutturale"; per il momento limitiamoci ad usare il concetto in modo intuitivo.

Compreso questo punto, possiamo realizzare che i vari "supporti di messaggi" che, come detto, costituiscono ciò che si può produrre nell'ambiente in modo da stimolare opportunamente i sensi del destinatario, possono essere classificati in almeno due diverse categorie.

La prima sarà composta da oggetti o fenomeni che non hanno "corrispondenze strutturali dirette" con il contenuto cognitivo che devono comunicare; per essa abbiamo usato la terminologia di "segni convenzionali" o di "simboli".

La seconda sarà invece composta da oggetti o fenomeni che possiedono invece delle oggettive corrispondenze con ciò che rappresentano. Si tratta ora di trovare il termine adatto per indicarla. Potrebbe andare bene la parola "modello", o anche la parola "rappresentazione". Tuttavia in questo momento il termine "simulazione" possiede un'accezione di significato in più rispetto alle altre, che consiste nell'idea di poter fungere da "sostituto esperienziale". Per il momento quindi faremo riferimento a questa parola. Più avanti avremo modo, dopo che saranno stati introdotti alcuni concetti, di cercare una terminologia più adatta.

Esempi di simulazioni di questo tipo sono: un disegno, una scultura, una fotografia, ma anche cose come un film, un modello in scala ridotta di un certo apparato, una simulazione al calcolatore dell'evoluzione di un certo fenomeno.

Le simulazioni possono essere usate per comunicare con gli altri, ma anche come "sostitute dell'esperienza diretta". Ciò è possibile perché esse possiedono le menzionate "relazioni strutturali" con ciò che rappresentano. Questa è una delle proprietà fondamentali che caratterizzano le simulazioni: esse possono fungere da **"sostituto esperienziale"**.

Con un disegno, o con una fotografia, possiamo compiere l'esperienza visiva di una certa cosa senza che questa sia effettivamente presente. Con una simulazione si può "sperimentare" l'accadere di un certo fenomeno senza che questo avvenga realmente.

L'esperienza effettuata con la simulazione non sarà, in genere, perfettamente coincidente con quella che si può avere con l'oggetto che essa sostituisce, ma sarà solo simile a essa, e lo sarà proprio in funzione del fatto che tra tale simulazione e l'oggetto rappresentato vi sono delle corrispondenze di struttura. Più forti saranno tali corrispondenze maggiormente simili saranno le esperienze effettuate sulle due cose. In genere, nell'utilizzare una simulazione, si sfrutta la non perfetta corrispondenza in maniera vantaggiosa, selezionando solo quelle similitudini che sono utili e scartando le altre. Di fatto, in tal modo, con l'utilizzo della simulazione possiamo evitare di incorrere nelle eventuali conseguenze negative che l'esperienza reale potrebbe produrre. Conseguenze che nel caso limite potrebbero essere pericolose o addirittura fatali.

Ritorniamo a esaminare il processo di comunicazione. Contrariamente a quanto può forse apparire a prima vista, l'uso dei simboli richiede un artificio maggiore ed è per questo motivo "meno naturale" rispetto all'uso delle simulazioni. Il loro utilizzo si basa, infatti, sostanzialmente su di un "trucco". Tale trucco è di avere preventivamente associato in maniera artificiale, da qualche parte nella mente sia di chi fa da sorgente, sia del destinatario, la percezione dell'oggetto (o del fenomeno) che fa da simbolo con la memoria di un comune particolare contenuto cognitivo.

Quest'associazione in linea di principio può essere completamente arbitraria! Essa costituisce un artificio nel senso che non è assolutamente necessario che vi sia alcuna corrispondenza di struttura tra simbolo e il contenuto cognitivo associato. Essenzialmente tutto ciò che si richiede per utilizzare dei simboli è che essi siano allo stesso modo condivisi tra più persone. In genere un simbolo deve anche possedere la caratteristica di essere facilmente producibile per la sorgente e facilmente rilevabile per il destinatario.

Sembra quasi legittimo a questo punto affermare che il "significato di un simbolo" è proprio quel particolare "contenuto cognitivo" al quale esso è convenzionalmente associato, in modo comune, nelle menti di chi lo utilizza. Quest'affermazione non è ancora completamente legittimabile poiché per il momento non sono ancora state formulate delle ipotesi precise su cosa sia questo "quid" che chiamo "contenuto cognitivo".

Per utilizzare dei simboli nella comunicazione è prima necessario aver concordato un certo linguaggio, ossia, in altre parole, aver concordato la corrispondenza tra i simboli, gli oggetti e i fenomeni, reali o immaginari, ai quali sono associati. Si noti bene che tale corrispondenza è in linea di principio completamente convenzionale. Per usare le parole del linguista Ferdinand de Saussure: "**Non c'è nessuna relazione naturale tra il significante e il significato**."

Diversamente, per usare le simulazioni nella comunicazione non è necessario utilizzare artifici di questo genere perché in molti casi non hanno bisogno di alcun preaccordo artificiale e convenzionale di sorta, essendo invece dotate proprio di una "relazione naturale" con il loro referente. Tale relazione è costituita dal fatto che esse presentano le menzionate corrispondenze strutturali con ciò che rappresentano.

Si può dunque affermare che le simulazioni, o meglio le rappresentazioni strutturali, costituiscono "un sistema diretto per comunicare", mentre, come abbiamo visto, l'uso dei simboli è in questo senso "artificiale".

Ma perché allora utilizziamo naturalmente e da svariati millenni un sistema simbolico? Sostanzialmente per due ragioni. La prima è che la comunicazione simbolica è in molte circostanze assai più efficiente. I suoni del parlato sono cose molto facili da produrre, mentre realizzare una buona simulazione è spesso tutt'altro che semplice. Nella maggior parte dei casi produrre simboli è assai più semplice che produrre simulazioni. La seconda ragione consiste nel fatto che vi sono dei limiti intrinseci a ciò che può essere comunicato tramite simulazioni. In particolare questi limiti diventano evidenti quando il contenuto da comunicare è un concetto astratto!

Ad ogni modo è anche interessante notare che vi sono circostanze nelle quali la comunicazione basata su simulazioni è più efficace di quella simbolica. Ad esempio tutti sappiamo che una fotografia o un disegno possono essere in taluni casi molto più efficaci di una descrizione verbale.

Riassumendo abbiamo dunque visto che:
- Per la comunicazione tra persone bisogna sempre produrre qualcosa nell'ambiente che può essere classificato o come simbolo o come simulazione.
- Le simulazioni a differenza dei simboli hanno sempre delle corrispondenze strutturali con ciò che rappresentano.
- Le simulazioni fungono da "sostituto esperienziale" di ciò che rappresentano.
- L'associazione tra simbolo e suo significato è convenzionale.
- E' sensato pensare che per utilizzare un certo sistema di simboli debbano esserci, entro un sistema cognitivo, degli appositi apparati nei quali sono associate le memorizzazioni (delle rappresentazioni) degli oggetti che fungono da simbolo con i relativi significati. Ciò costituisce un artificio in più, che non è necessario per la comunicazione che utilizza invece simulazioni.

Queste considerazioni portano a formulare l'ipotesi, che intendo esplorare in questo lavoro, che siano "queste simulazioni" a fungere da supporto fondamentale per le rappresentazioni primarie che costituiscono la conoscenza.

Preciso questo punto con la seguente congettura:
**Sono le rappresentazioni strutturali (che funzionano da simulazioni), e non i simboli, ad avere le proprietà necessarie per fungere da supporto primario per la conoscenza.**

Con questa congettura non intendo affermare che i simboli non svolgano un ruolo importante. Anzi, come vedremo, essi sono comunque fondamentali entro le rappresentazioni strutturali, anche se vanno inquadrati in modo diverso da quello abituale.
Ho affermato che le simulazioni fungono da sostituto del soggetto che rappresentano. Se la conoscenza si basa effettivamente su delle simulazioni, si può pensare che una delle sue funzioni sia quella, in certo qual modo, di sostituirsi alla realtà, di offrire appunto una specie di sostituto esperienziale. L'idea è che in questo modo si possano generare delle esperienze virtuali senza incorrere nelle conseguenze negative, che al limite potrebbero essere anche letali, che invece quelle reali potrebbero comportare. Come disse Karl Popper, ciò consentirebbe "alle nostre ipotesi di morire al posto nostro"!
L'idea di considerare il nostro sistema nervoso alla stregua di un generatore di modelli è stata proposta già nel 1943 da Kenneth Craik. Secondo Craik la macchina cerebrale funziona come un "simulatore" che dà al pensiero "il suo potere di predire gli avvenimenti", di anticipare lo svolgimento dei fatti sulla freccia del tempo.

Questo modo di vedere le cose induce anche a pensare che l'attività di ragionamento consista, in fin dei conti, nell'esplorazione, a livello di simulazione mentale, delle varie possibili evoluzioni delle situazioni, in particolare in funzione delle nostre possibilità di agire. Si tratterebbe quindi di una sorta di esplorazione entro lo spazio delle ipotesi o, se vogliamo, entro lo spazio "simulato" delle possibilità.

Altro punto notevole è che queste esplorazioni sembrano, in un certo senso, ricondurre i processi di base che costituiscono la cognizione a un approccio primariamente empirico. In effetti, avremo modo di vedere che spesso la "realtà macroscopica" è assolutamente troppo complessa perché si possa pensare di simularne l'evoluzione basandosi sui principi primi, e/o utilizzando solo le leggi della fisica. Se si dimostra vera l'ipotesi che il nostro sistema nervoso simula il mondo esterno, sicuramente non lo fa risolvendo complicatissime equazioni differenziali. Deve quindi ricorrere ad altre strategie. Ragionando su queste cose si arriva all'ipotesi che spesso esso usi, in qualche modo, la strategia di "ripetere qualcosa" che, in qualche modo, "è già contenuto" entro quelle che sono le esperienze empiriche dirette.

## 2.5   Sui limiti fondamentali del conoscibile

Le prime idee che mi hanno condotto a questo lavoro sono emerse da un tentativo di analisi dei limiti fondamentali di ciò che è conoscibile.

Si possono proporre varie argomentazioni a favore della tesi che il fenomeno della conoscenza si basi sulla possibilità di costruire rappresentazioni interne delle cose e dei fenomeni della realtà esterna. In merito a queste rappresentazioni congetturo che esse siano sempre relative alle **strutture** degli oggetti e dei fenomeni del mondo circostante e alle **operazioni** che si possono compiere su di esse. Penso sia legittimo proporre anche la congettura che non esiste alcun'altra possibilità di conoscere la realtà esterna (in modo razionale), se non facendo riferimento alle strutture delle cose e dei fenomeni, e alle operazioni possibili su di esse.

Ritengo che la precisazione razionale di questa congettura richieda l'analisi dei concetti di struttura, di operazione e, come vedremo, la loro unione nel concetto di schema. Essa richiede inoltre l'individuazione degli strumenti matematici adatti alla precisa definizione di questi concetti.

Quest'analisi permette a sua volta di rendere evidenti alcuni fenomeni interessanti:

- le strutture si prestano naturalmente ad operazioni di derivazione gerarchica;
- con esse, assieme a poche operazioni di base, è possibile rappresentare

ogni sistema computazionale;
- al loro interno è possibile esprimere gli oggetti matematici fondamentali e , per certi aspetti, il concetto di struttura appare "più primitivo" rispetto ai concetti di insieme, di varietà e di numero;
- con questi strumenti è possibile, cosa davvero importante, proporre una definizione precisa per i concetti di regola, di regolarità e di proprietà emergente.

**Alcune precisazioni linguistiche.**

In generale per "**sistema cognitivo**" possiamo intendere la generalizzazione di qualunque sistema che sia in grado di cognizione della realtà.
Con quello di "sistema cognitivo" si può anche introdurre il concetto di "**realtà esterna**". Per il momento possiamo intendere quest'ultima semplicemente come il complesso delle entità reali di cui si può avere cognizione.
In generale "la realtà esterna" è per un sistema cognitivo la fonte dei soggetti delle proprie rappresentazioni interne.
Fino ad ora non sono ancora stati introdotti gli strumenti che permettono di spiegare, con sufficiente precisione, cosa sia un sistema cognitivo. Per questo motivo queste appena proposte vanno intese, per il momento, solo come delle utili precisazioni linguistiche.

## 2.6  Prima congettura di riferimento

*"La scienza può solo farci conoscere i rapporti tra le cose e non le cose in quanto tali: al di là di questi rapporti non c'è alcuna realtà conoscibile".*
Questa è una delle conclusioni cui giunse Henri Poincarè nel suo libro La science et l'Hypothèse, pubblicato nel 1902.
Uno dei concetti del senso comune che più di altri sembra in grado di esprimere questa idea dei "rapporti tra le cose" come l'unica proprietà conoscibile della realtà esterna, è quello di struttura.
Intuitivamente la struttura si occupa proprio dei rapporti, delle interrelazioni tra le cose, o meglio, tra "**le parti che costituiscono le cose**". La struttura in qualche modo si riferisce a relazioni matematicamente ben descrivibili che intercorrono tra le parti costituenti.
Avendo in mente una precisazione razionale del concetto di struttura, che descrivo nel prossimo capito, credo sia utile proporre una reinterpretazione dell'idea espressa da Poincarè con la congettura che segue.

Congettura
**Degli oggetti e dei fenomeni della realtà esterna a un sistema cognitivo, tutto ciò che è rappresentabile all'interno dello stesso si limita alle loro strutture e alle operazioni possibili su di esse.**

Questa congettura può anche essere espressa con altre parole che, pur non cambiando il significato, aiutano l'intuizione, nel modo seguente: "di ogni cosa esterna alla nostra mente, solo la sua struttura è rappresentabile entro di noi e può quindi essere oggetto di attività cognitiva". Oppure anche affermando che "il conoscibile si limita alla struttura delle cose".

## 2.7  Alcune note di approfondimento

Concetti di questo tipo sono in realtà già stati espressi più volte sia nell'ambito della scienza cognitiva sia in altre discipline e, in effetti, se ci si limita ad utilizzare il concetto intuitivo di struttura, le affermazioni fatte sopra non aggiungono particolari contenuti innovativi. Queste affermazioni hanno un significato non banale, ma senza un'adeguata analisi del concetto di struttura, vale a dire di cosa si debba intendere con esso e di come lo si può descrivere in termini matematicamente precisi, si perde la possibilità di capire cose molto importanti per la comprensione dei fenomeni cognitivi. Nei testi più validi sulla scienza cognitiva, e in generale in quei lavori nei quali si affronta la problematica delle rappresentazioni cognitive (intese come una generalizzazione dell'idea di rappresentazioni mentali), il concetto di struttura è spesso tirato in ballo ma, almeno per quanto è a mia conoscenza, non nel modo opportuno.
Spesso in scienza cognitiva si utilizzano concetti come quelli di approccio simbolico, connessionista e ibrido. Come illustrato nell'introduzione, alcuni parlano in termini di approccio simbolico e subsimbolico. In "mereotopologia", ad esempio, si cerca di costruire delle teorie formali utilizzando come strumento di rappresentazione la logica del primo ordine.
A mio avviso si utilizzano metodi non adatti a mettere in luce le proprietà più importanti delle rappresentazioni strutturali. In tutti questi approcci mancano alcuni strumenti fondamentali. Sono necessari gli strumenti che, tra le altre cose, permettono: di eseguire delle operazioni di derivazione strutturale, di ordinare le rappresentazioni strutturali in modo gerarchico, di definire i concetti di regolarità e di regola, e che consentano di definire un criterio di emergenza con il quale poter stabilire quando una data operazione di "derivazione strutturale" è lecita o no. Il tutto va fatto secondo modalità che siano in grado di cogliere in modo diretto, e senza forzature, le proprietà naturali delle strutture degli oggetti e dei fenomeni reali. Infatti, se deve essere stabilito un ordine gerarchico tra le varie tipologie di rappresentazioni strutturali, è bene che questo emerga in modo naturale dall'analisi delle strutture di base e dei percetti sensoriali primari. Il senso di tutto questo sarà chiaro nei prossimi capitoli.
Esiste già una formalizzazione matematica del concetto di struttura in quelle che sono chiamate "strutture relazionali". Credo che essa sia poco adatta per descrivere gli oggetti e i fenomeni delle nostre percezioni. L'idea di struttura relazionale è stata sviluppata per essere applicata agli oggetti particolari di cui si occupa solitamente la matematica (come ad esempio gli insiemi infiniti dei numeri naturali, reali ecc.. e delle loro possibili funzioni). Nel far questo è maturata un'impostazione che non è immediatamente applicabile nel modo migliore alle strutture degli oggetti della nostra quotidianità. In matematica si parla delle strutture di insiemi che sono molto spesso infinitamente grandi, i cui elementi sono associati gli uni agli altri da varie operazioni di composizione interna e confrontati a coppie secondo particolari relazioni binarie. Per definire queste operazioni e queste relazioni, si usa una nozione di insieme che viene spesso data come primitiva e che viene pensata come "onnipotente" e senza limiti. Queste idee non corrispondono alle proprietà delle strutture degli oggetti e dei fenomeni della nostra quotidianità. I metodi utilizzati per trattare le strutture relazionali non si prestano bene per alcune fondamentali operazioni di analisi strutturale che, come vedremo, sono cruciali nei processi cognitivi.

Ciò che serve è sostanzialmente un approccio che, pur mantenendo rigore e precisione, usi un linguaggio e degli strumenti più adatti all'analisi delle strutture degli oggetti concreti. Serve un approccio che, nella sostanza, sia in grado di precisare quello intuitivo che utilizziamo naturalmente per descrivere le cose e i fenomeni della nostra quotidianità.

# 3 Strutture. Operazioni fondamentali sulle strutture. Schemi.

## 3.1 Introduzione: Il problema del confronto

Il capitolo che segue è molto tecnico, e per questo motivo potrebbe risultare noioso da seguire. Per capirne il senso penso possa essere utile cominciare con un problema che contiene alcuni aspetti molto importanti. Essi ci possono aiutare a comprendere di quali strumenti abbiamo bisogno per capire la cognizione.

Supponiamo di avere dei fogli bianchi con disegnate sopra delle figure geometriche, ad esempio dei poligoni regolari, e di volere confrontare fra loro queste figure per capire se hanno o non hanno qualcosa in comune. Il punto cruciale è che ci poniamo il problema di far fare questo a un calcolatore; in altre parole ci chiediamo come questo **problema di confronto di strutture** possa essere affrontato concretamente dal punto di vista computazionale.

Supponiamo quindi di avere una telecamera digitale, connessa con un calcolatore, in grado di inquadrare i fogli e di acquisire le immagini. Nel caso specifico il problema di "visione" è alquanto semplice. Le linee risaltano molto bene sullo sfondo, e per descrivere lo stato dei singoli pixel bastano in realtà due bit: 0 per il colore bianco e 1 per il nero. Le intere immagini corrispondono allora a matrici, quindi, in linguaggio informatico, a vettori bidimensionali di bit, che possono assumere i due valori: 0,1.

Come si possono confrontare due immagini? Partiamo dalla relazione di "uguaglianza": quando possiamo dire che due figure sono uguali?

Nella geometria di Euclide, per verificare se due figure sono uguali, si suggerisce di muoverle con **movimento rigido,** che non le deformi, fino a farle combaciare: se tutti i punti si **sovrappongono** allora le due figure sono uguali.

Far fare la stessa cosa a un calcolatore non è banale, ma nemmeno impossibile, anzi l'algoritmo necessario è abbastanza semplice. Si tratta di prendere una delle due matrici e traslarla e ruotarla a piccoli passi in tutti i modi possibili: sono tanti ma non infiniti. Se si trova una configurazione dove tutti i bit **corrispondono contemporaneamente**, allora le due figure geometriche rappresentate sono uguali; se non si trova questa combinazione allora significa che le due figure devono essere diverse almeno per un bit, e quindi per almeno un "pixel equivalente" (trascuriamo pure eventuali problemi dovuti alla discretizzazione delle immagini). Invito il lettore a riflettere per proprio conto su come dovrebbe essere implementato l'algoritmo che esegue questa verifica dell'eventuale coincidenza (e su quanto poco efficiente sia).

Si noti anche la seguente cosa: quando è un operatore umano che deve eseguire l'operazione di movimento rigido, non proverà tutte le possibili posizioni spostando passo per passo la figura, ma muoverà direttamente le figure nella direzione giusta che consente di sovrapporre da subito elementi importanti quali le linee e gli angoli simili.

Chiediamoci: cosa succede se le due figure hanno dimensioni diverse? Supponiamo di ritrarre due triangoli equilateri, ma con i rispettivi lati di dimensioni diverse. In questo caso il nostro primo algoritmo, che trasla e ruota le immagini, non sarà mai in grado di trovare una **sovrapposizione completa** tra i bit a 1.

Una soluzione potrebbe essere quella di provare molti "cambiamenti di scala". In teoria le variazioni di scala possibili sono infinite e, da questo punto di vista, se non si parte nella direzione giusta si corre il rischio di scrivere un algoritmo che non si ferma mai. Ad ogni modo con un po' di accortezza è comunque possibile scrivere un algoritmo che, procedendo per gradi ed eseguendo molte più prove, riesce comunque alla fine a trovare di nuovo delle buone sovrapposizioni.

Ma chiediamoci ora: cosa succede se i due triangoli non hanno angoli uguali? Supponiamo che in un foglio ci sia un triangolo rettangolo e nell'altro un triangolo equilatero. Nessuna combinazione di operazioni di variazione di scala, di rotazione e di traslazione è in grado di far sovrapporre fino a far combaciare le due figure.

Nonostante questo, noi comprendiamo che le due figure sono strutturalmente simili. Comprendiamo che hanno qualcosa in comune e che questo qualcosa riguarda proprio le loro strutture, proprio quelle strutture che ho congetturato costituiscano ed esauriscano quanto può essere colto dalla cognizione. Se davvero la cognizione si basa sulle corrispondenze strutturali che sussistono tra le nostre rappresentazioni interne e le cose del mondo esterno, allora comprendere in cosa consistono queste corrispondenze è fondamentale!

Un buon programmatore che si occupa di pattern recognition sarà in grado di ideare, per questo problema specifico (confronto di figure geometriche semplici composte di linee nere su sfondo bianco), degli algoritmi in grado di cogliere, in qualche modo, alcune delle similitudini che ci possono essere tra triangoli, o anche tra altri poligoni. Ma sarà in grado di scrivere un algoritmo generale che va bene in tutti i casi e che è in grado di cogliere tutte le similitudini che ci possono essere tra due strutture?

Il problema illustrato è volutamente molto semplificato. La nostra mente è in grado di cogliere la presenza di somiglianze strutturali tra entità molto più complesse di "semplici" figure geometriche regolari, e questa sua capacità non si limita ai problemi di confronto visivo, o di confronto delle altre tipologie di informazioni sensoriali. La nostra mente è capace di riconoscere in modo molto efficiente similarità che sono presenti anche tra "strutture molto astratte".

Se vogliamo comprende quali sono i segreti della conoscenza dobbiamo capire come si fa a cogliere le corrispondenze strutturali in generale. Dobbiamo capire quale è, se esiste, il "trucco generale" di questa capacità, ma soprattutto dobbiamo capire quale è la sua logica profonda.

Penso di essere in grado di proporre idee molto interessanti su questo problema. Ma per illustrarle ho bisogno di introdurre una serie di strumenti nei prossimi capitoli. Lungo l'esposizione ritornerò su questo problema del confronto tra strutture, ma solo dal quarto capitolo avremo gli strumenti per focalizzare la questione.

Rielaborando il pensiero di Poincarè ho proposto la congettura che il conoscibile del mondo esterno si limiti alla struttura delle cose e alle operazioni che sono possibili su queste strutture.
In questo capitolo illustro una metodologia che permette di trattare in modo preciso le **strutture**, le **proprietà strutturali**, e le **operazioni** possibili su queste, in termini matematicamente ben definiti e che si adattano bene alle necessità dell'attività cognitiva.
Vedremo nei prossimi capitoli che con questi strumenti è possibile ottenere dei risultati notevoli, tra i quali la definizione rigorosa del concetto di regola e la spiegazione del fenomeno dell'emergenza.

Propongo di distinguere tre tipologie di "oggetti strutturali" che indicherò come:
- **strutture di prima specie,**
- **strutture di seconda specie,**
- **proprietà e relazioni strutturali non autonome** (di prima e seconda specie) .

A questi oggetti vanno aggiunte le **operazioni** sulle strutture, che saranno trattate nella seconda parte del capitolo.
Gli oggetti che chiamo **strutture di prima specie** sono particolarmente importanti poiché appaiono essere l'unico tipo d'informazione strutturale direttamente memorizzabile. Vedremo che gli elementi basilari della computazione sono costituiti da questo tipo di strutture e dalle operazioni che si possono eseguire su di esse.
Vedremo che le strutture di prima specie costituiscono il punto di riferimento. Esse si possono memorizzare, e per il loro riconoscimento può essere sufficiente un'operazione di confronto tra quanto memorizzato e il nuovo input.
Utilizzando la funzione simbolica appare possibile mettere assieme strutture di prima specie e operazioni per ottenere un unico oggetto matematico che possiamo chiamare **schema**, o anche **struttura di seconda specie**. Con esso è possibile definire la "struttura" di un complesso di operazioni, quindi di un algoritmo.
Vedremo che le strutture di seconda specie hanno bisogno di utilizzare la funzione di associazione simbolica tra alcuni degli elementi che definiscono la struttura stessa e dei congegni in grado di eseguire fisicamente le operazioni di computo elementari.

Alcuni altri oggetti, comunque identificabili tramite procedure algoritmiche, costituiscono delle **proprietà e delle relazioni strutturali non autonome**. Esse possono essere pensate come delle "**quasi strutture**", nel senso che le informazioni che le definiscono non sono in genere sufficienti per identificare in maniera precisa e completa una singola struttura (di prima o seconda specie). L'idea è che questi oggetti, che consistono in **proprietà** e **relazioni strutturali,** utilizzano comunque una parte delle stesse informazioni che servono per definire delle strutture complete. Grazie a questa loro caratteristica spesso possono essere utilizzati per descrivere un'intera classe di strutture: tutte quelle che possiedono quelle particolari proprietà.

Penso si possa ben argomentare che in ogni caso una struttura è qualcosa che dipende da come alcune parti sono distinguibili tra loro e da come stanno le une rispetto alle altre. Questa impostazione si focalizza su quei "rapporti tra le cose" che come affermava Poincarè, costituiscono l'unica realtà conoscibile, in modo razionale, del mondo esterno.

Credo che questo sia vero sostanzialmente perché questi rapporti tra le cose sono probabilmente l'unico aspetto della realtà che può esserci in comune tra entità fisicamente diverse.

Penso che il punto di forza della trattazione che propongo consiste nel fatto che mette a disposizione gli strumenti per **confrontare le strutture** e per identificare le loro analogie, in modo particolarmente efficiente. Questa possibilità è fondamentale poiché permette di identificare le eventuali regolarità presenti al loro interno. A loro volta, come vedremo, le regolarità e le regole sono indispensabili e costituiscono il "motore della cognizione".

Nelle pagine che seguono baserò la descrizione del concetto di struttura focalizzando l'attenzione sulle idee di "**distinguibilità interna**" e "**distinguibilità esterna**" tra le parti che compongono un oggetto.

Questo modo di impostare le cose comporta una serie di vantaggi e sembra riflettere il nostro modo naturale di organizzare le informazioni sul mondo. Vedremo che impostando le cose in questa maniera è possibile compiere importanti operazioni che chiamo **derivazioni strutturali** e che permettono di ordinare gerarchicamente le rappresentazioni delle strutture della realtà e soprattutto di confrontarle in modo efficiente.

**Prima parte: Strutture di prima specie**

### 3.2 Una metodologia per la rappresentazione delle strutture di prima specie

Per focalizzare le prime idee, partiamo da alcune osservazioni apparentemente semplici, ma non per questo banali. Il lettore tenga presente che lo scopo delle pagine che seguono è quello di identificare una serie di strumenti formali che ci permettano di trattare con precisione il concetto di struttura. Quando ci si cimenta nel problema dell'identificazione degli strumenti di base è in generale necessario porre molta attenzione su questioni di dettaglio, anche quando possono sembrare banali.

Osservazione:
- affinché si possa parlare di struttura di un oggetto è necessario che questi sia composto da una "**molteplicità di parti**".

Assumiamo quindi che la "molteplicità di parti" sia richiesta per ogni struttura, anche quando questa è composta da un solo elemento (è solo apparentemente contraddittorio).

Osserviamo inoltre che:
- affinché vi possa essere una molteplicità di parti queste devono essere in qualche modo tra loro "**distinguibili**".

Questo è un principio generale, e non banale, a cui si farà in seguito riferimento. Ricordiamo quindi che: ogni "molteplicità" richiede sempre che sia data a priori la possibilità di "distinguere" tra loro degli elementi.

Notiamo anche che:
- in ogni struttura le parti della stessa stanno le une rispetto alle altre in certe "relazioni reciproche".

A volte può essere che una "relazione" sia qualche cosa di riducibile ulteriormente, ma a volte capita anche che la relazione sia semplicemente una distinguibilità tra le parti che non sappiamo ridurre ad alcunché di più fondamentale. Come sarà più chiaro fra non molto, propongo di rappresentare tali relazioni non riducibili con dei **rami di grafo**.

La metodologia che dopo vari tentativi mi è sembrata più funzionale per precisare in senso matematico il concetto di struttura di prima specie consiste nel pensare alle strutture come ad oggetti matematici composti da tre elementi base: un insieme di elementi che costituiscono "le parti della struttura", un certo "sistema di distinguibilità interna tra le parti", e un certo "complesso di relazioni esterne". Tutti questi tre punti sono precisabili rigorosamente con linguaggio matematico come verrà illustrato nelle pagine che seguono.

Questa metodologia si è rivelata pertinente perché si presta naturalmente per descrivere le strutture degli oggetti della nostra quotidianità e soprattutto perché consente in maniera naturale di eseguire operazioni di astrazione. Come vedremo essa può essere applicata anche per descrivere gli oggetti matematici standard.

Quando osserviamo un oggetto fisico nella nostra mente tendiamo in modo naturale a suddividerlo nelle sue parti componenti. Una sedia, ad esempio, è composta dalle gambe, dal piano di seduta, dallo schienale. Tutte queste parti stanno le une rispetto alle altre in reciproche relazioni geometriche riconoscibili e classificabili entro un insieme finito, anche se talvolta molto ampio, di possibilità. Se le gambe fossero attaccate direttamente allo schienale in modo da formare una struttura bizzarra, non saremo più di fronte ad una sedia, ma a un altro oggetto, anche se composto dagli stessi elementi. Notiamo che spesso le parti componenti sono a loro volta degli oggetti con la loro propria struttura interna.

Propongo quindi la seguente definizione:

In generale una **struttura di prima specie** è univocamente identificabile da:
- **un dato insieme di parti;**
- **un sistema di distinguibilità interna tra le parti;**
- **un complesso di relazioni esterne tra le parti.**

Questi tre punti sono tutti definibili in modo preciso, come sarà illustrato fra poco, e costituiscono gli elementi necessari per definire e rappresentare una certa struttura "statica" generica.

Questo implica quanto segue.

Due strutture di prima specie sono da considerarsi "**strutturalmente isomorfe**" se si verificano le seguenti tre condizioni:
- **esiste una corrispondenza biunivoca tra i rispettivi insiemi di parti;**
- **coincidono i rispettivi sistemi di distinguibilità interna;**
- **coincidono i rispettivi complessi delle relazioni esterne.**

Questa **proprietà d'isomorfismo** strutturale è fondamentale poiché permette di confrontare due strutture di prima specie e di dire se sono o no uguali[4].

In generale possiamo dire che due strutture di prima specie **coincidono strutturalmente** se sono **isomorfe.**

In questo lavoro l'uso del termine "parte" è più restrittivo del suo significato nel linguaggio comune. Spesso per parte di una cosa s'intende anche una sua porzione. In questo lavoro preferisco tenere separati questi due concetti; uno dei motivi di questa scelta potrà essere compreso tra qualche pagina, quando introdurrò il concetto di struttura quoziente. La giustificazione completa potrà invece essere chiara solo più avanti.

---

[4] In realtà avremo modo di vedere che il concetto stesso di uguaglianza può essere opportunamente analizzato e interpretato, e vedremo che può essere pensato come funzione degli effetti che una certa entità complessa ha sul mondo esterno.

Si noti ancora che il linguaggio proposto è leggermente differente da quello usato comunemente in matematica. In questo lavoro preferisco indicare come "parte" ciò che usualmente in insiemistica è invece chiamato "elemento". In matematica con l'espressione "insieme delle parti" s'intende la collezione di tutte le possibili porzioni di un certo insieme di partenza. In questo lavoro utilizzo tale terminologia con significato differente.

Come vedremo, data una certa struttura, risulta in genere possibile **derivare da essa delle altre strutture**. Più avanti, per riferirmi ad alcune specifiche tipologie di esse, userò termini quali: "struttura quoziente", "struttura morfismo", "struttura porzione" e altri.

Quindi, data una certa struttura A, si possono spesso da essa ottenere, con opportune operazioni, delle altre strutture B, C, D ecc.. che si diranno: "**strutture derivate da A**". Allo stesso modo potranno esserci strutture derivate dalla struttura B, che è stata a sua volta derivata da A e cosi via (in genere non indefinitamente).

E' possibile ordinare gerarchicamente una certa famiglia di strutture in base ai rispettivi rapporti di derivazione. Vedremo questo più avanti.

Indichiamo qui con le lettere minuscole le parti di una certa struttura. Quindi se A è la struttura, le sue parti potranno ad esempio essere gli elementi a, b, c, d.

Notiamo che l'insieme { a, b, c, d } **non individua univocamente la struttura A.** In effetti, strutture diverse possono essere costituite dallo stesso insieme di parti.

Definizione
In generale data una certa parte "a", essa può avere o non avere a sua volta una sua propria struttura che si dirà: **struttura interna di "a"**.

Definizione
Se una o più parti di una certa struttura A, hanno a loro volta una loro struttura interna, allora la struttura A sarà "riconducibile" ad almeno un'altra struttura $A^{-1}$ di livello gerarchico inferiore.

$A^{-1}$ sarà data dalla struttura che si ottiene considerando il complesso delle strutture delle parti di A nelle relazioni esterne che esse hanno in A stessa.

Congettura
Si ammette per ipotesi la possibilità di strutture che non sono riducibili ad altre di livello inferiore. Tali strutture si diranno: "**strutture base**". Per ipotesi le parti di una struttura base sono prive di struttura interna.

Nota:

In un tal senso dunque si può dire che A^-1, quando esiste, è più "basilare" di A.
Ho scritto sopra che una generica struttura A risulta definita da:
- l'insieme delle sue parti;
- il sistema di distinguibilità interna tra le parti;
- il complesso delle relazioni esterne tra le parti.

Vediamo ora di precisare questi punti.

### 3.2.1  L'insieme delle parti

Sia A una struttura. Ad essa risulta associabile l'insieme che ha per elementi le sue parti: $\{a_1, a_2,.... a_i,....a_n\}$.
Possiamo usare la seguente simbologia **Idp(A)**: semplicemente la sigla sta per: "insieme delle parti della struttura A". Quindi Idp(A)= $\{a_1, a_2,.... a_i,....a_n\}$
In genere interessano strutture finite e quindi si può parlare del numero delle parti di una struttura A. In tal caso quindi si tratta di un numero naturale e lo possiamo indicare con N°(A).
Comunemente in matematica, e in particolare nella teoria degli insiemi, si presuppone che gli elementi di un dato insieme siano sempre a priori ben distinguibili gli uni dagli altri. Nella teoria che propongo questo passaggio è più delicato. Penso convenga parlare di "**distinguibilità interna** e **distinguibilità esterna**". Si possono avere casi nei quali due parti di struttura possono essere **indistinguibili internamente**.

### 3.2.2  Il "sistema di distinguibilità interna" tra le parti

Ho affermato che le parti possono avere, a loro volta, una propria struttura interna. Però tali strutture interne in molti casi **non sono "accessibili",** il che implica che un sistema cognitivo **può non avere modo di acquisire informazioni complete su di esse**. Ciò nonostante è possibile comunque definire con precisione la struttura A, a condizione però di avere l'informazione che dice come le parti di A sono tra loro **distinguibili**.
Spesso può capitare che alla distinguibilità tra le parti di A si possa far corrispondere un numero naturale M°(A), che può essere minore o al massimo uguale ad N°(A).
Se risulta M°(A)<N°(A) significa che ci sono almeno due parti di A che non sono tra loro distinguibili internamente. Ciò significa che **scambiando tra loro queste due parti si ottiene una struttura isomorfa alla prima e quindi cognitivamente indistinguibile da essa.**
Attraverso questa proprietà si può ben definire la "distinguibilità interna tra le parti":

Definizione

Due o più parti di una certa struttura A sono "internamente indistinguibili" se con le loro permutazioni entro A si ottengono sempre strutture isomorfe tra loro (e indistinguibili per permutazione).

In generale se M°(A) è il numero associato alla distinguibilità interna delle parti di A significa che si possono individuare M° "tipi" di parti.
Se due o più parti di struttura sono tra loro distinguibili internamente allora significa che tali parti hanno a loro volta struttura interna. Il sistema di distinguibilità interna può essere anche molto complesso.
Ovviamente in una struttura base le parti costituenti non sono distinguibili internamente.
La relazione di indistinguibilità tra due parti di Idp(A) è una relazione di equivalenza. Attraverso essa è definito un insieme quoziente rispetto tale relazione che è l'insieme dei tipi di parti della struttura A. Indico tale insieme con la scrittura: Itp(A).

### 3.2.3 Il "complesso delle relazioni esterne"

Si consideri una struttura B avente delle parti che non sono tra loro distinguibili internamente. In tal caso vi sono solo due possibilità: o le parti in questione non hanno struttura interna oppure tali strutture sono tra loro isomorfe. Tali parti però sono comunque molteplici. Poiché, come rilevato in precedenza, ogni molteplicità è sempre accompagnata dalla possibilità di fare delle "distinzioni", vi deve essere qualcosa che distingue tra loro le parti, ma che non fa riferimento alla loro struttura interna. **A tale distinguibilità deve allora provvedere il complesso delle relazioni esterne.** Quindi, in generale, per ogni struttura il complesso delle relazioni esterne determina sempre una **distinguibilità esterna** tra le parti. Questa distinguibilità esterna è qualcosa che possiamo pensare come ciò che rende ad esempio i punti di uno spazio distinguibili gli uni dagli altri, o più in generale possiamo vederla come ciò che rende distinguibili gli elementi di una generica varietà (anche costituita da un numero finito di elementi). In realtà non siamo sempre in grado di ridurre la distinguibilità stessa a qualcosa di più basilare; dobbiamo però riconoscere la sua **necessità inalienabile** affinché si possa parlare di qualsiasi struttura e di qualunque molteplicità di parti.
Per alcune strutture, ma non per tutte, alla distinguibilità associata al complesso delle relazioni esterne tra le parti vi può essere anche, in aggiunta, una distinguibilità interna tra le parti stesse. In generale però ammettiamo la possibilità di strutture le cui parti non presentano distinguibilità interna di sorta e per le quali quindi, la capacità di distinguere comunque le parti che le costituiscono, è dovuta al complesso delle relazioni esterne. Chiaramente strutture di questo tipo non sono riducibili e sono quindi delle **basi**.
Quali metodi abbiamo allora per rappresentare con precisione il complesso delle relazioni esterne tra le parti di una certa struttura generica?

In parte la questione è stata già affrontata con la **geometria**. La geometria ci fornisce gli strumenti per rappresentare le strutture di tipo topologico. Tuttavia questo tipo di trattazione non si presta bene per le strutture tipiche che sono rappresentate in un sistema cognitivo, perché queste ultime sono discrete.

Si prestano invece molto bene i metodi che derivano da una rappresentazione molto simile a quella possibile con i "**grafi**". A essa si farà riferimento in questo lavoro.

Una struttura il cui complesso delle relazione esterne è del tipo "**a grafo**" può essere rappresentata semplicemente elencando le coppie di parti che sono adiacenti. Più precisamente si elencano tutte le parti della struttura, ad esempio: $a_0, a_1, a_2, \ldots a_n$. Si elencano quindi tutte le coppie che sono tra loro collegate direttamente, ad esempio: $(a_1, a_4), (a_4, a_3), \ldots (a_i, a_j) \ldots (a_x, a_n)$.

In quest'ultimo elenco una singola parte della struttura può anche apparire più volte: essa può quindi essere collegata direttamente con più di un'altra parte. Se due parti sono collegate direttamente allora si dicono "**adiacenti**". Nell'elenco delle coppie non può però mancare nessuna delle parti che costituisce la struttura. In tale caso, infatti, la parte esclusa sarebbe isolata e non apparterebbe alla struttura stessa[5].

Se il complesso delle relazioni esterne tra le parti di una certa struttura A è esprimibile attraverso un grafo allora si può pensare di indicarlo con la scrittura Grafo(A).

Quindi possiamo dire che in molti casi di interesse pratico una certa struttura risulta individuata dagli insiemi Idp(A), Itp(A), dal Grafo(A) e da due applicazioni: di Idp(A) a Itp(A) e di Idp(A) ai nodi di Grafo(A).

Si noti ancora che topologia, geometria e grafi si candidano per rappresentare direttamente strutture i cui elementi non hanno distinguibilità interna.

In generale se una struttura A è caratterizzata da un certo complesso di relazioni esterne, costituito ad esempio da un grafo, una struttura di livello gerarchico superiore deriverà, con metodi opportuni, il proprio complesso di relazioni esterne da quello di A.

Nota: riporto la definizione di grafo.
Si considerino due insiemi disgiunti V e S, rispettivamente l'insieme dei vertici e l'insieme degli spigoli. Si considerino due elementi appartenenti all'insieme dei vertici: x e y. Le coppie che contengono gli stessi due elementi sono da considerarsi equivalenti: (x,y) è equivalente a (y,x). Sia E tale relazione di equivalenza.

---

[5] Un punto da segnalare, è che in taluni casi potrebbe essere utile utilizzare dei grafi orientati, il che significa dati due elementi $a_x, a_y$ si distingue la coppia $(a_x, a_y)$ da quella $(a_y, a_x)$. In questo caso la struttura è però più complessa poiché è aggiunta distinguibilità interna alle coppie di elementi.

Sia g: S□ VxV / E un'applicazione. La terna ( V,S,g ) si dice "grafo" con V insieme dei vertici e S insieme degli spigoli. Se s□ S e g(s)=[(x,y)] allora si dice che x e y sono collegati attraverso s. Nel caso in cui x=y lo spigolo si dice cappio. Se g(s1)=g(s2) con s1 e s2 □□S e s1 □□s2, allora si parla di biangolo. Nella presente trattazione interessano grafi finiti senza cappi né biangoli; in tale caso in matematica si parla di grafi completi.

Non credo sia necessario, almeno a questo stadio di sviluppo del lavoro, spendere altro tempo su questioni di costruzione formale. Probabilmente non conviene pensare in termini di applicazioni tra i vari insiemi, ma conviene invece usare le idee esposte in modo intuitivo, in particolare per quanto concerne i concetti di distinguibilità interna e di complesso delle relazioni esterne. Vedremo fra non molto che per molte strutture di "livello gerarchico superiore" le relazioni esterne tra le singole parti diventano anch'esse "informazioni strutturali complesse". Assumo che esse derivino comunque sempre da un complesso di relazioni rappresentabile in maniera semplice attraverso concetti come quello di distanza geometrica o quello di ramo di grafo.

Nelle **strutture base** utilizzate nell'attività cognitiva abbiamo a che fare con strutture le cui relazioni esterne sono espresse in termini di adiacenza tra le parti componenti.

La relazione di adiacenza è quindi l'elemento fondamentale per descrivere il complesso delle relazioni esterne in strutture base.

Nota
Una relazioni esterna, di adiacenza, rappresentabile tramite un ramo di grafo è probabilmente la più semplice che possa essere definita (almeno con questi strumenti). Per le strutture che invece non sono di base, le relazioni esterne possono essere oggetti a loro volta complessi. Per questo motivo alcune relazioni esterne saranno delle entità nelle quali si possono riconoscere e rendere esplicite una serie di "caratteristiche strutturali". Credo che in generale le relazioni esterne sono "ereditate" da quelle delle strutture di livello gerarchico inferiore. Se pensiamo che esistano delle strutture di base, allora le strutture derivate da queste (attraverso l'operazione di quoziente) avranno delle relazioni esterne, che possono essere descritte a partire dai "grafi di base". Le strutture successive a loro volta avranno altre relazioni esterne, che dipenderanno da quelle di livello inferiore. Man mano che si passa da un livello a quello successivo le relazioni esterne possono diventare complesse da descrivere. Per il momento suggerisco di non preoccuparci troppo di questo fatto; credo infatti che questa complessità non cresca indefinitamente e che spesso, da un certo livello in poi, i metodi per descriverle "si stabilizzino".

Un'altra idea interessante è che per ogni struttura topologica, che interessa per aspetti pratici e non solo teorici, sia possibile trovarne una discreta sufficientemente densa che la approssimi. Fatto questo, si ha che ogni struttura topologica è approssimabile con un'altra nella quale la descrizione del complesso delle relazioni esterne è di tipo a grafo.

Possiamo pensare che tutti gli oggetti con cui abbiamo a che fare in concreto siano immergibili entro una varietà discreta sufficientemente fine. Sotto queste condizioni possiamo anche pensare che tutte le relazioni geometriche "emergenti" tra le parti componenti siano in linea di principio derivabili dal grafo di quella di base.

Un'altra nota interessante, è che la metodologia proposta per rappresentare le strutture di base, può essere considerata un'estensione del metodo utilizzato per i grafi (senza cappi e biangoli). In particolare propongo di aggiungere la possibilità di distinguere tra loro i vertici per delle proprietà interne, come avviene nei "grafi colorati", e di ammettere che possano esistere relazioni esterne più complesse della semplice adiacenza.

### 3.3   Alcuni primi approfondimenti

Gli oggetti della nostra quotidianità hanno strutture che sono composte da parti che stanno le une rispetto alle altre in specifiche relazioni reciproche. Se si parte dalla tesi, sostenuta nella prima congettura di riferimento, che siano solo le relazioni tra le cose ad essere rappresentabili, e quindi conoscibili, potrebbe sembrare che anche le parti di per se stesse non sono conoscibili se non nella misura in cui esse stesse hanno una propria struttura interna, vale a dire sono a loro volta costituite da altre parti in specifiche relazioni reciproche. È evidente che procedendo in questa maniera si rimanda sempre a parti di parti, quindi a strutture di strutture. Per dare un senso al tutto può convenire partire dal presupposto che siano possibili strutture basilari le cui parti non hanno una struttura interna.

Nella ricerca di un inquadramento teorico del concetto di struttura ci si imbatte in una serie di problemi difficili su aspetti della realtà che riguardano i limiti del razionalizzabile. Un approccio naturale in questa tipologia di problemi è quello riduzionista, che cerca, per quanto possibile, di eliminare tutto ciò che appare "superfluo" per cercare di cogliere ciò che è invece essenziale per la ricerca che si sta seguendo. Procedendo in questa maniera, nel nostro caso ci si rende presto conto che si deve fare appello a idee basilari, come quella di "distinguibilità di parti" o come quella, in fin dei conti ad essa vicina, di "distinguibilità geometrica dei punti di una varietà", che non sono facilmente riducibili a loro volta a qualcosa di più semplice e basilare. Come si può spiegare cosa rende distinguibili i punti dello spazio o gli istanti di tempo?

Si tratta di problemi per i quali, nella nostra cultura matematica e filosofica, non ci sono degli strumenti che permettono di ricondurre la questione a qualcosa di ancora più basilare. In matematica di fatto si aggira la questione utilizzando uno strumento che si è mostrato molto duttile ed efficace: la teoria degli insiemi. Ogni struttura appare potenzialmente descrivibile tramite un sistema di insiemi e di applicazioni (nel senso di funzioni matematiche) che vincolano le relazioni e la distinguibilità tra le parti costituenti.

### 3.3.1 Sull'indipendenza esistenziale delle parti di struttura

Si tratta di un argomento difficile e non essenziale per gli scopi generali di questo lavoro. Segnalo comunque la questione, perché mi sembra per sé interessante e perché consente di comprendere alcune delle idee che hanno ispirato il modo proposto per inquadrare il concetto di struttura.
Consideriamo una struttura base. Per ipotesi le sue parti sono prive di struttura interna e per esse quindi non si applica il concetto di "distinguibilità interna". In questo caso, come detto, l'onere della distinguibilità di tali parti passa al complesso delle relazioni esterne. A questo punto credo possa essere interessante chiedersi se tali "parti" possano essere considerate come entità dotate di una propria indipendenza esistenziale. Possiamo chiederci se è lecito pretendere di definire il loro insieme e considerare questo come una costruzione più primitiva della struttura stessa. Chiaramente non esistono problemi a considerare come insiemi delle collezioni di oggetti complessi e macroscopici, come quelli che fanno parte della nostra quotidianità (ad esempio, l'insieme degli oggetti che sono sopra la mia scrivania in questo momento). Ma quando si considerano strutture limite, che rappresentano la riduzione massimale cui si può pervenire (si pensi, ad esempio, al concetto di varietà), il concetto d'insieme è ancora applicabile o costituisce una forzatura rispetto alla realtà delle cose?

### 3.4 Indistinguibilità per scambio di elementi strutturali corrispondenti

La prossima definizione può essere utile per affrontare il problema degli eventuali limiti del concetto di struttura. La sua giustificazione potrà essere chiare al lettore più avanti, quando tratterò il fenomeno delle strutture emergenti.
Può aver senso introdurre un concetto più forte di quello già proposto per la distinguibilità tra le parti associate di due o più strutture differenti.

Siano date due strutture finite A e B che siano isomorfe. Ciò significa, in base alla definizione proposta più indietro, che in A e in B è possibile individuare:
- lo stesso numero di parti,
- lo stesso complesso di relazioni esterne tra queste parti,
- lo stesso sistema di distinguibilità interna.

Abbiamo però anche visto che un sistema di distinguibilità interna resta definito dalla proprietà di permutazione tra le parti di una struttura.
Ora se tali permutazioni riguardano solo le parti della stessa struttura, cioè parti di A che sono permutabili con altre parti di A stessa, e parti della struttura B che sono permutabili con altre parti appartenenti sempre a B, si dice semplicemente che le strutture sono **isomorfe**.
Possiamo però proporre la seguente definizione che va oltre il concetto di isomorfismo.

Definizione
Se la distinguibilità permane, per tutti gli effetti osservabili, anche dopo lo scambio tra parti di A e di B corrispondenti, allora possiamo dire che si ha **indistinguibilità fisica per permutazione** (o scambio) **di parti associate**.

Quindi, se prendendo parti che appartengono a A e parti, corrispondenti nell'isomorfismo, che appartengono a B, e se accade che scambiandole tra di loro (quindi portando le parti di A in B e viceversa), si ottengono altre due oggetti A' e B' che non sono distinguibili per "tutti gli effetti fisici esterni" da A e da B, allora si può parlare di "indistinguibilità per scambio di parti associate".

### 3.5 Strutture derivate

Per comprendere il senso di quanto propongo nelle pagine che seguono, credo sia importante tenere presente che, per capire la cognizione, serve una teoria per le rappresentazioni strutturali che permetta, tra l'altro, di fare due cose: rispettare in modo preciso il naturale ordine gerarchico che esiste tra gli oggetti che costituiscono il mondo, e passare da rappresentazioni di oggetti fisici concreti ad altri più astratti.
Per quanto riguarda l'ordine gerarchico tra gli oggetti, è evidente che molte entità che possono essere oggetto di cognizione sono costituite da una pluralità di parti che sono a loro volta delle "cose a sé stanti". In maniera analoga, ma in senso inverso, è anche vero che oggetti complessi sono spesso le "parti componenti" di oggetti ancora più complessi. Ne consegue che una metodologia per rappresentare le strutture del mondo reale deve essere concepita in modo tale da rendere quest'aspetto facilmente trattabile. Per questo motivo, quando opportuno, una struttura deve poter essere considerata come una **parte componente** di un'altra struttura di **ordine gerarchico superiore** (per "parte componente" intendo lo stesso concetto di parte utilizzato nei paragrafi precedenti).
Per quanto riguarda invece la necessità di generalizzare, vale a dire di passare da rappresentazioni specifiche ad altre "più generiche", possiamo notare che deve essere in qualche modo possibile agire sia a livello delle parti componenti, sia a livello delle relazioni tra le parti. In ambedue i casi è importante che sia possibile passare, in modo semplice e naturale, da cose "specifiche" a cose meno "specifiche" e più generali. A questo punto si tratta di capire come impostare una teoria per le strutture che renda semplice e naturale questo tipo di passaggi. In questo e nei paragrafi che seguono espongo alcune idee su queste cose.

Nella costruzione che propongo un concetto fondamentale è quello di **struttura derivata**. Ho già accennato al fatto che sono possibili strutture che possono essere gerarchicamente derivate da altre.

Vedremo che, in generale, da una certa struttura A si possono spesso "derivare" una serie di altre strutture e di "proprietà strutturali". Queste strutture saranno sempre in qualche specifica e ben definibile relazione con quella da cui derivano. Vedremo che le strutture derivate corrispondono cognitivamente a operazioni di "**astrazione**" e di "**generalizzazione**". Vedremo inoltre che molte operazioni di derivazione consentono di "**rendere esplicite proprietà e relazioni strutturali notevoli**".

Accade molto spesso che da due strutture A e B che non sono isomorfe, sia possibile derivarne altre, A' e B', che invece lo sono. Questo fatto è molto importante poiché contribuisce a fornire gli elementi che permettono, come avremo modo di vedere, di proporre una definizione per il concetto di regola.

Uno dei punti di forza dell'impostazione qui proposta consiste nel fatto che le operazioni di derivazione strutturale appaiono naturali e non impongono forzature concettuali. Vedremo inoltre che esse contribuiscono a determinare il naturale ordinamento dell'informazione entro un sistema cognitivo.

In generale credo sia corretta la seguente affermazione: sono operazioni di analisi di struttura tutte quelle che esplicitano informazioni strutturali senza aggiungere del contenuto informativo.

## 3.6 Le principali operazioni di derivazione strutturale

### 3.6.1 Le porzioni di struttura

Il concetto di **porzione** di struttura è semplice: data una struttura di partenza, se ne considera un'altra composta da una "porzione della prima".

Sia data una struttura A. Abbiamo visto che essa è individuata dalla collezione delle sue parti, dal sistema di distinguibilità interna di queste, e dal complesso delle relazioni esterne.

Una porzione di struttura è quella struttura che risulta considerando non tutte le parti che costituiscono la struttura originaria A, ma solo alcune.

Se consideriamo l'insieme che ha per elementi le parti di A, allora evidentemente una **porzione di A** corrisponde a un suo sottoinsieme.

Notiamo però che differentemente da quanto avviene nella teoria degli insiemi, non tutte le porzioni sono ugualmente delle strutture derivabili, quindi non a tutti i sottoinsiemi dell'insieme delle parti della struttura A, si può far corrispondere una struttura indipendente. In genere (ma non necessariamente) possiamo dire che, per costituire una porzione valida, le parti di A in questione devono stare in una concatenazione di relazioni esterne adiacenti. In parole più semplici tali parti devono essere tra loro "vicine" senza che sia lecito saltarne nessuna che si "trovi in mezzo".

In un'operazione di "porzione semplice" sono conservati sia il complesso delle relazioni esterne e sia il sistema di distinguibilità interna che le parti hanno nella struttura madre.

E' facile vedere che esistono più modi di scomporre una data struttura in sue porzioni, ma vedremo più avanti che, dal punto di vista cognitivo, alcuni di questi sono significativi, mentre altri non lo sono.

E' anche facile vedere che una struttura è data dalla composizione di un sistema completo (anche ridondante) di sue porzioni.

Per indicare una porzione di una struttura A potremmo usare una scrittura come: Por(i)(A)

Per quanto detto deve quindi, ovviamente essere $Idp(Por(i)(A)) < Idp(A)$.

La (i) sta ad indicare che le porzioni possibili sono più di una.

Esempio. Consideriamo il caso di una figura a forma di poligono disegnato su di un piano e intesa come un certo insieme di punti. In tal caso porzioni dell'intera figura sono strutture come: l'insieme dei punti che costituiscono un lato del poligono, l'insieme dei punti che costituiscono una coppia di segmenti adiacenti, ecc…

Data una struttura di partenza A, essa può essere scomposta in un insieme di porzioni tali che la loro unione dia A stessa. Questo insieme costituisce una **partizione** di A qualora i rispettivi insiemi delle parti delle singole porzioni siano tra loro disgiunti.

### 3.6.2   Le strutture quozienti

Una struttura quoziente si ottiene considerando la struttura che ha per parti un sistema di **porzioni** della struttura madre.

Sia data una struttura A e sia dato un sistema completo K, quindi una partizione, di sue porzioni.

Consideriamo una nuova struttura B le cui parti siano in corrispondenza biunivoca con l'insieme delle porzioni di A nel sistema completo K.

Ora date due porzioni di A queste possono essere tra loro in relazione di isomorfismo e di "indistinguibilità per permutazione". Se lo sono, diremo che le parti corrispondenti della struttura B sono tra loro non distinguibili internamente. Contrariamente saranno internamente distinguibili.

Le porzioni di A avranno in genere una certa estensione e le relazioni esterne tra le parti di B saranno quelle che sussistono tra questi oggetti estesi.

Detto questo, si ha che:
- **B è la struttura quoziente di A rispetto il sistema di porzioni K**.
- **La struttura B sarà di un livello gerarchico superiore ad A**.

Come si vedrà in seguito non tutte le strutture quozienti sono significative dal punto di vista cognitivo.
Come già accennato, le strutture quozienti ereditano le proprietà interne e le relazioni esterne tra le parti componenti, dalla struttura dalle quale sono state derivate. Queste proprietà e queste relazioni esterne possono essere sensibilmente diverse da quelle della struttura di partenza.

### 3.6.3 Operazioni di morfismo

In generale un'operazione di derivazione strutturale di tipo **morfismo** consiste nell'inibizione parziale di ciò che rende distinguibili le parti che costituiscono una certa struttura.
Si possono eseguire operazioni di **morfismo interno**, **morfismo esterno** e misto interno-esterno.
Per eseguire un'operazione di morfismo interno basta inibire alcune (o tutte) le distinguibilità interne tra le parti. Ad esempio, se abbiamo delle parti "colorate" in maniera diversa le une dalle altre, possiamo inibire questa forma di distinguibilità e considerarle come tutte dello stesso colore.
In talune situazioni, interessanti per la pratica cognitiva, è possibile eseguire anche operazioni di morfismo esterno, che consiste nell'inibire parte di ciò che rende distinguibili le parti in base al complesso delle relazioni esterne.
Queste operazioni sono possibili in vari casi. In particolare, quando si ha a che fare con strutture che non sono di tipo base, è spesso possibile "descrivere le relazioni esterne" in modo tale che si possa distinguere in qualche modo i vari elementi di distinguibilità. Vedremo che questo si può fare quando si passa da una rappresentazione base ad una che ne esplicita le "proprietà strutturali salienti". La logica generale di queste operazioni, e i motivi per cui vanno fatte, saranno chiari più avanti, nei capitoli 4 e 5 dove sono illustrate, tra le altre, alcune idee sul fenomeno dell'emergenza.

Potrebbero essere utili anche le seguenti definizioni.

Definizione.
Se data una struttura A, si considera un'altra struttura B che ha lo stesso numero di parti di A, ma che presenta un sistema di distinguibilità interna più semplice tra le sue parti (ma compatibile con l'originale), allora la struttura B si dirà essere un **morfismo interno** di A.

Come detto per alcune strutture non primitive non solo le parti possono avere struttura interna, ma anche le stesse relazioni esterne tra le parti possono avere a loro volta strutture complesse.

Definizione.
Se data una struttura A di tale tipo, si considera un'altra struttura B che ha lo stesso numero di parti di A, ma prendendo per relazioni esterne delle sottostrutture delle relazioni esterne che ci sono tra le parti di A, allora la struttura B si dirà essere un **morfismo esterno** di A.

Come detto si possono anche derivare strutture C che sono sia morfismi interni che esterni di A. In genere, come vedremo, i morfismi corrispondono ad operazioni di generalizzazione.

3.6.3.1   Ancora sul problema del confronto

Le strutture derivate sono particolarmente importanti per rendere possibile il confronto tra strutture.
Per com'è stata definita una struttura, utilizzando operazioni di morfismo e/o di quoziente, è spesso possibile passare da strutture che non sono isomorfe, ad altre che invece lo sono.
Non è raro avere a che fare con due o più strutture (poniamo X,Y,Z) che non sono in relazione di isomorfismo, ma che **intuiamo essere simili**.
In questi casi è spesso possibile ottenere da queste, altre strutture derivate, cognitivamente significative, (X,' Y,' Z') che coincidono.

Esempio: Consideriamo due triangoli di tipo diverso nel piano. Le due strutture, considerate come il luogo geometrico dei punti che costituiscono i due poligoni, non sono sovrapponibili e non sono isomorfe. Le cose cambiano se si passa a considerare le strutture quozienti e i morfismi di queste. Se ci "disinteressiamo" del fatto che i segmenti che compongono le due figure hanno lunghezze diverse, allora le due strutture quozienti possono risultare isomorfe.
Il senso di queste operazioni potrà essere chiaro più avanti quando avrò modo nel capitolo 4 di illustrare alcuni importanti concetti sulle strutture emergenti.
Un altro punto importante è il seguente: in generale compiendo delle operazioni di morfismo si possono ottenere sia strutture di prima specie, che come detto sono memorizzabili in modo autonomo, sia "proprietà strutturali" che invece non lo sono.
Avremo modo più avanti di tornare su questi punti muniti di strumenti che permetteranno di comprendere le cose con più chiarezza.

### 3.7 Proprietà e relazioni strutturali non autonome

Sono importanti le **proprietà strutturali non autonome**. Si tratta di entità che sono definibili specificando informazioni che provengono da almeno uno dei tre punti che definiscono una struttura: insieme delle parti, sistema di distinguibilità interna, complesso delle relazioni esterne (o loro combinazioni); ma che da sole non sono in grado di generare una struttura, di base o derivata, autonoma.

Vedremo che le proprietà strutturali sono, in un certo qual modo, delle "quasi strutture", nel senso che esse sono definite da un insieme d'informazioni di tipo strutturale che non sono però da sole sufficienti a definire una struttura autonoma. Ne consegue che queste "proprietà strutturali" non possono esistere autonomamente ma devono sempre essere presenti entro qualche struttura (di prima o seconda specie) che le "contiene". Tuttavia, come avremo modo di vedere più avanti, esse possono essere riconosciute entro le strutture che le contengono e la loro presenza può essere opportunamente segnalata attraverso delle operazioni che sono in grado di produrre un'informazione univoca associata alla loro presenza. Vedremo anche che la loro identificazione è utile e necessaria poiché ci possono essere dei fenomeni fisici che dipendono dalla loro presenza. Vedremo che in questo senso le proprietà strutturali sono oggetti emergenti, anche se non possono essere memorizzate autonomamente.

Il senso di questi concetti sarà più chiaro nel capitolo 4, dove vedremo che le proprietà e le relazioni strutturali non autonome sono identificabili solo attraverso delle "operazioni di analisi strutturale" e usando dei dispositivi che ne "esplicitano la presenza".

Spesso una proprietà strutturale, una volta definita, non individua una singola struttura ma un'intera classe di queste: tutte quelle che la possiedono.

Preferisco però in questa trattazione evitare di utilizzare il concetto di classe di strutture per definire le proprietà strutturali.

Fino ad ora abbiamo visto quelle che indico come "strutture di prima specie" e alcune delle principali operazioni di derivazione strutturale possibili su di esse. Credo che questa tipologia di strutture sia molto importante perché possono essere agevolmente memorizzate. L'impostazione proposta, che pone l'accento sulle distinguibilità interne ed esterne, è pensata proprio per permettere le operazioni di derivazione, in particolari quelle di quoziente e di morfismo. Considero queste operazioni molto importanti perché appaiono ben applicabili agli oggetti della nostra quotidianità per compiere le prime operazioni di astrazione. L'essere in grado di astrarre, passando da rappresentazioni dettagliate e univoche (nel senso che possono essere applicate solo a particolari oggetti) ad altre più generalizzanti che possono indicare intere classi di oggetti, è fondamentale per l'attività cognitiva. Con le operazioni di derivazione descritte è possibile eseguire un tipo di astrazione che possiamo indicare come strutturale. Vedremo che sono possibili altre tipologie di astrazione, ma questa costituisce la base per le altre.

**Seconda parte.**

Le operazioni di computo basilari sono rappresentabili con strutture di prima specie?

### 3.8  Due congetture interessanti

Le strutture illustrate fino a questo punto sono sostanzialmente "oggetti statici". Con esse non abbiamo ancora completato il quadro delle possibilità esistenti di rappresentare tutti quei "rapporti tra le cose", ai quali, oso pensare si riferisse Poincarè quando parlava di ciò che è conoscibile del mondo che ci circonda.
Agli elementi che sono stati presentati fino ad ora vanno aggiunti altri ingredienti fondamentali che permettono di "rappresentare e trattare l'aspetto dinamico", "i mutamenti" delle cose. Penso che questi ingredienti consistano nelle **operazioni di computo basilari**.
Nelle pagine che seguono assumo valida la congettura, peraltro largamente accettata, che tutte le operazioni computazionali possibili possono essere ricondotte a un numero veramente molto ristretto di operazioni elementari. L'idea centrale è che con opportune combinazioni di queste operazioni di base è possibile eseguire ogni computazione complessa. Questa idea è uno degli assunti che stanno alla base della tesi di Church-Turing sulla capacità di una macchina di Turing universale, e di tutte le macchine computazionali equivalenti, di poter potenzialmente implementare ogni funzione calcolabile.
Prima però di proseguire su questa strada propongo due congetture che considero molto interessanti. In realtà non sono certo della loro validità, anche se le ritengo molto probabili e, se corrette, hanno una serie di importanti implicazioni.

Congettura.
**Solo le strutture di prima specie sono memorizzabili senza ambiguità**.

Congettura.
**Non è possibile rappresentare senza ambiguità tutte le operazioni di computo di base tramite solo strutture di prima specie.**

Potrebbe anche essere vero che **nessuna operazione di computo può essere rappresentata senza ambiguità tramite strutture di prima specie**.

Non è semplice verificare o confutare la validità di queste due congetture. Anche la semplice operazione di memorizzazione implica in realtà una serie di operazioni fisiche sugli oggetti che "supportano la memoria"; alcune di queste, come vedremo, rientrano in quelle che possono essere considerate le "operazioni di base" necessarie per il computo. Ciò nonostante è probabile che vi siano importanti differenze nelle operazioni necessarie per memorizzazione delle semplici strutture di prima specie, e quanto serve invece per "memorizzare" delle operazioni di computo.

Le strutture di prima specie sono caratterizzate dalla "staticità", mentre le operazioni di computo implicano il mutamento; potrebbe essere, ma anche questa è solo una congettura, che "il mutamento" implichi una dinamicità intrinseca che non può essere memorizzata.

Se queste congetture si dimostrassero corrette, implicherebbero un problema: risulterebbe, infatti, che non esiste un modo per memorizzare, in "modo diretto" e "senza ambiguità", le operazioni di computo senza usare "l'artificio" dell'associazione simbolica.

Ad ogni modo vedremo che questo problema può essere facilmente superato utilizzando il concetto di schema, che introdurrò fra poco, e sfruttando il fatto che le operazioni di computo fondamentali sono poche e molto semplici.

Invito il lettore a riflettere sul fatto che la funzione di memoria è indispensabile per la cognizione e che senza di essa non potrebbe esistere alcuna attività cognitiva.

Come detto le strutture di prima specie descrivono entità statiche, mentre le operazioni implicano il mutamento. La funzione di memoria ha senso se riesce a conservare le informazioni, quindi può essere applicata a oggetti che rimangano uguali a se stessi nel tempo. Banalmente, ad esempio, non avrebbe senso implementare come memoria un "contenitore" dove è inserito un oggetto che muta nel tempo: tra una lettura e quella successiva di una "memoria" di questo tipo, potremmo avere risultati differenti (e non prevedibili).

Si noti che, quando memorizziamo l'evoluzione di un fenomeno dinamico, lo trasformiamo in un oggetto statico; nei fatti trasformiamo la dimensione temporale in una spaziale. Per rigenerare la dimensione temporale usiamo l'accortezza di rileggere la memoria rispettando una sequenza che ripristini il corretto scorrere del tempo. Non è detto che questo espediente sia sufficiente per implementare la memorizzazione di tutte le operazioni di computo.

Forse le cose più vicine a delle "rappresentazioni statiche" dello "svolgersi dinamico" di una data operazione di base, sono quelle rappresentazioni mutuamente associate, che intendono rappresentare "la situazione" prima e dopo che una certa operazione è stata eseguita, vale a dire che ritraggono la stessa struttura prima e dopo l'azione dell'operazione. Un esempio sono le tavole di verità che rappresentano gli input e gli output di un'operazione logica elementare per tutti i casi possibili che si possono presentare. Il punto è che anche in questo caso non sembra esistere la maniera per distinguere a posteriori, in modo non ambiguo, quando la rappresentazione in oggetto tratta di due o più strutture mutuamente associate che si riferiscono a due "tempi diversi", oppure quando invece rappresenta una singola struttura statica costituita dalla composizione delle due.

### 3.9  Operazioni fondamentali e strutture di seconda specie: gli schemi

Anche ammettendo l'ipotesi che non sia possibile memorizzare (senza ambiguità) le operazioni computazionali di base, è palese che siamo in grado di costruire e memorizzare algoritmi, e costruire descrizioni "statiche" di processi dinamici. È quindi palese che in qualche maniera siamo comunque in grado di costruire delle memorizzazioni che sappiamo associare, in maniera rigorosa, alle operazioni di computo. Come si spiega questo fatto?

Una possibile risposta è che usiamo lo "stratagemma" di ricorrere all'uso di simboli associati a particolari "congegni dinamici" in grado di eseguire le operazioni fondamentali.

Ricordo che un simbolo esplica la sua funzione se è in qualche maniera funzionalmente associato a ciò che rappresenta. Si possono quindi usare dei simboli per rappresentare le operazioni minimali se questi sono associati (direttamente o in un secondo tempo) in modo univoco a dei congegni che eseguono le operazioni in oggetto.

Come detto, assumo valida la congettura che tutte le operazioni, per quanto complesse, siano sempre realizzabili componendo opportunamente una serie di operazioni di base elementari.

Quali sono le operazioni elementari indispensabili?

Una risposta potrebbe essere che sono quelle svolte dalla macchina di Turing o da macchine computazionali di capacità equivalente. Notiamo che una macchina di Turing deve essere in grado di:

- muoversi lungo il nastro in modo predeterminabile (per direzione e numero di spostamenti);
- deve essere in grado di leggere e scrivere sul nastro;
- deve essere in grado di confrontare la lettura con la propria memoria e di agire secondo le regole codificate: l'azione consiste in uno spostamento e in una scrittura.

Di seguito propongo una possibile rielaborazione delle operazioni di computo basilari.

#### 3.9.1  Funzione di memoria

Nella macchina di Turing il nastro implementa **la funzione di memoria**. Si noti che questa funzione è necessaria per registrare "le istruzioni" che dicono alla macchina "quali azioni compiere", tra quelle (poche) che è in grado di svolgere. La funzione di memoria è fondamentale, non possono esistere sistemi cognitivi che ne sono privi. Secondo la teoria degli automi, un sistema privo di memoria presenta dei limiti nelle funzioni che può calcolare.

### 3.9.2 Operazione di confronto

Una macchina computazionale deve essere in grado di eseguire delle operazioni di confronto. La macchina di Turing, dopo che si è spostata lungo il nastro, e dopo aver "letto" il valore che punta in un certo istante, deve **confrontare** questo con le "istruzioni" che determineranno il suo comportamento per il passo successivo.
Quest'operazione di confronto è indispensabile per compiere operazioni complesse. Nei circuiti digitali le operazioni di confronto, tra stringhe di bit, sono solitamente effettuate con dispositivi di tipo EXNOR, le cui risposte convergono in un singolo dispositivo di tipo AND.

### 3.9.3 Operazione di movimento lungo la struttura (di trasporto di informazione)

La macchina di Turing deve essere in grado di muoversi lungo il nastro. Un macchina computazionale deve essere sempre in grado in qualche modo di compiere l'equivalente del "muoversi lungo una struttura". Questo ad esempio per essere in grado di confrontare i vari elementi che la compongono ed eventualmente per poterli modificare. In un calcolatore l'unità di elaborazione deve essere in grado "di spostarsi" entro lo spazio di memoria.

### 3.9.4 Operazione di copia di elementi strutturali

L'operazione di "copia" di elementi strutturali è fondamentale per modificare o aggiungere "nuovi elementi strutturali".
Solitamente quando si presenta la macchina di Turing, si afferma che essa deve essere in grado di "scrivere un valore nel nastro". Ma se riflettiamo su quest'operazione di scrittura, è facile rendersi conto che essa non può essere arbitraria poiché altrimenti la macchina sarebbe non deterministica. La scrittura quindi deve obbedire a delle regole, e da qualche parte deve essere già presente il simbolo che si va a scrivere. Il simbolo in oggetto è quindi, alla fin fine, copiato.
Per il computo strutturale, nel caso più generale, non è sufficiente essere in grado di agire solo "sullo stato" delle "distintinguibilità interne" delle parti di struttura (che è quanto avviene nella macchina di Turing), ma può essere anche necessario essere in grado di modificare la struttura aggiungendovi delle parti. Affinché queste appartengano alla struttura in oggetto, è necessario che queste siano in qualche modo collegate alle altre. È necessario che siano quindi definite anche le relazioni esterne; questo può, ad esempio, essere fatto aggiungendo un "nuovo ramo di grafo".

### 3.9.5 Operazione di associazione simbolica funzionale

Come visto, potrebbe essere vero che le operazioni di computo minimali non si possano rappresentare, in maniera non ambigua, usando solo i metodi illustrati per definire le strutture di prima specie. Ho proposto anche la congettura che solo queste ultime potrebbero essere memorizzabili. Siamo però in grado di memorizzare in qualche modo anche le operazioni dinamiche. Un algoritmo scritto sulla carta utilizzando un qualunque linguaggio di programmazione è, di fatto, una rappresentazione memorizzata di operazioni computazionali complesse. Come si riesce a fare ciò?

Come detto, propongo che si debba ricorre all'associazione simbolica. L'idea è che nei linguaggi di programmazione si utilizzano dei simboli che sono associati all'azione di particolari congegni in grado di eseguire fisicamente le operazioni descritte.

Questo meccanismo di associazione funzionale tra simbolo e congegno fisico reale è probabilmente fondamentale per tutti i processi di computo e per la cognizione. Può essere utile generalizzare questo concetto e indicarlo come "**operazione di associazione simbolica funzionale**".

Si possono associare funzionalmente elementi di struttura, simboli, operatori e, come vedremo tra non molto, interi **schemi procedurali**.

Se si riflette sulla questione, non credo sia difficile convenire che l'utilizzo dei simboli presuppone sempre che siano eseguite delle operazioni. Queste operazioni possono andare dalla semplice associazione tra un simbolo e una rappresentazione strutturale, fino all'associazione con congegni in grado di eseguire operazioni anche molto complesse.

Ad esempio in un programma per calcolatore, l'associazione tra i simboli che in esso compaiono e le operazioni relative, avviene in fase di scrittura nella mente di chi lo scrive, e in fase di esecuzione attraverso i dispositivi fisici (in genere presenti nella CPU) che eseguono fisicamente le istruzioni. Quando è solo sulla carta, o fermo nella memoria di un computer, un programma di calcolatore è, di per se, una "struttura statica di prima specie" come quelle che ho illustrato nelle pagine precedenti.

### 3.10 Coincidenza tra operazioni

Penso si possa proporre la seguente definizione:
**Due operazioni coincidono se, agendo su strutture isomorfe, generano sempre strutture che sono tra di loro ancora isomorfe**.

### 3.11 Gli elementi base del computo strutturale

Le **strutture di prima specie**, e le **operazioni di computo basilari**, come quelle descritte sopra, costituiscono gli elementi base di ciò che penso si possa chiamare "**computo strutturale**".

L'idea è che queste due tipologie di oggetti siano essenziali per ogni attività computazionale e per ogni attività di rappresentazione di conoscenza. Vedremo fra poco come, mettendo assieme i concetti di struttura di prima specie e le operazioni di illustrate, sia possibile "estendere" il concetto di struttura.

### 3.12  Concetto di "congegno operatore"

Credo possa essere utile il concetto di "**congegno operatore**" (o **dispositivo operatore**).
Un dispositivo operatore è un congegno fisico in grado di eseguire almeno una delle operazioni minimali descritte. Un certo dispositivo operatore può ad esempio eseguire l'operazione di copia di un elemento di struttura. Oppure può eseguire un'operazione di trasporto, che può essere spezzata in due operazioni di copia (o copia e cancellazione). Un altro tipo di dispositivo può eseguire un'operazione di confronto proponendo in output un bit di informazione che dice se le entità confrontate coincidono o meno.
Si possono definire anche operatori molto più complessi di quelli di base, ma le cui azioni possono tutte essere implementate attraverso una serie di opportune ripetizioni delle operazioni minimali.

### 3.13  Strutture di seconda specie

Veniamo dunque al concetto di algoritmo. Che cosa è un algoritmo secondo le idee e il linguaggio fin qui proposto?
Come detto è possibile implementare associazioni di tipo simbolico tra oggetti che fanno da simbolo e operatori. Ma questi oggetti che svolgono la funzione di simbolo possono nello stesso tempo essere anche degli "elementi" costituendi di una struttura. Ad esempio, la funzione di simbolo può essere associata allo stato di distinguibilità interna ad alcune delle parti componenti di una struttura di prima specie. Con ciò si perviene al concetto di schema, che possiamo anche chiamare: **struttura di seconda specie**.

### 3.14  Concetto di schema ( o schema procedurale)

Definizione.
Uno **schema** è una struttura nella quale alcuni degli elementi strutturali componenti (in genere le parti), svolgono anche la funzione di simbolo associato ad operazioni di computo.

Queste associazioni in genere si possono implementare facendo in modo che il riconoscimento del simbolo causi l'attivazione di un congegno fisico in grado di eseguire le operazioni in oggetto.
Lo schema quindi è un oggetto misto che mette assieme le strutture e i simboli.

I simboli presenti entro uno schema possono essere associati sia a congegni di computo elementari, sia ad altri schemi. Quindi possiamo avere schemi le cui parti componenti funzionano da simboli che rimandano ad altri schemi (come avviene nei linguaggi di programmazione di alto livello).

Se lo schema è **ben definito**, esiste sempre la possibilità di ricostruire da questo uno "**schema di base**" che non rimanda ad altri schemi.

Condizione necessaria perché uno schema procedurale sia "di base" è che le operazioni associate alle sue parti che fanno da simbolo siano operazioni di base.

Si noti che questa definizione del concetto di schema è comunque utile a valida a prescindere dalla validità delle due congetture proposte più indietro, vale a dire a prescindere dalla validità dell'ipotesi che le operazioni di base non siano rappresentabili con le sole strutture di prima specie.

Congettura
Ogni operazione complessa, quindi ogni algoritmo, può sempre essere rappresentata da uno schema.

Si possono rappresentare schemi ciclici e annidati, così come avviene per gli algoritmi nei linguaggi di programmazione. Tutti gli schemi annidati sono ricomponibili in uno schema di base, ovviamente più lungo, costituito da operazioni di computo basilari.

Penso che, dal punto di vista operativo, i concetti di schema e quello di algoritmo sostanzialmente coincidano. In questo lavoro preferisco utilizzare una terminologia appositamente dedicata per non fare confusione e per invitare il lettore a considerare gli aspetti che derivano dall'impostazione proposta per il concetto di struttura. Penso che il concetto di schema che qui propongo sia per certi aspetti più ricco di quello intuitivo di algoritmo, qualcosa di più di una sua mera precisazione.

Agli schemi si possono estendere molte dei concetti visti per le strutture.

Il concetto di **coincidenza strutturale** si può applicare anche a due o più **schemi. Due schemi coincidono se:**
- sono isomorfe le loro strutture,
- gli elementi che assumono la funzione di simbolo sono associati a congegni operatori che eseguono operazioni coincidenti.

Anche a molti schemi si possono applicare operazioni di derivazione strutturale.

### 3.15 Alcune riflessioni sul concetto di struttura di seconda specie

Una struttura di seconda specie è in un certo senso una struttura "più ricca" rispetto a quelle di prima specie e può essere quindi considerata un'estensione di questo concetto.

Pensiamo ad un algoritmo: è un oggetto complesso e in quanto tale possiede sempre una struttura di prima specie. Esso è quindi identificabile come oggetto matematico composto di un insieme di parti, con un certo sistema di distinguibilità interna e un dato sistema di relazioni esterne. In questo modo però non si riesce a esprimere e rappresentare tutta l'informazione contenuta entro l'algoritmo, manca qualcosa.

In certo senso possiamo dire che la "struttura di prima specie" costituisce la parte "memorizzabile di un algoritmo", e essa è in effetti quella che viene memorizzata entro la memoria di un calcolatore! Ma per diventare attivo un algoritmo ha bisogno di girare entro una macchina computazionale, entro una CPU. Deve quindi essere connesso con dispositivi in grado di eseguire concretamente, dinamicamente, come divenire (temporale), la sequenza delle operazioni che sono in esso rappresentate.

Si noti ancora che spesso, ma non necessariamente, uno schema procedurale agisce su una o più strutture modificandole. Quindi spesso uno schema può essere visto come un operatore che agisce su strutture in "input", e produce strutture in "output".

Uno schema può agire sia su strutture di prima che di seconda specie, quindi si possono descrivere algoritmi che modificano algoritmi e anche algoritmi che modificano se stessi. Tuttavia questi ultimi non sono in genere interessanti poiché non se ne conserva la memoria (il che li rende molto difficili da gestire e capire).

Si noti ancora che ogni procedura algoritmica, ogni legge fisica, e, come avremo modo di vedere, una vasta classe di regole, possono essere espressa per mezzo di schemi.

Penso che una delle assunzioni implicite nell'idea che la realtà sia rappresentabile con i metodi della computazione classica è che essa sia, in ogni istante, completamente rappresentabile attraverso una struttura di prima specie e che le sue leggi fondamentali siano rappresentabili tramite strutture di seconda specie.

In fisica si utilizzano strutture continue e operatori differenziali. Esistono anche, come avremo modo di vedere, "realtà emergenti" il cui stato è rappresentabile in maniera completa tramite strutture discrete di prima specie che mutano nel tempo secondo una temporizzazione anche'essa discreta. Si pensi ad esempio allo stato interno delle memorie e dei registri di un calcolatore.

### 3.16 I simboli sono memorizzabili?

Più indietro ho proposto la congettura che siano solo le struttura di prima specie ad essere "facilmente" memorizzabili. Ammettendo la validità di tale ipostesi penso ci possa chiedere cosa succede per i simboli. In generale possiamo subito notare che quanto utilizziamo dei simboli per rappresentare delle cose, all'atto pratico li riconosciamo attraverso la loro struttura. Ad esempio, una lettera dell'alfabeto viene riconosciuta dal nostro sistema visivo grazie alla struttura del carattere tipografico utilizzato. La stessa cosa vale anche per i suoni che costituiscono il parlato, anche se la cosa potrebbe a prima vista apparire un po' più complicata. Quando il nostro orecchio traduce un suono puro in segnali nervosi, esegue un processo di analisi della "struttura temporale" con la quale varia la pressione istantanea dell'aria (passando per una analisi spettrale). Se si riflette sulla questione, si può comprendere per quale motivo non è possibile utilizzare direttamente la funzione simbolica senza passare attraverso la verifica della coincidenza strutturale. La questione cruciale è che l'oggetto che fa da simbolo deve poter essere riconosciuto anche quando si trova su supporti diversi. Il supporto utilizzato dentro la nostra mente, dentro i circuiti di neuroni e connessioni sinaptiche, o qualunque altra cosa ci sia nel nostro cervello, è un supporto diverso da quello in cui lo stesso simbolo si presenta quando è rappresentato su un foglio di carta o quando viaggia nell'aria come sequenza di onde di pressione. I supporti sono diversi, ma ciò che coincide è, a livello di stimolo prossimale, proprio e solo le loro strutture, o meglio quegli insiemi di proprietà strutturali che una vota rese esplicite ne permettono nei vari casi il riconoscimento. Quindi **anche i simboli sono riconosciuti attraverso delle corrispondenze strutturali**. Una certa rappresentazione mnemonica, riconoscibile in base all'analisi della sua struttura, acquisisce il ruolo di simbolo quando essa è associata in maniera funzionale a un'altra "entità" che essa rappresenta. Questa entità, come visto, può essere a sua volta o un'altra rappresentazione strutturale (di prima o seconda specie), o un congegno che esegue delle operazioni di computo, o, come avremo modo di vedere, "delle astrazioni" che costituiscono "la rappresentazione cognitiva" di un certo soggetto specifico.
I simboli, quindi, sono effettivamente memorizzabili (com'è ovvio), ma questo è possibile solo grazie alle loro strutture.
Uno stesso simbolo può anche essere rappresentato con strutture diverse. Ad esempio, banalmente, lo stessa lettera dell'alfabeto può essere rappresentata con caratteri tipografici molto diversi tra loro, in corsivo, in stampatello, maiuscole, minuscole ecc.…Quando leggiamo, riconosciamo ogni una di queste strutture e la associamo allo stesso suono.

**3.17 Concetto di operatore generalizzato.**

Può essere utile definire il concetto di operatore generalizzato. Un operatore (generalizzato) consiste in una serie di operazioni che possono essere applicate su una data struttura d'input e che producono una struttura in output. La serie delle operazioni che vanno applicate è esprimibile in modo algoritmico. Quindi a ogni operatore è associabile uno schema procedurale ben definito.
Se la lettera O indica un operatore, A la struttura di input, B la struttura di output. Possiamo allora usare la notazione B = Ox(A), che è comune in matematica.

Si noti che nella "pratica reale" di computo, sia A che B sono sempre strutture discrete. Anche l'operatore è dato da uno schema che ha struttura discreta, vale a dire che il numero di operazioni effettivamente eseguibili è sempre finito. Tuttavia in matematica si definiscono anche strutture continue. Vedremo che si possono definire anche "operatori ideali" che indicano un'infinità di operazioni. Un esempio è l'operatore di integrazione che è la generalizzazione al continuo di quello di sommatoria (vedremo queste cose tra non molto).

**3.18 Concetto di sistema di computo strutturale.**

Particolarmente importante penso sia il concetto di "**sistema di computo strutturale**", che possiamo utilizzare per indicare qualunque sistema dinamico finito e deterministico. Il concetto può essere esteso a sistemi non finiti.

Possiamo definire in generale come **sistema di computo strutturale** ogni sistema per il quale è vero che:
- Il suo "stato istantaneo" è rappresentato da almeno una struttura di prima specie che lo descrive completamente.
- La sua evoluzione, quindi il passaggio da uno stato al tempo t1, a quello al tempo t2, è computabile tramite opportune operazioni che agiscono sulla sua struttura al tempo t1.
- Queste operazioni sono riducibili ad un insieme finito di operazioni di base, quelle eseguibili da una macchina computazionale universale (e che sono esprimibili tramite uno o più operatori generalizzati).

Quando le procedure da utilizzare, quindi le operazioni da eseguire, sono uniche e completamente prestabilite, allora abbiamo a che fare con un **sistema di computo strutturale deterministico**.

Sistemi di questo genere sono quindi costituiti da: una struttura globale di partenza e da un certo insieme di operazioni che si devono compiere.

Ci possono essere sistemi di computo strutturale che possono funzionare indefinitamente, e sistemi che invece si "**bloccano**" poiché nelle loro evoluzioni sono soggetti a pervenire in situazioni nelle quali agiscono almeno due regole procedurali che sono tra di loro in conflitto.

Si possono definire varie tipologie di sistemi di computo strutturale: deterministici e non deterministici, discreti o continui. Alcuni lasciano la possibilità di "scegliere" quali operazioni eseguire tra un certo insieme di possibilità. Alcuni quindi si evolvono in un solo modo, altri invece possono evolversi in molti modi diversi.

Come ultima nota, invito il lettore a riflettere su cosa è il **calcolo simbolico formale**. Non è forse vero che ogni formalismo si concreta, alla fin fine, nella "manipolazione formale" di stringhe di simboli? In questo lavoro propongo di considerare queste "stringhe come delle strutture". Non è difficile convenire che ogni stringa è, a tutti gli effetti, una struttura: i simboli che la compongono ne sono le parti, e le relazioni esterne sono date dalle adiacenze e dal loro ordinamento. Le operazioni di manipolazione formale cosa altro sono se non operazioni di computo eseguite su queste strutture?

Penso che considerare le cose dal "punto di vista strutturale" arricchisca il modo di concepire queste cose. Penso che la consapevolezza che quando si fa manipolazione formale, in realtà si agisce su delle strutture modificandole, possa condurci ad una concettualizzazione più vicina alla realtà.

In cosa consistono le "manipolazioni formali"? Si tratta di spostare, copiare, eliminare, sostituire simboli e stringhe di simboli. Si parte da certe stringhe di partenza, e si utilizza un insieme di regole di manipolazione per modificarle. Le scritture dei simboli stessi fanno da supporto, quindi da memoria, alle stringhe stesse. Le operazioni sono ben codificate e sono associate ad alcuni simboli particolari che dicono come si deve o si può agire nelle manipolazioni. Le stringhe sono, di per se stesse, delle strutture di prima specie; quando descrivono delle procedure sono degli schemi.

### 3.19 Reti NAND (o NOR).

Può essere interessante come approfondimento quanto segue.

Più sopra ho descritto, o semplicemente menzionato, alcune delle operazioni di computo fondamentali, che ci permettono di esaminare un po' più nel dettaglio le azioni che vengono svolte da una macchina di Turing o da qualunque automa universale.

Può avere senso chiedersi se è possibile procedere ulteriormente nell'approccio riduzionista e ricondurre queste operazioni a qualcosa di ancora più semplice. La teoria delle reti digitali dice che ci sono tre operazioni logiche fondamentali AND, OR, NOT, e quelle composte come le NAND o le NOR . Nello stesso tempo è possibile dimostrare he si può fare tutto utilizzando un solo tipo di dispositivo. È ben noto che ogni funzione logica è implementabile componendo solamente, in modo opportuno, varie porte NAND (o solamente porte di tipo NOR) a due ingressi.

Riflettendo su com'è fatto un calcolatore digitale, si può concludere che ogni processo di elaborazione di informazione è realizzabile utilizzando solo tre elementi base:

- memorie (anche a singolo bit),
- canali di comunicazione (nei circuiti digitali sono i fili che collegano i vari elementi),
- dispositivi NAND.

Le memorie, insieme ai canali di comunicazione, servono ad implementare, nel modo minimamente necessario, i due aspetti di una struttura: la distinguibilità interna (le memorie) e le relazioni esterne (il modo in cui i canali di comunicazione sono interconnessi). Questo per quanto riguarda **l'aspetto statico** delle strutture (quindi di prima specie).

L'aspetto dinamico è implementato con la funzione di trasporto dei canali di comunicazione, dalle operazioni dei dispositivi NAND, e ovviamente dal fatto che le memorie possono cambiare stato. I canali di comunicazione consentono di trasportare le informazioni "sullo stato" degli elementi di struttura, da e verso i dispositivi NAND, dove le informazioni "**interagiscono**"!

Una riflessione sulle caratteristiche di questa tipologia di "reti deterministiche", capaci di implementare potenzialmente ogni funzione calcolabile, mostra delle analogie con i sistemi fisici reali, almeno secondo la concezione classica della fisica.

Nei sistemi fisici ci sono corpi che si muovono e che possono **interagire**. I corpi (le particelle) "mantengono memoria di certe informazioni: le loro proprietà fisiche (massa, impulso, posizione ad un certo istante, carica elettrica… e altre ancora)". Essi si muovono e quindi "**trasportano queste informazioni**" e infine "**interagiscono**". Si noti che l'interazione è di fondamentale importanza, se non ci fosse avremo uno strano sistema dinamico di particelle che si muovono ma non si urtano mai. In un sistema di questo tipo non sarebbe possibile alcuna forma di attività computazionale e anche nessuna forma di attività cognitiva. Altra cosa importante da notare è che, almeno secondo la concezione classica, le interazioni avvengono localmente (in un "punto" dello spazio-tempo). Questo fatto è importante, ed è da tener presente per comprendere le modalità con le quali possono aver luogo i processi di elaborazione fondamentali per la cognizione. Più avanti riprenderò questo concetto illustrando quello che indico come "principio di convergenza delle verifiche" e che dipende dal fatto che le interazioni sono pensate essere sempre locali.

Con delle reti costituite da memorie, canali di comunicazione e dispositivi NAND si può implementare ogni funzione computabile. Risulta quindi possibile realizzare entro queste reti ogni macchina di Turing e ogni automa a stati finiti.
In genere, all'atto pratico, conviene osservare le cose un po' più dall'alto e invece di raffigurarsi reti di porte NAND interconnesse, è utile pensare direttamente ad operazioni leggermente più complesse, come quelle descritte nelle pagine precedenti.

**Alcuni approfondimenti.**

Congruenza della teoria strutturale proposta con gli oggetti matematici standard.

### 3.20 Numeri naturali nella teoria delle strutture finite.

Probabilmente si possono pensare vari modi per descrivere le strutture e le operazioni possibili su di esse. Quello che ho proposto nelle pagine precedenti è solo uno dei possibili e non posso certo escludere che sia migliorabile e sostituito da qualcosa d'altro. A ogni, per questo lavoro, è necessario che esso funzioni bene nel descrivere e trattare le strutture della nostra quotidianità. Avrò modo di approfondire quest'aspetto nei capitoli che seguono. Ma esiste anche un'altra condizione essenziale: l'inquadramento proposto per i concetti di struttura e di schema deve essere applicabile anche per rappresentare gli oggetti matematici standard.

Entro il contesto delle idee fin qui proposte, il concetto di "numero" corrisponde ad una notevole operazione di morfismo. Ho accennato al fatto che un morfismo di una certa struttura corrisponde ad un operazione di generalizzazione. Eseguendo un morfismo spesso si mantiene il numero della parti ma si procede limitando la distinguibilità tra le parti stesse.
Il numero è quel morfismo che tra tutti gli elementi che definiscono una particolare struttura mantiene solo l'insieme delle parti.
Si prende quindi un insieme campione e gli si associa un nome (secondo l'opportuno sistema di nomenclatura). Tale nome costituisce il simbolo associato al morfismo numero.
Presa un'altra struttura, ed eseguite le stesse operazioni di soppressione delle distinguibilità, si potrà o no ottenere un insieme di parti coincidente con quello di partenza. Se ciò avviene, diremo che le parti sono equinumerose. Come ben noto i numeri naturali si basano sulla possibilità di costruire biiezioni tra insiemi.

Questo vale per i numeri naturali. In matematica però si utilizzano anche altri tipi di numeri.

### 3.21 Numeri reali.

Un'altra delle strutture più semplici, ed interessanti, è quella costituita da un certo numero di parti, non distinte internamente, che nel complesso delle relazioni esterne sono disposte secondo una **grafo a catena**. Si tratta di un caso semplicissimo da immaginare. In tale struttura ogni una delle parti componenti, tranne quelle che si trovano agli estremi, "comunica" solo con altre due parti che le sono quindi adiacenti. Se inoltre, date due parti *a* e *b* si distingue la relazione che porta da *a* a *b* da quella che porta da *b* ad *a* si ottiene un orientazione nel grafo a catena.

Consideriamo ora una struttura a catena A di lunghezza infinita. Se ammettiamo per ipotesi che le parti di questa struttura, possiedano struttura interna, e siano a loro volta delle catene, in questo caso finite, tutte di lunghezza uguale, disposte in modo tale che l'inizio dell'una coincida con la fine di quella successiva, allora la struttura $A^{-1}$ sarà anche essa una catena di lunghezza infinita (la struttura A è quoziente rispetto la $a^{-1}$). Se ammettiamo inoltre, sempre per ipotesi, che quest'operazione di derivazione inversa possa continuare all'infinito, si ottiene una famiglia infinita di strutture: { A, $A^{-1}$, $A^{-2}$, $A^{-3}$,…} che è numerabile. Si noti che il numero delle parti di ogni una di queste strutture, può essere messo sempre in biiezione con l'insieme dei naturali **N**. Sappiamo, grazie a Cantor, che il prodotto cartesiano di due insiemi numerabili è anche esso numerabile. Così invece non è per l'insieme delle parti di un insieme infinito numerabile (qui uso il concetto "di insieme delle parti" usato comunemente in matematica, vale a dire, nel caso specifico, all'insieme di tutti i sottoinsiemi di **N**). Immaginiamo ora di costruire una rappresentazione 2grafica virtuale", di queste famiglie di strutture a catena, in questo modo: si rappresenta la prima catena A secondo il modo tipico dei grafi, segnando lungo una retta con dei pallini i punti corrispondenti ai vari nodi, quindi alle parti di A. Si rappresenta la struttura $A^{-1}$ subito sotto la prima e così si procede indefinitamente per $A^{-2}$, $A^{-3}$ ecc… Avremmo quindi tanti grafi lineari, disposti l'uno sotto l'altro, e man mano si scende saranno sempre più densi. Come detto, per ipotesi, l'insieme totale dei pallini, che si estenderanno per un intero semipiano, è numerabile. Immaginiamo a questo punto di tracciare degli altri grafi unendo questa volta i vari pallini non in orizzontale ma bensì procedendo sempre dall'alto verso il basso. Non è difficile vedere che tra tutte le infinite famiglie di grafi in questo modo **potenzialmente** tracciabili, ve ne è una che corrisponde all'insieme **R** dei reali. Insieme che come **P(N)** è di cardinalità maggiore di **N**.

### 3.22 Il passaggio al continuo.

Le strutture che sono state descritte nei paragrafi precedenti sono strutture discrete. Sappiamo che in matematica è in fisica si fa largo uso del concetto di continuità e dell'idealizzazione di entità e funzioni che sono costituite da un insieme infinito di parti infinitesime. Nella pratica non è possibile costruire rappresentazioni esplicite di numeri incommensurabili, o di quantità che non sono finite. La matematica permette, però, di associare dei simboli anche quantità che non sono finite, o esprimibili con numeri razionali, di scrivere equazioni in cui queste compaiono, di utilizzarle per dimostrare teoremi, e per ottenere in molti casi, attraverso di esse, risultati finali che sono di nuovo finite e razionali. L'utilizzo di quantità non finite e non razionali è una necessità per la descrizione degli enti geometrici, che sono comunque, alla fin fine degli oggetti astratti ideali.

In questa trattazione, non ho difficoltà ad ammettere la possibilità di costruire simboli associabili a **infiniti potenziali**. Questo può essere fatto definendo un algoritmo che, se non terminasse, genererebbe, potenzialmente, una struttura, e quindi il suo morfismo numero, composta da un numero infinito di parti.

Le cose sono però più delicate per quanto riguarda il concetto di **infinito attuale**. Gli strumenti che ho introdotto in questo lavoro non permettono di costruire delle strutture che sono realmente costituite da un infinità di elementi. Non so se esistono infiniti attuali.

### 3.23 Le principali operazioni aritmetiche come operazioni su strutture.

Su strutture di prima specie possiamo definire delle operazioni che possono essere considerate come la generalizzazione di quelle aritmetiche fondamentali.
Le prime operazioni facilmente definibili sono le generalizzazioni di quella di somma che possiamo chiamare operazioni di composizione di strutture.
Un'operazione di composizione è quella che date due strutture A e B, ne ottiene un altra C, in cui A e B sono due sue porzioni. A e B assieme costituiscono un ricoprimento di C. Si noti però che ci possono però essere molti modi per eseguire la composizione di A con B. Vale a dire che in molti casi si possono comporre varie strutture C1, C2, C3.... ecc, diverse le une dalle altre, attaccando A e B in modi diversi. Si noti comunque che tutte le strutture Cx, avranno lo stesso numero di parti e la stesso sistema di distinguibilità interna tra le parti.
Ne risulta che tutti i "morfismi numero" delle varie Cx coincidono. Quindi considerando i "morfismi numero" l'operazione di composizione di strutture coincide con l'operazione di somma tra interi.
Possiamo definire anche operazioni di differenza tra due strutture: A-B. Essa è pero possibile se B risulta essere isomorfa ad almeno una porzione di A. Se inoltre accade che le porzioni di A che sono isomorfe a B sono più di una , allora sono possibili diverse operazioni di differenza.

Anche in questo caso è facile vedere, che considerando i morfismi numero, si ottiene l'ordinaria operazione aritmetica di sottrazione tra interi. Sappiamo che tra interi deve essere B<=A, il che significa appunto chiedere che B sia isomorfa ad una porzione di A..

Un'altra operazione che è interessante introdurre è quella di "convoluzione". Siano date due strutture A e B e sia C la convoluzione di A con B. Non sempre tale operazione è possibile. Essa si ottiene sostituendo le parti di B con delle strutture interamente isomorfe ad A. Si ottiene una struttura che ha un numero di parti pari a N(A) x N(B).

Solo in casi speciali tale operazione è, in effetti, fattibile. Deve innanzitutto essere che i complessi delle relazioni esterne delle due strutture siano tra loro compatibili. Ad esempio, in taluni casi dovrà essere vero che la distanza minima tra le parti di B sia comunque maggiore del raggio massimo di A.. I concetti di distanza minima e di raggio massimo corrispondono a quelli usati in geometria. Se ne possono definire di equivalenti anche quando il complesso delle relazione tra le parti delle due strutture in questione è descritto da dei grafi.

Esiste una variante a tale operazione che è molto usata in matematica. Essa si usa in genere su strutture costituite da funzioni continue . In questa variante non è necessario che la distanza tra le parti di B sia maggiore del raggio massimo di A. Infatti, quando accade che due parti di A vengono a sovrapporsi si prende come parte della struttura risultante la somma delle due strutture interne delle parti di A in questione.

Si noti però che il morfismo numero di tale variante non coincide con l'operazione di moltiplicazione di numeri naturali cioè non si ha che N(C)=N(A) x N(B).

In modo analogo possiamo anche definire la generalizzazione dell'operazione di quoziente.

### 3.24 Concetto di struttura continua.

Il concetto di struttura continua può essere considerato un'estensione di quello di struttura discreta, qualora si accetti l'ipotesi della continuità. Quest'ultima consiste nell'ammettere l'esistenza di enti infinitamente divisibili. Quando concepiamo l'idea di segmento, diciamo che è un oggetto che ha lunghezza, ma non ha spessore (è interessante notare come le idee intuitive di Euclide mostrino tuttora una certa utilità e validità..). Questo concetto credo sia strettamente legato proprio all'idea che un segmento sia infinitamente divisibile, in altre parole che si può continuare a spezzare ottenendo parti che sono sempre più piccole. Intuitivamente affinché le cose non generino contraddizioni, sembra necessario anche ammettere che il segmento sia un oggetto composto da parti prive di lunghezza: i punti. Il punto sarebbe, secondo questo modo intuitivo di pensare, una parte di struttura priva di dimensioni. Sospetto che in realtà, a ben guardare, anche quest'approccio conduca a contraddizioni. Le "varie stranezze" che queste idee comportano, sono note fin dall'antichità. Si pensi, ad esempio, ai paradossi di Zenone. In sostanza è difficile capire come si possa ottenere un oggetto di lunghezza finita mettendo assieme tanti elementi che ne sono privi. Francamente sospetto che a tuttora non sia stata trovata una soluzione chiara a questo tipo di problemi, a questi strani "pasticci d'idee".

In matematica per affrontare il problema della continuità si utilizzano sostanzialmente degli stratagemmi come quello di "passaggio al limite". Quest'ultimo si poggia a sua volta sui concetti di successione e di serie infinita, che appaiono intuitivamente meno problematici di quello di infinitesimo.

Se una delle strutture finite e discrete più semplici da concepire è il grafo a catena, un segmento può essere pensato come un oggetto simile qualora si accettino alcune ipotesi. Nel grafo a catena le parti componenti sono i singoli vertici. In un segmento le parti componenti sono invece gli altri segmenti in cui può esser suddiviso. Con il passaggio al contino si pensa che questi segmenti possano essere suddivisi in elementi "infinitamente cortissimi", così corti che non si riuscirà mai a trovare dei segmentini, che per quanto piccoli, siano ancora più corti. Ma per costruire un oggetto di lunghezza finita devono allora diventare moltissimi, tanto numerosi quanto sono corti, quindi: infiniti. Francamente mi sembra che quest'approccio sia comunque soggetto a una serie di contraddizioni. Comunque sia, la matematica mette a disposizione una serie di strumenti per trattare le strutture continue. Questi strumenti sono i numeri reali, il concetto di infinito potenziale, la teoria delle serie e quella dei limiti. Con questi strumenti è possibile definire le funzioni continue.

### 3.25 Operatori differenziali ottenuti tramite il "passaggio al limite" di operatori discreti.

Non è difficile inquadrare il concetto di operatore discreto. Di esso ho già parlato, quando ho illustrato il concetto di operazione computazionale elementare, quello di schema procedurale, e quello di operatore generalizzato. Con degli schemi opportuni si possono rappresentare le operazioni aritmetiche fondamentali e le loro generalizzazioni strutturali. Utilizzando il concetto di limite è i concetti di ciclo infinto, e di infinito annidamento, è possibile definire degli operatori continui. Ad esempio la derivata è un operatore continuo: per definirla è necessario applicare l'operazione di passaggio al limite sul rapporto incrementale. A sua volta, il passaggio al limite richiede di associare uno schema a una successione infinita, o meglio di associarlo ad un altro schema che è in grado di generare, potenzialmente, la detta successione, continuando a "reiterare" indefinitamente.

# 4 Seconda congettura di riferimento. Definizione di regola

## 4.1 Introduzione

Torniamo a considerare l'idea di Poincarè:
*La scienza può solo farci conoscere i rapporti tra le cose e non le cose in quanto tali: "al di là di questi rapporti non c'è alcuna realtà conoscibile".*
Con quanto proposto fino ad ora ho cercato di tradurre questa idea di "rapporti tra le cose" in qualcosa di preciso e rigoroso. Per far questo ho cercato di precisare i concetti di struttura e di schema.

Abbiamo visto che le strutture di prima specie sono entità che possono essere oggettivate con linguaggio matematico e che possiedono la caratteristica di essere effettivamente memorizzabili. Questa loro caratteristica è ovviamente irrinunciabile per ogni attività cognitiva. Abbiamo però anche visto che, seppur fondamentali, le strutture di prima specie, da sole, non esauriscono tutti quei "rapporti tra le cose" che costituiscono la materia del conoscibile. All'appello mancano alcune possibilità, tra le quali le operazioni di computo. È possibile (e probabile) che queste ultime non siano rappresentabili senza ambiguità utilizzando solamente strutture appartenenti al primo gruppo. Ho illustrato l'idea che per rappresentare la struttura "di operazioni complesse", sia necessario utilizzare la funzione di associazione simbolica e affidarsi, alla fin fine, a congegni, quindi a meccanismi fisici, che siano in grado di eseguire realmente le operazioni fondamentali. Per fortuna le operazioni di base sono poche e ben codificabili. Quando si compiono operazioni complesse si opera ripetendo sequenze opportune di queste operazioni elementari. Tali sequenze, essendo oggetti composti, possiedono una struttura. Si perviene quindi al concetto di schema che mette assieme la funzione di associazione simbolica con le operazioni elementari e le strutture di prima specie. Gli schemi possiedono sempre una struttura. Spesso le parti dello schema fungono da simboli che sono riconoscibili in base alla loro distinguibilità interna, e quindi in base alle loro "strutture interne". Queste ultime sono però strutture di livello gerarchico inferiore rispetto alla struttura dello schema. La struttura dello schema è in genere una struttura quoziente rispetto al livello delle strutture interne dei suoi simboli.

Muniti degli strumenti di concetto di struttura, di schema, di operazione elementare, di associazione simbolica, della definizione di isomorfismo e più in generale di coincidenza strutturale, e i vari altri proposti nelle pagine precedenti, è possibile cominciare a dare delle risposte precise ad alcune questioni che sono cruciali per comprendere i meccanismi della cognizione. In particolare ci sono due domande che sono particolarmente importanti:

- Che cosa è precisamente una regola ?
- Che cosa sono, precisamente, quelle che sono chiamate "proprietà emergenti"?

## 4.2 L'importanza delle regole

Che cosa è una regola? Tutti sanno usare il concetto intuitivo di regola, ma chi sa darne una definizione precisa?

La realtà nella quale viviamo mostra tutta una serie di regolarità, di simmetrie, di proprietà che si conservano e che in certa misura si ripetono, sia nel tempo che nello spazio. Il comportamento dei sistemi fisici obbedisce a delle leggi che sono ben codificabili in termini matematici. Anche in quei sistemi che manifestano comportamenti caotici è possibile descrivere i singoli micro fenomeni componenti inquadrandoli tutti entro un insieme finito di leggi fondamentali, vale a dire di regole che descrivono le operazioni da compiere per prevedere il divenire dei fenomeni.

Ragionando su questo tema appare chiaro che le regole hanno un ruolo fondamentale nella possibilità di conoscere. Nella sostanza, in ultima analisi, la cognizione stessa appare possibile nella misura in cui la realtà è soggetta a delle regole. Affinché l'attività cognitiva sia di qualche utilità, è necessario che si determini la possibilità di riutilizzare la conoscenza acquisita e questo richiede appunto che vi siano aspetti della realtà che tendano, in qualche modo, a ripersi nel tempo e nello spazio. Se dovessimo confrontarci con una realtà che fosse, per ipotesi, priva di ogni regolarità, non avremmo possibilità alcuna di esercitare utilmente la nostra capacità di memorizzare le cose. Che senso avrebbe memorizzare le rappresentazioni di cose, situazioni e fenomeni che non si ripresentano mai, in nessun modo "uguale a se stesse" in qualche loro aspetto strutturale? Se non ci fossero delle "ripetizioni di qualche tipo" in ciò che osserviamo, come faremmo a riutilizzare quanto memorizziamo?

Non è difficile convenire che in assenza di ordine e di regolarità il concetto stesso di conoscenza perde di senso.

Ma che cosa è l'ordine? Che cosa è una regolarità? Cosa è una legge fisica o una legge logica? Se andiamo a vedere su un vocabolario o su un'enciclopedia, troviamo molti modi per definire questi concetti. Si tratta realmente di definizioni complete e rigorose? Sono in grado di cogliere l'essenza del fenomeno "regola"? Chi è in grado oggi di dare una definizione matematicamente rigorosa del concetto di regola?

Non è strano che non esista ancora una definizione rigorosa di un concetto così fondamentale per ogni scienza e per l'esistenza stessa della cognizione?

Ritengo che l'analisi accurata e la precisazione razionale del concetto di regola costituiscano un passaggio cruciale per comprendere i principi che stanno alla base della conoscenza.

Vedremo che, con la possibilità di imbrigliare il concetto di regola tramite una precisa definizione, si rendono anche disponibili gli strumenti per riuscire a dare un inquadramento preciso ad altri concetti di notevole importanza. Tra questi spicca quello di proprietà emergente.

### 4.3 Seconda congettura di riferimento

Nel secondo capitolo ho proposto una congettura che credo sia di notevole importanza per la costruzione di una teoria razionale della cognizione: tutto ciò che è conoscibile razionalmente si limita alle strutture dei fenomeni e alle operazioni possibili su queste strutture.
Ritengo che questa congettura possa avere un ruolo di "riferimento" poiché da essa si possono derivare buona parte delle idee su cui si fonda il presente lavoro. Un ruolo altrettanto importante ha allora anche la seguente, che può essere considerata la "seconda congettura di riferimento" per le idee che propongo.

Seconda congettura di riferimento:
**Ogni regolarità corrisponde sempre alla possibilità di identificare delle ripetizioni strutturali.**

In altre parole, ogni volta che identifichiamo la presenza di una regolarità lo facciamo attraverso il riscontro di almeno un isomorfismo (totale o parziale) tra strutture. Secondo i casi queste strutture possono essere di prima o di seconda specie, quindi possono essere anche degli "schemi", e possono riferirsi sia a strutture di base sia ad astrazioni.
Credo che questa congettura sia corretta, ma per mostrare che essa effettivamente funziona in tutti i casi possibili sono necessari ulteriori strumenti che saranno illustrati nei capitoli che seguono. Non sono in grado di fornire una dimostrazione formale della sua correttezza, ma ho verificato che funziona in tutti gli esempi che sono riuscito a formulare.

Questa congettura può anche essere espressa nel seguente modo:
**In una serie di informazioni esiste una regolarità se, e solo se, ci sono almeno due strutture, o due caratteristiche strutturali, non distinguibili.**

Secondo quanto visto nel capitolo precedente, la coincidenza strutturale è ben definibile in termini matematici.
La congettura significa anche che l'unico modo che abbiamo per accorgerci che vi è una regolarità è quello di identificare almeno una coincidenza strutturale. Se questa non è subito palese, significa che dobbiamo fare qualche tipo di operazione che ci consenta di "estrarre" da quelle date delle altre strutture (di prima o di seconda specie) che siano tra di loro coincidenti.

In molti casi identificare queste "coincidenze strutturali" è relativamente facile, in altri è invece difficile. In alcuni casi le difficoltà dipendono dal fatto che la "regolarità" si manifesta a livello delle attività di pensiero, quindi per fenomeni che avvengono all'interno della nostra mente.

Quando la regolarità si manifesta in "fenomeni esterni ben osservabili", vale a dire in fenomeni per i quali siamo in grado di costruirci delle rappresentazioni complete delle loro strutture, si possono avere vari casi. Credo che i principali siano quelli qui riassunti:

- Primo caso (il più semplice): è quello banale, dove si ha una serie di oggetti, di fenomeni o di operazioni, con porzioni nelle loro strutture che sono effettivamente identiche.
- Secondo caso. Può capitare che le porzioni non siano del tutto identiche ma si assomigliano molto, nel senso che, "modificando di poco" le strutture che si confrontano, queste vengono a coincidere. Chiaramente deve essere affrontato il problema di definire il concetto di "piccolo cambiamento" che è qualcosa che dipende dal contesto specifico, ma che in genere indica che la modifica necessaria riguarda una percentuale contenuta degli elementi che definiscono una struttura. Resta comunque il fatto saliente che la parentela tra le varie strutture è identificabile grazie alla proprietà di isomorfismo.
- Terzo caso. Le strutture di base non coincidono ma è possibile eseguire delle operazioni di derivazione strutturale, come quozienti e morfismi, dalle quali si ricavano strutture coincidenti. Si noti che le derivazioni eliminano parte del contenuto informativo delle strutture di partenza (ma si noti anche che non ne aggiungono di nuovo!).
- Quarto caso. Si ha una serie di strutture tra loro non coincidenti, per nessun aspetto del loro "contenuto informativo interno". Ma intuiamo lo stesso che siamo in presenza di una regolarità poiché esiste una "stessa operazione" che consente di passare da una struttura all'altra lungo la serie che si sta esaminando. In questo caso sono le operazioni da eseguire ad essere sempre le stesse. Abbiamo visto che una sequenza di operazioni è rappresentabile tramite una struttura di seconda specie, vale a dire uno schema. In questo caso la coincidenza riguarda proprio la struttura di questo schema. Tutte le operazioni che sono eseguite nel passare da una struttura a un'altra hanno schemi coincidenti. Può generare un attimo di confusione il fatto che siamo abituati a pensare "la legge", "la formula", "l'operatore" utilizzato come una "cosa unica". Diciamo che una legge fisica, esprimibile analiticamente, è sempre la stessa, perché la formula che utilizziamo è unica. In realtà possiamo anche vedere le cose in questo modo: a ogni occorrenza, quando passiamo da una struttura della serie a quella successiva, usiamo delle formule che sono tutte isomorfe tra di loro e soprattutto isomorfe a una rappresentazione comune che abbiamo memorizzato da qualche parte. Regolarità di questo tipo sono "esterne" al contenuto d'informazione della sequenza che si sta esaminando. Vedremo più avanti cosa significa.

L'idea generale è, quindi, che per quanto nascosta e non subito palese, la regolarità presente, entro una serie d'informazioni, diventa esplicita quando è possibile verificare la presenza di una coincidenza strutturale.

Questo per quanto riguarda il fenomeno delle "regolarità", che sono in genere il risultato dell'applicazione di "regole". Credo sia corretto anche quanto segue:

**Ogni regola consiste in una "prescrizione strutturale" o sulle operazioni che si compiono, o sui risultati che si ottengono**.

Nella sostanza una regola descrive la struttura delle operazioni da compiere o la struttura dei risultati da ottenere.
Uso qui il concetto di "operazione" in senso generalizzato: sono operazioni quelle elementari eseguite da una macchina computazionale (che ho cercato di descrivere nel capitolo precedente), ma anche le azioni fisiche concrete, i comportamenti complessi e le "operazioni formali". In questo senso **sono operazioni tutti quegli atti che generano dei cambiamenti**.
Per quanto visto nel capitolo precedente assumo che ogni comportamento e ogni operazione, per quanto complessa o astratta, possa essere ricondotta all'esecuzione di una sequenza (anche molto complessa) di operazioni di computo elementari.
Affermare che una regola consiste in una prescrizione strutturale significa che, se la si applica più di una volta, allora devono essere identificabili delle strutture che coincidono. Applicando più volte una regola si ottiene una regolarità.
Come detto, quando ci riferiamo a regole e a regolarità che si manifestano all'interno dei processi cognitivi, può essere difficile scorgere con chiarezza la presenza di "corrispondenze strutturali". Spesso usiamo regole che prescrivono cosa si può fare e cosa non si può fare, utilizzando per la loro formulazione dei concetti molto astratti che si riferiscono ad attività che riguardano il pensiero e la sua gestione. Con il linguaggio siamo in grado di comunicare queste regole, ma ciò nonostante non siamo in grado di rappresentare con assoluta precisione e in modo completo le loro strutture[6]. In questi casi può non essere facile verificare che una regola consiste effettivamente in "prescrizioni vincolanti" sulle strutture dei comportamenti o dei processi cognitivi da adottare. Sospetto che questo accada proprio perché non abbiamo ancora a disposizione una "teoria completa" della cognizione e dei suoi processi. Credo che con gli strumenti adeguati sia possibile mostrare che anche le regole più astratte sono riconducibili a prescrizioni strutturali.

---

[6] In effetti la logica, che dovrebbe essere la disciplina che si occupa di codificare le "leggi del pensiero", è in grado di descrivere solo una parte dei processi di inferenza.

### 4.4  Concetto di regola e concetto di regolarità

Nel linguaggio comune i termini "regola" e "regolarità" possono avere diverse accezioni in funzione del contesto nel quale sono applicati. Ad ogni modo, in genere, le regolarità sono "fenomeni che accadono nella realtà". Un sistema cognitivo ha quindi un ruolo passivo rispetto ad esse. Diversamente, il concetto di regola è legato a un ruolo attivo. La regola, in genere, è qualcosa che disciplina un'azione o un'operazione (anche formale), è qualcosa che vincola sulle possibilità di agire sul mondo. Il soggetto che compie l'azione, o l'operazione (generalizzata), può essere una persona, un animale o un congegno, ma può essere anche un ente astratto, come lo è una legge fisica, o un operatore matematico. Nel linguaggio e nel pensiero comune tendiamo psicologicamente ad associare le azioni a un soggetto che le compie; tendiamo quindi a dire che le leggi fisiche sono "entità che agiscono", regolando il divenire degli eventi.

Il concetto di regola implica dunque un approccio attivo da parte di un sistema cognitivo. Queste attività possono essere sia "elaborazioni interne al sistema" di informazione di vario genere, sia azioni fisiche, vale a dire "operazioni concrete", che si compiono interagendo con l'ambiente esterno.

Quando utilizziamo una regola produciamo delle regolarità; se è corretta la seconda congettura, questo implica che l'utilizzo di una regola deve passare sempre attraverso la verifica di una qualche coincidenza strutturale.

In generale possiamo dire che tale coincidenza costituisce un **vincolo** che deve essere rispettato.

Secondo i casi questo vincolo può agire o direttamente sul modo di operare, oppure sui risultati che si ottengono.

Il vincolo costituisce una regolarità da rispettare; dunque, per la seconda congettura, questo consisterà sempre in una coincidenza di struttura di qualche genere.

Si possono presentare due casi fondamentali. Nel primo la regola dice sostanzialmente **come** si deve procedere e quindi possiamo parlare di **regole procedurali**. Nel secondo, invece, la regola dice **cosa** si deve ottenere; in questo caso possiamo parlare **di regole vincolanti sui risultati**.

Vediamo più in dettaglio:
- **Regole procedurali**. La regola vincola il modo nel quale si procede. In questo caso la regola descrive le operazioni da compiere, vale a dire lo schema procedurale (quindi la struttura di seconda specie), che le descrive. Si possono avere a loro volta altri sottocasi: da quelle che vincolano totalmente le azioni da compiere, e che quindi non lasciano libertà alcuna (un esempio sono le leggi fisiche fondamentali, che vanno semplicemente sempre applicate), a quelle che invece permettono di scegliere entro un certo insieme di operazioni permesse (un esempio è il gioco degli scacchi, in cui si può appunto scegliere tra un certo insieme di mosse legali). In generale in tutti questi casi le regole dicono come si deve procedere; i risultati da ottenere, invece, non sono esplicitamente predeterminati.

- **Regole vincolanti sui risultati**. Possiamo avere casi nei quali la regolarità vincola non il modo in cui si deve agire, ma i risultati che devono essere ottenuti. Si lascia quindi libertà di azione, non si dice quali mosse si devono compiere, quali operazioni sono da fare. Nello stesso tempo s'impongono però dei vincoli su quello che deve essere ottenuto. Questi vincoli, se è vera la seconda congettura, avranno allora la forma di corrispondenze strutturali che si devono manifestare in quello che si ottiene con le azioni, o le operazioni, eseguite.

Come accennato, un'altra possibile classificazione delle regole consiste nel distinguerle tra quelle **totalmente vincolanti** e quelle **parzialmente vincolanti.** Approfondiremo quest'aspetto nel capitolo 7.

Un'altra possibile e utile classificazione è quella che distingue le regolarità che possono essere presenti in una sequenza di strutture, in **interne** e **esterne**.
Una regolarità è **interna** quando la coincidenza strutturale è contenuta direttamente nelle "informazioni interne alle strutture" prese in considerazione. Questa coincidenza può essere non subito palese e, al fine di estrarla, si devono compiere delle operazioni di analisi strutturale. In questo caso la regolarità può essere resa palese attraverso una serie di operazioni che comunque non **aggiungono alcun "contenuto informativo interno", che non sia già implicitamente presente entro la sequenza data**.
La regolarità è invece **esterna** quando la coincidenza strutturale riguarda un operatore comune, ma esterno alla sequenza stessa. È il caso visto sopra, che si ha quando una serie di strutture è generata tramite una sequenza di operazioni comuni, che, agendo sulla struttura n-1 della serie, generano quella successiva. In questo caso l'informazione sulla regolarità non è contenuta internamente alla sequenza di strutture, neppure in maniera implicita, ma è invece esterna. Credo che in situazioni di questo tipo non esista un metodo diretto (generale) per scoprire la detta regolarità e si debba procedere per tentativi, quindi per

"congetture e confutazioni", per dirla alla Popper. Talvolta può essere molto difficile individuare regole esterne.

Ho fatto menzione al concetto di "**contenuto informativo interno**". Con esso intendo un'idea molto affine a quelle di informazione di Shannon e di complessità algoritmica di Chaitin Kolmogorov. Questo concetto è legato alla "quantità di regolarità interne" presenti nella struttura in esame. In funzione di queste la struttura può essere compressa in un algoritmo (quindi una struttura di seconda specie) di lunghezza più breve che è però in grado di generare quella di partenza. Propongo la congettura che la compressione è possibile solo quando sono presenti queste regolarità interne. Su queste cose ritorneremo nel prossimo capitolo.

Penso che la classificazione più importante per gli scopi di questo lavoro sia quella che distingue le regole in **operazionali** e **associative**.

Avrò gli strumenti per spiegare, con maggiore chiarezza, i motivi e la logica di questa distinzione nel capitolo 7. Per il momento posso dire che le regole **operazionali** sono quelle che consistono in una serie di operazioni da eseguire su una o più strutture di partenza, e che le "trasformano" in altre. Sono quelle regole che operano su quantità ben definite e che spesso consistono in una serie di calcoli, e/o in una serie di operazioni algoritmiche, che agiscono direttamene su ciò che definisce le strutture. Questa tipologia di regole comprende quindi le leggi della fisica e tutte quelle per le quali si utilizzano espressioni matematiche o algoritmi di una certa complessità; ma possiamo anche includere in essa delle regole operazionali intese in senso generalizzato, come quelle che descrivono dei comportamenti complessi o molto astratti, o dei processi di pensiero.

Le regole **associative** operano sul riconoscimento di un certo insieme di "premesse" associandovi delle "conclusioni", senza seguire direttamente le operazioni che trasformano quelle di partenza in quelle finali. Come accennato, mi sarà possibile chiarire meglio queste cose nel capitolo 7. Vedremo anche che le regole associative sono particolarmente importanti nell'attività cognitiva: in molte situazioni il loro utilizzo prevale nettamente rispetto a quello delle regole operazionali.

I concetti di regola e di regolarità sono fondamentali per comprendere con quale "logica" devono essere strutturati tutti i processi cognitivi. Un principio generale da tener presente è il seguente: **si deve cercare di massimizzare la capacità di sfruttare in maniera vantaggiosa le regolarità alle quali la realtà si dimostra soggetta**.

Un sistema cognitivo è essenzialmente un sistema che cerca di individuare **tutte le regolarità** che sono presenti nelle informazioni che riceve in input dai propri apparati sensoriali, e che agisce in maniera da estrarre da queste delle **regole affidabili per** riuscire a prevedere l'evoluzione degli eventi, per pianificare le azioni, e in generale per simulare in maniera affidabile i vari scenari possibili del reale.

# 5   Il fenomeno delle strutture emergenti. Strutture, schemi e logiche emergenti.

Cosa sono le proprietà emergenti? Quale è la loro logica? Quale ruolo hanno nella possibilità di conoscere?

## 5.1   Introduzione.

La fisica subnucleare descrive il mondo come composto da "particelle elementari in interazione" entro lo spazio-tempo (e che hanno anche effetti su quest'ultimo). Questi "piccoli pezzi" di materia ed energia sono tutt'altro che entità stabili e ben definite, sono soggetti a processi di creazione ed annichilazione e le leggi quantistiche che governano la loro esistenza appaiono difficili da inquadrare secondo gli strumenti cognitivi del nostro senso comune. A rigore non sappiamo con sicurezza se le particelle che sono state fino ad ora individuate siano davvero elementari o se vi sia qualche altra struttura più semplice sotto di esse. Anzi, per quanto ne sappiamo, la nostra stessa "fisica elementare" potrebbe anche essere una proprietà emergente di un altro substrato la cui fisica potrebbe essere per noi virtualmente insondabile. Comunque sia, a prescindere da questioni così "fondamentali", per fissare le idee possiamo inquadrare buona parte degli oggetti macroscopici grossomodo come costituiti da "insiemi di particelle elementari" che si riuniscono in atomi e molecole, quindi in sistemi che hanno una buona stabilità nel tempo.

Noi siamo però abituati a pensare alle cose del nostro mondo quotidiano, ai vari oggetti macroscopici con i quali interagiamo, come ad esempio al tavolo che abbiamo davanti, o alla sedia su cui siamo seduti, senza preoccuparci di pensare che essi in realtà sono fatti di atomi e di molecole.

Per comprendere il senso e la logica dell'idea di "entità emergente" credo sia importante cercare di rispondere a questa domanda: qual è il criterio generale che fa sì che, in un certo istante, particolari insiemi di particelle possano essere considerati come oggetti macroscopici autonomi, mentre altri insiemi no?

In realtà non si tratta solo di una questione di individuazione di particolari "sottoinsiemi" di ciò che compone le cose. Si consideri, infatti, che molti degli oggetti macroscopici che noi abitualmente consideriamo come cose a sé stanti, e che sembrano permanere nel tempo, spesso non corrispondono mai allo stesso insieme di particelle. E' famoso l'aforisma di Eraclito: "Non si può entrare due volte nello stesso fiume, perché si è bagnati da acqua sempre nuova". La questione diventa ancora più interessante se si considera il fatto che spesso usiamo definire degli oggetti che non corrispondono proprio ad alcun insieme di atomi. Si pensi ad esempio agli enti geometrici astratti, o quando indichiamo come oggetto a sé stante qualcosa che non è fatto da alcunché: cose come un'apertura, o un foro nel mezzo di un solido.

Esistono tutta una serie di problematiche non banali che sono connesse con questi fenomeni Si tratta di questioni sulle quali si sono cimentati molti pensatori nel passato e che sono ancor oggi dibattute anche in ambito scientifico. Accenno di seguito ad alcune di queste, utilizzando in parte il linguaggio introdotto nei capitoli precedenti:

- Alcuni pensatori si sono posti il problema di come un complesso di cose possa essere considerato come una singola cosa;
- di come un certo complesso di parti microscopiche possa dare origine a qualche cosa di macroscopico, la cui cognizione prescinde totalmente dal fatto che tale oggetto sia costituito da quel tipo di parti;
- di quali siano i criteri secondo i quali, tra tutti gli insiemi possibili di atomi, solo alcuni sono considerabili come oggetti macroscopici;
- di come sia possibile che alcuni oggetti macroscopici siano cognitivamente considerabili sempre gli stessi sebbene capiti, in molti casi, che gli atomi che li compongono cambino nel tempo;
- di come sia possibile avere cognizione pertinente di cose che non sono costituite da alcunché di fisico.

Ho proposto che degli oggetti e dei fenomeni esterni possiamo costruire delle rappresentazioni che riguardano solo le loro strutture e le operazioni di computo possibili su queste; ho inoltre illustrato una possibile metodica per descrivere queste ultime in modo oggettivo. Secondo lo spirito delle idee esposte, non ha molta importanza se le parti che compongono queste strutture corrispondono o no a qualcosa che "ha sostanza". È sufficiente che le parti siano in qualche modo tra loro distinguibili. Anzi, a rifletterci bene, seguendo questo modo di vedere le cose, il concetto stesso di "sostanza" non ha una reale giustificazione. Si consideri che chi studia le basi della fisica teorica sa bene che, procedendo con metodo riduzionista, si finisce con l'avere a che fare con entità che sono "varietà geometriche astratte", vale a dire delle "pure strutture", e non delle "sostanze".

Più indietro ho anche proposto che da delle strutture di base è possibile estrarne delle altre eseguendo una serie di operazioni, in particolare quelle di derivazione strutturale. Ricordo inoltre che ho fatto menzione all'esistenza di "proprietà strutturali" che possono essere riconosciute con metodi computazionali pur non costituendo delle "strutture autonome" di prima o di seconda specie.

Continuando su questa linea di pensiero mi pare naturale proporre che le famose "proprietà emergenti" siano anche loro delle strutture e delle proprietà strutturali, di prima o di seconda specie, che possono essere in qualche modo "**estratte**" da altre più basilari.

Il punto cruciale della faccenda è che non tutte le strutture derivabili, e le proprietà strutturali ben definibili, sono necessariamente delle "**legittime entità emergenti**"! Non tutte le operazioni di derivazione strutturale producono delle strutture "**interessanti**". Quando decidiamo di separare dal contesto un certo sottoinsieme di parti e di considerarlo come oggetto a sé stante, lo facciamo seguendo delle regole, anche se inconsapevoli; non mettiamo assieme parti a caso.

Quale è allora la logica che permette la "separazione dal contesto" di certe particolari porzioni e non altre? Quand'è che un'operazione di derivazione produce una struttura derivata legittima?

Intuitivamente si potrebbe essere indotti a pensare che ha senso separare una porzione di una struttura dal resto quando ci sono "delle proprietà" che sono comuni solo a tutte le parti che appartengono a quella specifica porzione. Per estensione d'idee, si potrebbe pensare che ci debbano essere criteri analoghi per tutte le operazioni di derivazione lecite. Credo che in quest'approccio ci siano effettivamente degli aspetti importanti, che come vedremo sono connessi a ciò che possiamo chiamare "**contenuto informativo interno**" delle strutture, ma non credo sia questo l'approccio corretto per capire la logica del fenomeno dell'emergenza.

Credo che il criterio corretto sia un altro e che dipenda da ciò che può essere "fisicamente rilevato". Penso che un'operazione di derivazione di una nuova struttura o di una proprietà strutturale non autonoma, sia lecita se, e solo se, quanto viene prodotto è qualcosa che è in grado di generare "effetti fisici rilevabili".

Ma in cosa possono consistere questi effetti fisici? Se è vera la prima congettura, essi devono consistere in qualcosa che "cambia la struttura di qualche altra cosa", poiché partiamo dall'ipotesi che sono le strutture a esaurire ciò che possiamo "rilevare" del mondo che ci circonda!

Nelle prossime pagine propongo sostanzialmente che ciò che intendiamo per "proprietà emergenti" consistano in strutture derivate e proprietà strutturali che rispondono a quello che indico come "**criterio di emergenza**". Ritengo inoltre che per la loro identificazione sia essenziale procedere con particolari "operazioni di analisi", tra le quali sono particolarmente importanti quelle di derivazione strutturale, soprattutto quelle di quoziente e di morfismo.

Propongo quindi, come vedremo, che le proprietà emergenti corrispondano ai concetti di **struttura emergente** e di **proprietà strutturale emergente.**

### 5.2   Interdipendenza funzionale tra strutture, criterio di emergenza

Sappiamo che molti fenomeni naturali sono mutuamente legati da relazioni d'interdipendenza funzionale. Chiaramente quando lo stato di un sistema fisico, che a livello macroscopico può essere considerato come entità a sé stante, dipende dallo stato di un altro, significa che vi è interazione fisica tra i due a livello dei loro costituenti elementari. Ciò nonostante molto spesso compaiono delle regolarità ben codificabili che coinvolgono elementi strutturali macroscopici. Ad esempio, banalmente, le dimensioni e la geometria di un corpo solido determinano come questo può essere incastrato con altri. Nel momento dell'interazione fisica sono, in ultima analisi, molti singoli atomi che vengono a interagire tra loro nei punti di contatto. Nei due corpi però le rispettive collettività atomiche sono tra loro legate secondo relazioni geometriche che tendono a conservarsi nel tempo e che oppongono una certa resistenza qualora si cerchi di forzarne l'ordine reciproco. Ne consegue che gli "elementi strutturali macroscopici" determinano cosa si può fare e cosa non si può fare, quello che può accadere e quello che non può accadere. In un caso di questo genere, da un punto di vista cognitivo, questi elementi strutturali costituiscono dei legittimi "soggetti a sé stanti".

Usando il concetto di dipendenza funzionale può essere proposto il seguente criterio che stabilisce quanto un certo "elemento strutturale" è distinguibile dal contesto come struttura (o proprietà strutturale) emergente e quando non lo è.

**Criterio di emergenza.**
- Un certo insieme di parti, in certe specifiche relazioni, emerge come "entità unica" se, e solo se, nella realtà esistono delle altre strutture, o degli altri elementi di struttura, il cui stato può dipendere funzionalmente proprio da tale insieme, preso per intero, e non da sue parti o da sue porzioni considerate singolarmente.

In questo caso diremo **emergente** le strutture, di prima o seconda specie, o le proprietà strutturali che partecipano, per intero, in tali fenomeni di dipendenza funzionale.
Possiamo quindi parlare di **strutture emergenti** e di **proprietà strutturali emergenti**.
Ipotizzo, quindi, che se non si verifica un fenomeno di dipendenza funzionale come quello menzionato, allora non esiste alcun motivo per considerare un certo insieme di parti come una struttura emergente.

In effetti, se per ipotesi non vi è alcuna entità del mondo fisico che dipende da un certo complesso strutturale allora, ovviamente, non esisterà niente e nessuno in grado di segnalarne la sua presenza. Se per assurdo abbiamo un certo insieme qualunque di parti (che nella pratica possono essere cose di vario genere: molecole, oggetti più o meno grandi, insiemi di punti dello spazio-tempo ecc..) ma se nello stesso tempo non vi è alcunché nella realtà il cui stato dipende da tale insieme, considerato in tutto il suo complesso, allora non vi sarà nessun fenomeno che per così dire si "accorgerà" dell'esistenza dell'insieme in questione. Esisteranno le singole parti, ma quello specifico insieme non avrà alcun peso per nessun fenomeno e quindi non avrà modo di essere rilevato.

Nel nostro mondo fisico accade continuamente, in una varietà enorme di casi, che lo stato e l'evoluto di certi insiemi di particelle elementari, di certi complessi atomici (da singole molecole fino ad arrivare ad aggregati anche enormemente grandi), dipenda dallo stato di altri complessi di particelle elementari.

Tale dipendenza può concretarsi fisicamente in maniera complessa e non essere semplicemente la "risultante della somma delle parti costituenti".

Il concetto di dipendenza funzionale menzionato ha forti analogie con quello di funzione usato in matematica e nelle scienze fisiche, che a sua volta è legato a quello di regola. In genere la presenza di un legame tra due o più variabili è indice dell'esistenza di qualche correlazione fisica tra le due. Uno degli assunti del metodo riduzionista è che ogni correlazione funzionale è in linea di principio scomponibile in una serie di meccanismi elementari che, per quanto complessi, possono sempre essere espressi in modo analitico, vale a dire tramite descrizioni matematiche. Ad ogni modo, il concetto di funzione prescinde dal fatto che sia possibile o meno esprimere con precisione il legame che esiste tra due o più variabili. Molto spesso si usa per indicare l'esistenza di una relazione che si sa esistere ma che non si conosce nel dettaglio. Situazioni nelle quali si conosce solo in modo approssimativo il legame che lega due o più fenomeni, e soprattutto quelle in cui questo legame è, di fatto, per noi osservabile solo a livello macroscopico, sono molto frequenti nella normale attività cognitiva.

Vedremo che il criterio di emergenza svolge anche un ruolo importante per stabilire quando, entro un sistema di rappresentazione della realtà (e i sistemi cognitivi ricadono ovviamente in questa categoria), una certa struttura derivata può essere considerata, o non essere considerata, un "soggetto a sé stante". Si noti però che questo criterio di emergenza è una proprietà indipendente dall'attività cognitiva; in genere i fenomeni di dipendenza funzionale si manifestano nella realtà fisica a prescindere dal fatto che vi sia qualcuno in grado di osservarli.

### 5.3   Un criterio di emergenza di validità meno ampia

Ho affermato più indietro che la cognizione è affamata di regolarità. Credo che uno dei leitmotiv dei processi cognitivi sia la ricerca delle regolarità e il loro utilizzo sotto forma di regole. Le regole sono fondamentali, sono il motore dell'attività cognitiva.
Il criterio di emergenza proposto sopra sembra avere validità generale e penso sia coretto. Tuttavia credo sia utile considerarne anche un'altra versione di validità leggermente più limitata.

**Secondo criterio di emergenza:**
- Sono emergenti ogni struttura, ogni schema e ogni proprietà strutturale che contribuiscano a individuare almeno una regola valida.

Significa che se una struttura o una derivazione strutturale o un certo schema, o qualunque proprietà strutturale esplicitabile, partecipa nella definizione di una regola di qualche tipo effettivamente utilizzabile, allora essa va considerata emergente. Probabilmente si tratta di una condizione sufficiente ma non necessaria.

### 5.4    Alcuni punti importanti sulle strutture emergenti

Nel criterio di emergenza appena proposto ci sono alcuni aspetti particolarmente importanti sui quali vorrei richiamare l'attenzione del lettore.
Poniamo l'attenzione sul fatto che una struttura non è costituita da un'unica entità elementare, ma è costituita da una pluralità di parti. Quando una struttura, o una proprietà strutturale, emerge, lo fa come oggetto complesso. Essa deve quindi essere presente per intero, con tutto ciò che serve per definirla:
- insieme delle parti,
- sistema di distinguibilità interna tra le parti,
- complesso delle relazioni esterne,
- e, per le strutture di seconda specie: sistema di associazione simbolica tra parti di struttura e operazioni elementari;

Il criterio di emergenza dice che ha senso considerare questo complesso d'informazioni come qualcosa che agisce come "**un tutto unico**" se e solo se esiste qualcosa nella realtà in grado di "accorgersi" che questo ente complesso esiste; quindi, come detto, se c'è qualche fenomeno che dipende da tale struttura presa nella sua complessità e non solo da un sottoset di ciò che la definisce.
Se esiste un fenomeno che dipende da un particolare sottoinsieme delle informazioni che definiscono una certa struttura, allora sarà quest'ultimo sottoinsieme a essere emergente, anche nel caso che questo non costituisca propriamente una struttura autonoma.

### 5.5    Sul concetto di proprietà strutturale non autonoma

Ho usato più volte il concetto di proprietà strutturale non autonoma; ora è possibile precisare meglio questa nozione proprio grazie al criterio di emergenza.  Consideriamo il concetto di "cinque" o l'idea di "convessità". Possiamo individuare innumerevoli strutture costituite da cinque parti, così come possiamo individuarne tantissime che presentano, nella loro geometria, qualche forma di convessità. Ma siamo in grado di produrre una struttura di prima o di seconda specie che sia solo "cinque" o solo "convessità"? Chiaramente questo non è possibile, ma ciò nonostante esse sono indubbiamente "dei legittimi soggetti dell'attività cognitiva". Non è difficile trovare innumerevoli casi in cui l'accadere o meno di un certo fenomeno dipende dall'esistenza o meno di entità di questo tipo. Appare quindi del tutto legittimo considerare anche queste come "proprietà della realtà" che sono oggettivamente emergenti. Ma se non sono strutture a sé stanti, di cosa si tratta? E com'è possibile costruire delle loro rappresentazioni?

Penso si possa affermare che una **"proprietà strutturale non autonoma"** è costituita da un complesso di "informazioni strutturali" che sono emergenti, secondo il criterio appena esposto, ma che da sole non sono sufficienti ad individuare una struttura autonoma.

Spesso le proprietà strutturali corrispondono a notevoli morfismi non autonomi (ad esempio nel caso dei numeri naturali); o a particolari relazioni tra le parti costituenti, che sussistono in certe porzioni della struttura un esame (cadono in questa categoria proprietà come la convessità, la spigolosità, e tante altre). Tutte queste proprietà possono essere rese esplicite attraverso un'opportuna serie di operazioni di computo strutturale. Le operazioni che conducono "all'esplicitazione" della singola proprietà strutturale sono a loro volta rappresentabili tramite uno schema; esse producono un risultato che, nel caso più generale, sarà qualche tipologia di struttura o, nel caso limite (ma tutt'altro che raro), sarà una singola informazione elementare (quindi **un singolo bit**), che con il suo valore "certifica" la presenza di quella determinata proprietà.

Un'altra caratteristica comune di tutte queste proprietà strutturali è che l'algoritmo che le identifica non aggiunge informazioni che non siano già implicitamente contenute entro la struttura di partenza.

Una **proprietà strutturale** implica quindi l'esistenza di un algoritmo (nel senso di una serie di operazioni), rappresentabile con uno schema procedurale (vale a dire con una struttura di seconda specie), che sia in grado di renderla esplicita.

Quest'algoritmo non deve aggiungere nuova informazione[7] che non sia già implicitamente contenuta dentro la struttura di partenza.

Come detto **le proprietà strutturali non sono strutture autonome;** questo implica che **esse richiedono sempre la presenza di qualche struttura che "le contenga" o di un algoritmo in grado di generarle.**

Questo passaggio può sembrare banale ma è importante. Significa, ad esempio, che ogni qualvolta abbiamo a che fare con oggetti molto comuni quali i numeri, ci serve in realtà una qualche struttura che "li contenga", oppure qualche algoritmo che sia in grado, almeno potenzialmente, di generarli.

---

[7] Qui per "informazione" e in altri punti, per "contenuto informativo" intendo un concetto analogo a quello di complessità algoritmica introdotta (per quanto è a mia conoscenza) da Kolmogorov, Chaitin, Solomonoff. Una "struttura estesa" può essere generata da un algoritmo, anche più corto della struttura stessa, qualora in essa siano presenti delle regolarità. L'algoritmo più corto che la può generare sarà rappresentato da uno schema e avrà quindi una propria struttura non comprimibile. Sospetto che la lunghezza di questo schema sia in relazione con il contenuto di informazione presente entro la struttura di partenza. Si tratta di argomenti interessanti che penso meritino opportuni approfondimenti.

La questione è sottile; noi siamo in grado di riferirci al concetto di numero, e in modo analogo a molti altri, anche in maniera molto astratta, utilizzando per queste "rappresentazioni cognitive" una serie di riferimenti ad ulteriori riferimenti. Ma credo che in ultima analisi debba sempre esserci, ad un certo punto dei vari passaggi, o la struttura che "li contiene" o, come detto, l'algoritmo in grado di generarla.

### 5.6 Fenomeno delle regole e delle logiche emergenti

Molto spesso i fenomeni che coinvolgono strutture emergenti possono essere simulati tramite gruppi di regole che hanno un aspetto ben diverso dalle leggi fisiche. Queste regole sono anch'esse, a tutti gli effetti, dei fenomeni **emergenti**.
Possiamo quindi riferirci ad esse dicendo che sono delle "**regole emergenti**".
La nostra realtà è caratterizzata dalla comparsa di molte **regolarità** e molte **regole** di questo tipo. Il concetto di **regola emergente** è fondamentale per comprendere la logica della cognizione. Noi utilizziamo continuamente regole emergenti per fare previsioni, per generare inferenze, per pianificare azioni e comportamenti. Pressoché tutta la nostra attività cognitiva dipende da regole di questo genere. Senza il fenomeno naturale "della comparsa di regole emergenti" l'intelligenza non sarebbe mai potuta evolversi.

Penso si possa definire emergente ogni regola che può essere formulata in maniera indipendente dalla conoscenza delle strutture e delle altre regole, di livello gerarchico inferiore, che ne costituiscono il "substrato portante".

Ricapitolando, in molte situazioni si determinano dei fenomeni di dipendenza funzionale tra strutture (e proprietà strutturali) emergenti. Spesso accade che questi siano governati da un insieme di regole che possono essere espresse in maniera completa anche **senza avere conoscenza alcuna delle strutture di partenza dalle quali quelle in oggetto derivano**. In tal caso si determinano dei sistemi di computo strutturale rappresentabili e simulabili in maniera indipendente dal substrato che, appunto, non serve conoscere.
Penso che per tali paradigmi si possa utilizzare, forse con un piccolo abuso di linguaggio (che però consente di rendere in maniera efficace alcune idee importanti), la locuzione di "**logiche emergenti**."
Possiamo quindi dire che:
- con "logica emergente" indichiamo un insieme di regole (emergenti) mutamente legate, applicabili a un certo complesso di strutture emergenti.

Come detto, gran parte della nostra attività cognitiva riguarda proprio questo genere di strutture e di logiche emergenti. Uno dei punti salienti di questo fenomeno è che in molte situazioni, quando si rappresenta una logica emergente, non solo non è strettamente necessario rappresentare la logica (vale a dire il complesso delle regole) del substrato da cui questa deriva, ma non è neppure conveniente.

Spesso, utilizzando direttamente le regole che costituiscono una logica emergente, si possono facilmente ottenere dei risultati che sarebbero invece difficili (e forse impossibili) da trovare operando sui substrati di livello gerarchicamente inferiore! Credo che questo sia un punto di notevole importanza da tenere ben presente. Sospetto che pressoché la totalità dei fenomeni che rappresentiamo nella nostra mente riguardino logiche emergenti.

Un sistema di rappresentazione e simulazione in grado di identificare ed utilizzare le regole emergenti può essere estremamente più funzionale rispetto ad un altro che non lo fa. Questo è probabilmente **uno dei "trucchi fondamentali" dei sistemi intelligenti**.

Più avanti, nel capitolo 7, espongo alcuni importanti approfondimenti su questi argomenti. Vedremo che, in ultima analisi, è proprio il fenomeno delle regole e delle logiche emergenti a determinare come deve essere codificata e organizzata l'informazione all'interno di un sistema cognitivo. È importante comprendere che un sistema cognitivo non può permettersi di ignorare le regole emergenti: esso deve sfruttare tutte le regolarità e tutte le regole che possono essere utili. Vedremo che è proprio il fenomeno delle regole emergenti ad imporre che in un sistema cognitivo la realtà debba essere rappresentata non ad un unico livello, ma bensì tramite una **gerarchia di rappresentazioni**.

## 5.7 Oggettività dei fenomeni di emergenza

Secondo quanto visto fino a ora, le strutture e le logiche emergenti non sono solo fenomeni che riguardano la rappresentazione della realtà entro un sistema di conoscenza, ma presentano una forma di oggettività indipendentemente dal fatto che siano "pensate" da qualcuno o che siano rappresentate in qualsiasi sistema cognitivo.

Si pensi a un programma di calcolatore. Esso può essere rappresentato in termini di bit e in termini di istruzioni macchina in maniera strettamente deterministica. Si può essere in grado di simulare con precisione l'evoluzione dello stato dei circuiti logici di un calcolatore a prescindere dalla conoscenza delle leggi fisiche che governano i dispositivi allo stato solido con cui sono realizzati. Il programma e il suo divenire costituiscono dunque una logica emergente che può essere simulata in maniera indipendente dal substrato.

Riassumendo, in generale si può dire che una logica emergente consiste in un complesso di strutture e in un complesso di regole da applicare a queste. In un tipico processo di simulazione cognitiva, le rappresentazioni delle strutture emergenti sono collegate le une alle altre da regole emergenti in modo tale che sia possibile tracciare un percorso che le comprenda tutte. Tale sistema di computo strutturale, come detto, può manifestarsi a alto livello di derivazione e costituire un sistema che non necessariamente dialoga con i livelli sottostanti. In tal senso costituisce una "logica" che può essere utilizzata in maniera parzialmente indipendente dal substrato (in taluni casi anche totalmente). Come si avrà modo di vedere più avanti, in molti contesti tale substrato è di fatto sconosciuto, non si ha cioè modo di conoscere quali siano i meccanismi microscopici che ne determinano l'evoluto. In taluni casi logiche di questo genere possono apparire deterministiche e complete, come avviene ad esempio in un circuito digitale. In altri casi, che sono quelli notevolmente più frequenti nell'ambiente in cui viviamo, le regole in gioco sono tutt'altro che deterministiche e tutt'altro che complete. Naturalmente vi sono anche situazioni per le quali il complesso di regole emergenti individuabili concretamente non consentono di prevedere granché.

Le logiche emergenti possiedono alcune caratteristiche peculiari, come **la possibilità di computare in negativo** (vale a dire sulla mancanza di qualcosa), cosa che invece non appare possibile a livello di regole base. Spesso le regole emergenti permettono di operare una drastica contrazione della complessità del sistema che si sta simulando. Avremo modo di vedere che molte regole emergenti permettono di compiere previsioni a lungo termine, cosa che spesso è assai difficile ai livelli di base, proprio per problemi di eccessiva complessità, e spesso anche per eccessiva sensibilità alle condizioni iniziali.

Più avanti, nel capitolo 6, avremo modo di vedere che molte delle regole emergenti che utilizziamo normalmente sono di tipo associativo.

### 5.8 Concetto di Analisi di struttura.

Le idee presentate nei paragrafi precedenti mostrano come sia fondamentale per l'attività cognitiva procedere a processi di analisi che permettano di "accorgersi" della presenza **di strutture, di proprietà strutturali e di regole che sono emergenti**.

La nostra cognizione si esplica normalmente a livello di oggetti macroscopici e spesso, in tale contesto, il rapporto profondo tra cause ed effetti non è comprensibile direttamente, ma costituisce invece un "dato di fatto" che si ottiene "sperimentalmente", vale a dire dall'osservazione diretta di come avvengono le cose.

Questo implica che, nella maggior parte dei contesti pratici, la presenza di un rapporto funzionale tra due "strutture macroscopiche" può essere di fatto scoperto e oggettivato solo in base all'esperienza concreta. Come sarà più chiaro nei capitoli seguenti, questo fatto implica che non sia possibile pensare a un processo generale di analisi diretta in grado di decidere a priori se una certa struttura è o no emergente. Nella sostanza l'unica possibilità per testare se si ha o no emergenza è di procedere in due fasi: prima generando delle potenziali "strutture candidate", e poi verificando se queste sono o non sono correlate ad altre tramite qualche tipo di regola. La correlazione è reale quando le strutture in oggetto contribuiscono alla formulazione di almeno una regola d'inferenza valida, oppure quando, in conformità ad altre regole d'inferenza già acquisite, è possibile descrivere dei meccanismi di dipendenza funzionale che legano tra di loro le parti che le costituiscono.

Se è vero che quella di essere emergente è una proprietà che spesso può essere verificata solo a posteriori, e nella maggior parte dei casi solo su base prettamente empirica, ciò non toglie che siano utili, e spesso necessari, dei processi di analisi che devono avvenire prima di tale verifica e la cui funzione è di isolare dal contesto e proporre quelle che sono le "**candidate**" strutture emergenti. Penso sia utile indicare una parte importante di questi processi con il termine "***analisi di struttura***".

Per costruirci una prima idea, possiamo pensare all'analisi di struttura come a un'attività che, partendo da una certa data rappresentazione strutturale di partenza, procede effettuando tutte le derivazioni strutturali pertinenti in modo da rendere espliciti tutti i "contenuti informativi latenti e potenziali" che sono in essa contenuti.

Il prodotto dell'analisi di struttura è quindi una molteplicità di altri oggetti, che sono a loro volta strutture a sé stanti derivate da quella di partenza, o che sono proprietà e relazioni strutturali non autonome. In buona parte dei casi queste entità consistono in **porzioni**, **quozienti e morfismi** di quella di partenza.

Una caratteristica generale di questi oggetti è che tutti hanno un contenuto informativo che è sempre minore di quello della struttura di partenza.

**L'analisi di struttura (ideale) quindi non aggiunge mai contenuto informativo interno.**

Possiamo sintetizzare dicendo che, dato un certo insieme di rappresentazioni, l'analisi di struttura consiste in quel processo che si occupa di individuare e di **esplicitare** tutte le strutture da esse utilmente derivabili e tutte le proprietà strutturali, che sono potenzialmente emergenti. Il tutto **senza aggiungere nuova informazione**.

### 5.9   Strutture emergenti macroscopiche

Particolarmente importanti per l'attività cognitiva che si esplica nell'ambiente naturale sono le **strutture emergenti macroscopiche**. Il concetto di **macroscopico** si applica a quelle classi di oggetti e di fenomeni che si manifestano ad una scala di grandezza tale da risultare sensibile ai nostri sensi. Nel significato comune del termine sono macroscopiche tutte quelle entità che sono sufficientemente grandi, anche se in taluni casi non direttamente osservabili (le galassie sono oggetti macroscopici anche se per vederle abbiamo bisogno di ricorrere a particolari strumenti). Le strutture macroscopiche sono importanti in quanto in natura costituiscono per un sistema cognitivo "l'unico mondo esterno accessibile ai sensi".

## 5.10  Rendere esplicite le strutture: il principio di convergenza delle verifiche

Per com'è stato proposto, il fenomeno delle strutture emergenti appare oggettivo, indipendentemente dal fatto che queste siano o no rappresentate all'interno di qualche sistema. In questo lavoro siamo però interessati alla cognizione e quindi siamo interessati a capire quali debbano essere le modalità per rappresentare, all'interno di un sistema cognitivo, l'equivalente dei fenomeni emergenti della realtà esterna.

Una struttura è costituita da un complesso di parti. Affinché tale struttura sia esplicita in un sistema di conoscenza quest'ultimo deve essere in grado di prendere delle decisioni su tale complesso. Deve quindi in qualche modo essere implementata una funzione equivalente al "sapere" se la struttura è stata, o non è stata, riconosciuta (spesso, se ne è stata riconosciuta una sua approssimazione).

Dato che le parti che costituiscono la struttura in oggetto sono tante, e poiché solo se esse sono tutte presenti in certe specifiche relazioni, allora la struttura stessa, in quanto tale, può dirsi presente nel contesto in oggetto, se ne ricava che è necessario produrre un'**unica informazione** in cui converga la verifica della presenza di tutte le parti in tutte le loro specifiche relazioni.

Probabilmente questo principio di "**convergenza delle verifiche**" ha validità generale (almeno fino a che si rimane in ambito computazionale classico). Si può pensare di esprimerlo come segue:

- Per rendere esplicito il riconoscimento di una certa struttura, o di una certa proprietà strutturale, all'interno di un sistema cognitivo, è sempre comunque necessario far convergere verso un'unica informazione tutte le verifiche degli elementi che la definiscono.

Un modo per attuare il criterio di convergenza è di realizzare un dispositivo che sia in grado di produrre, in modo dedicato, l'informazione dell'avvenuto riconoscimento della struttura in oggetto. Ad esempio si può pensare un dispositivo che fornisce 1 quando la struttura è stata riconosciuta e che fornisce 0 in caso contrario.

L'entità in oggetto può anche essere una proprietà strutturale non autonoma.

Il principio appena illustrato implica che, entro un sistema di conoscenza, è necessario avere molti dispositivi dedicati al riconoscimento delle singole strutture e delle singole proprietà strutturali emergenti. Questi non necessariamente devono essere dispositivi fisici, ma possono anche essere dispositivi algoritmici e devono essere in grado di produrre informazioni che si attivano, in modo specifico, ogni qualvolta vengono riconosciute le strutture e/o le proprietà strutturali alle quali sono associati.

Si noti che per realizzare quanto appena richiesto i singoli dispositivi devono a tutti gli effetti svolgere, almeno in parte, le funzioni di memoria e di decodifica che sono necessarie per il **riconoscimento**. Si tratta di una sorta di funzione di memoria di tipo attivo: il dispositivo non si limita a detenere passivamente l'immagazzinamento dell'informazione, ma svolge anche la "**funzione attiva**" di confrontare il proprio contenuto con quanto è presentato ai propri input, fornendo poi come output un valore proporzionale alla corrispondenza di tale input con quanto è in esso memorizzato. Quando necessario, oltre alla funzione di memoria, il dispositivo deve eseguire tutte le operazioni che servono per la decodifica di una specifica particolare proprietà strutturale.

L'output di questi dispositivi può essere un valore strettamente discreto del tipo 1 o 0, corrispondente, secondo una logica binaria, all'avvenuto o meno riconoscimento dell'entità in questione; ma può anche essere costituito da un altro valore (in genere un numero compreso tra 0 e 1) che, qualora non vi sia corrispondenza totale, ma solo parziale, tra la struttura in memoria e quella in input, rappresenta comunque il grado di somiglianza tra le due.

Può accadere che il processo di convergenza in un'unica informazione avvenga attraverso passaggi intermedi nei quali si ha l'esplicitazione di convergenze parziali. Questo può avvenire ad esempio qualora esistano porzioni, a loro volta emergenti, della struttura in questione. In tal caso si dovrebbe passare prima per l'esplicitazione della verifica dell'esistenza di tali porzioni emergenti e poi procedere a operazioni equivalenti al loro prodotto logico per l'esplicitazione di tutta la struttura.

### 5.11 Ancora sul problema del confronto tra strutture

Riprendiamo il problema proposto all'inizio del secondo capitolo.
Ci si ponga il problema di classificare delle figure tracciate con un segno a penna nera su dei fogli di carta tutti uguali tra loro. Il numero di disegni possibili è ovviamente elevatissimo. Supponiamo di riprendere il tutto con una telecamera, e di studiare degli algoritmi che permettano ad un calcolatore di procedere in maniera automatica ad una serie di classificazioni che per noi osservatori umani sono alquanto naturali. Supponiamo che il problema sia di riconoscere quando il disegno tracciato appartiene ad alcune categorie semplici, ad esempio a quelle dei poligoni, dei poligoni regolari, degli esagoni. Vogliamo inoltre che quest'algoritmo sia in grado di rendere esplicite le regolarità che sono presenti entro queste figure.

La struttura di partenza, che è una struttura base (si ricordi il concetto di struttura base illustrato nel secondo capitolo), è in questo caso data da un insieme di parti (i singoli pixel) che possono assumere, per ipotesi, solo due valori, quindi solo due stati di distinguibilità interna, corrispondenti al colore bianco e a quello nero. Il complesso delle relazioni esterne è dato dal grafo che rappresenta le adiacenze tra un singolo pixel e quelli immediatamente vicini.

Supponiamo di avere a che fare con disegni che sono costituiti solo da linee dritte o curve, ben separate l'una dall'altra, che possono al massimo intersecarsi in pochi punti.

Non è difficile scrivere algoritmi in grado di **riconoscere** punti e linee. Non è neppure difficile fare in modo che questi algoritmi siano in grado di distinguere tra linee dritte e linee curve, e tra coppie di segmenti che si toccano in qualche punto generico, in particolare ai vertici. È anche possibile scrivere algoritmi che sono in grado di riconoscere, senza ambiguità, quando un insieme di segmenti dritti sono uniti in modo da formare una figura chiusa che sarà quindi un poligono.

Il problema che ci poniamo è quello di trovare le similitudini che si possono presentare nei vari disegni, quindi le regolarità che sono presenti in essi.

Ancora una volta il caso più semplice è quando confrontiamo due figure che sono identiche punto per punto. Anche le due strutture di base sono allora direttamente isomorfe. In questo caso scrivere un algoritmo che sia un grado di identificare questa coincidenza è quasi banale.

Le cose diventano però più complesse quando le figure non coincidono più perfettamente. Se i disegni che stiamo esaminando mostrano ambedue due esagoni, ma di dimensioni diverse e ruotati in qualche maniera, come facciamo a costruire un algoritmo che sia in grado di rilevare le regolarità che sono presenti nelle due figure?

Fin tanto che si tratta di esagoni regolari, si può sempre procedere con operazioni di cambiamento di scala, rotazioni e traslazioni, e con queste trasformare le figure diverse in due oggetti che coincidono a livello di struttura base. Una persona che osserva le due figure non ha certo difficoltà a capire di quale entità e in quale verso va fatto l'opportuno riscalamento di dimensioni, nonché l'opportuna rotazione e la giusta traslazione. Ma se vogliamo scrivere un algoritmo che sia in grado di trovare da solo la combinazione giusta, come dobbiamo procedere? Non è un problema banale!

Supponiamo ora che la situazione sia più complicata, supponiamo di essere sempre in presenza di due esagoni, ma questa volta non regolari e fatti in maniera tale che, per quanto si provi, non esista alcuna sequenza di rotazioni, traslazioni e riscalamenti uniformi, in grado di far coincidere le due figure. Per l'osservatore umano che guarda le due figure è semplice intuire che esistono delle corrispondenze, quindi delle regolarità di qualche tipo tra le figure rappresentate; non è invece affatto banale scrivere un algoritmo che sia in grado venire a capo del problema.

Ma in cosa consistono queste regolarità? Nei casi precedenti riuscivamo sempre ad ottenere alla fine due strutture isomorfe effettuando delle opportune operazioni di rotazione, spostamento e cambiamento di scala. L'isomorfismo si presentava sull'intera struttura di base, quindi sull'intera struttura della matrice di pixel che costituisce l'immagine. Se è corretta la seconda congettura di riferimento che ho proposto, dovrebbe essere possibile ricondurre le regolarità, che intuiamo essere presenti entro le due immagini, a delle corrispondenze tra strutture. È possibile fare questo? Se sì, come si deve procedere?

Il problema può essere affrontato in maniera diversa rispetto a quanto illustrato fino ad ora. Invece di cercare di far coincidere le intere strutture di base, possiamo cercare se esiste la possibilità di eseguire delle **operazioni di analisi strutturale**, ad esempio derivando da quelle di partenza delle altre strutture, e verificare poi se queste sono isomorfe!

Ritengo che operando in questo modo passiamo da un livello di rappresentazione ad un altro. Invece di considerare la struttura base, che ha per parti i singoli pixel, andiamo a considerare le strutture che si ottengono considerando come parti componenti gli interi singoli segmenti che sono presenti nella figura. Per fare questo dobbiamo affrontare il problema di trovare gli "elementi che definiscono la nuova struttura secondo le modalità viste nel secondo capitolo".

Le nostre nuove parti di struttura, come detto, sono ora gli interi segmenti, e la struttura che stiamo considerando è "**quoziente**" rispetto a quella di base (o meglio è **quoziente** rispetto ad una certa **porzione** della struttura base, quella costituita solo dai pixel di colore nero). Ma cosa possiamo allora dire della distinguibilità interna e delle relazioni esterne di queste nuove parti di struttura?

Le nuove parti di struttura non sono più singoli pixel che possono avere solo due tipi di distinguibilità interna, ma sono a loro volta oggetti complessi e in quanto tali possono presentare una serie di **proprietà strutturali** che li caratterizzano. Nel caso specifico dell'esempio abbiamo a che fare con segmenti. Di un segmento possiamo esprimere la lunghezza, e dobbiamo verificare la proprietà di "drittezza" che lo differenzia da tutti gli altri tipi di spezzoni di linea. Queste "**caratteristiche della parte**" costituiscono ore le proprietà interne e vanno quindi a codificare il sistema con cui i vari segmenti sono "**distinguibili internamente gli uni dagli altri**". Le relazioni esterne dovranno invece codificare "**come sono messi**" i vari segmenti. Si dovranno quindi specificare le distanze e le orientazioni relative, l'angolo risultante qualora due segmenti si tocchino in qualche punto, e l'eventuale parallelismo.

Procediamo allora a rilevare queste proprietà e a catalogarle opportunamente. Non è difficile scrivere algoritmi che siano in grado di portare a termine queste operazioni.

La nuova struttura che si ottiene, che è un poligono, sarà in questo caso definita da:

- Il numero delle sue parti.
- La tipologia delle parti (le loro proprietà interne): quindi il fatto che sono segmenti (linee dritte), le rispettive lunghezze, gli eventuali angoli di inclinazione.

- Le relazioni esterne tra le parti: quindi il fatto che si tocchino a coppie, che formano a coppie un determinato angolo, che tutto l'insieme costituisce un poligono chiuso, le loro orientazioni nello spazio.

Specificando ed **esplicitando** in maniera completa tutte queste informazioni si individua di volta in volta una struttura poligono particolare.
Due poligoni così definiti saranno isomorfi se e solo se avranno in comune tutte le proprietà descritte sopra. Ne consegue che sono da considerarsi completamente isomorfi solo quei poligoni che si trovano nella stessa posizione ed hanno le stesse identiche dimensioni.
Ma cosa succede questa volta quando abbiamo a che fare con poligoni non sovrapponibili?
Essendo ora le informazioni sulle caratteristiche strutturali **date in forma esplicita,** si può procedere a vedere cosa succede considerando i vari morfismi possibili. Per far questo dobbiamo quindi via via **attenuare** le proprietà che caratterizzano le strutture. Supponiamo quindi di attenuare le distinguibilità interne tra le parti. Si proceda non distinguendo più sulle lunghezze specifiche. Ci ritroviamo allora a codificare con la medesima rappresentazione tutti i poligoni che hanno lo stesso numero di lati e che determinano lo stesso angolo tra i lati stessi. Viste in questo modo diventano isomorfe tutte le strutture che corrispondono a poligoni regolari con lo stesso numero di lati e che hanno una certa orientazione. Possiamo anche agire diversamente e decidere di non considerare come proprietà distintiva le orientazioni e gli angoli specifici. Generiamo quindi un'altra rappresentazione più permissiva e così facendo diventano in questo caso isomorfi tutti i poligoni costituiti da uno specificato numero di lati, ma di dimensioni e di orientazione qualunque.
Vediamo quindi che passando a considerare una particolare struttura quoziente e **traducendo in maniera "esplicita"** le proprietà che la definiscono in quanto struttura, e quindi successivamente, procedendo a sopprimere (o a seconda dei casi: semplicemente ad ignorare) le varie proprietà caratterizzanti, diventa semplice identificare degli isomorfismi strutturali, o varie altre forme di indistinguibilità strutturali, e quindi delle regolarità, che sono presenti nelle strutture in oggetto.
Si noti che, passando dalla rappresentazione estesa di una struttura alla sua versione esplicita, non è sufficiente generare dei semplici lunghi elenchi di tutte le proprietà che sono state identificate, ma è strettamente necessario essere in grado di associare le proprietà ai rispettivi "elementi strutturali". Si consideri, ad esempio, che nell'esecuzione di un quoziente generiamo delle "**nuove parti di struttura**" (che sono porzioni di quella originaria), e di queste ultime dobbiamo rendere esplicite sia le "proprietà interne" sia le "relazioni esterne" che esse hanno con le altre "nuove parti di struttura". Le elencazioni di proprietà esplicite devono quindi essere raccolte in sottoinsiemi e associate alle specifiche entità strutturali alle quali si riferiscono. Nel passaggio dalla rappresentazione estesa a quella composta dall'elenco delle proprietà esplicite è quindi fondamentale trovare il modo di "conservare i riferimenti", che spesso appunto sono relativi a "nuovi elementi strutturali", assenti nella struttura di partenza.

Penso sia quindi sbagliato pensare che con le operazioni di esplicitazione si passi da rappresentazioni strutturali a rappresentazioni simboliche. Le esplicitazioni sono necessarie per il "riconoscimento" delle singole "entità". Ma quando esse sono generate in un sistema cognitivo, devono mantenere una "rete di riferimenti" che, di fatto, conservano la "struttura alla quale esse si riferiscono", che, ripeto, può essere diversa da quella di partenza.

Nota. Ho affermato che il problema dell'identificazione delle varie caratteristiche strutturali dell'immagine dell'esempio sopra illustrato non presenta particolari difficoltà. Questo è vero nel caso specifico, perché ho supposto che si tratti di immagini in bianco e nero, semplici, ben illuminate ecc… Mentre in generale il problema dell'analisi delle immagini "naturali" in condizioni realistiche è molto più difficile.

Come avremo modo di vedere, le operazioni di "analisi di struttura" qui accennate costituiscono solo uno dei modi possibili, anche se basilare, per procedere con "operazioni di astrazione". Vedremo che queste ultime sono fondamentali per cogliere le "similitudini" nelle varie situazioni possibili.

Riassumo alcuni concetti importanti.

Abbiamo visto nell'esempio specifico, volutamente semplificato, che in taluni casi è possibile trovare delle regolarità sotto forma di isomorfismi, o di altre corrispondenze strutturali, procedendo con delle opportune operazioni di derivazione. Con queste operazioni cambiamo la rappresentazione e **rendiamo esplicite** alcune proprietà strutturali importanti della struttura in esame. Ma nel far questo **non aggiungiamo mai nuova informazione che non sia già implicitamente contenuta nei dati originali**.

Ho affermato che, se si passa dalla rappresentazione della struttura di base ad un'altra, dove le varie nuove parti e le varie loro proprietà sono rese esplicite, allora le operazioni di morfismo diventano molto semplici. Con queste operazioni si "trasforma" la nostra struttura di partenza in altre, "**rinunciando**" ad alcune delle **proprietà interne e relazioni esterne** che rendono le nuove parti di struttura distinguibili le une dalle altre.

Se da una struttura di partenza eseguiamo un'operazioni di quoziente, e in qualche modo ammettiamo di riuscire a rendere esplicite tutte le **caratteristiche strutturali che la caratterizzano,** ma, invece di eseguire delle operazioni di morfismo, consideriamo il **prodotto logico** di **tutte queste caratteristiche**, ciò che otteniamo è semplicemente un quoziente dell'intera struttura di partenza. In questo caso la nostra capacità di trovare isomorfismi strutturali non è aumentata, è rimasta la stessa. Se invece rinunciamo a qualcosa otteniamo delle altre rappresentazioni più tolleranti. Con questa operazione, che ho appunto chiamato morfismo, continuiamo a conservare qualcosa della struttura in oggetto, ma non tutto, e in questo modo costruiamo della altre rappresentazioni che sono in ora in grado di individuare intere classi e non oggetti ben specifici.

### 5.12 La relatività della relazione di uguaglianza e il senso dell'analisi di struttura

Quanto descritto nei paragrafi precedenti credo sia interessante, ma lascia aperte alcune domande.

Come può essere che date due rappresentazioni di oggetti diversi, esistano delle operazioni che le fanno diventare **"uguali"**? Procedere in questo modo è davvero corretto o si tratta di "magheggi" ingiustificati? Che cosa giustifica il procedere con le operazioni di derivazione descritte e con quelle che ho indicato come operazioni di esplicitazione?

Per rispondere a queste domande credo sia in realtà necessario riflettere sul significato del concetto di "uguale" e cercare prima di rispondere alla seguente domanda: cosa significa affermare che due oggetti sono uguali?

Credo che una risposta operativamente corretta sia la seguente: **due oggetti complessi possono essere considerati uguali se, qualora scambiati, determinano gli stessi effetti sul resto del mondo**.

Porre le cose in questo modo implica definire la relazione di uguaglianza tra due o più oggetti come "non distinguibili per gli effetti fisici esterni" in seguito ad operazioni di scambio.

Per "effetti esterni" intendo tutto ciò che può essere fisicamente rilevato, sia da un essere intelligente, ma anche da parte di "qualunque fenomeno" (sia che si tratti di fenomeni che si manifestano a livello della fisica di base, sia di qualche effetto che si manifesta a livello macroscopico).

Chiaramente ogni effetto macroscopico dipende dalla sua fisica di base; ma un sistema cognitivo può costruire rappresentazioni corrette che riguardano il solo aspetto macroscopico delle cose, ignorando cosa avviene a livello microscopico.

Questo modo di vedere le cose comporta che la proprietà di "indistinguibilità" può essere pensata non come assoluta, bensì come "relativa" alla classe dei fenomeni emergenti che si prendono in considerazione. In effetti non tutti i fenomeni fisici della realtà esterna sono "sensibili" a tutti gli elementi che definiscono la "struttura vera", vale a dire quella di base (se esiste) di un certo specifico oggetto o fenomeno.

Chiaramente con ciò non posso affermare che non sia possibile determinare in modo assoluto se due oggetti sono oppure no "uguali" per permutazione reciproca.

Ho illustrato nel capitolo 2 che le idee proposte in questo lavoro per descrivere le strutture, possono essere applicate anche agli oggetti "standard" della matematica e quindi anche ai metodi per "descrivere la realtà" che si utilizzano in fisica. Per la fisica esiste una "descrizione strutturale massima" da associare a un oggetto, che dovrebbe, almeno potenzialmente, essere in grado di rendere conto di tutti gli effetti che tale oggetto può comportare sul resto del mondo. Questa descrizione strutturale dovrebbe essere quella più vicina, tra tutte quelle possibili, alla "vera" natura dell'oggetto o del fenomeno che si sta considerando. Dovrebbe inoltre essere anche una struttura di base.

In realtà si può speculare molto sulla reale possibilità di definire questa struttura "vera" di un oggetto. Basti pensare ai limiti che la meccanica quantistica impone alla possibilità di misurare contemporaneamente tutte le grandezze che servono per descrivere fisicamente un oggetto, o all'enormità del numero di variabili che si dovrebbero prendere in considerazione. Inoltre, personalmente, penso esistano dei limiti di natura "logica" che impediscono di pensare che sia veramente possibile definire la "vera struttura" di un oggetto con gli strumenti standard che stanno a fondamento della matematica moderna (e che sono comunque equivalenti a quelli che ho usato anche in questa teoria).

Comunque sia, impostare la trattazione del concetto di "indistinguibilità" come appena proposto è per certi versi un po' delicato e, se non si fa attenzione, si può incorrere in paradossi. Ad esempio, se "per ipotesi" abbiamo a che fare con "due" oggetti, e non uno solo, allora in qualche modo essi devono essere a priori distinguibili, proprio perché sono due e non uno! La fisica classica sembra dirci che due "entità idealmente identiche" sono comunque distinguibili perché occupano posizioni spaziali diverse. Ma l'idea alla base del concetto di uguaglianza è che sia concepibile poterle "scambiare senza introdurre perturbazioni". Il presupposto è: se le "descrizioni strutturali massime coincidono" allora gli effetti di due oggetti sul resto dell'universo dovrebbero essere gli stessi e non dovrebbe quindi esserci nulla in grado di accorgersi che si è verificata questa permutazione. In questo senso le due entità possono essere considerate indistinguibili: appunto per invarianza a "tutti gli effetti esterni" a seguito di un'operazione di scambio.

Sembrerebbero allora essere le "descrizioni strutturali massime", quelle che dovrebbero essere confrontate per dire se due cose sono o no uguali. Ma allora come mai usiamo le strutture derivate e le varie caratteristiche strutturali per descrivere il mondo?

Il concetto è che non tutti i fenomeni reali sono "sensibili" a tutti gli elementi che definiscono la struttura fisica di base di un oggetto. Accade invece che molti fenomeni e strutture emergenti sono sensibili solo ad alcuni di questi elementi. Addirittura spesso sono intere logiche emergenti, vale a dire interi complessi di regole reciprocamente legate, a essere sensibili solo a un certo sottoinsieme di tutto ciò che contribuisce a definire le strutture di base. Per queste regole possono risultare uguali, nel senso di non distinguibili, classi di oggetti che dal punto di vista delle strutture fisiche di base sono invece ben distinguibili.

### 5.13 Il meccanismo della chiave

Un esempio particolarmente significativo, direi emblematico, per comprendere il senso di queste idee, è quello che possiamo indicare come "il meccanismo della chiave".

Una chiave è un oggetto che possiede una piccola struttura particolare, la cui presenza o assenza può causare differenze macroscopiche, talvolta enormi, nell'evoluzione degli eventi. Con la giusta chiave si può essere in grado di aprire una porta, far partire il motore di una macchina, lanciare un missile balistico.

Ciò che è importante in una chiave è la particolare struttura geometrica delle sue dentellature (e ovviamente ciò che consente di applicare gli effetti di tale struttura!). L'uso di una chiave si basa sul fatto che ci sono congegni che sono sensibili proprio a quella particolare struttura, che diventa a tutti gli effetti, per questo motivo, una struttura derivata emergente. Si possono realizzare molte chiavi di modelli molto diversi per impugnatura, fattezze, materiali di costruzione, ecc.. ma di esse, per il meccanismo in grado di innescare il fenomeno emergente in oggetto, ha importanza solo la particolare struttura nella dentellatura. Quest'ultima è spesso solo una piccola porzione, apparentemente insignificante, della struttura complessiva "dell'oggetto chiave". Ciò nonostante, dal punto di vista della funzione eseguita dal quel particolare meccanismo, appaiono di fatto uguali, nel senso di non distinguibili, tutte le chiavi che possiedono solo quella data sottostruttura.

Credo sia anche importante notare come questo fenomeno di "meccanismo della chiave" non compare solo in sistemi artificiali inventati dall'uomo, ma svolge invece un ruolo importantissimo in molti fenomeni naturali, tra i quali spiccano quelli biologici. Il funzionamento della vita si basa su molecole che "hanno la forma giusta" per incastrarsi con altre e rendere possibile l'attivazione di varie serie di processi biochimici!

### 5.14 Analisi di struttura e riconoscimento

L'analisi di struttura ha un ruolo fondamentale nel riconoscimento. Quando combinata con l'analisi delle informazioni sensoriali può essere utilizzata per riconoscere gli oggetti e i fenomeni che ci circondano.

È evidente che riconosciamo gli oggetti in base al loro aspetto, quindi, secondo le idee esposte in questo lavoro, in base allo loro struttura. Abbiamo però anche visto che non è pensabile cercare di riconoscere le cose semplicemente confrontando le loro strutture estese, vale a dire cercando di "sovrapporre" le loro immagini o le loro ricostruzioni tridimensionali.

Allora come riusciamo a riconoscere gli oggetti?

Probabilmente ci riusciamo in **base a combinazioni delle loro caratteristiche strutturali**.

È sicuramente possibile scrivere algoritmi che, data una rappresentazione strutturale di partenza, sono in grado di identificare e renderne esplicita la presenza di molte delle caratteristiche strutturali che questa possiede. Questi algoritmi devono sostanzialmente cercare, entro la struttura in analisi, quali sue porzioni possono essere considerate come "entità a se stanti", devono quindi provvedere a classificarle e a rendere esplicite le loro relazioni reciproche. Negli oggetti concreti queste ultime sono principalmente relazioni di tipo spaziale, ma in generale si possono però anche identificare, classificare e segnalare, relazioni di tipo temporale e di movimento.

Algoritmi di questo tipo sono già ampiamente utilizzati in campi quali la visione automatica, il riconoscimento del parlato e altri ancora. Ci sono inoltre buone evidenze sperimentali che esistano gruppi di neuroni nella nostra corteccia visiva in grado di identificare entità di questo tipo.

A rigore non è ancora stato dimostrato che sia possibile con metodi computazionali, quindi con algoritmi, identificare ogni caratteristica strutturale importante. Ad ogni modo ritengo ci siano forti indicazioni a favore del fatto che almeno una parte importante di queste sono effettivamente identificabili con metodi di questo tipo.

Nello spirito delle idee fin qui presentate segnalo che il concetto di caratteristica strutturale può essere considerato una generalizzazione di quelli di proprietà e di relazione strutturale. Le **proprietà** si riferiscono **all'aspetto interno delle parti di struttura**, mentre **le relazioni riguardano l'aspetto esterno**. Ciò è in accordo con la strategia generale che ho proposto per rappresentare le strutture e per rendere agevoli le operazioni di derivazione strutturale. Ritengo molto probabile che questa impostazione sia corretta, ciò nonostante devo raccomandare di tenere un atteggiamene flessibile. Non posso escludere a priori che esistano delle caratteristiche strutturali (delle "features") che non rientrano in questi due casi.

Le osservazioni su come il cervello riesce nel riconoscimento visivo, mostrano che è oggettivamente probabile che dagli oggetti concreti sia possibile estrarre delle combinazioni di loro caratteristiche strutturali che sono sufficientemente univoche da permetterne il riconoscimento. Credo sia importante notare che in genere dalla struttura di un singolo oggetto sono estraibili contemporaneamente più combinazioni caratterizzanti e non una sola. Questo implica che ci possono essere varie opportunità di riconoscere attraverso di esse un singolo oggetto.

L'idea, quindi, è che queste combinazioni costituiscano delle specie di "firme" che sono associabili ad oggetti specifici in modo sufficientemente univoco.

Credo sia a questo punto importante notare che queste combinazioni di caratteristiche strutturali si manifestano sotto forma di **regolarità** in ciò che viene prodotto con l'analisi di struttura. Possiamo allora capire che **uno dei "trucchi fondamentali", per apprendere a riconoscere le singole cose,** **consiste nel cercare le regolarità che si manifestano nei prodotti dell'analisi strutturale**.

In effetti credo che una delle attività fondamentali nel quale si deve impegnare un sistema cognitivo sia proprio quella di cercare le regolarità che si manifestano nei prodotti delle varie attività di analisi, sia quelle condotte sulle informazioni sensoriali primarie, sia quelle condotte sulle ricostruzioni delle strutture estese.

Credo sia importante riflettere sul fatto che quando osserviamo il mondo esterno sono contemporaneamente presenti molti oggetti, che cambiano spesso di posizione e di forma. Quindi in un singolo "atto di osservazione" acquisiamo in realtà contemporaneamente molte informazioni che riguardano entità diverse. Il problema di identificare quali caratteristiche appartengono a specifici oggetti non è affatto semplice. Per capire la logica del tutto è utile mettersi dal punto di vista di un sistema cognitivo che riceve in input delle informazioni strutturali e che su queste esegue una serie di operazioni di analisi. Cosa potrà ricavare con queste? Se non possiede già una "conoscenza delle forme delle cose", sarà in grado di identificare solo una congerie di singoli particolari strutturali, e potrà al massimo, accorgersi, con tecniche di analisi statistica, che ci sono particolari insiemi di queste che tendono a presentarsi con regolarità.

Come detto, ritengo che un sistema cognitivo debba utilizzare queste regolarità per il riconoscimento dei singoli oggetti. Le singole caratteristiche strutturali che fanno parte di questi particolari insiemi devono quindi essere computate, tramite opportune funzioni, come "condizioni" che, opportunamente combinate, concorrono all'identificazione di singoli oggetti e fenomeni.

In realtà si può vedere che il problema del riconoscimento degli oggetti, sia da informazioni visive, sia, più in generale, da informazioni strutturali generiche, è spesso molto difficile. Ciò per più ragioni. Spesso le informazioni sensoriali sono "intrinsecamente di cattiva qualità", nel senso che non corrispondono direttamente a buone ricostruzioni delle strutture reali degli oggetti ma sono invece solo delle proiezioni frammentarie e distorte di frammenti di queste, spesso mischiate assieme in modo complicato. Per questo motivo accade che in realtà l'analisi delle informazioni sensoriale deve spesso andare oltre alla "pura analisi strutturale" e si deve procedere ad aggiungere del contenuto di informazione non direttamente presente nella rappresentazione che si sta analizzando.

Un altro dei motivi per cui il riconoscimento risulta difficile è dovuto al fatto che per riuscire a riconoscere i singoli oggetti, è necessario mettere appunto moltissime regole specifiche che riconoscono le dette combinazioni e le associano all'oggetto specifico. Si tenga anche presente che, come visto, spesso queste sono più di una per ogni oggetto (o per ogni classe di oggetti). Ma per costruire le corrette associazioni sono necessari vari passaggi che richiedono moltissime elaborazioni. È necessario riuscire a identificare le varie caratteristiche strutturali attraverso una serie di operazioni di analisi, che spesso devono essere ripetute molte volte sia nelle diverse porzioni delle medesime informazioni sensoriali (ad esempio lungo il campo visivo), sia per stratificazioni gerarchiche. Va inoltre considerato che spesso si deve procedere alla cieca, non è infatti possibile predeterminare a priori quali operazioni riusciranno a identificare caratteristiche strutturali, e loro combinazioni, realmente utili, e quali invece saranno inutili. Si consideri che ci sono molti modi possibili per eseguire operazioni di analisi e per mettere assieme le caratteristiche identificate: in genere problemi di questo tipo sono caratterizzati dalla crescita esponenziale delle possibilità! Un'altra difficoltà è dovuto al fatto che le opportunità di identificare le combinazioni giuste possono presentarsi con una frequenza relativamente bassa.

Credo sia possibile adottare delle strategie di apprendimento che consentono di aggirare parzialmente alcuni di queste problemi.

Come ultima nota raccomando di non confondere le combinazioni di caratteristiche strutturali che sono utili per il **riconoscimento iniziale** di un oggetto, con le rappresentazioni che costituiscono la "**conoscenza**" degli oggetti. Penso che quanto appena illustrato sia corretto per quanto riguarda il problema del **riconoscimento primario**, ma non credo che la rappresentazione degli oggetti e dei fenomeni si limiti a identificare e codificare le regolarità nelle combinazioni di loro caratteristiche strutturali. Non credo che la rappresentazione di un oggetto si limiti nella identificazione di un particolare "pattern di feature". Esse sono utili, anzi probabilmente fondamentali, per eseguire quel riconoscimento iniziale che permette di attivare conoscenze e altri processi che a loro volta permettono di costruire delle rappresentazioni ben più complete.

## 5.15 Alcuni approfondimenti

Ritorniamo a considerare il concetto di analisi di struttura. Penso che esso possa essere applicato sia per i processi di elaborazione che avvengono a ridosso degli stimoli prossimali, sia in fase più avanzata, quando si devono analizzare le "ricostruzioni" delle strutture reali degli oggetti macroscopici. Con ciò non intendo affermare che tutti i processi di analisi degli stimoli prossimali rientrano nelle specifiche elencate quando ho illustrato il concetto di analisi di struttura. Come accennato, ritengo, infatti, probabile che, in queste prime fasi di elaborazione delle informazioni sensoriali, sia spesso necessario aggiungere del contenuto informativo che non è presente nello stimolo prossimale. Ad esempio, spesso, quando si analizzano le immagini visive, è importante completare delle linee di bordo, sebbene queste non siano subito derivabili nella loro intera lunghezza con tecniche di filtraggio delle immagini. Riflettendo sulle tipologie di informazioni che devono essere rese esplicite nelle prime fasi di elaborazione delle informazioni sensoriali, mi sembra evidente che è spesso necessario aggiungere molta informazione strutturale che non è direttamente contenuta nei dati iniziali. Ciò nonostante mi sembra plausibile che una parte di questi processi corrispondano effettivamente a genuine attività di analisi strutturale.

Penso sia particolarmente interessante il caso delle informazioni di tipo visivo. A mio parere i processi di visione devono avere almeno due obiettivi principali.Si tratta del riconoscimento diretto degli oggetti osservati in base alle loro immagini bidimensionali, e la ricostruzione delle geometrie tridimensionali delle loro superfici. Questi due aspetti sono probabilmente spesso fortemente correlati, ma credo che, in linea di principio, possano essere analizzati in maniera indipendente.

Si noti che siamo spesso in grado di costruire delle buone rappresentazioni delle geometrie tridimensionali di oggetti che vediamo per la prima volta, quindi di oggetti che non siamo in grado di riconoscere. Si noti che in questi casi non siamo in grado di costruirci, a partire da una prima singola immagine bidimensionale, come una fotografia, una rappresentazione tridimensionale completa anche delle parti che non sono immediatamente visibili. Per far questo dobbiamo aver modo di ruotare l'oggetto, o di girarci attorno, in modo da poterlo osservare da tutti i suoi lati. È con tutta probabilità vero che anche queste nostre capacità di ricostruire e di collocare correttamente nello spazio le superfici di oggetti che vediamo per la prima volta, utilizzano una cospicuo numero di regole specifiche per interpretare i risultati dell'elaborazione degli stimoli prossimali. Ed è probabile che queste regole richiedano anche molta conoscenza specifica per essere in grado di agganciare gli stimoli visivi con le geometrie corrette. In questo senso è dunque probabile che anche in questi casi utilizziamo in realtà dei processi di "riconoscimento" di forme tipiche, anche se non corrispondono ad interi oggetti noti. In questo ambito sono sicuramente interessanti le idee proposte da Irving Biederman con i geoni[8].

Ad ogni modo penso che sia comunque particolarmente utile, ai fini di comprendere la logica generale della visione, pensare ai due obiettivi indicati (riconoscimento diretto e ricostruzione tridimensionale) come ad attività, almeno in parte, distinguibili.

Si noti che la ricostruzione delle geometrie nello spazio degli oggetti osservati corrisponde proprio alla ricostruzione di "**strutture macroscopiche emergenti**". Infatti le geometrie delle superfici tridimensionali delle cose sono strutture emergenti oggettive e sono ovviamente macroscopiche. In questo senso potremmo pensare che la nostra mente è anche in grado di mettere a disposizione una specie di **teatro virtuale**, dove ricostruisce al suo interno queste strutture, quindi le geometrie tridimensionali delle cose osservate, ma anche di quelle immaginate. Queste rappresentazioni sono completate da varie

---

[8] Secondo una teoria sviluppata da Irving Biederman il nostro cervello ricostruisce le forme degli oggetti componendo un set di "forme standard" che egli chiama appunto geoni (geons)

altre informazioni quali i colori e le ombreggiature delle superfici. In questi teatri virtuali si è anche in grado di seguire i movimenti degli oggetti in tempo reale, ma anche di spostarli "mentalmente", realizzando quindi, delle **simulazioni interne**. Per certi versi queste simulazioni assomigliano a quelle utilizzate ad esempio nei moderni videogiochi. Ma si noti bene, e questo è il punto fondamentale, da sole queste simulazioni ritraggono la realtà solo a "basso livello". In un moderno videogioco non sono rese esplicite le caratteristiche strutturali emergenti.

Propongo l'idea che per realizzare un sistema cognitivo intelligente sia invece necessario eseguire, su queste rappresentazioni, le operazioni di analisi di struttura necessarie a rendere esplicite tutti gli elementi strutturali derivabili che, come vedremo nel prossimo capitolo, possono essere dei "**soggetti a sé stanti**" per i processi cognitivi. Solo in questo modo un sistema può avere accesso ad quel patrimonio di regole emergenti che costituisce la "base portante" della nostra conoscenza semantica.

Quindi, secondo questo modo di vedere, la ricostruzione delle strutture macroscopiche, come lo sono le geometrie degli oggetti reali, sono solo delle rappresentazioni di base, appunto di basso livello. Su di esse è necessario eseguire varie operazioni di analisi di struttura per rendere espliciti tutti quegli elementi strutturali (che in letteratura molti chiamano "features" e "pattern"), che costituiscono a loro volta entità emergenti. Come visto, propongo che siano emergenti tutte quelle strutture derivate, proprietà strutturali, relazioni reciproche tra oggetti, ecc… che contribuiscono a codificare qualche regola utile.

Non credo che le operazioni di analisi strutturale siano di per se stesse sufficienti ad individuare tutte le tipologie di informazione che un sistema cognitivo deve essere in grado di riconoscere. Ma credo che queste costituiscano la base per tutti gli altri processi di categorizzazione che possono avvenire a valle.

Per comprendere meglio la logica dei concetti che sto cercando di illustrare, può essere utile seguire un esempio concreto, anche se molto semplificato.

Consideriamo la rappresentazione tridimensionale estesa della geometria di un oggetto concreto, ad esempio di una sedia. Le parti che la costituiscono possono essere raccolte in una partizione di **porzioni** importanti, che corrispondono a elementi come: le gambe, la seduta, lo schienale, e altre ancora. Queste entità hanno delle proprie strutture che possono essere rappresentate in modo autonomo, e sono anch'esse a tutti gli effetti, come vedremo, dei legittimi "soggetti cognitivi". L'intera sedia è, rispetto a queste parti, un'entità strutturale di livello più elevato, che corrisponde ad un'operazione di **quoziente**. In questa nuova struttura le parti componenti stanno le une rispetto alle altre entro un insieme ben specificabile di **relazioni esterne:** di distanza reciproca, angoli reciproci, relazioni di perpendicolarità e varie altre. Queste relazioni sono **caratteristiche strutturali** che possono e devono essere rese esplicite tramite l'utilizzo di opportuni algoritmi. Tra le proprietà strutturali interessanti ci sono informazioni del tipo che le gambe sono quattro, che sono parallele tra di loro, che sono perpendicolari al piano di seduta, ecc... Anche molte di queste informazioni (forse tutte) costituiscono a loro volta dei legittimi soggetti cognitivi.

È dunque **l'associazione delle dette parti, nelle specifiche relazioni, ad identificare il soggetto di ordine più elevato.**

Se tutte queste relazioni sono "rigide" e ben specifiche, nel senso ad esempio che le gambe devono avere particolari rapporti tra lunghezza, larghezza, distanze relative, che la seduta deve essere di un certo specifico colore ecc… il soggetto identificato sarà uno specifico modello di sedia. Quindi la nostra nuova rappresentazione sarà in grado di identificare solo quel particolare modello di sedia, e solo qualora esso sia "integro" nella sua forma originale. Non sarà quindi un "concetto" flessibile, capace di identificare un'intera classe di oggetti.

Per poter generalizzare è necessario passare ai suoi "**morfismi**", e questo si ottiene "inibendo le richieste sulle distinguibilità specifiche". Se costruiamo un'altra rappresentazione, che di tutte le proprietà specifiche di quella iniziale ne mantiene solo alcune opportunamente selezionate, o che comunque si "accontenta" di certificare una corrispondenza parziale, possiamo essere in grado con essa di riconoscere, ad esempio, tutte le "sedie di tipo classico": quelle con quattro gambe, una seduta, uno schienale e poco altro. Inibendo quindi le distinguibilità, cosa che otteniamo nel caso specifico "accontentandoci" solo di un sottoinsieme delle proprietà strutturali iniziali, riusciamo a costruire rappresentazioni in grado di generalizzare e di identificare intere classi di oggetti.

Questo tipo di rappresentazione, che penso sia di "medio livello di astrazione e generalizzazione", basato su un certo insieme selezionato di proprietà strutturali costanti, non può ancora essere paragonato alla nostra capacità di concettualizzare, ma penso sia comunque un primo rudimento di questa funzione. Se non abbiamo ancora definito il "concetto sedia", nel senso più generale, abbiamo comunque prodotto un soggetto cognitivo legittimo, di livello intermedio, per il quale è soddisfatto il criterio di emergenza.

Ha chiaramente senso chiedersi che cosa ci permetta di riconoscere **"a priori"** un certo insieme di parti strutturali, come appunto le gambe, la seduta, lo schienale, e le loro relazioni reciproche, come elementi che vanno correlati reciprocamente per proporre un primo prototipo di un "quasi concetto" di ordine più elevato. Credo che nella prima costruzione della conoscenza del mondo, quindi nelle prime fasi di apprendimento, si tratti essenzialmente delle "regolarità statistiche" con le quali le parti, e le relazioni menzionate, si sperimentano quando si interagisce con l'oggetto in questione.

Uno specifico modello di sedia è un oggetto fisico che in genere conserva nel tempo la propria struttura. In tutte le nostre esperienze esso continuerà a mantenere certe proprietà strutturali, anche se è spostato, ribaltato ecc. Se ammettiamo che il nostro sistema cognitivo sia in grado di ricostruire la struttura spaziale delle varie superfici che lo compongono, e di effettuare dei processi di analisi, capaci di codificare e memorizzare come stanno queste superfici le une rispetto alle altre, allora ripetendo questi processi ad ogni osservazione identificheremo in esso una serie di regolarità.

Molte di queste relazioni e proprietà strutturali sono indipendenti dalla particolare posizione da cui si osserva una sedia, dal fatto che essa sia o no ribaltata ecc. Queste proprietà risulteranno quindi degli **invarianti strutturali**, delle regolarità, che andranno a costituire un soggetto a sé stante, di livello più elevato, che sarà il primo abbozzo del "quasi concetto" che individua quella particolare sedia (o quelle uguali ad essa se sono più di una).

Per funzionare bene è importante che un sistema cognitivo sia costruito in modo tale da andare sempre alla ricerca di questo tipo di regolarità, di fatto deve esserne "avido". Penso che una delle direttive fondamentali che guidano l'attività cognitiva sia proprio la ricerca di regolarità, che si manifestano nelle esplicitazioni degli elementi strutturali e delle loro relazioni reciproche (anche di gerarchia).

Quello appena illustrato è un esempio volutamente semplificato; con tutta probabilità un sistema cognitivo reale deve essere in grado di analizzare una quantità molto maggiore di elementi strutturali e delle loro relazioni reciproche. Probabilmente sono possibili varie altre operazioni di analisi strutturali, oltre a quelle illustrate. È inoltre molto probabile che si debbano integrare vari processi di feedback, che coinvolgono "riconoscimenti parziali", con "tentativi" di analisi e di esplicitazione di elementi strutturali, che non danno riscontri utili. Credo però che le idee illustrate possano aiutare a comprendere alcuni aspetti di questi processi.

Riassumendo, propongo di considerare la ricostruzione della geometria tridimensionale di un oggetto come una rappresentazione di base. Se essa è di buona qualità, contiene al proprio interno una serie di elementi che permettono di derivare rappresentazioni più astratte, ma proprio per questo più potenti. Queste rappresentazioni sono in genere dei quozienti rispetto alle rappresentazioni di base. In generale un oggetto concreto è composto da "elementi componenti" che sono a loro volta dei legittimi "soggetti cognitivi". Questi devono essere riconosciuti e a loro volta analizzati strutturalmente; la loro composizione costituisce il soggetto di ordine più elevato, quoziente della struttura di base, che identifica una prima astrazione strutturale. Prendendo in considerazione vari morfismi di strutture quozienti di questo tipo, la capacità di generalizzazione aumenta. Di fatto, ai primi livelli di rappresentazioni astratte, molti oggetti sono riconoscibili proprio come astrazioni strutturali.

Come detto, le parti componenti di queste strutture quozienti sono a loro volta delle legittime "entità rappresentazionali" a sé stanti. Nell'esempio della sedia le parti componenti sono le gambe, la seduta, lo schienale… ecc. Questi elementi possono e devono essere a loro volta analizzati strutturalmente, identificando in essi altre porzioni e altre caratteristiche strutturali notevoli, che possono essere a loro volta, dei legittimi "soggetti autonomi" per i processi cognitivi.

Si può procedere ulteriormente. A loro volta anche le rappresentazioni astratte di singoli oggetti, o di categorie di oggetti, possono costituire le parti di entità più complesse. L'identificazione di strutture quozienti può essere reiterata più volte per definire astrazioni di ordine più elevato.

Esempio: un motore è un oggetto composto da pistoni, bielle, testata ecc… Prese da sole queste cose non sono un motore. E non lo sono nemmeno considerate come insieme se non si trovano nelle corrette relazioni spaziali reciproche. Un motore può essere considerato un oggetto di ordine superiore rispetto ai suoi componenti, e precisamente una struttura quoziente, quando questi stanno gli uni rispetto agli altri secondo relazioni precisabili. Un motore di per sé non è un automobile e neppure quattro ruote lo sono. Ma se mettiamo assieme il numero minimale di pezzi giusti otteniamo un altro oggetto, che dal punto di vista cognitivo è un oggetto di scala superiore. A loro volta più automobili, in funzione di come sono disposte, possono costituire altri soggetti ancora: un raduno automobilistico, un ingorgo stradale, il carico di un treno merci, l'esposizione di una rivendita di auto o un'azienda di autodemolizioni.

### 5.16 Separazione dal contesto sulla base delle "irregolarità interne". Concetto di "contenuto informativo interno"

Spesso accade che da un dato vettore di informazioni a una, due, o più dimensioni, è possibile estrarre, e quindi separare dal contesto, alcune strutture derivate sfruttano le regolarità e le irregolarità presenti entro il vettore stesso. Per quanto visto nel capitolo 4, una regolarità consiste nella coincidenza e quindi nella "ripetizione di qualche elemento strutturale". Se, ad esempio, in un'immagine (che è un vettore a due dimensioni) ci sono dei pixel adiacenti che hanno tutti lo stesso colore e la stessa luminosità, sarà naturale metterli assieme e considerarli come appartenenti alla stessa "unità strutturale". Si noti che per i pixel contenuti in questa porzione non ci sono "informazioni interne" all'insieme considerato che giustifichino di separare in qualche modo questo insieme in ulteriori porzioni. Se invece troviamo delle discontinuità nell'immagine, e quindi la "rottura di una o più regolarità", allora è sensato separare altre porzioni e provare a proporle come "unità strutturali a sé stanti". In modo analogo si può procedere considerando anche ripetizioni di elementi strutturali più complessi dei singoli pixel, come ad esempio vari piccoli segmenti in una linea di bordo, o porzioni di "texture" all'interno di un area.

Se si riflette sulla questione non credo sia difficile convenire che in generale è lecito separare, a priori dal contesto un certo elemento strutturale generico se si verifica una qualche rottura nelle regolarità interne. Questo fatto sembra essere una proprietà generale di processi di analisi che possono essere eseguiti a priori su alcune importanti classi di strutture.

Noto che la possibilità di operare in questa maniera sembra legata a quello che possiamo chiamare "contenuto di informazione" presente all'interno della strutture in esame". Uso qui il termine "informazione" nell'accezione di Shannon. In qualche modo il "**contenuto informativo interno**" di una struttura è legato alla possibilità di scomporla in porzioni che possono avere una loro autonomia. Quando l'informazione è minima, quindi quando la struttura è massimamente uniforme (regolarità massima), non ci sono "giustificazioni interne" per compiere alcuna scomposizione. Diversamente, quando il "contenuto d'informazione interna alla struttura" è alto, significa che ci sono delle discontinuità (rottura di regolarità interne) e che quindi è giustificato dividere il complesso in porzioni.

Qui la questione è un po' difficile: ogni struttura, per se stessa, è qualcosa di costituito da una pluralità di parti, e se appunto sono più di una, significa che in qualche modo devono pur essere distinguibili le une dalle altre; deve quindi comunque esserci sempre un certo "contenuto di informazione". Il punto è che, a parità di numero di parti componenti, ci possono essere strutture uniformi, vale a dire con parti non distinguibili internamente, e con relazioni esterne tra loro uguali (ma comunque tali da garantire la "distinguibilità esterna" delle singole parti le une dalle altre); e ci possono essere strutture che presentano delle disuniformità. Le strutture uniformi sono anche quelle che sono "meglio comprimibili", poiché possono essere generate da un algoritmo di lunghezza minima: ad esempio se la struttura in oggetto consiste in una sequenza di N parti tutte uguali (con N non troppo piccolo), allora può essere generata da una macchina computazionale attraverso la ripetizione di N cicli, e da un algoritmo breve. Ricordo che a sua volta un algoritmo può essere considerato una struttura di seconda specie, quindi una struttura nella quale alcune parti sono associate a specifiche singole operazioni della macchina computazionale stessa.

Diversamente, una struttura non uniforme, pur composta dallo stesso numero di parti, potrà essere generata solo da un algoritmo più lungo; in questo senso essa appare intrinsecamente più complessa, quindi dotata di un maggiore contenuto di informazione interna.

Riassumendo, la possibilità di distinguere porzioni di struttura in base alle regolarità/irregolarità interne eventualmente presenti in quella di partenza, appare legata al concetto di contenuto di informazione alla Shannon, e ai concetti di complessità algoritmica (o computazionale) di Chaitin - Kolmogorov.

### 5.17 Le strutture derivate non bastano

Quanto abbiamo visto finora non è ancora sufficiente per classificare tutte le tipologie di informazione che possono essere presenti all'interno di un sistema cognitivo. Le tecniche illustrate in questo capitolo permettono di rendere esplicito solo quello che possiamo chiamare "contenuto informativo interno" ad un certo insieme di rappresentazioni strutturali di partenza. Ma questo è probabilmente solo il substrato di partenza delle classificazioni che la nostra mente è in grado di fare. Vedremo che molte entità reali possono essere utilmente e legittimamente classificate anche in funzione di "**proprietà acquisite**", che derivano dai ruoli che esse assumono in un altro tipo di rappresentazioni di scala superiore a quanto abbiamo visto finora. Ad esempio abitualmente classifichiamo le cose di tutti i giorni in funzione dell'uso che ne possiamo fare, o in funzione delle effetti che possono determinare, o delle implicazioni che possono avere. Queste nuove proprietà non sono in alcun modo estraibili dal contenuto informativo interno alle loro strutture.

Per questo e anche per altri motivi, abbiamo bisogno di un ulteriore concetto, che descrivo nel prossimo capitolo.

# 6  L'esplicitazione della complessità: i soggetti della cognizione

## 6.1  Introduzione

Finora abbiamo visto alcuni primi elementi che possono suggerire alcune idee su come possono essere codificate e organizzate le informazioni all'interno di un sistema cognitivo. Particolarmente importanti a questo scopo sono le idee espresse nel capitolo precedente sulle strutture e sulle proprietà strutturali emergenti, e sulla necessità di procedere alla loro "esplicitazione".
Credo si possa mostrare che in un sistema cognitivo è davvero necessario procedere con processi di analisi di struttura il cui scopo è quello di mettere in evidenza quelle rappresentazioni che si riferiscono a "reali entità emergenti". Credo inoltre che questi processi costituiscono la base per ogni altra costruzione di rappresentazioni astratte. Ciò nonostante ritengo che sia possibile, e necessario, rappresentare astrazioni che vanno oltre quello che possiamo chiamare "contenuto strutturale interno". Pensiamo, ad esempio, al concetto di veicolo: si tratta di una classificazione che dipende dall'uso che si può fare di un oggetto, e non dalla sua specifica struttura, anche se è comunque sempre quest'ultima ciò che permette di riconoscerlo.
Nel capitolo precedente abbiamo anche visto che è necessario essere in grado di "riconoscere" in modo univoco, quelle porzioni di informazione complessa, che costituiscono delle "entità a sé stanti". Credo sia opportuno cercare di generalizzare questa funzione cognitiva di identificazione e riconoscimento dei "singoli oggetti", in modo indipendente dai concetti di struttura e proprietà strutturale emergente. Credo che il filo conduttore che permette di cogliere nella sua essenza questa funzione sia legato al concetto di regola emergente.

Le regole svolgono un ruolo centrale nell'attività cognitiva, costituiscono il motore della cognizione: ogni processo cognitivo ha senso se contribuisce al corretto utilizzo di almeno una regola valida.
Un fatto importante è che nella pratica dell'attività cognitiva le regole dominanti sono di tipo associativo. Contribuiscono anche le regole di tipo operazionale, ma queste ultime per essere applicate hanno comunque il bisogno del supporto di regole associative. Illustrerò nel prossimo capitolo alcuni dei motivi di questa dominanza delle regole associative.
Nella pratica, semplificando, si può affermare che le regole associative consistono, come dice il nome, nell'associazione di "cause ed effetti". Si può allora capire che, affinché le regole stesse siano codificabili e utilizzabili, è necessario rendere in forma esplicita le "entità" che all'interno del sistema cognitivo contribuiscono alle rappresentazioni delle cause e degli effetti.

L'esplicitazione del riconoscimento di una "singola entità" svolge dunque un ruolo fondamentale nell'attività cognitiva ed è opportuno utilizzare per questa funzione una terminologia specifica. Propongo quella di **soggetto cognitivo**.

Un soggetto cognitivo è, semplificando un po', ogni tipologia d'informazione che deve essere riconosciuta come "entità a sé stante" e il cui riconoscimento deve essere segnalato, quindi **reso esplicito,** al complesso dei processi che costituiscono l'attività cognitiva.

Vedremo che i soggetti cognitivi possono essere definiti come i "punti di applicazione delle regole" e che svolgono la loro funzione proprio perché capaci di contribuire alla codifica di regole valide.

### 6.2 La terminologia

Per il concetto che intendo illustrare in questo capitolo si pone un problema di terminologia, cioè di scegliere tra le parole disponibili nel linguaggio comune, o tra le loro combinazioni, quella più adatta per designarlo. Il problema non è semplice, probabilmente perché il linguaggio si riferisce ad un patrimonio concettuale, condiviso tra i membri di una comunità, che spesso non è adatto per descrivere alcuni aspetti della realtà che sono, per varie ragioni, non subito accessibili al senso comune. Di seguito illustro i motivi che mi hanno spinto a scegliere la terminologia "soggetto cognitivo".

Nella maggioranza dei casi quando si scatta una fotografia, o si disegna qualcosa, lo si fa inquadrando un "soggetto" ben determinato. Spesso esso compare assieme a molte altre cose che fanno parte della scena ritratta, ma l'attenzione del fotografo, o dell'artista, è indirizzata su qualcosa di particolare. Si usa in tale contesto il termine "soggetto della fotografia", o del disegno, e si usa la parola "soggetto" intesa non nel senso, più ristretto, che essa ha in grammatica di "chi o cosa compie l'azione", ma nel senso di "ciò che è oggetto dell'attenzione".

Il "soggetto" è in un certo senso "**il protagonista**" della situazione.

In una fotografia il soggetto è ciò che il fotografo ha selezionato dal contesto e sul quale ha focalizzato la propria attenzione. Anche se sono presenti altri elementi, ne interessa solo uno in particolare. Il soggetto della fotografia è di norma a sua volta composto da varie parti di figure. Se fotografiamo una persona, la sua figura sarà composta dalle aree che costituiscono la testa, il corpo, gli arti, ecc.. Quando la nostra attenzione si focalizza sull'immagine della persona, solo questi elementi vengono presi in considerazione, mentre tutto il resto presente nella fotografia passa in secondo piano, diventando "figure di sfondo".

A tuttora non sono particolarmente soddisfatto di questa terminologia. Ho pensato ad alcune alternative. Si potrebbero usare termini del tipo "**soggetti delle rappresentazioni**", oppure il generico "oggetti", o "oggetti cognitivi", oppure ancora "atomi cognitivi", o "esplicitazioni emergenti" o altri ancora. Per il momento mi sembra che quello di "soggetto cognitivo", da intendersi appunto come quella particolare informazione che è per determinati processi il "soggetto" di taluni processi cognitivi, sia quello che rende meglio il senso del concetto che intendo esprimere.

In generale possiamo dire che può essere "soggetto cognitivo" ogni porzione di informazione che è legittimamente separabile dal contesto e considerabile come una rappresentazione di una entità, anche astratta, a sé stante.

Vedremo che i "soggetti cognitivi" si prestano naturalmente ad un'organizzazione per livelli gerarchici. Ve ne saranno di basso livello, di vari livelli intermedi e di alto livello. Alcuni di questi ultimi somigliano fortemente ai nostri concetti. Personalmente parto del presupposto che la presente è in fin dei conti una teoria "solo computazionale" della cognizione e, a rigore, non esiste la prova che anche le nostre facoltà mentali superiori, i nostri pensieri, possano essere ricondotti, nella loro interezza, a "pura computazione classica". Se così fosse, allora effettivamente molti soggetti cognitivi di medio e di alto livello corrisponderebbero a quelli che sono per noi i concetti.

È anche vero che a basso livello di rappresentazione si possono individuare soggetti cognitivi molto semplici, come ad esempio il riconoscimento di una particolare categoria di linee di bordo, o di particolari ombreggiature, o particolari "texture" ecc.. Se utilizzassi il termine concetto per quelli di livello medio alto, dovrei utilizzare un termine del tipo "sub-concetto" per quelli di livello inferiore.

Ad ogni modo, il rispetto che nutro per il valore della consapevolezza umana e per i misteri che comunque permangono sulla sua natura, mi spingono a utilizzare una terminologia diversa. Nonostante alcuni risultati indubbiamente incoraggianti che ho ottenuto, continuo a considerare i soggetti cognitivi di livello più elevato, che come vedremo sono anche quelli più astratti e più somiglianti agli elementi che costituiscono le nostre facoltà superiori, come delle "emulazioni" delle nostre capacità di concettualizzare.

### 6.3 Un primo approccio intuitivo

Ad una concezione intuitiva dell'idea che sto proponendo si può arrivare in maniera empirica da una semplice analisi introspettiva di come tutti noi pensiamo. Tutti abbiamo esperienza di come naturalmente tendiamo a suddividere il mondo conosciuto in singole **cose,** ognuna delle quali può in linea di massima essere considerata come entità a sé stante. Si può notare come sia una caratteristica intrinseca di tale processo il fatto che il pensiero della cosa in questione deve essere discriminabile, e quindi isolabile in modo univoco, dalla pluralità di tutti i contesti in cui appare.

Esprimiamo questo fatto dicendo appunto che queste rappresentazioni delle singole cose costituiscono dei "soggetti autonomi". È chiaro però che al momento sto parlando primariamente di una funzione mentale, quindi cognitiva, prima ancora di una proprietà del mondo reale. In questo senso si tratta di soggetti relativi alla cognizione e quindi, appunto, di "soggetti cognitivi". Questo nostro modo di suddividere la conoscenza è un processo naturale che ognuno di noi mette in atto in maniera del tutto automatica, senza avere precisa cognizione di come esso avvenga. È chiaro che se vogliamo comprendere i fondamenti della cognizione è importante capire quale sia la sua logica, quali siano i principi su cui si basa. Già il fatto stesso di pensare che si tratti di una suddivisione di informazione costituisce un'ipotesi sulla natura di questa nostra facoltà. Serve però anche capire quale sia la sua funzione. L'idea che viene sostenuta nelle pagine che seguono è che essa sia sostanzialmente quella di permettere l'utilizzo della nostra conoscenza delle "regole del mondo". L'idea sottesa a questa interpretazione è che i soggettivi cognitivi fungano da "**centri di riconoscimento**" delle informazioni e che debbano con ciò anche assumere il ruolo dei **" punti di appoggio" per la decodifica, l'implementazione e l'utilizzo di pressoché tutte le regole che costituiscono la conoscenza.**

## 6.4  Approccio ontologico

Da un punto di vista ontologico il concetto di soggetto cognitivo può essere posto in relazione, almeno in molti casi, con quello di struttura emergente. I soggetti cognitivi sono, almeno in buona parte, l'equivalente, a livello di rappresentazione interna, delle strutture e degli schemi emergenti della realtà esterna. Le "cose concrete del mondo", i singoli oggetti che vediamo e nominiamo sono, per un sistema cognitivo, strutture, proprietà strutturali e schemi emergenti.
È naturale che le rappresentazioni interne vadano distinte le une dalle altre in funzione di come sono naturalmente distinguibili le corrispondenti strutture della realtà, ovvero, appunto, in base alle loro proprietà di emergenza. Ricordo che considero l'essere emergente una proprietà oggettiva.
Vanno fatte alcune precisazioni.
In primo luogo in un sistema cognitivo reale vi possono essere varie imperfezioni nella capacità di conoscere e può quindi capitare che siano isolati dei soggetti cognitivi illusori che non corrispondono a reali strutture emergenti.
Un'altra nota da fare è che molti soggetti possono riferirsi non solo a proprietà della realtà esterna, ma anche a stati e proprietà che sono interni allo stesso sistema di conoscenza.
Il sistema stesso può, in molti contesti, essere considerato come facente parte integrante della realtà che si vuole rappresentare. Si possono generare della "**meta rappresentazioni**" che si riferiscono all'attività stessa di costruire ed elaborare rappresentazioni. Inoltre possiamo costruire rappresentazioni di mondi verosimili ma non reali, o di "mondi" che osservano solo una parte delle regole a cui è soggetta la nostra realtà. In altre parole possiamo usare la fantasia.

Una possibile definizione "ontologica" del concetto di "soggetto cognitivo" potrebbe essere che esso corrisponde all'**esplicitazione** di quelle rappresentazioni che si riferiscono a strutture, proprietà strutturali e schemi emergenti, sia della realtà esterna oggettiva, che delle realtà possibili, ma anche di stati interni degli stessi processi cognitivi.

### 6.5 Approccio funzionale

Esiste però un altro approccio alla questione che è probabilmente più importante da un punto di vista operativo. Si deve infatti tenere conto che, per quanto varie possano essere le attività di elaborazione interna di un sistema cognitivo, non è difficile convenire che esse devono **tutte basarsi sull'applicazione di regole**. Per convincersi di ciò basta, ragionando per assurdo, provare ad identificare un processo di inferenza che possa essere condotto senza ricorrere al loro utilizzo.

Non è difficile concludere che in pratica non ne esistono.

Ne consegue che l'informazione che è presente all'interno di un sistema cognitivo è utile, e trova con ciò il suo motivo di essere, se permette di codificare e utilizzare delle regole di qualche tipo.

La funzione primaria di queste suddivisioni, classificazioni e codificazioni di informazione che chiamo soggetti cognitivi è consentire la codifica e l'utilizzo delle regolarità che si manifestano nell'ambiente (ai vari diversi livelli di emergenza).

Si può proporre la seguente definizione:
- **In generale i soggetti cognitivi corrispondono a quelle suddivisioni e codificazioni dell'informazione che consentono di implementare delle regole utili**

Questa definizione appare particolarmente importante poiché è quella che individua, da un punto di vista operativo, quella che penso sia la vera funzione dei soggetti cognitivi: **I soggetti cognitivi fungono da punti di riferimento per l'implementazione e l'utilizzo delle regole.**

Questa definizione non è in contrasto con il punto di vista ontologico. In effetti, secondo il principio enunciato nella sezione precedente, una struttura della realtà esterna è emergente se per essa si manifesta un fenomeno di dipendenza funzionale con altre strutture.

In ultima analisi, ogni fenomeno di dipendenza funzionale dovrebbe essere riconducibile alle leggi fisiche che ne determinano l'evoluto. Tali leggi sono a loro volta "le regole" cui è soggetta la realtà. Vedremo tuttavia che esistono una serie di difficoltà che impediscono, di fatto, che vi possa essere una corrispondenza completa tra le leggi fisiche e le regole che effettivamente un sistema cognitivo è in grado di identificare e di utilizzare. Vedremo quindi che, all'atto pratico, un sistema cognitivo sarà in grado di far corrispondere un soggetto cognitivo a una struttura emergente solo nella misura in cui questa dà luogo a regole accessibili ed effettivamente utilizzabili.

Con il proseguire dell'esposizione sarà chiaro che un sistema cognitivo è in grado di dare un senso all'informazione che riceve in input nella misura in cui riesce ad associarla a delle regole utili.

Da quanto appena visto consegue che il problema dell'individuazione e della decodifica dei soggetti cognitivi è strettamente legato al problema della decodifica ed implementazione delle regole utili. Come avremo modo di vedere, ciò ha un peso rilevante nell'apprendimento, vale a dire nella problematica generale di individuare dei meccanismi e delle strategie che consentano ad un sistema di "costruirsi" una base di conoscenza efficiente.

### 6.6    Alcune proprietà dei soggetti cognitivi

In questo paragrafo illustro alcune proprietà significative dei soggetti cognitivi che possono contribuire alla loro caratterizzazione.

Non è difficile convenire che la rappresentazione strutturale che costituisce un certo soggetto cognitivo può svolgere la sua funzione entro un sistema di conoscenza solo qualora esista un meccanismo in grado di produrre un'unica particolare informazione "che la rende esplicita in quanto ente complesso", e che le deve essere associata in maniera univoca. Solo in questo modo, infatti, può essere attuata la funzione di riconoscimento. Si tratta dell'applicazione del **principio di convergenza delle verifiche** enunciato nel capitolo precedente: "per rendere esplicita una certa struttura, o una certa proprietà strutturale, all'interno di un sistema cognitivo, è sempre comunque necessario far convergere verso un'unica informazione tutte le verifiche della presenza degli elementi che la definiscono".

Si noti che questa proprietà svolge, nell'ambiente cognitivo, una **funzione equivalente al criterio di emergenza nella realtà esterna**. In effetti, secondo questo criterio una certa struttura (o una proprietà strutturale) è emergente qualora sia in grado, con la sua presenza (quindi con il complesso di tutto ciò che la definisce), di generare una dipendenza funzionale con altre strutture. In modo analogo una certa rappresentazione svolge la sua funzione di soggetto cognitivo, entro un sistema di conoscenza, quando è in grado di far accadere qualcosa a livello di elaborazione interna.

La certificazione della presenza di rappresentazioni di strutture e proprietà strutturali emergenti è, come detto, un passaggio obbligato nell'elaborazione cognitiva dell'informazione. Questo è vero perché, se e solo se si certifica la presenza di questi "oggetti", diviene allora possibile la decodifica delle regole emergenti. L'insieme delle rappresentazioni delle strutture, delle proprietà strutturali e delle regole emergenti, costituisce la base portante di ogni attività cognitiva. Un sistema cognitivo è fatto di queste cose!

L'operazione di rendere esplicita, con un'informazione univoca, qualcosa che è di per se stessa un'entità complessa (nel senso di composta da una pluralità di parti), è dunque cruciale e onnipresente nell'attività cognitiva.

Pongo l'accento sui seguenti punti:

- Molto spesso l'informazione da segnalare è selezionata entro la rappresentazione di una situazione più ampia dove compare mescolata a molte altre. Si tratta quindi di un processo di evidenziazione e separazione di informazione.
- Queste informazioni sono di norma riutilizzabili, spesso potenzialmente all'infinito, quindi ha senso memorizzarle come "soggetti" a sé stanti.
- Nell'implementazione pratica dei processi di elaborazione d'informazione è utile (in pratica necessario) predisporre dei dispositivi (che non necessariamente devono essere fisici) che si occupano proprio di riconoscere nei dati che ricevono in input, e di segnalare al resto del sistema cognitivo, l'avvenuto riconoscimento (o la chiamata in causa) della presenza di quella specifica informazione complessa. Secondo i casi, questi dispositivi risponderanno quindi a specifiche strutture o a specifici schemi (che dovranno avere memorizzato a loro interno), a specifiche proprietà e relazioni strutturali (e quindi dovranno essere in grado di computare gli algoritmi necessari alla loro identificazione), oppure a specifici morfismi (e saranno quindi in grado di riconoscere intere categorie di strutture o di proprietà strutturali, o entrambe le cose). Potranno anche corrispondere a classificazioni che vanno oltre il contenuto strutturale interno. L'informazione specifica che il singolo dispositivo è in grado di selezionare, riconoscere e segnalare, costituisce il "**soggetto**" della sua specifica attività.

### 6.7 Ordinamento gerarchico dei soggetti cognitivi

I soggetti cognitivi si prestano in modo naturale a un ordinamento gerarchico. Possono essere soggetti cognitivi legittimi anche insiemi, o meglio particolari **"insiemi strutturati",** di altri soggetti. I primi si possono considerare allora di "**livello gerarchico superiore**" rispetto ai secondi.

Sono spesso degli utili soggetti cognitivi quelli che costituiscono un'**astrazione** e/o una **generalizzazione** di altri di livello gerarchico inferiore. Spesso le operazioni di astrazione e generalizzazione si eseguono attraverso delle operazioni di derivazione strutturale, in particolare con operazioni di quoziente e di morfismo.

Le operazioni di quoziente cambiano "la scala" alla quale si rappresenta la struttura di partenza, mentre le operazioni di morfismo inibiscono parte di ciò che rende distinguibili le sue parti componenti.

Credo che la gerarchia dei soggetti cognitivi costituisca un principio di ordinamento naturale per l'informazione all'interno di un sistema cognitivo. Per forza di cose, per come l'informazione cognitiva è definita, è necessario passare attraverso quest'ordinamento.

In realtà penso esistano almeno due principi generali di ordinamento: il primo è stabilito sostanzialmente dalle modalità con le quali i singoli soggetti possono essere riconosciuti eseguendo l'analisi delle "informazioni sensoriali" e ciò che ad essa segue; il secondo dipende dalla "logica di utilizzo" di queste informazioni.

In generale l'informazione all'interno di un sistema cognitivo è utile se permette di utilizzare delle regole. Credo che le regole, a loro volta, permettono sostanzialmente di fare due cose: prevedere l'evoluzione della realtà e pianificare in modo finalizzato le azioni che si possono compiere.

Ma le informazioni devono anche essere prima riconosciute, e non è detto che le caratteristiche strutturali che ne permettono il riconoscimento siano le stesse che determinano la "logica", vale a dire il complesso delle regole utili da associare all'informazione individuata.

Per questo motivo può accadere che i principi di ordinamento gerarchico seguano almeno due direzioni distinte: una di analisi strutturale per il riconoscimento delle informazioni, e un'altra per l'implementazione funzionale delle regole.

In questo capitolo ho puntato l'attenzione sul fatto che l'informazione deve essere opportunamente suddivisa e classificata. L'idea è che ogni suddivisione di informazione che goda di una qualche forma di autonomia, costituisce di per sé un "soggetto" che può essere "degno di attenzione" da parte di qualche processo cognitivo. Buona parte di questi soggetti sono reciprocamente legati: sia attraverso delle regole che permettono il loro utilizzo, sia da rapporti gerarchici.

### 6.8 Dalle rappresentazioni strutturali ai concetti

Per come sono stati definiti, alla categoria dei soggetti cognitivi appartengono cose molto semplici come lo sono molti elementi strutturali che corrispondono a dettagli, per esempio: la classificazione della curvatura delle linee di bordo, la presenza di spigoli, la presenza di una concavità, la sua direzione nello spazio, ecc.. Nel contempo sono incluse le rappresentazioni complete degli oggetti reali, nella loro forma spaziale tridimensionale; sono inoltre considerabili soggetti cognitivi anche molte rappresentazioni molto più astratte e generali che assomigliano fortemente ai nostri **concetti**.

Un oggetto concreto, realmente esistente, come una sedia o un tavolo, ha sempre ovviamente una struttura particolare. Ma questa struttura non costituisce la nostra rappresentazione del concetto generale di sedia o di tavolo. Cosa sono allora i concetti?

L'idea che propongo è che tutte le cose che sono sedie o tavoli hanno delle caratteristiche in comune, che possono essere sia proprietà strutturali sia, e questo è un punto molto importante, delle proprietà funzionali che risultano definite dal complesso delle regole ad essi applicabili.

Sulla base di quanto abbiamo visto finora, possiamo avanzare delle ipotesi sulle proprietà strutturali comuni che devono avere (quasi) tutte le istanze specifiche di un soggetto concreto, come una sedia. Possiamo infatti dire che, "molto spesso", per avere una sedia, ci deve essere un piano orizzontale di forma grossomodo quadrata, il quale deve avere dimensioni contenute entro certi limiti, non deve essere troppo lungo o troppo largo (per distingue ad esempio la sedia da una panchina), e deve trovarsi ad una certa altezza da terra. Ci devono inoltre essere in genere, ma non necessariamente, quattro gambe e uno schienale, posto anche questo ad una certa altezza, e che forma con la seduta un angolo vicino alla perpendicolarità, ecc…

Utilizzando una logica sfumata possiamo pensare di mettere assieme tutte queste proprietà strutturali e utilizzarle per proporre una prima definizione di qualcosa che si avvicina al concetto di sedia. Se si riflette sulla questione non è difficile rendersi conto che una costruzione di questo tipo del "concetto sedia" mostra una certa limitata funzionalità, ma in molte situazioni non risulta applicabile. La questione è complessa e dipende dal fatto che le nostre concettualizzazioni sono formulate non solo in base alle proprietà strutturali che gli oggetti concreti hanno, ma anche in funzione delle implicazioni che possono avere, e in particolare dall'uso che se ne può fare. I soggetti cognitivi sono spesso definiti in base alle loro "**proprietà funzionali**".

Credo che le funzioni che un certo oggetto può assumere siano associabili alle rappresentazioni dello stesso, utilizzando parte del complesso delle regole che lo stesso contribuisce a definire. Ritengo comunque che le proprietà strutturali siano **essenziali per poter riconoscere tutte le altre.** Cerco di spiegare meglio questa idea.

Quando osserviamo una certa cosa, o un certo fenomeno, i nostri sensi ci forniscono essenzialmente solo delle informazioni sulle proprietà strutturali dell'oggetto. Sono però le nostre conoscenze preesistenti del mondo e soprattutto delle sue regole, che ci permettono di associare a queste informazioni strutturali anche le rappresentazioni delle proprietà funzionali dell'oggetto. Le proprietà funzionali dipendono dalle regole che sono associate all'oggetto e che spesso riguardano le "azioni" e le "attività", che con esso si possono compiere. In certi contesti possiamo chiamare "sedia" una cosa che possiede proprietà strutturali molto distanti da quelle di una "sedia standard". Se ci troviamo all'aperto, in mancanza di meglio possiamo chiamare sedia anche una pietra, qualora questa possa svolgere questa determinata funzione.

Una cosa interessante da notare è che le capacità di concettualizzare dipendendo fortemente dalla nostre conoscenze del mondo, e per tale motivo esse sono soggette ad evolversi nel tempo. Svariati studi sullo sviluppo cognitivo dei bambini, a partire da quelli di Piaget, mostrano gli aspetti salienti dell'evoluzione della capacità di concettualizzare durante l'apprendimento. Per un bimbo, che è un esploratore cognitivo del mondo alle prime armi, i concetti sono formulati in modo semplice. Per un sistema cognitivo "alle prime armi", il concetto di sedia, costruito in base al riconoscimento delle sole proprietà strutturali, potrà essere un buon primo prototipo di riferimento.

Un'idea importante è che concetti che riguardano astrazioni di comportamenti che sembrano difficili da definire in termini puramente computazionali, potrebbero essere implementati, e quindi definiti, sulla base di operazioni di classificazione e di astrazione di schemi comportamentali standard, nonché delle stesse operazioni di "gestione interna" dei processi cognitivi. Semplificando un po' le cose credo sia corretto pensare all'attività cognitiva come essenzialmente proiettata all'azione e che esista il modo di rappresentare, in maniera computazionale, i passaggi obbligati di ogni attività di pianificazione della azioni che sono destinate ad uno scopo. Queste rappresentazioni consistono in **schemi** che rappresentano i **comportamenti**. Sono quindi, per l'appunto, degli "**schemi comportamentali**". Per gli schemi, come detto, valgono molte delle cose viste per le strutture di prima specie. Anche sugli schemi si possono eseguire operazioni di derivazione. In questo modo si possono ricavare astrazioni degli schemi comportamentali di base. Un'idea importante è che anche concetti astratti e apparentemente difficili da "imbrigliare" con precisione, come ad esempio quello di "impedimento", o quello di "vincolo", possono in realtà essere definiti con opportune operazioni di analisi a partire proprio dalla generalizzazione e astrazione di schemi che rappresentano la "gestione delle fasi salienti" degli stessi processi cognitivi.

Un altro concetto, sul quale ritornerò sul prossimo capitolo, riguarda il ruolo "motivazionale" che particolari soggetti cognitivi possono assumere. Penso che vari "**obiettivi da raggiungere**" e "**pericoli da evitare**" possano essere formulati in termini di soggetti cognitivi ai quali è attribuita una speciale funzione di "desiderabilità" (che può essere positiva o negativa). Questa funzione potrebbe fornire un modello per comprendere come finalizzare l'attività cognitiva e renderla utile indirizzandola verso degli scopi.

Un obiettivo da raggiungere potrebbe consistere in una situazione caratterizzata dalla presenza di specifici soggetti cognitivi (spesso in specifiche relazioni), che è valutata, per qualche motivo, come desiderabile. Parallelamente si possono definire delle situazioni indesiderabili. È probabile che se queste situazioni sono particolarmente importanti, possano essere classificate opportunamente, e costituire a loro volta dei singoli soggetti cognitivi a sé stanti di medio o alto livello di astrazione.

Prima di passare al capitolo successivo penso sia utile far notare che quanto visto per i soggetti cognitivi, e in particolare:

- la necessità di produrre una singola informazione elementare che

segnala e certifica la loro presenza nella situazione rappresentata,
- il fatto che queste singole informazioni possono essere usate per codificare altri soggetti cognitivi di livello gerarchico superiore e per codificare regole valide, implichi che all'interno di un sistema cognitivo l'informazione dovrebbe essere organizzata secondo una struttura a **reti. I nodi** di queste reti potrebbero corrispondere alle informazioni che esplicitano il riconoscimento dei singoli soggetti cognitivi.

A questo punto mi sembra naturale segnalare che la funzione di questi nodi potrebbe essere svolta, nel nostro cervello, da singoli neuroni della neocorteccia, o, più probabilmente, da gruppi di questi che costituiscono dei singoli moduli funzionali.

### 6.9  Insiemi strutturati di soggetti cognitivi

Un altro concetto importante è che si possono mettere assieme più soggetti cognitivi per costruire rappresentazioni (talvolta molto "compatte") di singole situazioni. Per questo motivo può essere utile il concetto di "insieme strutturato di soggetti cognitivi". Questa costruzione sembra utile perché spesso molte singole regole sono sensibili proprio a queste rappresentazioni compatte, quindi proprio a questi insiemi (strutturati) di soggetti cognitivi, che possono essere anche molto astratti.

Ritengo che anche queste rappresentazioni siano sostanzialmente di tipo strutturale, anche quando i soggetti specifici che vi compaiono servono per esplicitare classificazioni alquanto astratte (che non dipendono dal "contenuto informativo interno"); si pensi all'esempio già citato del concetto di veicolo.

Anche in queste casi si può continuare ad utilizzare la strategia di distinguere tra le "proprietà interne" e le "relazioni esterne" dei soggetti che fanno parte della rappresentazione. Si può quindi continuare ad utilizzare la metodologia generale che consiste nello specificare:
- quali soggetti appartengono alla rappresentazione (l'equivalente dell'insieme delle parti),
- quali sono le loro proprietà interne,
- e quali sono le loro relazioni esterne.

Si tenga presente che i vari soggetti che possono essere messi assieme per costruire queste "rappresentazioni compatte", in realtà compaiono insieme a moltissimi altri. Si tratta di quelli che sono stati riconosciuti, o attivati per altre vie, da poco tempo e che fanno ancora parte "del presente". Inoltre, punto fondamentale, al di sotto di questi soggetti astratti devono essere presenti (in modo attivo) molte altre rappresentazioni di livello gerarchico inferiore, e quindi molto più concrete. In generale i soggetti cognitivi astratti, che possiamo mettere assieme per costruire le nostre rappresentazioni compatte di alto livello, sono attivi proprio perché stanno esplicitando proprietà che sono vere per le rappresentazioni più concrete sottostanti. Senza questo substrato molte rappresentazioni astratte non potrebbero esistere!

In casi particolari è pensabile, ad esempio nell'interpretazione di un messaggio linguistico, attivare direttamente rappresentazioni astratte, senza che queste siano state chiamate in causa da altre di livello inferiore. Potrebbe però essere vero che anche in questi casi torni utile, e sia talvolta necessario, generare delle rappresentazioni di più basso livello (talvolta molto schematiche) per dare supporto a quelle più astratte.

Quindi in generale, se si "sceglie" di mettere assieme dei soggetti molto astratti per rappresentare una certa situazione, saranno in genere disponibili le informazioni che consentono di attribuirgli delle proprietà interne, e di descrivere le loro relazioni esterne.

Teniamo presente che anche le relazioni esterne che possono intercorrere tra i soggetti "protagonisti" che, nell'insieme strutturato, costituiscono le parti della nostra rappresentazione, sono a loro volta soggetti cognitivi. Esse possono essere alquanto semplici, come la specificazione di adiacenze (che sono descrivibili come ramo di un grafo). Possono consistere in relazioni spaziali e temporali, che corrispondono a concetti quali: essere sopra, sotto, di lato, accanto, vicino, lontano, contemporaneo, ecc… Possono essere inoltre relazioni di tipo logico, come l'essere in rapporto di causa ed effetto, l'essere una condizione necessaria, una condizioni sufficiente, costituire un impedimento….

**Alcuni approfondimenti**

**6.10 Sui concetti di feature e di pattern recognition**

Attualmente, nel campo degli studi sul machine learning (apprendimento automatico) sono utilizzati i concetti di "feature" (caratteristica), di trasformata nello "spazio delle features", e di pattern recognition.

Questi concetti sono indubbiamente utili e sono molte le applicazioni che dimostrano la loro funzionalità. Ciò nonostante, a mio parere, essi soffrono della mancanza di un inquadramento teorico completo che sia in grado di spiegarne pienamente il senso e la funzione. Non mi sembra che questo sia disponibile in letteratura. Mi ha colpito che alcuni ricercatori facciano riferimento alla definizione proposta nel 1985 dal fisico giapponese Watanabe: egli definisce un pattern come "l'opposto del caos" e come "un'entità vagamente definita alla quale può essere assegnato un nome".

Secondo il mio modo di vedere, le features sono sostanzialmente delle caratteristiche strutturali. Il concetto di caratteristica strutturale può essere pensato come una generalizzazione e un'estensione di quello di struttura derivata e di proprietà/relazione strutturale non autonoma. In molte situazioni le operazioni di estrazioni delle features e dei pattern possono essere assimilate a ciò che in questo lavoro chiamo analisi di struttura. In effetti, analizzando cosa si fa nel concreto in molte attività di pattern recognition, si può vedere che quando si estraggono le features, si parte da una struttura di partenza e si rendono esplicite "entità" che corrispondono a porzioni di questa, alle loro proprietà interne e a loro relazioni esterne. Molti pattern corrispondono quindi a strutture quozienti e ai loro eventuali morfismi.

Chiaramente, in questo lavoro preferisco utilizzare una terminologia diversa rispetto a quella "standard", e invece di parlare in termini di features e di pattern, preferisco, quando possibile, usare i concetti di porzione, quoziente, morfismo, ma anche quello di "proprietà" o "caratteristica strutturale" non autonoma. Mi sembra che la terminologia che propongo sia più accurata, ma non possono escludere che esistano situazioni nelle quali concetti più "elastici" come quelli di features e di pattern, possano essere più efficaci, proprio perché meno vincolanti.

Penso che per le features valgano i concetti espressi per ogni proprietà strutturale che può essere resa esplicita, giacché sono "quasi" la stessa cosa; quindi, tra l'altro, penso che la loro esplicitazione abbia senso qualora esse possano contribuire a implementare una qualche regola utile.

In molti studi, e in varie realizzazioni pratiche, si utilizzano degli oggetti matematici chiamati "spazi delle features". Sono stati sviluppati dei metodi interessanti e molto potenti per alcuni problemi di separazione "di porzioni" di questi spazi, che permettono di identificare "entità di scala superiore": "i pattern". In questo senso i "pattern" corrispondono a speciali collezioni di sottoinsiemi di punti nei relativi spazi delle features.

Alcune features sono esplicitabili tramite solo delle informazioni binarie, mentre ad altre può essere associato un numero. Ad esempio una features "quantificabile" importante può essere la lunghezza massima di un oggetto. Mettendo assieme più features numeriche si ottiene una varietà continua, vale a dire uno spazio. Quest'oggetto matematico è però particolare: per esso non si possono usare gli stessi concetti e gli stressi strumenti applicabili a un normale spazio lineare. La distanza tra due suoi punti serve sostanzialmente a valutare la loro "distinguibilità relativa". Le metriche di questi spazi dipendono dal pattern specifico che esse permettono di definire. Due punti possono essere "assolutamente distanti", nel senso di completamente distinguibili, per un certo pattern, e avere invece "distanza nulla" per un altro.

Possiamo pensare agli spazi delle features anche come a degli "spazi delle condizioni" che possono essere necessarie e/o sufficienti, anche in modo molto complesso, per identificare un soggetto cognitivo di livello superiore (il pattern).

## 6.11 Codificare soggetti di scala superiore

Quanto appena visto ci introduce alla problematica di comprendere come si possono codificare soggetti di scala superiore partendo da quelli di livello inferiore. Credo che il problema possa essere diviso in due parti: individuare quali operazioni di codifica devono essere eseguite e quali sono i criteri per stabilire se quanto codificato è, o non è, un legittimo soggetto cognitivo.

Per il secondo punto credo di avere già indicando i criteri generali: sono soggetti cognitivi legittimi quelli che contribuiscono a identificare regolarità e regole utili. Il problema è che spesso l'utilità di una regola può essere stabilita solo a posteriori, provando ad usarla e verificando che funzioni. Questo comporta che spesso la "supervisione", che certifica la validità di una features o di un pattern o di qualunque soggetto cognitivo, può avvenire **solo dopo vari passaggi**. Ciò contribuisce a rendere i problemi di apprendimento intrinsecamente difficili. Per questo motivo in genere non è possibile stabilire a priori come devono essere implementate nel dettaglio le singole operazioni di codifica. Spesso ciò che possiamo dire a priori si esaurisce a come dovrebbero essere le "forme tipiche" che queste operazioni possono assumere.

Per generalizzare considero "i pattern" come dei soggetti cognitivi di livello "N" che sono definiti tramite un certo numero di altri di livello "N-i": "le features".

Un soggetto cognitivo di livello N-i può, per quanto riguarda il suo contributo alla codifica di uno di livello N, risultare:
- una condizione sufficiente, ma non necessaria;
- una condizione necessaria, ma non sufficiente;
- costituire un contributo che viene computato con una particolare funzione. Quest'ultima può essere anche alquanto complessa.

Se il soggetto N-i risulta essere necessario, ma non sufficiente, alla identificazione del soggetto N, significa che l'informazione sulla sua presenza va computata come OR logico insieme ad altri. Esso quindi è sufficiente a individuare il soggetto di livello N, ma non in maniera esclusiva, poiché vi possono essere anche altri soggetti di livello N-i che sono in grado di fare altrettanto. Il soggetto di scala superiore è quindi, in questo caso, una generalizzazione di quelli inferiori.

Se N-i è invece necessario, ma non sufficiente, significa che l'informazione sulla sua presenza va computato in AND con altre informazioni (che possono, a loro volta, essere il risultato di altre funzioni computate su altre features).

È interessante notare che sia le operazioni di AND che quelle di OR possono essere realizzate con lo stesso procedimento matematico: eseguendo un prodotto scalare delle variabili in input con un vettore di pesi appropriati, seguito da una funzione che fornisce 0 o 1, a seconda che il risultato del prodotto sia minore o maggiore di una certa soglia prefissata. Se i valori in input sono binari e se i pesi sono tutti a 1, allora si ottiene un operazione di OR se si utilizza come soglia la condizione che il risultato del prodotto sia solo maggiore di 1. Si ottiene invece un'operazione di AND se si richiede che il risultato del prodotto sia uguale all'intero corrispondente al numero degli input.

Variando la soglia e i pesi si ottengono "operazioni intermedie" che assomigliano tanto più a un operazione di OR quanto la soglia è bassa (ma maggiore o uguale a 1…), e si avvicinano invece ad una di AND quando la soglia è alta.

Se al posto di una funzione di soglia discriminante se ne utilizza una continua, si possono computare moltissime altre funzioni. È interessante notare che queste operazioni sono proprio quelle eseguite dai singoli dispositivi di una tipica rete neurale.

Operazioni di questo tipo si possono comporre potenzialmente in un numero infinito di varianti, e possono essere usate anche per costruire funzioni in grado di separare le popolazioni entro gli spazi delle features. Il lavoro difficile consiste nella messa appunto delle funzioni corrette. A questo fine sono stati condotti molti studi e sviluppati interessanti metodi, come quelli che si basano sulle tecniche Support Vector Machines (SVM), o come i molti che si basano su reti neurali. Il punto critico è che molti di questi metodi funzionano nelle condizioni che l'apprendimento sia "supervisionato", quindi quando esiste il modo di stabilire a "priori", almeno per un certo numero di casi, che ciò che si ottiene è la risposta giusta; ciò ci rimanda al problema della "verifica" che, come detto, penso possa in realtà esser fatta solo a posteriori.

### 6.12 Sull'apprendimento in profondità e la stratificazione delle rappresentazioni

Con gli studi sull'apprendimento profondo (deep learning) ci si è resi conto che è necessario rappresentare la realtà per stratificazioni gerarchiche. Gli studi in questo campo sono molto importanti e gli strumenti che si stanno sviluppando appaiono molto promettenti. Credo che l'idea di rappresentare la realtà per gerarchie di astrazioni di livello via via crescente sia vincente. Tuttavia, a mio parere, si sta procedendo senza una visione chiara dei principi e delle motivazioni profonde che stanno alla base di questo approccio. Ritengo che alcuni dei problemi di machine learning che oggi appaiono particolarmente difficili, possano essere affrontati in modo più semplice con una comprensione più chiara del senso dei vari passaggi.

Credo di poter mostrare che il processo complessivo di costruzione delle varie gerarchie di rappresentazioni deve seguire "diverse fasi" dove prevalgono, di volta in volta, obiettivi differenti. Nei prossimi capitoli descriverò alcune di queste fasi. Tra le altre cose, vedremo che si possono distinguere fasi dove si analizzano gli stimoli sensoriali primari, che sono intrinsecamente informazioni strutturali di "cattiva qualità", da altre dove si procede con operazioni di analisi di struttura condotte già su "buone ricostruzioni" di reali strutture emergenti. Vedremo anche che si possono distinguere le operazioni di astrazione in "interne" e "esterne", in funzione che ci si limiti ad analisi che sfruttano solo il "contenuto strutturale interno", o che invece si usino anche informazioni che dipendono dai "ruoli funzionali" che i vari soggetti cognitivi possono assumere. Vedremo inoltre che questi ruoli dipendono da quali "obiettivi" si perseguono, e che per le astrazioni più spinte è necessario costruire rappresentazioni che si riferiscono alla "gestione globale" degli stessi processi cognitivi.

Vedremo che entro alcune di queste singole fasi si presentano effettivamente problemi che richiedono l'utilizzo di tecniche di apprendimento profondo. Questo accade proprio perché, come accennato nel paragrafo precedente, spesso si deve passare attraverso più gerarchie di soggetti prima di poter verificare la loro "legittimità". Come detto, spesso il test finale della correttezza di una sequenza di soggetti cognitivi può avvenire solo verificando che essi permettano di codificare regole che funzionano davvero. Per eseguire questa verifica è necessario fornire al sistema la possibilità concreta di "**testare la validità delle regole**" che genera.

Uno dei concetti centrali è che nell'apprendimento è molto importante legare fin da subito i vari processi di "pattern recognition", con quelli di codifica di regole valide. È importante che le attività di analisi e quelle di implementazione e test delle regole avvengano in sinergia.

# 7 Alcuni approfondimenti sulle regolarità e sulle regole

## 7.1 Introduzione

La seconda congettura di riferimento, la quale afferma che ogni regolarità consiste sempre in coincidenze strutturali, appare molto promettente. Essa sembra funzionare da principio di riferimento in grado di risolvere una parte importante delle problematiche che si incontrano nella scienza cognitiva, in intelligenza artificiale e in varie altre discipline. In modo particolare essa si dimostra molto potente quando è combinata con le idee illustrate sul fenomeno dell'emergenza, quindi con le idee di struttura e di logica emergente, con le operazioni di derivazione strutturale (vale a dire con i concetti di struttura derivata e di proprietà strutturale), e con il concetto di soggetto cognitivo.

Una delle idee fondamentali è che partendo dal basso, vale a dire dalle rappresentazioni dei sistemi di computo strutturale di base, sia possibile procedere in una serie di operazioni di derivazione strutturale capaci di mettere in luce la comparsa di logiche emergenti. Queste ultime consistono in complessi di regole affidabili che coinvolgono strutture e schemi derivabili da quelli di base. È intrigante l'ipotesi che siano queste logiche emergenti a costituire il patrimonio più importante per l'attività cognitiva.

Un aspetto molto importante di questo fenomeno è che procedendo verso l'alto, lungo il processo di derivazione strutturale, compaiono regole, e complessi di regole, che sono tutt'altro che facilmente deducibili da quelle di base.

I processi di derivazione strutturale consentono di stratificare le varie rappresentazioni del mondo, producendo diversi livelli di rappresentazione. Man mano che si procede dal basso verso l'alto, si passa spesso da rappresentazioni che sono molto contestualizzate, e per le quali sono applicabili insiemi di regole di validità solo locale, a rappresentazioni di "entità astratte" che sono molto più generali e consentono l'individuazione di regole che, pur essendo applicabili solo ad alto livello, sono spesso molto più generali e potenti. Un processo di derivazione di rappresentazioni strutturali che procede dal basso verso l'alto, accompagnato da processi che cercano di individuare le regolarità che si presentano ai vari livelli, sarà soggetto naturalmente anche a trovare le analogie che spesso si manifestano in situazioni differenti. Due o più situazioni sono "analoghe" quando dalle due è possibile estrarre una struttura o uno schema comune. Sospetto che questo avvenga in maniera automatica quando si procede a una descrizione della situazione a un livello più alto. In una parte importante dei casi ciò avviene perché nella descrizione di alto livello si costruiscono, in maniera legittima, delle rappresentazioni che, di fatto, **inibiscono una serie di differenze che sono presenti in quelle di livello inferiore.**

Altro aspetto fondamentale è che questo modo di procedere per derivazioni successive, partendo da rappresentazioni concrete per generarne altre più astratte e generali, è imposto in maniera naturale dallo stesso manifestarsi di regole emergenti utilizzabili. Sono queste, infatti, che danno senso alle operazioni di derivazione. Non sarebbe di alcuna utilità procedere con un'operazione di quoziente per individuare una certa astrazione, se questa non fosse soggetta a qualche regolarità che rende tale rappresentazione riutilizzabile al fine di poter con essa compiere, in ultima analisi, delle previsioni ed inferenze utili.

## 7.2 Regole e regolarità

In generale possiamo dire che, dal punto di vista dell'attività cognitiva, il concetto di **regolarità** sottende a un approccio **passivo**: le regolarità sono fenomeni che accadono nella realtà; esse, in un certo senso, sono semplicemente osservate da un sistema cognitivo senza che ci sia un approccio "attivo" da parte del sistema stesso. Si manifestano, come detto, sotto forma di qualche tipo di coincidenza strutturale.
Diversamente le **regole** implicano un ruolo **attivo** da parte di un sistema cognitivo: una regola è qualcosa che dice come si deve procedere (o che risultati si devono ottenere).

## 7.3 Le regole utili vincolano ma non troppo

Penso che una buona definizione operativa del concetto di regola, quando applicata al dominio delle rappresentazioni cognitive, è quella che si appoggia al concetto di vincolo.
In generale possiamo dire che una regola è costituita da uno o più "vincoli" che devono essere rispettati. Questi vincoli sono esprimibili, in ultima analisi, in termini di coincidenza tra strutture, o tra schemi, che devono comparire in qualche punto dell'attività di elaborazione che il sistema cognitivo esegue (anche se talvolta tali coincidenze sono parziali, il concetto resta valido). Questa definizione quindi è ben compatibile con la seconda congettura di riferimento.
Come già affermato, credo la realtà sia comprensibile, e quindi conoscibile, nella misura in cui essa è soggetta a regolarità.
**A livello emergente, e solo in esso**, benché queste regolarità costituiscano dei vincoli, nello stesso tempo permettono spesso una certa libertà di scelta di azione: quindi, a livello emergente, **le regole vincolano ma non troppo!**
Questo è un punto assai importante perché se non ci fosse questa libertà di "scegliere come agire" la stessa attività cognitiva non avrebbe senso!
Le nostre rappresentazioni interne della realtà sono utili se ci consentono di fare delle previsioni su come può evolvere la situazione in funzione delle nostre azioni possibili. Noi possiamo eseguire delle simulazioni interne dei possibili scenari e in funzione di queste "scegliere" cosa ci conviene fare. Possiamo "decidere" quali azioni è utile intraprendere affinché la realtà evolva verso quella che giudichiamo essere "la migliore tra le situazioni che ci sembra possibile ottenere". Se, per assurdo, a ogni livello di emergenza, la realtà fosse soggetta a regole "totalmente vincolanti", che non lasciano libertà di azione, quale utilità avrebbe la capacità di conoscere? È facile convenire che in una simile situazione le facoltà intellettive umane non si sarebbero mai evolute poiché non avrebbero offerto alcun vantaggio.

Un fatto sorprendente è che questa libertà di scelta sembra manifestarsi solo a livello emergente, mentre a livello di fisica di base, almeno secondo la concezione "classica" della fisica, le cose appaiono molto diverse. In effetti, a questo livello le regole sono strettamente e totalmente vincolanti, non sembra esserci alcuna libertà di scelta!

Questo fatto, per certi versi paradossale, è solo uno degli aspetti che differenziano le regole che valgono ai livelli di base rispetto a quelle che si manifestano ai livelli emergenti.

È dunque molto importante distinguere tra **regole di base** e **regole emergenti**.
In generale, le regole dei livelli di base manifestano queste interessanti proprietà:
- Sono totalmente vincolanti.
- Sono spesso "subite", le persone sono passive di fronte ad esse, sono quindi fenomeni che succederanno indipendentemente dalla nostra volontà!
- Sono **regole operazionali.**
- Non possono computare in negativo, nel senso che non possono essere definite "sulla mancanza di qualcosa".
- Sono regole certe, che funzionano sempre.

Solo procedendo dal basso verso l'alto lungo i processi di derivazione strutturale e di astrazione, compaiono regole emergenti che hanno proprietà diverse:
- Spesso sono solo parzialmente vincolanti.
- Noi possiamo avere un ruolo attivo di scelta!
- Possono essere formulate in maniera "**associativa**" e non solo operazionale.
- Posso computare anche in negativo, ovvero sulla mancanza di qualcosa.
- Ci possono essere regole utili che non sono "sicure al 100%", ma che hanno solo una certa **probabilità** di funzionare.

### 7.4 Regole procedurali e regole vincolanti nei risultati

In generale possiamo dire che una regola consiste nell'esistenza di alcuni vincoli, che possono essere sempre espressi, in ultima analisi, in termini di coincidenza strutturale (o anche di "somiglianze" strutturali).
Secondo i casi, il vincolo può consistere o nel modo nel quale si deve procedere, o nei risultati che si devono ottenere.
Se i vincoli che determinano la regola agiscono a livello di "cosa si deve fare", possiamo dire di essere in presenza di una **regola procedurale**. In questo caso è quindi il modo con il quale si deve procedere ad essere in qualche misura vincolato, e, a seconda dei casi, può esserlo totalmente o solo parzialmente.
Diversamente ci possono essere situazioni nelle quali è lasciata libertà di azione sul come affrontare un certo problema, ma il vincolo che viene imposto riguarda i risultati da ottenere. In questo caso possiamo quindi parlare **di regola vincolante sui risultati**.

### 7.5 Le regole della fisica e dei sistemi strettamente deterministici

In fisica lo stato istantaneo di un sistema può essere rappresentato in termini strutturali, o più precisamente con funzioni a molte variabili che possono essere interpretate, per estensione, come strutture continue. L'evoluzione del sistema può essere calcolata tramite degli operatori, più precisamente tramite degli operatori differenziali, che a loro volta possono, alla fin fine, essere espressi in termini di schemi che rappresentano la successione delle operazioni differenziali elementari da compiersi (che a rigore andrebbero applicate sul continuo, su grandezze infinitesime per un numero infinito di volte). In genere un operatore differenziale consiste in un'espressione che contiene, tra gli altri, operatori di derivazione e integrazione. Con questi possiamo scrivere delle equazioni che consentono di calcolare, in linea di principio, come il valore di una singola variabile si evolve nel tempo.
Queste idee possono essere estese ad ogni sistema che può essere, almeno potenzialmente, rappresentato in maniera completa ed esaustiva in termini di computo strutturale. In questi casi lo stato istantaneo del sistema è rappresentato da strutture, e la sua evoluzione può essere computata da operatori.
In generale, le rappresentazioni strutturali e gli operatori esauriscono tutta l'informazione necessaria per determinare l'evoluzione di un sistema strettamente deterministico.
Possono esistere sistemi strettamente deterministici che sono emergenti rispetto al substrato fisico sottostante. Anzi molto spesso questi sistemi possono essere implementati su substrati alquanto differenti. Ad esempio un circuito logico digitale può essere realizzato in molti modi differenti: con dispositivi elettronici, con congegni meccanici o idraulici, e in moltissime altre maniere. Le evoluzioni temporali di questi sistemi sono governate da regole emergenti.

Per i sistemi strettamente deterministici le regole da applicare sono totalmente vincolanti, non lasciano quindi alcuna possibilità di scelta.
In questi casi i vincoli determinano come si deve procedere, quali operazioni sono da compiere. Abbiamo quindi a che fare con regole di tipo procedurale.

### 7.6 Regole operazionali e regole associative

Un altro importante modo per classificare le regole è quello che si basa sulla distinzione tra **regole operazionali** e **regole associative**.
Per **regole operazionali** intendo sostanzialmente tutte quelle che sono implementate tramite funzioni che **calcolano** il proprio valore di output, e che non si basano principalmente su "associazione già memorizzate" tra variabili di ingresso e risultati da produrre in uscita. Le regole operazionali possono essere "ben definite" in maniera rigorosa e precisa, a partire dalle operazioni di computo strutturale di base, per arrivare, tramite combinazioni di queste, alle operazioni matematiche sofisticate. Ricordo che, come mostrato nel capitolo 2, comprendono anche quelle che possono essere considerate la generalizzazione, nell'ambito della teoria strutturale, delle operazioni matematiche classiche. Sono regole operazionali quelle che utilizzano equazioni, ma rientrano nella categoria anche molti algoritmi anche complessi.
In genere, alle regole operazionali è possibile associare un **singolo operatore**.
Le **regole operazionali** consistono dunque in una successione ben definita di operazioni di computo strutturale. Tale successione di operazioni è un algoritmo e costituisce "lo schema dell'operatore" da utilizzare. Le regole operazionali in generale agiscono su strutture di prima specie producendo altre strutture di prima specie. In taluni casi possono agire anche sulla struttura di operatori producendo altri operatori, ma anche in questi casi si tratta sempre di azioni riconducibili ad operazioni effettuate su strutture. Quando si utilizza una regola operazionale deve essere specificato lo stato di partenza, quindi l'informazione che lo definisce, che è costituita dalla rappresentazione della struttura di partenza e dell'operatore che viene applicato. Abbiamo una situazione di questo tipo:
- Stato iniziale costituito da: struttura di partenza A e schema dell'operatore Op da applicare. Quindi si applica Op(A).
- Stato finale costituito dalla struttura B che si ottiene: ( B = Op(A) )

In generale è ovvio che per utilizzare una regola operazionale è necessario conoscere la sequenza delle operazioni da compiere. In altre parole deve essere rappresentato lo schema che definisce l'algoritmo da utilizzare; si deve inoltre disporre dei congegni fisici che siano in grado di espletare le singole operazioni di computo strutturale.

Ha senso però chiedersi se esistono altri possibili modi di procedere. In particolare se è possibile implementare delle regole in modo diverso: senza necessariamente dover conoscere, o dover applicare, gli algoritmi e i calcoli specifici: vale a dire senza utilizzare delle regole operazionali.

La risposta a questa domanda è positiva. Questo avviene ogni qualvolta si può costruire direttamente un'associazione tra "la situazione iniziale" e " la situazione finale" senza dover conoscere l'operatore che sta agendo.

Consideriamo il caso in cui **ogni volta** che si presenta la situazione che indico con A allora sistematicamente si ottiene la situazione B: si può allora procedere ad "associare" direttamente B come "conseguenza" di A.

Abbiamo quindi l'**associazione diretta** tra due situazioni A e B . In questo caso posiamo dire che "A implica B". Si noti che è qualcosa che assomiglia ad una forma, forse grezza, del "modus ponens" utilizzato in logica: $[(A \rightarrow B) \land A] \vdash B$ . Anche se va tenuto presente che stiamo trattando rappresentazioni strutturali e non proposizioni.

Credo si possa mostrare che, almeno entro un contesto deterministico, quello associativo e quello operazionale possano alla fine dei conti essere considerati come due aspetti dello stesso fenomeno.

Sia data una struttura iniziale A e sia vero che su di questa agisce in modo regolare un certo operatore specifico Op che genera la struttura B. Ciò significa che ogni qualvolta si presenta la struttura A su essa agisce sempre e comunque Op e viene generata B. Si consideri ora il complesso della struttura A +(t)B (A composta, o associata, con B, nel tempo o in taluni casi anche nello spazio). Tale struttura costituisce dunque una "regola" nel senso indicato più sopra. Dato che per ipotesi ogni volta che si presenta A sicuramente viene generata anche B si può sfruttare tale fatto per effettuare inferenze.

La composizione A +(t)B sarà allora una "**regola associativa**".

Semplicemente vengono "associate" A e B in una relazione funzionale di causa ed effetto. Quest'argomentazione vale fin tanto che si opera con strutture esattamente ben definite e entro un ambiente dove l'operatore Op è sempre lo stesso. Esistono molti sistemi fisici dove è in pratica impossibile partire da situazioni fisicamente del tutto identiche. In realtà le regole associative molto raramente sono applicabili per prevedere l'evoluzione di sistemi reali rappresentati a "basso livello di astrazione". Ma si può verificare che esistono moltissime situazioni pratiche nelle quali si possono costruire delle rappresentazioni più astratte dove le regole associative funzionano molto bene.

In fisica e nelle scienze esatte siamo abituati ad utilizzare regole espresse in forma operazionale. Queste hanno in genere un campo di validità molto ampio e possono essere espresse in forma compatta; richiedono cioè una quantità limitata di memoria per essere rappresentate. Nella simulazione scientifica le regole operazionali svolgono sicuramente il ruolo più importante.

Ma nella pratica concreta dell'attività cognitiva le regole operazionali possono essere applicate con successo solo in domini particolari: quando si ha una conoscenza dettagliata e precisa della struttura della situazione che si sta esaminando, e quando il sistema del quale si vuole prevedere l'evoluzione non è troppo complesso. Inoltre si deve avere a che fare con sistemi che sono poco sensibili alle piccole differenze nelle condizioni iniziali. In caso contrario gli inevitabili "errori di misura" che si commettono in partenza rendono di fatto impossibile ogni previsione che possa funzionare per tempi lunghi.

Quando si passa dalle scienze esatte ai problemi che dobbiamo affrontare nel quotidiano, le regole operazionali continuano ad avere un ruolo sicuramente importante, per esempio nei processi percettivi e di coordinamento senso-motorio. Ma non appena passiamo a rappresentazioni di livello medio alto, che sono quelle che adoperiamo più frequentemente, è l'aspetto associativo a prevalere in modo netto.

A livello di strutture e di logiche emergenti diventa spesso molto difficile poter applicare regole di tipo operazionale, mentre risulta più semplice ed efficiente l'utilizzo di regole associative, anche se queste richiedono la memorizzazione di molte più informazioni.

Altro aspetto molto interessante è che a livello emergente si possono definire regole valide che computano o associano anche in negativo (sulla mancanza di qualcosa). Per implementare una regola di questo tipo è necessario passare attraverso la funzione di memoria, giacché ci deve essere qualcosa che sia in grado di "accorgersi" che "manca qualcosa". Regole che computano in negativo non sembrano possibili a livello di fisica di base.

Consideriamo un esempio. Ogni agricoltore sa bene, fin dall'alba della civiltà, che in mancanza di acqua le piante sono destinate a rinsecchire. Questa regola di ampia validità, e di fondamentale importanza per la nostra sopravvivenza, consiste nella capacità umana di associare due fatti, due concetti e quindi, secondo le idee esposte, una serie di soggetti cognitivi di livello medio alto. In questa regola sono associati direttamente alcuni piccoli insiemi di soggetti cognitivi. Anzi, più precisamente, sono associate le esplicitazioni degli avvenuti riconoscimenti dei soggetti menzionati.

In particolare notiamo che nella regola specifica si associa in negativo: si associa il fatto che il soggetto "acqua" non è stato riconosciuto (da un certo periodo di tempo), con la comparsa del soggetto "piante rinsecchite", il quale può avere, in taluni casi, un effetto drammatico per la possibilità di nutrirsi.

Supponiamo di possedere un ipotetico super calcolatore, in grado di simulare alla perfezione le reazioni biochimiche che avvengano all'interno delle cellule delle piante. Procedendo con queste simulazioni si realizzano delle configurazioni molto complesse, ma che osservate solo a "basso livello" non hanno alcun significato particolare. Solo un osservatore che sia in grado di riconoscere in esse la presenza delle varie proprietà emergenti, che corrispondono a soggetti cognitivi di livello più elevato, può "differenziare" le varie situazioni e attribuirgli, nel caso, qualche valenza positiva o negativa. Notiamo anche che a livello di simulazione biochimica, per simulare il processo in cui la pianta rinsecchisce, non è affatto necessario introdurre la decodifica esplicita del "fatto che manca acqua". La simulazione può procedere senza alcun problema computando, e applicando quindi le varie regole operazionali, solo in positivo. Non è necessario codificare in negativo. Di fatto la fisica, e in generale i sistemi che usano regole di base, non lo fanno!

Uno dei punti essenziali da comprendere bene è che, a livello di base, le strutture emergenti non sono definite. Con la sola simulazione, per quanto precisa e ben eseguita, non ci accorgeremmo assolutamente del fatto che la pianta sta morendo; affinché ciò avvenga occorre codificare tutte le proprietà emergenti che costituiscono le rappresentazioni del "sistema pianta" pensato come un tutt'uno. In una simulazione effettuata davvero a livello di base, senza la decodifica delle strutture emergenti, tutto ciò che si osserva è come interagiscono i moltissimi singoli atomi o le moltissime singole molecole. Ma se non si cambia prospettiva di osservazione, questo enorme brulichio di reazioni chimiche non avrà alcun significato.

Solo se abbiamo la possibilità di definire soggetti emergenti possiamo, teoricamente, usare la simulazione al calcolatore per verificare che la regola associativa dell'esempio sopra funzioni. Con una simulazione possiamo compiere questa verifica in un modo che appare essere, nonostante tutto, ancora sostanzialmente empirico. È una specie di esperimento virtuale: si simula dentro un calcolatore invece di fare l'esperienza reale.

Agli esseri umani è concessa anche la facoltà di capire perché questa regola è vera! Questa nostra facoltà di "capire il perché" si basa probabilmente sulla capacità di seguire il "concetto" di acqua quando si procede ad reinterpretarlo, con approccio riduzionista, sulla base delle conoscenze di cui disponiamo. Riusciamo allora a raffigurarci che l'acqua consiste in molecole di un certo tipo, con una certa serie di proprietà che riusciamo a rappresentare. Riusciamo a comprendere che l'ambiente acquoso è proprio quello dove avvengono le reazioni chimiche all'interno della cellula, e in questo modo ad un certo punto ci appare chiaro il motivo per cui senza acqua non ci può essere vita. Francamente non so se effettivamente tutti questi passaggi siano sempre e comunque scomponibili in termini di rappresentazioni computazionali di strutture e logiche emergenti, e se quindi queste nostre capacità di "comprendere il perché" sia effettivamente riducibile a pura computazione, o se intervenga qualche altro fenomeno. Mi sembra però comunque che le idee che sto proponendo in questo lavoro possono dare un contributo a chiarire alcuni aspetti importanti di questi processi cognitivi.

Un coltivatore dell'antichità non poteva avere alcuna nozione sulle reazioni chimiche, non aveva microscopi per osservare le cellule; tuttavia era in grado, sulla base di una serie di osservazioni macroscopiche ripetute, di scoprire per induzione la verità di una regola importantissima.

Le regole di tipo associativo sono molto importanti ma non sono le uniche che utilizziamo. Come detto, penso che le regole operazionali abbiano un ruolo importante nei problemi di coordinamento senso motorio e in molti altri casi. È probabile che molte regole siano di tipo misto, sia associative sia operazionali. Ma è probabile che in questi casi sia la parte associativa a "decidere" quali operazioni applicare.

Per i compiti di "basso livello di astrazione", l'aspetto operazionale di queste regole "ibride" dovrebbe essere più marcato, ma probabilmente tende ad attenuarsi a favore di quello associativo man mano che si passa dalle rappresentazioni di basso livello a quelle più astratte.

### 7.7 Sull'importanza e sulla decodifica delle regole associative

L'esempio illustrato mostra che spesso (forse sempre) le regole associative compaiono quando si passa ad "osservare" un fenomeno "dall'alto", vale a dire utilizzando direttamente rappresentazioni di strutture e di proprietà strutturali emergenti. Spesso per riuscire a cogliere queste regole è necessario passare a rappresentazioni che non siano "sensibili" a tutto ciò che di volta in volta è presente, ma solo a ciò che è realmente essenziale per la codifica della regola in oggetto. In genere i soggetti cognitivi davvero essenziali, che partecipano direttamente alla regola, sono pochi e in genere sono "astratti", in quanto sono "generalizzazioni categoriali" di oggetti e di fenomeni.

Credo che nella pratica reale dell'attività cognitiva l'aspetto associativo emergente sia molto spesso, anche se non sempre, più facilmente da gestire e di maggiore utilità di quello operazionale. Credo che le regole associative svolgano un ruolo fondamentale per la cognizione. Come visto esse presentano una serie di differenze rispetto a quelle operazionali, non ultimo il fatto di poter inferire in negativo.

Ho usato il termine "associative" per enfatizzare le differenze con l'aspetto operazionale. Anche se a rigore non si tratta sempre di semplici associazioni dirette tra cause ed effetti, ma di combinazioni logiche e temporali che possono essere anche di una certa complessità. In genere queste combinazioni mettono assieme gli avvenuti riconoscimenti di particolari insiemi (strutturati) di soggetti cognitivi in specifiche relazioni temporali, spaziali e logiche, con altre composizioni di soggetti cognitivi che costituiscono il "risultato dell'inferenza". Con buona probabilità il modo corretto di utilizzare queste codifiche è, in taluni casi, più vicino alla logica fuzzy che a quella booleana standard.

Siano A, B, C, D….. informazioni che esplicitano l'avvenuto riconoscimento di specifici soggetti cognitivi, e siano X, Y, Z altri soggetti che possono contribuire a costituire possibili previsioni sull'evoluzione degli eventi nell'ambiente che si sta osservando.

I vari riconoscimenti A, B, C, D… possono non essere contemporanei, e saranno quindi caratterizzati da una variabile tempo $t_x$, valutata rispetto al presente. Quindi $t_x$ indica, spesso in modo discreto e non necessariamente preciso, quanto tempo è passato da quando il particolare soggetto è stato riconosciuto come presente nell'ambiente.

Avremo quindi dei riconoscimenti caratterizzati da valutazioni della variabile "tempo trascorso", In simboli: $A(t_a)$, $B(t_b)$, $C(t_c)$ ecc..

Una regola associativa potrebbe allora essere, ad esempio, in caso semplice, del tipo: $A(t_a) \Rightarrow X(t_x)$ (con $t_x - t_a$ compreso entro un certo intervallo).

Ma può anche essere una combinazione logica più complessa come, ad esempio:

$A(t_a)$ AND $B(t_b)$ AND NOT($C(t_c)) \Rightarrow X(t_x)$

Dove AND e NOT si riferiscono alle note operazioni di logica booleana e corrispondono ai nostri "e" e "non" utilizzati in italiano.

Come detto però la decodifica può anche seguire una logica di tipo fuzzy, e quindi possiamo anche avere riconoscimenti parziali e quindi regole di questo tipo:

Se $A(t_a)$ è stata riconosciuta con un affidabilità almeno del 70% assieme al soggetto $B(t_b)$, riconosciuto almeno all'80%, allora si può avere il verificarsi di X a $t_x$ con probabilità….

Spesso le regole associative non sono esatte, ma solo più o meno affidabili, quindi più o meno "probabili" secondo una probabilità di tipo frequentistica. Questo implica che al segno $\Rightarrow$ dobbiamo associare una certa probabilità che l'associazione accada effettivamente; ad esempio potremo avere:

$A(t_a) \Rightarrow (80\%) X(t_x)$, con significato evidente.

Molto spesso non sappiamo associare un numero preciso alla probabilità, ma possiamo solo dire che l'evento è "poco probabile" oppure "molto probabile".

Oltre agli operatori logici booleani AND, OR, NOT, XOR ecc… possiamo avere tutta una serie di relazioni particolari tra soggetti cognitivi. Queste relazioni sono però a loro volta altri soggetti cognitivi.

Ad esempio possiamo avere la relazione "essere vicino a", oppure "essere sopra a", venire prima, venire dopo ecc. Quindi una regola associativa può diventare del tipo:

Se il soggetto A (una pentola d'acqua) "è sopra" B (il fuoco acceso) $\Rightarrow$ C (l'acqua si scalda) "e dopo un certo tempo" $\Rightarrow$ D (l'acqua bolle).

### 7.8 Concetto di "microsituazione"

Può essere utile il concetto di "**microsituazione**".
Nell'attività cognitiva è molto importante riuscire a descrivere e rappresentare **le situazioni**. In generale **una situazione è uno "stato possibile della realtà"**.

In genere in un sistema cognitivo potranno essere attive contemporaneamente rappresentazioni di situazioni diverse. Ad esempio situazioni che si riferiscono al presente oggettivo, o che invece descrivono possibili previsioni di quanto può accadere, oppure anche rappresentazioni di situazioni che costituiscono degli "obiettivi da raggiungere", oppure delle "situazioni ipotetiche" che in quel particolare contesto risultano di qualche interesse.

In molti casi per descrivere le situazioni penso sia utile, e talvolta necessario, costruire delle rappresentazioni accurate delle strutture degli oggetti e dei fenomeni che "compaiono nelle scene" sotto esame. Ma credo anche che, nella maggior parte dei casi, ciò che assolutamente non può mancare è il semplice **riconoscimento** dei soggetti cognitivi presenti. Penso che in molti casi le singole situazioni possano essere rappresentate in "modo compatto" per mezzo di un "**insieme strutturato**" di **pochi** soggetti cognitivi.

Come illustrato nel capitolo precedente, un "**insieme strutturato**" di soggetti cognitivi consiste in un insieme di alcuni soggetti cognitivi, importanti per il problema in oggetto, dove spesso sono anche specificate (quindi rese esplicite) alcune delle relazioni che intercorrono tra questi (che possono essere di vario tipo). Proprio per questo fatto l'insieme è appunto "strutturato".

Credo che questo concetto sia importante appunto perché le "singole regole" in realtà si applicano non a tutto ciò che fa parte "della situazione globale della realtà presente", ma solo a particolari "microsituazioni" presenti in essa. Queste "micro-situazioni" contengono la rappresentazione o il semplice riconoscimento di solo quei soggetti cognitivi che sono effettivamente implicati nella singola regola in oggetto, nonché, come detto, la specificazione di alcune loro relazioni.

Rispetto a una rappresentazione "completa" di una situazione, queste "micro-situazioni" sono delle specie di sottoinsiemi che contengono solo i soggetti cognitivi effettivamente implicati nella specifica regola.

In genere, per utilizzare delle regole di tipo operazionale devono essere rappresentate, con un buon grado di dettaglio, le strutture estese (come ad esempio le loro geometrie tridimensionali) degli oggetti ai quali la regola si applica (o talvolta, di alcuni dei loro elementi strutturali). La regola stessa agisce su queste strutture producendone delle altre (che, ad esempio, possono consistere in previsioni dell'evoluzione nel tempo delle prime, quali delle simulazioni degli oggetti in movimento nello spazio).

Diversamente le regole di tipo associativo si applicano al riconoscimento di specifici soggetti cognitivi, spesso di alto livello, in specifiche relazioni, che come detto possono essere di vario tipo: spaziali, temporali, quantitative, logiche ecc… (si tenga presente che anche le stese relazioni sono, in genere, a loro volta dei legittimi soggetti cognitivi).

L'insieme strutturato di questi riconoscimenti costituisce la "microsituazione" di partenza alla quale la regola si applica. L'output della regola sarà l'indicazione di un altro insieme di soggetti cognitivi (anch'esso strutturato), che costituirà il risultato dell'inferenza di quella singola regola.

Le regole associative potranno, secondo i casi, associare "micro-situazioni" di diverso grado di astrazione.

**Approfondimenti**

### 7.9  Accenni sulle regole deduttive

Esiste un'altra importante classe di regole utilizzata in logica e in matematica: quella delle regole deduttive.

Probabilmente le regole deduttive non svolgono un ruolo di grande importanza in intelligenza naturale, è anzi probabile che siano utilizzate raramente. Un animale in genere non compie deduzioni, ma elabora le proprie pianificazioni e le proprie previsioni utilizzando essenzialmente regole comportamentali che provengono o dal patrimonio istintuale, oppure dalle proprie esperienze dirette. Queste regole in genere non sono "esatte", ma hanno solo una certa probabilità di funzionare.

Possiamo dire si è compiuta una **deduzione** quando si è certi che è impossibile che quanto stabilito sia in contrasto con le premesse dalle quali si è partiti. Questo può avvenire quando si è avuto modo di esplorare tutte le possibilità, oppure quando si è sicuri che le operazioni che sono state compiute siano esenti da errori.

Un sistema di computo strutturale, come descritto nel terzo capitolo, è costituito da un insieme di strutture di partenza e da un insieme di operazioni permesse su queste strutture. Eseguendo queste operazioni a rigore non si fa "logica", ma si eseguono dei calcoli, o meglio delle operazioni di computo. Banalmente un calcolo è esatto se non si commettono errori.

Alcuni sistemi fisici reali esibiscono spesso un comportamento deterministico. Questo avviene ad esempio nei circuiti di un sistema digitale. Conoscendo lo stato di un sistema di questo genere, nonché le "regole" con cui commutano le porte logiche e gli elementi di memoria che lo costituiscono, è possibile prevedere con esattezza l'evoluzione temporale dello stesso.

In questo senso è prevedibile ogni sistema che sia riducibile a computo strutturale nel quale non esiste possibilità di scegliere.

Anche in sistemi comunque ben definiti, ma dove esista la possibilità di scegliere tra un certo insieme di mosse permesse, è spesso possibile compiere delle deduzioni. Questo capita qualora il problema proposto è tale che, nonostante la libertà di scelta, in realtà le cose possono essere combinate in un solo modo. È il caso ad esempio di molti puzzle.

Un tipico caso dove è possibile compiere delle "deduzioni logiche" si ha quando si deve ricostruire la struttura di qualcosa sulla base di informazioni parziali ma che appunto possono essere combinate insieme, senza dar origine a incongruenze o contraddizioni, in un unico modo.

Come detto il computo strutturale e la logica sono due cose diverse. La logica si occupa del modo corretto di costruire rappresentazioni e di gestire delle regole valide che consentono di compiere inferenze su queste rappresentazioni.

L'argomento è in realtà complesso e per essere spiegato in modo esauriente richiede una trattazione che va oltre l'ambito di questo capitolo.

Sintetizzando e semplificando, possiamo pensare il computo strutturale come qualcosa che sta a un livello più basso della logica. Un sistema di computo strutturale è governato da un insieme di operazioni permesse. Di solito questo insieme è minore di quelle che un sistema cognitivo è in grado di utilizzare. In questo senso le "regole del computo strutturale" vincolano le mosse permesse. Ci sono sistemi di computo strutturale totalmente vincolati, come ad esempio la simulazione dell'attività di un circuito digitale, e sistemi che invece permettono di scegliere, ad esempio nel gioco degli scacchi. In entrambi i casi le operazioni permesse sono solo alcune di quelle che un sistema cognitivo è in grado di eseguire. Un sistema cognitivo potrebbe violare le regole delle tavole di verità delle porte logiche, oppure le regole di movimento dei vari pezzi sulla scacchiera, ad esempio movendo l'alfiere come la torre ecc…

In un sistema totalmente vincolato, come il circuito digitale, se il sistema cognitivo rispetta le regole, allora è in grado di fare previsioni corrette. Molti sistemi fisici reali sono prevedibili a patto di osservare le regole. In questi casi si eseguono delle simulazioni che si limitano al semplice calcolo.

La logica formale sta a un livello più complesso. Le regole della logica non si riferiscono direttamente a un sistema di computo strutturale, ma ai modi corretti di costruire rappresentazioni ed eseguire inferenze. Una rappresentazione è qualcosa che si riferisce a una certa altra cosa e che "pretende" di ricostruirne in modo corretto alcune delle strutture. Una rappresentazione può essere quindi o giusta o sbagliata a seconda che la sua struttura (in realtà si tratta quasi sempre di una sottostruttura) coincida o meno con quella del rappresentato.

# 8     Alcune idee per una definizione del concetto di problema

Che cos'è un problema?
Esiste la possibilità di definire questo concetto in modo sufficientemente preciso?

## 8.1   Breve introduzione alla teoria dei sistemi di produzione

In intelligenza artificiale si studiano delle situazioni operative nelle quali è possibile proporre una definizione ben precisabile per il concetto di "**problema**", sfruttando il fatto che molti **sistemi** sono caratterizzati sempre da "**stati ben definibili**". Questi sistemi sono talvolta indicati come "sistemi di produzione". La terminologia deriva dal fatto che queste idee sono nate nell'ambito di alcuni studi sulla possibilità di "produzione" automatica di espressioni linguistiche.
In questi studi il sistema che si sta esaminando è chiamato "**sistema universo**". Come detto, esso può assumere vari "stati distinti" ben definiti, e l'insieme di tutti quelli possibili è chiamato "**spazio degli stati**". La struttura di questo spazio è costruita in modo tale che se il sistema si trova in certo stato allora è possibile passare solo a quegli altri stati che sono eventualmente **collegati** al primo tramite una "**regola di produzione**". Non è quindi possibile saltare direttamente da uno stato ad un altro qualsiasi, ma si deve necessariamente transitare attraverso altri lungo un certo percorso.
Lo spazio degli stati ha quindi tipicamente una struttura a grafo orientato, e ogni ramo di questo grafo corrisponde all'esistenza di una regole di produzione valida.
Per fare un esempio pensiamo al gioco degli scacchi. In questo caso il sistema universo consiste nella scacchiera e nei vari pezzi, mentre i singoli stati consistono nelle varie disposizioni possibili dei pezzi sulla scacchiera durante una partita. Si noti che in modo analogo si può descrivere l'attività di una generica macchina a stati finiti provvista di memoria. Ne consegue che quanto qui illustrato è applicabile ad ogni sistema computazionale discreto e finito.
ome detto, entro lo spazio degli stati, a partire da un generico stato è possibile passare ad altri solo applicando le "**regole di produzione**" o "**mosse legali**". Negli scacchi le regole di produzione sono le mosse permesse. Ne risulta che nello spazio degli stati sono collegati solo gli stati connessi da una regola valida.

In genere un **problema** è quindi definito da uno **stato di partenza** e da uno **stato obiettivo**. **Risolvere il problema** significa cercare, entro lo spazio degli stati, un percorso che permetta di passare dallo stato di partenza a quello obiettivo utilizzando le regole di produzione permesse per passare da uno stato all'altro.

Spesso interessa solo trovare almeno un percorso possibile, ma in altri casi interessa invece trovarne uno che sia, per qualche motivo, particolarmente conveniente. Quest'ultimo può essere quello "più corto", ma anche quello che permette di evitare di dover passare per stati (quindi per "situazioni") che risultano per qualche motivo "indesiderabili" e quindi da evitare. In modo inverso, alcuni percorsi possono risultare preferibili perché alcuni stati sono invece "desiderabili" in quanto permettono di ottenere un vantaggio di qualche genere.

In tal senso la risoluzione di un problema può essere vista come un'attività di ricerca e selezione di percorsi possibili nello spazio degli stati.

Ci possono essere stati che costituiscono un "vicolo cieco", perché una volta che si finisce in essi non esiste alcuna regola applicabile per poterne uscire. Possono esistere sistemi caratterizzati da stati dove la "mossa da fare" è solo una ed è obbligatoria; ci sono anche sistemi, e sono quelli più frequenti nel mondo reale, nei quali le mosse permesse sono più di una per un certo insieme significativo dei loro stati possibili

I problemi che ammettono, per buona parte degli stati possibili, più regole di transizione tra le quali poter scegliere, si dicono "**esponenzialmente complessi**", poiché il numero di tutte le possibili varianti di percorso diverge esponenzialmente. Nella maggior parte dei casi pratici tale numero cresce così velocemente da diventare intrattabile già dopo pochi passi (almeno utilizzando computazioni di tipo classico).

La metodologia principale per affrontare i problemi esponenzialmente difficili consiste nell'associare ad un certo sottoinsieme degli elementi dello spazio degli stati, un valore di "preferibilità". Si ottengono in questo modo quelle che si chiamano "**funzione euristiche**". Tali funzioni hanno per dominio un sottoinsieme dello spazio degli stati e sono in genere associate al particolare problema in questione. Le funzione euristiche permettono di accorciare i tempi di ricerca in modo molto semplice: qualora si sia in un determinato stato e si hanno a disposizione più alternative, si sceglie quella che presenta il valore maggiore della funzione euristica. In questo modo si evita di esplorare tutti i percorsi possibili (quando l'euristica funziona). Una delle problematiche che si affronta in IA è appunto quella della ricerca di strategie per l'esplorazione euristica.

## 8.2  Una possibile definizione del concetto di problema

Abbiamo dunque visto che nell'ambito della "teoria dei sistemi di produzione"

il **concetto di problema** può essere definito come attività di ricerca di un percorso di mosse legali, che consentono di passare da uno stato di partenza ad uno stato obbiettivo.
Sebbene i sistemi di produzione possano essere visti come dei casi particolari, penso che questo modo di pensare possa essere esteso, con le dovute precisazioni, all'attività cognitiva in senso generale.

Per l'attività cognitiva gli equivalenti degli "stati del sistema" sono le possibili **situazioni della realtà**, sia attuali che potenziali, e penso che queste possano essere spesso rappresentate, come illustrato, con "**insiemi strutturati di soggetti cognitivi**". Vedremo inoltre che proprio in funzione dei soggetti cognitivi in esse presenti queste possono essere più o meno desiderabili e quindi costituire o meno delle "situazioni obbiettivo" oppure delle "situazioni sgradevoli o pericolose".
Detto questo appare allora possibile proporre un primo tentativo di definizione per il concetto di problema, in modo analogo a quanto si fa nella teoria dei sistemi di produzione. Possiamo proporre una "definizione operativa" come segue:
- Un problema consiste nel cercare il modo per passare, attraverso l'applicazione di una serie di azioni permesse (anche nel senso di fisicamente possibili), da una data situazione di partenza ad un'altra che costituisce l'obbiettivo da raggiungere.

Per quanto questo tentativo di imbrigliare il concetto di problema possa sembrare in parte una semplificazione, credo che in realtà possa avere una certa utilità.
In molti contesti questa definizione può essere resa concreta qualora sia possibile identificare e rappresentare **le situazioni di partenza**, **quelle obiettivo**, nonché il complesso delle azioni possibili (che in genere consistono in "regole da osservare").
Secondo le idee esposte in questo lavoro, un sistema cognitivo dovrebbe presentare un'importante differenza rispetto a quanto avviene nei sistemi di produzione. In effetti, in **un sistema cognitivo si può rappresentare la medesima realtà di base, a diversi livelli di astrazione.** Il vantaggio consiste nel fatto che **a queste diverse rappresentazioni è spesso possibile applicare diversi insiemi di regole emergenti.**
Come accennato, le situazioni di partenza e le situazioni obiettivo possono essere caratterizzate in base ai soggetti cognitivi che contengono. Credo che taluni obiettivi possano essere definiti in maniera molto astratta, quindi attraverso soggetti cognitivi di alto livello. Per questa ragione spesso la "risoluzione di un certo problema" potrà essere soddisfatta contemporaneamente da "molte situazioni concrete diverse": tutte quelle che, viste ad un opportuno livello di astrazione, contengono al loro interno proprio quei soggetti astratti che costituiscono gli obiettivi da raggiungere.

Spesso tutte queste situazioni concrete, quando sono "osservate" da un punto di vista più astratto attraverso le opportune operazioni di derivazione strutturale, non risultano più distinguibili e vengono a coincidere.

Esempio: il problema che mi pongo è: "ho fame e desidero mangiare". Posso risolvere questo problema in molti modi concreti diversi; posso nutrirmi con diversi tipi di cibo, posso andare in un ristorante o comprare qualcosa al mercato, posso nutrirmi all'aperto, posso cacciare o pescare ecc… Tutte queste situazioni tra loro diverse, osservate da un punto di vista più astratto, coincidono con quella in cui "mi nutro", che costituisce appunto la "situazione obbiettivo" alla quale voglio arrivare, e quindi la soluzione del problema che mi pongo.

Spesso, quando si definisce un problema, anche la "situazione di partenza" può essere descritta in modo astratto, e in taluni casi si può anche fare a meno di "dichiararla" in maniera esplicita. Spesso, infatti, essa è facilmente arguibile dal problema che viene proposto. Se ad esempio mi pongo il problema di conoscere l'orario di partenza di un treno, significa che la situazione obiettivo è quella astratta nella quale vengo, in qualche modo tra i tanti possibili, a conoscenza dell'orario di partenza del treno, e quella di partenza è implicitamente quella dove non ho questa informazione. Anche questa descrizione della situazione di partenza è ovviamente astratta: ci possono essere, nel concreto, "moltissimi modi di stare" che la soddisfano.

Ad ogni modo esiste comunque una situazione di partenza particolarmente importante, che è quella del presente oggettivo, vale a dire quella che costituisce lo stato della realtà in questo istante. Ovviamente, ogni volta che si affronta un problema per forza di cose si deve partire dalla situazione del presente. Alla fine dei conti ogni sistema cognitivo che deve operare entro un contesto reale (dove deve spesso sopravvivere) non può permettersi di non dedicare molta attenzione alla "situazione dello stato del presente".

Vedremo che anche la "situazione globale del presente" può essere rappresentata contemporaneamente a molti livelli diversi di astrazione. Se da un lato è vero che in generale, per descrivere un problema dato, la situazione di partenza può essere definita in modo molto astratto, è nello stesso tempo vero che, nel concreto, si deve sempre partire dalla conoscenza della situazione del presente oggettivo.

### 8.3 Regole di previsione e regole per la pianificazione delle azioni

Ho parlato in precedenza della differenza tra regole operazionali e regole associative. Sicuramente sono possibili altri modi per distinguere e classificare le regole utili per l'attività cognitiva. Uno di questi, particolarmente importante, è quello che distingue tra regole che possiamo chiamare di "pura previsione" e

"regole per la pianificazione delle azioni".

Ho illustrato più indietro che una delle caratteristiche più importanti della nostra "realtà emergente" consiste nel fatto che in essa esiste la possibilità di "scegliere". Questo fatto si concreta nella possibilità di eseguire delle azioni sull'ambiente fisico che possono essere "scelte" da un certo insieme di possibilità.

Noi siamo dotati di un corpo e con questo possiamo agire in diversi modi sulla realtà esterna, principalmente con movimenti muscolari che possono essere organizzati per produrre molti tipi di azioni fisiche diverse. Possiamo anche agire producendo dei suoni, che in taluni casi possono essere "ordini simbolici" impartiti ad altre persone o ad apparati artificiali. Queste azioni fisiche producono delle conseguenze e, in genere, una persona adulta che ha già acquisito un certo grado di capacità di pianificare il proprio agire (passando attraverso opportune fasi di apprendimento), è in grado di "prevedere", con un certo grado di precisione, quale ne sarà l'effetto. Le azioni fisiche quindi obbediscono a delle regole che in genere sono ben conosciute da chi agisce.

È utile tenere presente il seguente punto: quanto si mettono in atto dei comportamenti, lo si fa spesso avendo già in mente una qualche rappresentazione, anche se astratta e parziale, dei risultati a cui porteranno. È importante comprendere che quando si "pianificano delle azioni" si affronta in genere un problema che è in un certo senso l'inverso di quello che si affronta quando si cerca di prevedere come evolverà la realtà.

Le regole di pura previsione possono essere sia associative che operazionali, e possono essere applicate per prevedere come evolveranno le cose nel futuro. Esse sono quindi (in genere) applicate in avanti nel tempo. Il problema consiste nell'esaminare la "situazione di partenza" e formulare per questa delle previsioni. Si cerca quindi di prevedere, senza conoscerle in anticipo, quali saranno le situazioni della realtà in futuro.

Diversamente le regole (e le strategie) per la "pianificazione delle azioni" affrontano un problema parzialmente inverso: si sa già a cosa si vuole arrivare, è già formulata una "rappresentazione della situazione futura desiderata", e si cerca invece di trovare un insieme di comportamenti che porti ad essa.

Ovviamente anche le azioni che si possono compiere sono soggette alle stesse regole fisiche, chimiche ecc. che valgono per gli altri fenomeni naturali; una volta che è stata stabilita quale sarà la sequenza di azioni che si compiranno, si potrà cercare di prevedere l'evoluzione delle cose usando in buona parte gli stessi procedimenti, e sostanzialmente le stesse regole, che si applicano per le "previsioni pure". Una cosa quindi è l'attività di "previsione pura" in sé, che può essere applicata anche per prevedere gli effetti delle azioni attuate da un generico "agente"; un'altra è il problema inverso: la ricerca, tra l'insieme delle azioni possibili, di quelle che hanno "buona probabilità" di portare verso una determinata situazione obbiettivo.

Penso possa essere utile affermare che i problemi di "pura previsione" sono **problemi diretti**: essi, infatti, seguono la freccia del tempo; al contrario, i problemi di ricerca di una serie di azioni che portano ad una determinata situazione sono **problemi inversi.**

Una delle caratteristiche importanti dei **problemi inversi**, è che spesso sono più difficili da affrontare di quelli diretti poiché, per riuscire a risolverli, è necessario compiere un processo di ricerca tra molte alternative possibili. Mentre nei problemi diretti una volta descritta la situazione di partenza, esiste spesso, almeno potenzialmente, un unico procedimento da applicare (che deriva dalle leggi fisiche sottostanti), nei problemi inversi esiste la possibilità di scegliere tra varie possibilità, e per questo la complessità da affrontare tende a divergere.

Nella pratica il problema di identificare "per tentativi" una sequenza efficace di azioni consentite può essere così complesso che diventa impensabile affrontarlo direttamente. Una delle strategie per venire a capo di questa esplosione esponenziale delle possibilità, consiste nell'identificare e collezionare, durante opportune fasi di apprendimento, un cospicuo insieme di "soluzioni già pronte" da utilizzare per i vari problemi specifici (a volte si tratta di soluzioni parziali che hanno solo una "certa probabilità" di funzionare). Le associazioni tra i "problemi dati" e le relative soluzioni funzionanti, vale a dire le "sequenze di comportamenti" in grado di affrontarli, costituiscono a tutti gli effetti delle regole di tipo associativo. In altre parole, accade spesso che le regole di "pianificazione delle azioni" siano, di fatto, delle regole associative tra ciò che definisce il problema che si deve affrontare e la sequenza dei comportamenti che ne costituisce la soluzione.

Penso che una parte consistente del nostro apprendimento consista nell'acquisire un buon patrimonio di queste regole associative che connettono la formulazione dei problemi con delle soluzioni collaudate.

Altro punto importante è che la maggior parte dei problemi che si deve imparare ad affrontare, sono tali per cui non è possibile associarvi direttamente una sequenza di azioni fisiche, predefinite nel dettaglio, che sia in grado di portare alla risoluzione. Questo avviene perché non è possibile prevedere in anticipo l'evoluzione delle cose nei particolari. Nonostante ciò è però spesso possibile associare ai vari problemi delle soluzioni di medio o alto livello di astrazione, che dovranno poi essere, al momento dell'esecuzione concreta, tradotte in azioni più specifiche. Credo che ai problemi complessi convenga associare non una sequenza di comportamenti ben definiti, ma una serie di "strategie" risolutive, dove le soluzioni sono formulate appunto attraverso rappresentazioni astratte.

### 8.4 Alcuni punti importanti

Secondo le idee proposte in questo lavoro, è spesso possibile descrivere la medesima situazione usando molte rappresentazioni sovrapposte. Ovviamente tutte queste rappresentazioni devono essere tra di loro compatibili giacché descrivono la stessa realtà. La medesima realtà può essere descritta e rappresentata a diversi livelli, in funzione di come si riconoscono ed esplicitano le eventuali strutture derivate emergenti in essa presenti e, in generale, in funzione di come si procede ad effettuare operazioni di astrazione. La stessa "situazione base" può essere rappresentata con differenti insiemi strutturati di soggetti cognitivi che appartengono a livelli gerarchici diversi. Tutte queste rappresentazioni sono quindi tra di loro sovrapposte e ritraggono le medesime "situazioni di base" a vari livelli di astrazione. In funzione dei casi, potrà essere più utile rappresentare le cose a un certo livello di dettaglio, piuttosto che a un altro. In genere, ad esempio, quando si compiono dei movimenti, è necessario rappresentare in modo preciso le strutture tridimensionali degli oggetti, e in taluni casi, quando "si lavora di fino", è importante porre molta cura nella rappresentazione dei dettagli. Al contrario, per programmare attività a lungo periodo è spesso utile rappresentare le cose in maniera molto più compatta ed astratta e diventa importante la possibilità di tralasciare i dettagli, demandandoli ad altre competenze già collaudate.

Spesso i vari livelli di rappresentazione sono soggetti a regole diverse ma comunque tra di loro compatibili (quando questo non avviene, significa che ci sono degli errori a qualche livello).

Abbiamo già visto che a livello di fisica di base agiscono regole deterministiche totalmente vincolanti che non lasciano spazio di azione. Mentre a livello emergente compaiono regole che lasciano la possibilità di scegliere quali azioni eseguire tra più opzioni possibili.

Questa possibilità di scegliere è, come già detto, essenziale. Se essa non ci fosse la stessa attività cognitiva non avrebbe senso.

In generale, chiaramente, per la normale attività cognitiva non interessa rappresentare la realtà a livello dei suoi costituenti fisici fondamentali ed ha davvero poca importanza se a questo livello le regole sembrano (forse apparentemente) totalmente vincolanti. Le rappresentazioni alle quali siamo interessati sono invece quelle che consentono di applicare in maniera utile regole che permettono di fare previsioni e soprattutto di pianificare delle azioni. Anche sotto queste condizioni esiste la possibilità di descrivere la realtà a diversi livelli nei quali intervengono complessi di regole specifiche e parzialmente indipendenti. Se dobbiamo programmare un viaggio rappresentiamo le cose in maniera compatta: decidiamo dove andare, quale mezzo prendere, dove alloggiare ecc… Queste sono il genere di scelte che facciamo, e il campo delle possibilità è dato dalle informazioni che possiamo recuperare da varie fonti e che ci dicono quali scelte possiamo fare sui mezzi di trasporto, sui posti dove alloggiare, i giorni nei quali partire ecc... In queste rappresentazioni non descriviamo nei particolari quali azioni specifiche faremo, di che colore sarà l'autobus che prenderemo, il modello del taxi e via dicendo.

Sono tutte informazioni che non interessano e per la nostra rappresentazione basta utilizzare soggetti che rappresentano le cose ad alto livello, in maniera astratta e poco dettagliata. Nel mettere a punto queste pianificazioni ci affidiamo a capacità già acquisite e collaudate di saper far fronte ai vari problemi specifici che dovremmo affrontare. Se dobbiamo programmare il viaggio via internet, ci affidiamo alle nostre capacità già acquisite di saper utilizzare un computer. Se dobbiamo usare l'auto, ci affidiamo alle nostre capacità già acquisite di saper guidare. Quando poi ci apprestiamo a fare effettivamente le varie cose, produrremo una serie di altre pianificazioni soggette a regole specifiche. Quando guidiamo, dobbiamo eseguire di continuo previsioni a breve termine su come l'auto si comporta quando affrontiamo le curve, quando acceleriamo, ecc.. Nell'esecuzione dei movimenti dobbiamo utilizzare un complesso di regole che sono spesso operazionali e molto diverse da quelle che utilizziamo per pianificare a lungo termine. Ma queste ultime pianificazioni sono in genere proprio delle astrazioni, organizzate per gerarchie, che si affidano e si basano, in fin dei conti, sulle nostre capacità di eseguire e rappresentare i singoli movimenti, di collezionare poi sequenze di questi per definire, ad livello più elevato, delle "singole azioni", di mettere insieme sequenze di queste altre per definire dei "singoli compiti", e via salendo. Nel passare da un livello a quello successivo rappresentiamo i soggetti dei livelli inferiori in maniera molto compatta. In questo modo le rappresentazioni delle azioni e dei comportamenti si possono compattare e stratificare per gerarchie.

Abbiamo dunque visto che la stessa realtà può essere descritta con diverse categorie di soggetti cognitivi in funzione dei diversi livelli possibili alla quale può essere rappresentata. Un punto importante è che alcuni di questi soggetti servono anche a caratterizzare dal punto di vista "motivazionale" le situazioni, nel senso che queste possono essere più o meno desiderabili (o indesiderabili) proprio in funzione di alcuni dei soggetti che sono in esse identificabili.

In generale penso che affermare che "c'è un problema da risolvere", significa che si desidera passare dalla situazione attuale ad un'altra nella quale sono presenti particolari soggetti desiderabili (o sono assenti altri soggetti indesiderabili). Risolvere il problema significa individuare un comportamento, quindi una serie di azioni da fare, per passare dalla situazione presente a una situazione obbiettivo. Queste azioni sono soggette ad una serie di regole che devono essere rispettate. Queste regole servono a suggerire quello che si può fare e quello che non si può fare.

In generale l'attività cognitiva è proiettata al futuro. Il suo scopo è di anticipare i possibili scenari con simulazioni interne, e di esplorare lo spazio delle possibilità al fine di trovare in esso un percorso che consenta di passare dalla situazione attuale a quella obbiettivo.

Un punto importante è che spesso l'attività di esplorazione di tutti i percorsi possibili può divergere in maniera esponenziale, e questo può costituire un enorme problema. È importante trovare degli accorgimenti che permettano di contenere questa divergenza esponenziale. Come già affermato, uno di questi accorgimenti consiste nel collezionare, durante l'apprendimento, molte **soluzioni pronte**. Quando individuiamo un percorso che sembra funzionare bene, lo memorizziamo e lo utilizziamo come strategia per affrontare le situazioni future: quando ci troviamo di fronte ad un problema che coincide o che presenta delle analogie con uno già sperimentato, tentiamo di riutilizzare la stessa strategia di risoluzione. In genere tentiamo di adattare soluzioni già sperimentate al nuovo problema. Solo se nessuna delle soluzioni già sperimentate funziona, proviamo a cercare nuovi percorsi nello spazio delle possibilità.

# 9  Un possibile modello di sistema cognitivo

Alcune anticipazioni di cosa si può fare con gli strumenti illustrati nei capitoli precedenti

## 9.1  Introduzione

Con le idee illustrate nei capitoli precedenti mi è stato possibile sviluppare un modello per il funzionamento generale di un sistema cognitivo. In questo capitolo illustro, in modo conciso, alcuni dei suoi aspetti salienti. L'intento di queste spiegazioni è mostrare che il modello può funzionare, nello stesso tempo non ritengo opportuno in questa sede scendere nei dettagli su alcune tematiche. Questo sia per i possibili risvolti commerciali sia per altri motivi.
Come anticipato nella presentazione al presente volume, le spiegazioni che seguono sono pensate principalmente per un lettore che ha già affrontato queste tematiche.

L'esposizione dei prossimi argomenti non è semplice per una serie di motivi. In primo luogo mi trovo nella necessità di introdurre una serie di concetti, a volte non semplici, dovendo utilizzare una terminologia non pienamente adeguata poiché nel vocabolario comune non sono disponibili le parole adatte. Spesso per questo motivo userò il "virgolettato" per segnalare che i termini usati vanno interpretati secondo delle estensioni un po' diverse da quelle usuali. Ma la maggiore difficoltà deriva dal fatto che il modello che intendo proporre funziona tramite la sinergia di una serie di processi e di fasi che collaborano mutuamente al funzionamento del tutto. Il problema è che nell'esposizione queste cose devono essere introdotte una alla volta, e le sinergie con ciò che è descritto più avanti non possono essere pienamente colte prima di aver completato il quadro.
Un'altra difficoltà è probabilmente costituita dai miei limiti personali nella capacità di esporre per iscritto queste idee. Ho cercato una strategia per semplificare l'esposizione e ho reimpostato quest'ultima più volte.
Nell'esposizione che segue mi occupo dapprima della descrizione di un sistema minimale capace di alcune forme d'intelligenza naturale. Questo sistema è pensato all'inizio "già maturo" (uso il virgolettato…), presumo quindi sia già passato per le fasi di apprendimento necessarie per mettere assieme le conoscenze di cui dispone.

## 9.2  Punti generali

Un buon sistema cognitivo dovrebbe essere capace di analizzare le informazioni sensoriali e di costruire, sulla base di queste, una serie di rappresentazioni stratificate della realtà circostante. Dovrebbe essere in grado di riconoscere gli oggetti, i fenomeni e le situazioni che si presentano. Dovrebbe essere in grado di utilizzare al meglio le regole disponibili per generare previsioni e per cercare percorsi, entro lo spazio delle azioni possibili, che permettano di passare dalle varie situazioni di partenza contingenti a quelle poste come obiettivo.

Un sistema cognitivo ideale dovrebbe essere in grado di rappresentare al meglio la realtà del presente, del passato e tutte quelle "raggiungibili", e dovrebbe essere in grado di affrontare ogni problema potenzialmente risolvibile.

I sistemi cognitivi reali (che, almeno fino ad ora, sono solo quelli che la natura ha generato) hanno in realtà capacità limitate. Solo con l'uomo, e in tempi relativamente recenti rispetto alla storia dell'evoluzione, sono comparse facoltà cognitive "superiori".

Penso sia utile per i nostri scopi suddividere i livelli di intelligenza che si riscontrano in natura in almeno in tre categorie:
- **intelligenza naturale**: quella presente in varia misura in alcune specie animali, compresi i primati;
- **intelligenza linguistica**: quella che caratterizza l'uomo e la società umana prima dello sviluppo dei metodi e degli strumenti della filosofia analitica e soprattutto della scienza;
- **intelligenza avanzata**: quella, caratterizzata dai metodi e dagli strumenti concettuali della scienza, a partire da quelli utilizzati in matematica.

Ritengo che l'intelligenza linguistica e le facoltà avanzate abbiano bisogno, per esistere, delle facoltà più basilari. Quindi non penso sia possibile implementare forme di intelligenza superiore se prima non è stata acquisita un buona base di intelligenza naturale (eccetto casi particolari che riguardano domini limitati).

Nell'esposizione che segue mi occupo principalmente della descrizione di un sistema minimale capace di alcune forme d'intelligenza naturale. Questo sistema è pensato all'inizio "già maturo".

Un sistema cognitivo di tipo generale deve essere in grado di acquisire autonomamente, durante opportune fasi di apprendimento, una parte importante delle proprie conoscenze. Sono concepibili vari tipi di sistemi, ma sicuramente i più interessanti da studiare sono quelli dotati dell'equivalente dei nostri principali organi di senso e dell'equivalente di "un corpo con arti" in grado di spostarsi e agire sul mondo esterno.

Un sistema cognitivo di questo tipo deve essere in grado di acquisire la maggior parte delle informazioni sulla realtà attraverso i propri organi di senso.

### 9.3 Per iniziare

Per introdurre le idee parto usando delle semplificazioni. Il modello che propongo nei prossimi paragrafi non è del tutto corretto, ma è utile per costruire una prima visione d'insieme.

Possiamo pensare, introducendo una serie di semplificazioni, che un sistema cognitivo funzioni, in alcune delle sue attività, in modo simile a un "videogioco speciale" che ricostruisce al suo interno una rappresentazione tridimensionale della scena che osserva attraverso gli "occhi equivalenti" di cui è dotato (come possono esserlo delle telecamere digitali). Come un videogioco mette a disposizione una "simulazione tridimensionale" di una realtà immaginaria, in modo analogo possiamo pensare che una delle "prime fasi" dell'attività cognitiva consista proprio nel generare, all'interno del sistema, una rappresentazione tridimensionale della realtà circostante, quindi delle cose che si ha attorno: le pareti della stanza, gli oggetti presenti, gli scenari di sfondo, la forma tridimensionale di persone e animali ecc… Insomma la ricostruzione tridimensionale di tutto ciò che vediamo. Queste ricostruzioni tridimensionali inoltre non saranno statiche, i vari "oggetti virtuali", proiezioni interne di quelli veri, dovranno muoversi, nella simulazione interna, come si muovono quelli reali nella realtà esterna. Queste simulazioni tridimensionali seguiranno quindi fedelmente (nei limiti del possibile) l'evoluzione temporale di quanto osservato. In questo modo all'interno del nostro sistema cognitivo avremo a disposizione una specie di "**teatro virtuale**" tridimensionale che mette a disposizione una ricostruzione del mondo esterno.

È necessario premettere che in realtà implementare un sistema artificiale, capace di tradurre le informazioni che riceve dai propri sensori visivi (quindi da delle telecamere) nelle ricostruzioni tridimensionali corrette delle cose osservate, è una cosa tutt'altro che banale; è, anzi, un problema molto difficile. Questo problema richiede la messa a punto di buone capacità visive. La visione può essere vista, come già accennato, in due modi: come la capacità di ricostruire le geometrie degli oggetti, e come la capacità di riconoscere gli stessi, spesso in modo diretto, senza passare necessariamente per la ricostruzione delle loro geometrie tridimensionali.
Credo si possa mostrare che ambedue questi compiti sono in realtà difficili e richiedono una cospicua attività di analisi strutturale.
Le informazioni visive primarie devono essere analizzate per individuare in esse quelle caratteristiche strutturali che hanno buona probabilità di corrispondere anche a proprietà delle reali strutture macroscopiche (come lo sono le geometrie tridimensionali). Attraverso varie tecniche di analisi è possibile individuare cose come linee di bordo, angoli, aree uniformi e vari altri particolari di questo genere. Essi devono essere opportunamente classificati, in base anche ad informazioni quantitative (cose come la lunghezza, l'inclinazione, le estensioni delle aree, la colorazione, e molto altro...). Alcuni sottoinsiemi di queste informazioni, come descritto nel capitolo 5, si ripeteranno con regolarità e potranno permettere il riconoscimento di oggetti.

Il tutto richiede l'utilizzo di moltissime regole specifiche per agganciare i vari insiemi di elementi strutturali con le memorie dei reali oggetti macroscopici e delle loro ricostruzioni tridimensionali.

Ad ogni modo, per procedere nell'esposizione, chiedo al lettore di partire dall'ipotesi di lavoro che sia in qualche modo possibile risolvere i problemi della visione, e dotare il nostro sistema cognitivo elementare della capacità di ricostruire le "simulazioni tridimensionali" delle cose esterne.

Un punto importante è che da sole, simulazioni di questo tipo non sono ancora sufficienti per realizzare delle reali "forme di conoscenza" sul mondo esterno. Mancano alcune facoltà essenziali. Un moderno videogioco, per quanto realistico, non è ancora un sistema cognitivo. Ha senso chiedersi che cosa manca a sistemi di questo tipo per poter diventare dei sistemi intelligenti.

Usando le idee introdotte nei capitoli precedenti, potremmo rispondere che, tra le altre cose, manca la capacità di riconoscere le singole entità, quindi i vari soggetti cognitivi, e manca la capacità di riconoscere le situazioni che essi determinano. Manca inoltre la capacità di applicare a questi riconoscimenti le regole che permettono di generare previsioni e di pianificare i comportamenti in modo finalizzato. Procediamo però con ordine.

Notiamo che le rappresentazioni tridimensionali delle forme degli oggetti sono a tutti gli effetti delle **strutture**, quindi su di esse possiamo applicare quanto illustrato nei capitoli 3 e 5. Possiamo ad esempio applicare delle operazioni di derivazione strutturale come quelle di quoziente e di morfismo. Infatti sicuramente un sistema cognitivo non si può limitare a rappresentare le strutture tridimensionali delle cose, ma su queste rappresentazioni deve eseguire subito una serie di **operazioni di analisi di struttura**.

Questo punto è fondamentale perché proprio grazie a queste operazioni si può cominciare a costruire i primi strati di rappresentazioni astratte.

Questa capacità è assente nei videogiochi e in pratica in ogni simulatore attuale (anche in quelli che non si occupano di ricostruzioni tridimensionali).

Con queste prime astrazioni si aggiungono, sopra alle nostre rappresentazioni di base, molte importanti informazioni sulle **proprietà** degli oggetti e sulle loro reciproche **relazioni**. Per esprimere queste relazioni e proprietà strutturali delle rappresentazioni tridimensionali è necessario, come illustrato nel capitolo 5, **rendere esplicita** la loro presenza. Quest'operazione di **esplicitazione** è fondamentale, e si attua principalmente attraverso dei dispositivi dedicati che servono a "**segnalare al sistema cognitivo**" che i vari singoli elementi strutturali sono presenti entro la porzione del mondo circostante osservata (e ricostruita all'interno del sistema).

Una volta rese esplicite queste informazioni, esse possono essere selezionate per comporre i vari possibili morfismi. Come spiegato più volte, un'altra operazione importante è quella di quoziente che consente di eseguire dei "cambiamenti di scala" e passare ad osservare strutture di scala più grande, le cui nuove parti consistono in porzioni di quelle di scala inferiore.

Con queste operazioni sicuramente non si generano tutte le rappresentazioni astratte possibili, ma si mette a disposizione un primo substrato essenziale. Da questo il sistema potrà comunque ricavare il materiale per implementare le **prime famiglie di regole** che costituiscono **il primo strato di** "**conoscenza semantica**" di come funziona il mondo.
Un sistema cognitivo deve utilizzare queste regole per fare principalmente due cose:
- generare previsioni e
- pianificare azioni e comportamenti.

La costruzione della "conoscenza semantica" costituita dalle "regolarità" e dalle "regole" che permettono di anticipare gli eventi e di gestire le proprie azioni è l'altro ingrediente portante dell'attività cognitiva. Gran parte dell'impegno di un sistema di conoscenza è speso nel cercare di individuare le **regolarità** che si manifestano nelle simulazioni interne della realtà, e nel riuscire a sfruttare una parte di queste come **regole utili** per generare previsioni e pianificare le azioni.
Come spiegato, quanto appena illustrato è solo la prima fase di ciò che è necessario fare per costruire una "conoscenza semantica" che sia in grado di anticipare e, per quanto possibile, pilotare l'evoluzione della realtà in modo efficace. Va tenuto presente che oltre alle astrazioni di tipo strutturale ne sono possibili delle altre che le completano, ma che possono essere apprese solo con l'acquisizione di molta esperienza e solo partendo dal "substrato" costituito da quelle strutturali.
Come anticipato, questa "prima visione" costituisce una forte semplificazione delle cose. Ci serve in questo momento per introdurre le idee, ma dovrà essere sostituita con delle descrizioni più appropriate.

### 9.4 La necessità di un sistema di reti

Nel capitolo 5 abbiamo visto il "principio di convergenza delle verifiche". Esso afferma che, per rendere esplicito l'avvenuto riconoscimento di un singolo soggetto cognitivo, è necessario far convergere in un singolo "bit equivalente" di informazione, tutte le operazioni di "verifica" della presenza di tutti gli elementi che definiscono una specifica struttura o una specifica proprietà strutturale.
Questo principio **impone** una particolare struttura all'organizzazione e all'elaborazione delle informazioni. Essa può essere rappresentata con dei grafi orientati e può essere implementata concretamente tramite un sistema di reti di dispositivi attivi.
Ai nodi di questi grafi devono essere implementate delle **funzioni,** nel senso di algoritmi**,** che, secondo i casi, si occupano principalmente:
- di rendere espliciti i riconoscimenti dei singoli soggetti cognitivi;
- di implementare regole, principalmente di tipo associativo, ma anche di

tipo operazionale.

In questo lavoro propongo un modello di sistema cognitivo composto da una serie di **reti di dispositivi attivi** (non necessariamente fisici) che funzionano attraverso degli automatismi implementati al loro interno.
Tutte queste reti sono per alcuni aspetti simili tra di loro. **In tutti i casi esse si occupano di contenere rappresentazioni della realtà**. Come detto, alcune di queste rappresentazioni potranno essere molto concrete e particolareggiate, mentre altre saranno più astratte e schematiche.
Il modello prevede che spesso, nello stesso **sottosistema di reti**, siano contenute sia le rappresentazioni concrete, sia buna parte di quelle astratte, e che si passi dalle une alle altre attraverso una serie di livelli intermedi.
Le rappresentazioni sono però anche distinguibili in base alla loro funzione cognitiva. Si possono distinguere quelle che si riferiscono "**alla situazione del presente**" da altre che si riferiscono a **situazioni ipotetiche**, o a **situazioni potenziali,** che consistono in previsioni, o obiettivi da raggiungere, o schemi per la pianificazione delle azioni e dei comportamenti.
È importante tenere presente che le rappresentazioni con una diversa funzione cognitiva devono essere implementate su reti (e sottosistemi) differenti. In genere in un sistema cognitivo saranno contemporaneamente attive più rappresentazioni di "realtà diverse": situazione attuale, situazioni previste, situazioni obiettivo e altre ancora. Il sistema deve sempre essere in grado di distinguere quelle che sono le rappresentazioni che si riferiscono alla "realtà oggettiva", che sono quelle che provengono dai sensi, e quelle che sono invece il frutto di elaborazioni interne.
Il modello che propongo consiste in varie reti, organizzate in sottosistemi, che devono essere opportunamente interconnesse ma dedicate a scopi parzialmente diversi.
Si possono distinguere varie tipologie di reti; tra le più importanti vi sono:
- reti per le prime analisi delle informazioni sensoriali,
- reti che fungono da teatri virtuali,
- reti di analisi strutturale e riconoscimento dei singoli soggetti cognitivi,
- reti di analisi e classificazione funzionale,
- reti per l'implementazione delle regole di previsione,
- reti per gli obiettivi e i pericoli,
- reti per la pianificazione delle azioni e dei comportamenti (tra queste sono particolarmente importanti le "**reti suggeritrici**").

Buona parte di queste reti, seppur ben distinguibili perché appunto dedicate a contenere rappresentazioni con funzione diversa, devono essere "**parzialmente parallele**", nel senso **che devono condividere molti dei medesimi soggetti cognitivi** . Credo che si possa mostrare che l'essere un particolare soggetto cognitivo costituisce, in un senso che sarà chiaro più avanti, una "**dimensione trasversale**" rispetto alle varie reti.

### 9.5 Schema di funzionamento

Come affermato nel capitolo 8, semplificando un po', si può pensare di finalizzare l'attività di un sistema cognitivo assegnando un valore di desiderabilità o di indesiderabilità a certe particolari rappresentazioni di situazioni e a certi particolari singoli soggetti cognitivi che assumono i ruoli di **obiettivi da raggiungere** o di **pericoli da evitare**. Questi valori di desiderabilità/in desiderabilità possono, in linea di principio, essere costituiti da dei semplici valori numerici[9]. Non tutti i soggetti cognitivi avranno associati questo valore di desiderabilità, alcuni potranno quindi risultare "neutri".

In genere un soggetto cognitivo che è anche un **obiettivo da raggiungere** (quindi con associato un valore positivo di desiderabilità) sarà **"attivo" quando non è soddisfatto**, vale a dire quando non è presente nella situazione del presente, o in quelle previste. In maniera inversa un soggetto indesiderabile, come può esserlo un pericolo, sarà **"attivo" quando è presente** nella rappresentazione della situazione del presente o nelle situazioni previste.

Possiamo quindi pensare, schematizzando all'estremo, che un sistema cognitivo funzioni nella maniera che segue.

Ci sono reti, o meglio sistemi di reti, dedicate alle rappresentazioni dello stato del presente che ricevono informazioni dagli organi di senso. Queste informazioni sono opportunamente analizzate, interpretate e utilizzate per costruire una serie di rappresentazioni, sovrapposte per livelli gerarchici, della situazione attuale. Queste rappresentazioni sono contenute nel complesso degli stati di attivazione dei dispositivi che si trovano ai nodi di queste reti.

Come risposta a queste rappresentazioni, altre reti opportune dovranno produrre le previsioni probabili di come si "evolverà in modo naturale" tale situazione. Queste previsioni saranno generate tramite l'utilizzo delle regole "conosciute", che dovranno essere implementate all'interno di opportuni dispositivi, i cui output, di fatto, coincidono con i nodi delle reti dedicate alle previsioni ( e in taluni casi entro dei "**teatri virtuali**" che eseguono simulazioni basate su regole operazionali).

Queste reti (e questi teatri virtuali) genereranno quindi le proprie previsioni su come è destinata ad evolversi la "situazione del presente".

A questo punto se accade che nella situazione del presente, oppure in quelle previste, viene riconosciuto un soggetto che costituisce un pericolo, allora

---

[9] In realtà è necessario gestire questi attraverso delle funzioni euristiche: il valore di desiderabilità può variare nel tempo, in funzione delle priorità, dei bisogni fondamentali da soddisfare ecc..

verranno attivati automaticamente i nodi relativi nelle reti degli obiettivi e dei pericoli.

Queste attivazioni costituiranno dei segnali che inducono l'attivazione automatica di altre opportune "**reti suggeritici**", dedicate alla pianificazione delle azioni. Queste ultime cercheranno di identificare un comportamento che determini un'evoluzione degli eventi dove il pericolo non è più presente nelle previsioni per il futuro.

In modo molto simile si può pensare di gestire gli "obbiettivi da raggiungere". Se nella rete opportuna (quella degli obbiettivi e dei pericoli) è attivo un obbiettivo e questo non è riconosciuto nella situazione del presente o nelle previsioni per il futuro, le reti suggeritrici, dedicate alla pianificazione, saranno stimolate a cercare un comportamento che permetta di fare in modo che il soggetto obiettivo diventi invece presente nelle previsioni associate alle azioni suggerite.

## 9.6    Alcuni primi punti sul funzionamento delle reti

Secondo questo modello un sistema cognitivo è quindi costituito da vari insiemi di reti che si scambiano informazioni.

Queste reti sono composte di dispositivi che ricevono input e che possono produrre degli output.

Questi dispositivi implementano delle funzioni. Uso qui il termine "funzione" con un significato molto vicino a quello che assume in matematica e in informatica. Una funzione costruisce una "mappa" tra lo spazio dei possibili input, e quello dei possibili output, implementata attraverso operazioni computazionali. In molti dispositivi l'output consisterà in un singolo bit, o in un valore che "misura il grado di una corrispondenza". Questo bit potrà essere utilizzato per **"esplicitare" l'avvenuto riconoscimento di uno specifico soggetto cognitivo**. In altri casi potrà costituire un "comando" da spedire a un singolo attuatore (o a un fascio muscolare), o ad apparati in grado di eseguire specifiche azioni concrete.

In altri "gruppi di dispositivi", riuniti in moduli, l'output non consisterà in un singolo bit, ma in un'informazione più complessa, quindi dotata di una propria struttura. Molti di questi moduli implementeranno al loro interno delle funzioni di memoria. Si tratterà in genere di memorie di tipo attivo, spesso a "indirizzamento per contenuto". Il singolo dispositivo, o il singolo modulo di memoria eseguirà delle operazioni di "verifica sugli input ricevuti" e in funzione di queste "deciderà" (in modo automatico attraverso la funzione in esso implementata) se generare o no il proprio output.

Un punto importante è che, per la maggior parte del tempo, molti di questi dispositivi rimarranno **attivi ma silenti**, nel senso che riceveranno degli input, eseguiranno al loro interno una serie di operazioni (in genere di confronto), ma solo in taluni casi reagiranno producendo il loro output.

In generale quindi, all'interno di un sistema cognitivo devono essere contemporaneamente presenti molte rappresentazioni di situazioni diverse. Alcune di queste saranno "**attive in moto esplicito**" entro le reti del sistema. Moltissime altre saranno invece "**attive in modo silente**". Le rappresentazioni che rimangono silenti in genere si trovano all'interno dei singoli dispositivi che costituiscono le reti, mentre quelle esplicite compaiono anche come attività dei nodi delle reti.

Le varie reti conterranno **le conoscenze e le memorie semantiche**, costituite principalmente da quanto serve per rappresentare e riconoscere soggetti cognitivi di varia tipologia e per implementare le regole. I soggetti cognitivi potranno essere dettagli strutturali, rappresentazioni complete di singoli oggetti concreti, rappresentazioni di situazioni e fenomeni in divenire, o astrazioni di vario tipo.

In un singolo "istante", o meglio nel singolo "stato di attività momentanea" del sistema cognitivo, la grande maggioranza di queste memorie saranno **attive ma silenti**. Le reti del sistema devono, infatti, essere implementate in modo tale che tutte (o quasi) le memorie contenute entro di esse siano **vagliate in continuazione**. Il contenuto di questo memorie deve essere confrontato continuamente con le "rappresentazioni attive in modo esplicito", nelle reti dedicate alla rappresentazione del presente (ma anche in altre), alla ricerca di corrispondenze, o, più in generale, alla ricerca di condizioni tali che autorizzino i singoli dispositivi (che possono anche essere raggruppati in opportuni moduli) a proporre in output il proprio contenuto, o più precisamente, i risultati della **funzione** in essi implementata.

Entro queste reti possono quindi essere presenti sia dispositivi che svolgono principalmente la funzione di "memoria attiva" (indirizzabile per contenuto), sia dispositivi che si occupano più in generale di computare funzioni generiche.

Come detto, la grande maggioranza delle rappresentazioni e delle funzioni dovranno essere comunque attive ma silenti! I singoli dispositivi dovranno confrontare gli input ricevuti con quanto contenuto al loro interno, e solo in casi particolari, quando si presentano ben determinate condizioni, dovranno proporre il loro output.

Gli output dei dispositivi e dei moduli che si saranno attivati produrranno altre rappresentazioni di situazioni della realtà. Queste ultime saranno quindi le **rappresentazioni attive in modo esplicito**.

Nel seguito, per non appesantire la terminologia, indicherò semplicemente come **rappresentazioni attive** quelle che lo sono in modo esplicito, le altre saranno invece silenti.

Come affermato, alcuni dispositivi potranno produrre dei comandi, come ad esempio degli impulsi che pilotano i movimenti e quindi le azioni concrete. Anche questi comandi possono essere oggetto di rappresentazione. In genere prima di "decidere" di fare una cosa, quindi di "passare all'azione", si può rappresentare in modo interno, vale a dire in modo virtuale, l'azione stessa senza compierla effettivamente.

Un concetto importante è che le rappresentazioni attive in modo esplicito servono per spingere altre reti, quindi altre parti del sistema cognitivo, a "occuparsi di esse", a **reagire** ad esse.

In un sistema cognitivo, durante la sua normale attività, potranno essere **contemporaneamente attive** in modo esplicito rappresentazioni di situazioni ben diverse, che si riferiscono sia alla **realtà oggettiva**, sia a **realtà ipotetiche** (o potenziali). Un sistema cognitivo deve rappresentare quello che sta accadendo in questo momento, ma deve generare anche previsioni per il futuro. Come visto deve inoltre generare rappresentazioni che costituiscono gli obiettivi da raggiungere e che quindi contribuiscono a definire i "problemi da affrontare". I sistemi più evoluti possono anche essere in grado di generare rappresentazioni che si riferiscono a "fatti e situazioni raccontati da altri" tramite il linguaggio, o anche rappresentazioni che costituiscono dei "prodotti di fantasia".

È necessario che un sistema sia sempre in grado di distinguere, in modo chiaro, ciò che è realtà oggettiva e ciò che invece è un prodotto delle proprie elaborazioni interne. Anche per questo motivo è utile che alcune delle varie rappresentazioni delle diverse "realtà" (o, per meglio dire, delle diverse situazioni reali o ipotetiche che il sistema sta "valutando" in un certo momento) siano contenute in reti (e in "teatri virtuali") ben distinte.

Per tracciare un primo modello del funzionamento globale di un sistema cognitivo può essere utile approfondire alcuni punti salienti di quanto fin qui presentato.

### 9.7 Teatri virtuali

Il compito del sistema di reti che acquisiscono ed elaborano le informazioni sensoriali primarie è quindi quello di ricostruire e di simulare la realtà a più livelli.

Alcune di queste rappresentazioni saranno molto concrete e potranno essere utilmente implementate all'interno di apparati che qui indico con la terminologia generica di "**teatri virtuali**".

Un esempio importante di teatro virtuale è un simulatore tridimensionale delle geometrie degli oggetti concreti mentre si muovono. Gli equivalenti di apparati di questo genere possono essere utilizzati all'interno di un sistema cognitivo per contenere le "**rappresentazioni estese**" delle "strutture emergenti macroscopiche". La ricostruzione tridimensionale della geometria di un oggetto concreto, come quella di una sedia, è proprio un esempio di **rappresentazione estesa di una struttura emergente macroscopica**.

Ricordo che, come visto nel capitolo 5, le strutture emergenti macroscopiche sono quelle che si manifestano a un livello di grandezza direttamente osservabile tramite i sensi. All'interno di un sistema cognitivo generico possono essere necessari non solo apparati per rappresentare le geometrie degli oggetti, ma anche apparati dedicati alle altre tipologie di "entità macroscopiche", quali i suoni, "la realtà olfattiva", o le rappresentazioni "sonar" dell'ambiente che utilizzano alcune specie animali, e varie altre ancora. In funzione della tipologia delle entità da rappresentare servono apparati diversi, accomunati dal fatto di contenere appunto le **rappresentazioni estese** (che sono quelle più basilari) delle **reali** strutture macroscopiche. Si noti che l'immagine bidimensionale di un oggetto, proiettata sulla retina o su una fotografia, non ne rappresenta in modo completo la struttura macroscopica, perché l'oggetto stesso è tridimensionale. Invece la ricostruzione della sua geometria tridimensionale, all'interno di un adeguato simulatore, corrisponde in modo molto più accurato alla sua reale struttura.

Per generalizzare utilizzo la terminologia di "**teatro virtuale**" per indicare ogni apparato, o porzione di rete, dedicato a contenere le rappresentazioni estese delle intere strutture macroscopiche di entità concrete. I teatri virtuali possono spesso essere usati come "**simulatori di basso livello**".

## 9.8   I limiti dei sensi

Idealmente in un sistema cognitivo perfetto (che non esiste) gli organi di senso dovrebbero fungere direttamente da teatri virtuali. Ad esempio, in un sistema ideale, gli organi di senso visivi dovrebbero fornire direttamente le ricostruzioni tridimensionali delle cose vicine e lontane che cadono sotto lo sguardo. Per motivi legati alla fisica del mondo in cui viviamo non appare possibile implementare sensi di questo tipo.

Nei sistemi reali si pone quindi l'ulteriore complicazione dovuta al fatto che le informazioni sensoriali sono spesso molto distanti dall'essere direttamente delle buone rappresentazioni delle strutture macroscopiche. Nella maggior parte dei casi, le informazioni sensoriali primarie sono solo degli "**indizi**", spesso confusi e parziali, delle reali strutture macroscopiche.

Per questo motivo buona parte dell'attività delle prime reti che raccolgono ed elaborano le informazioni sensoriali è dedicata alla trasformazione di queste nelle strutture macroscopiche e al "riconoscimento diretto" delle singole cose (quindi dei soggetti cognitivi corrispondenti). Questo problema è però in realtà molto difficile. Anche le reti di analisi primaria devono essere costituite da molti dispositivi che si occupano di analizzare e riconoscere singoli elementi strutturali. Queste reti dovranno essere organizzate per livelli. I dispositivi che si trovano in "basso", vale a dire a contatto con le prime informazioni che costituiscono gli stimoli prossimali dovranno in genere occuparsi di dettagli strutturali; salendo lungo la gerarchia si passerà gradualmente a dispositivi che si occupano dell'analisi e del riconoscimento di entità più complesse.

Come detto, spesso le reti di analisi primaria identificano combinazioni di caratteristiche strutturali che consentono un "**riconoscimento diretto**" che può avvenire senza passare per una ricostruzione completa della struttura macroscopica di quanto osservato. Spesso inoltre questo riconoscimento diretto può essere utilizzato per fornire informazioni agli stessi teatri virtuali.

Quindi le reti di analisi possono mandare informazioni sia verso i teatri virtuali sia verso le reti di riconoscimento, che a loro volta spediscono informazioni "all'indietro" verso i teatri virtuali (questo perché in genere le reti che eseguono i riconoscimenti si dovrebbero trovare "dopo" i teatri virtuali e dopo quelle di analisi di struttura).

## 9.9 Reti di analisi di struttura

In un sistema cognitivo **ad ogni teatro virtuale dovrà essere associata almeno una rete di analisi di struttura** (ma in genere saranno più di una). Il compito di queste reti è di eseguire sulle rappresentazioni generate entro i teatri virtuali (e talvolta a livello di stimoli sensoriali) una serie di operazioni di analisi, in buona parte costituite da derivazioni strutturali. Queste sono essenziali per riconoscere i singoli soggetti cognitivi e per passare dalla rappresentazioni concrete a quelle astratte.

I teatri virtuali stanno spesso quindi alla base dei processi di rappresentazione del mondo esterno, ma sopra di questi dovranno essere attive una serie di reti di analisi nelle quali si generano le rappresentazioni più astratte delle cose e delle situazioni.

Sottolineo ancora una volta che il riconoscimento dei singoli soggetti cognitivi, anche degli oggetti concreti, non può avvenire a livello delle rappresentazioni di base ricostruite dentro i teatri virtuali, **ma può avvenire solo sui prodotti delle reti di analisi;** quindi **l'analisi strutturale delle ricostruzioni di base è fondamentale per il riconoscimento delle "singole cose"**.

Alle rappresentazioni di base dovranno corrispondere quindi vari strati di altre rappresentazioni che, pur occupandosi delle stesse cose e delle stesse situazioni, le ritraggono a livello di astrazione via via crescente, passando prima per le astrazioni strutturali e poi per quelle funzionali e logiche.

Non è difficile concepire tecniche di analisi in grado di rendere esplicite le principali caratteristiche strutturali che possono essere presenti nelle rappresentazioni di base. Va detto che una cosa è comprendere che il problema è teoricamente affrontabile, identificando delle possibili strategie per condurre l'analisi, e un'altra è trovare degli algoritmi efficienti, in grado di riuscire a portare a termine il compito in modo efficiente, in tempi brevi e con un hardware ragionevole (e compatibile con i sistemi biologici.

Per alcuni aspetti i problemi da affrontare sono simili, e forse più semplici, di quelli che si incontrano nell'analisi delle informazioni visive. Si possono concepire e scrivere algoritmi che identificano e classificano cose come le linee di bordo, le aree, gli angoli, le lunghezze maggiori, le forme delle superfici ecc… e quantificano queste informazioni con misure appropriate.

Si tenga presente che in genere percepiamo contemporanea-mente gli effetti di più oggetti ed è quindi naturale che tipicamente con queste tecniche si identifichino molte caratteristiche strutturali che appartengono a vari oggetti diversi. Serve quindi trovare qualche strategia che permetta di identificare e "mettere assieme" le caratteristiche strutturali che appartengono al medesimo soggetto. Si possono pensare vari criteri pratici, ad esempio spesso le caratteristiche che si trovano vicine nello spazio hanno buona probabilità di appartenere al medesimo oggetto reale; ancora meglio se queste si muovono assieme. Ma credo esista un criterio molto generale, che consiste nel mettere assieme quelle caratteristiche che si dimostrano correlate, vale a dire quegli insiemi che sono "regolari", che tendono a ripetersi.

Questo criterio è particolarmente importante perché è strettamente connesso con il principio generale che, come visto nel capito 5, legittima i singoli soggetti cognitivi: il servire a identificare ed implementare delle regole utili. In effetti i singoli oggetti e i singoli fenomeni possono anche essere considerati come delle "regolarità" che si manifestano nella realtà. Un oggetto conserva la propria struttura nel tempo. Quindi ogni volta che lo si rappresenta, e si procede ad eseguire su di esso delle operazioni di analisi di struttura, ci saranno delle regolarità in alcune combinazioni delle rispettive caratteristiche strutturali, specialmente per alcuni morfismi.

Sono molto importanti le relazioni spaziali tra gli oggetti. Abbiamo visto infatti che è importante riuscire a descrivere le "situazioni".

Nei casi più semplici in un sistema cognitivo "rudimentale", vale a dire poco evoluto, le situazioni possono essere descritte semplicemente specificando quali entità concrete sono presenti e come sono relazionate le une rispetto alle altre. Quindi, per fare un esempio molto semplice, nel caso più basilare una situazione può consistere in una certa disposizione spaziale di particolari oggetti. Entro questo contesto, un "obiettivo da raggiungere" può consistere in un'altra situazione dove gli stessi oggetti assumono una disposizione diversa. Il compito del sistema è "risolvere il problema dato", che in questo caso consiste nel pianificare una serie di movimenti in grado di condurre dalla situazione di partenza a quella obiettivo, quindi di spostare gli oggetti e ridisporli in modo che si trovino nella disposizione voluta.

Credo sia importante notare che molto spesso le relazioni spaziali tra gli oggetti possono essere espresse anche senza specificare informazioni quantitative sulla loro posizioni assolute (valori numerici che ne esprimono le distanze e angolazioni), ma "dichiarando" relazioni "più qualitative", che corrispondono, ad esempio, a concetti come quelli di: stare sopra, essere vicino, stare di lato, essere dentro ecc.. Molto spesso, per i reali problemi pratici (per il complesso delle regole emergenti utili per pianificare i comportamenti) interessa semplicemente che l'oggetto X sia sopra l'oggetto Y, o dentro l'oggetto Y e non importano le coordinate precise che questi hanno gli uni rispetto agli altri. Da questo punto di vista ci sono moltissimi modi nei quali X può essere sopra a Y, o dentro a Z. Non importa in quale posizione precisa si trova la confezione di zucchero dentro la nostra borsa della spesa quando ci accingiamo a trasportare il tutto, è un'informazione che non ci interessa. Ci interessa semplicemente che sia vero il fatto che la confezione di zucchero è all'interno della borsa.

Sicuramente non è banale codificare e riconoscere queste relazioni spaziali qualitative, ma non è nemmeno impossibile, si possono pensare varie strategie algoritmiche potenzialmente in grado di identificarle.

Come detto quelli illustrati sono esempi di situazioni e di problemi molto basilari. Chiaramente un sistema cognitivo deve avere la capacità di rappresentare situazioni ben più astratte, deve essere in grado di riconoscere situazioni dove i "soggetti protagonisti" non sono necessariamente oggetti concreti specifici, ma sono invece classi astratte, e dove le loro relazioni non si limitano a quelle di "disposizione reciproca", ma includono relazioni di tipo logico e funzionale. Ma per descrivere situazioni di livello di astrazione così elevato bisogna avere prima acquisito la capacità di rappresentare le situazioni più concrete, e le capacità che permettono di risolvere i problemi più pratici, come quello di infilare degli oggetti dentro una borsa, di "comprendere" che in questo modo si possono trasportare più oggetti in una volta sola e che questo comporta un notevole vantaggio rispetto al trasportarli uno alla volta (ma questa in realtà è già una facoltà che appartiene all'intelligenza superiore).

### 9.10 Analisi che sfruttano il contenuto informativo interno

Un trucco molto potente per l'analisi di struttura, del quale ho già illustrato alcuni punti, consiste nello sfruttare direttamente il contenuto informativo interno per identificare a priori quali porzioni di struttura hanno buona probabilità di essere parti emergenti di strutture quozienti.

L'idea è di eseguire le varie classificazioni possibili di piccole porzioni di struttura, e di utilizzare queste per confrontare tra loro le porzioni adiacenti. Poiché si tratta di porzioni piccole, in genere le classificazioni possibili saranno un numero limitato (ma che in taluni casi può crescere velocemente). La tecnica consiste nel verificare se sono conservate o no delle regolarità interne: si tratta allora di verificare se a "piccola scala" le strutture si assomigliano o se invece sono diverse. Quelle che si assomigliano o coincidono, vanno composte assieme per generare la **proposta** di una porzione più grande, e si continua così fino a che non s'incontra una "**rottura nella regolarità**".

Un processo di questo tipo (con le ovvie varianti del caso) può essere utilizzato non solo per identificare porzioni di volumi e/o aree superficiali, ma anche per identificare un'altra serie d'importanti elementi strutturali e di loro informazioni quantitative significative. Si possono, ad esempio, identificare linee di bordo, si può contare il numero di occorrenze "dei pixel componenti" per esprimere le loro lunghezze, e si possono anche valutare il loro grado di drittezza e di curvatura.

Si noti che ai nuovi "mini elementi strutturali" che di volta in volta si generano, in funzione della loro forma, corrisponderanno delle "relazioni esterne" più complesse della semplice adiacenza. Ad esempio due porzioni di segmenti "contigui" staranno, l'uno rispetto all'altro, secondo particolari angoli. Le regolarità tra gli angoli di "mini segmenti contigui" può essere uno dei criteri per "metterli assieme" in una linea di "curvatura uniforme" e per caratterizzarla. Riflettendo sul senso generale di questo processo, si può capire che dovrebbe essere ripetuto per gerarchie di elementi strutturali e che tende a generare molte informazioni, forse troppe.

Si rifletta sui seguenti punti. In un oggetto solido reale i suoi lati, i suoi spigoli, i suoi bordi, sono elementi strutturali oggettivamente emergenti. Essi, infatti, determinano come l'oggetto può stare rispetto ad altri, determinano l'uso che se ne può fare, come può incastrarsi o accostarsi ad altri, se gli si può posare sopra qualcosa, eccetera. Questi elementi strutturali possono essere individuati con la corretta operazione di quoziente, passando dalle rappresentazioni estese della sua geometria, a quella dove sono evidenziate, e rese esplicite, le sue "nuove parti componenti". Questi elementi sono a tutti gli effetti entità emergenti. Sono, infatti, entità per le quali esistono molte regole che dipendono da esse. Queste parti devono essere opportunamente classificate, "misurate", e ben descritte in termini delle loro eventuali strutture interne.

Molti di questi elementi strutturali sono proprio quelli che è possibile individuare applicando le tecniche di analisi per contenuto informativo interno (forse potremmo usare il termine: "per composizione delle regolarità interne").

Sembrerebbe che una tecnica di questo genere applicata alle rappresentazioni tridimensionali, sia destinata a generare elementi strutturali che sono già "molto vicini" a quelli emergenti.

Questo è molto interessante ed è vero per le rappresentazioni delle geometrie tridimensionali degli oggetti concreti, ma per quanto riguarda le immagini bidimensionali cosa succede? In fondo le informazioni visive sono in buona parte a due dimensioni.

Nelle immagini, gli elementi separabili con le tecniche che sfruttano il contenuto informativo interno hanno solo una certa probabilità di costituire la proiezione visiva bidimensionale di un reale elemento strutturale. Molti dei contorni che si estraggono nell'analisi delle immagini sono illusori, altri sono incompleti, e spesso capita che sulle superfici degli oggetti siano disegnate delle figure, che presentano linee, superfici e altro, senza corrispondere a reali proprietà tridimensionali. Questi inconvenienti fanno si che quando si analizzano con queste tecniche delle immagini bidimensionali, le "caratteristiche" estratte costituiscono solo "**degli indizi**" sulle reali strutture tridimensionali.

Il problema di trovare le corrette correlazioni tra questi indizi e gli elementi strutturali tridimensionali effettivamente emergenti, è molto complesso. Nei primati e nell'uomo a questo problema è dedicata una parte rilevante della corteccia cerebrale.

Queste tecniche possono dunque contribuire a semplificare i processi di analisi, ma resta il fatto che esse sono caratterizzati da due aspetti cruciali: sono tali da generare spesso una "esplosione delle possibilità" e non esiste un vero criterio per potere stabilire a priori se una certa operazione identifica una reale proprietà strutturale emergente!

Questi fatti implicano che il problema dell'analisi di struttura sia intrinsecamente oneroso dal punto di vista delle risorse computazionali e di memoria richieste. Se dovessimo affrontare il problema per "forza bruta", dovremmo escogitare algoritmi in grado di rendere espliciti tutte le possibili caratteristiche strutturali e dovremmo cercare successivamente di selezionare a posteriori quelle che funzionano. Con buona probabilità far questo è troppo oneroso. È probabile che in natura sia stato il processo di selezione naturale a fare una parte di questo lavoro, in centinaia di milioni di anni.

### 9.11  Una possibile ricetta per l'analisi di struttura

Una ricetta per l'analisi di struttura che si occupa di quella che possiamo indicare come "astrazione interna" potrebbe essere la seguente: si esegue il passaggio dalla struttura di partenza a altre strutture quozienti. Nel far questo, si esplicitano tutte le classificazioni possibili delle proprietà interne delle sue parti e del complesso delle loro relazioni esterne. Si eseguono quindi i vari morfismi possibili su queste informazioni esplicite.

Spesso le porzioni che hanno buone probabilità di costituire le "nuove parti di struttura" per le varie operazioni di quoziente, si possono individuare a priori, sfruttando **il contenuto informativo interno**. Probabilmente anche in questo caso è necessario utilizzare particolari algoritmi di analisi, che nei sistemi biologici sono stati selezionati dall'evoluzione.

Probabilmente la separabilità per contenuto informativo interno di un certo elemento strutturale costituisce una **condizione necessaria, ma non sufficiente,** affinché questo sia anche un legittimo soggetto cognitivo. Quindi non tutti gli elementi strutturali, e le derivazioni così ottenute, saranno anche legittimi soggetti cognitivi. Lo saranno solo quelli che sono utilizzabili per codificare qualche regola valida, e questo potrà essere stabilito **solo a posteriori**.

Quest'ultimo punto è particolarmente importante, perché implica che alla base della messa a punto dell'analisi di struttura deve esserci un processo che procede sostanzialmente per tentativi e per verifica a posteriori. Si deve procedere generando le esplicitazioni delle "probabili" proprietà strutturali interessanti, si deve verificare se queste contribuiscono a identificare qualche regola valida, e ciò può avvenire solo provando a utilizzare concretamente queste regole. Se esse consentono di generare previsioni che funzionano, o di pianificare azioni che hanno successo, allora significa che sono corrette, e quindi lo sono anche i soggetti cognitivi che servono per la loro implementazione.

### 9.12 La rappresentazione complessiva della situazione globale del presente

Quanto abbiamo visto finora riguarda la rappresentazione dello stato della realtà esterna al presente. Riassumendo, l'insieme costituito dalle prime reti di analisi sensoriale, dai teatri virtuali, da tutte le reti di analisi strutturale che agiscono su questi ultimi, da reti (o estensioni di queste) che servono a contenere le memorie a breve termine del "passare del tempo" e "dei fatti contingenti" (e varie altre ancora), costituisce il complesso di reti ed apparati dove si realizzano le rappresentazioni dello **stato della situazione del presente**, vale a dire della **realtà oggettiva, attuale, esterna**, conosciuta dal sistema.

In un sistema cognitivo non rudimentale si può in realtà identificare un sistema di rappresentazioni ancora più esteso, che non contiene solo le informazioni sul mondo esterno, ma che comprende anche le informazioni sullo "**stato interno del sistema**": su cosa il sistema sta facendo e in particolare sugli obbiettivi e sui pericoli che in questo momento sono attivi.

Questa "**super rappresentazione globale**" è fondamentale e può essere considerata come il punto di partenza di tutte le altre attività. Pressoché tutte le altre reti del sistema non possono ignorare ciò che è "attivo esplicitamente" in questo complesso di reti e apparati: esse sono chiamate a **reagire** a queste rappresentazioni.

Entro questa rappresentazione globale sono in realtà distinguibili molte rappresentazioni "locali", che ritraggono sia aspetti diversi e specifici della realtà, sia le medesime "entità e situazioni", osservate però a diversi livelli di astrazione.

Molte porzioni di reti a valle degli apparati appena menzionati saranno sensibili solo a queste "rappresentazioni parziali" più locali, spesso astratte. Si tenga presente che in genere le singole regole sono sensibili a un numero limitato di soggetti cognitivi, che, stando gli uni rispetto agli altri in specifiche relazioni costituiscono delle "**microsituazioni**".

Alle nuove informazioni sensoriali, che servono ad aggiornare la rappresentazione della situazione del presente (o meglio: che aggiornano le varie rappresentazioni che nel loro complesso formano quella globale), il sistema cognitivo **dovrà reagire esplorandone le implicazioni**. Ciò significa che il sistema dovrà verificare se le nuove informazioni hanno implicazioni sul complesso degli "in soddisfacimento", se contengono dei nuovi pericoli cui far fronte o nuove prospettive per attivare nuovi obbiettivi.

Queste verifiche si fanno sia vagliando, con le nuove informazioni, le regole di gestione degli obbiettivi, sia generando previsioni sui "futuri probabili" che si possono determinare in assenza di azioni da parte del sistema stesso.

Queste attività saranno generate in modo automatico dalle reti preposte. Come detto le varie reti del sistema sono infatti sempre attive, anche se silenti; continuamente vagliano lo stato della realtà presente (e prevista), vale a dire lo stato dei nodi delle reti che contengono le relative rappresentazioni, e se trovano le condizioni opportune proporranno i propri output.

Quindi "in parallelo" alle reti e agli apparati dedicati alla rappresentazione della realtà al presente, ci saranno altri sistemi di reti che generano, se ci sono le condizioni, altre rappresentazioni. Come detto, ci saranno **reti di previsione**, **reti per la rappresentazione degli obbiettivi e dei pericoli attivi**, reti per la pianificazione delle azioni, e altre ancora.

### 9.13 La conoscenza semantica in un sistema cognitivo

La conoscenza semantica all'interno di un sistema cognitivo si compone di due componenti principali: ciò che serve a rappresentare la realtà e le sue situazioni, e ciò che serve per utilizzare in modo proficuo le sue regole.

Semplificando un po', ma non troppo, credo sia corretto affermare che, nell'ambito dell'intelligenza naturale, le regole sono utilizzate principalmente per due scopi: per generare previsioni e per pianificare (e guidare) azioni e comportamenti.

Possono essere utilizzate anche per "scoprire delle verità" generiche, quindi per aggiungere informazioni su come è fatto il mondo, ma questo avviene nelle forme di intelligenza più avanzata. Per il momento propongo di concentrarci sui primi due aspetti menzionati.

Va premesso che un sistema cognitivo può generare previsioni e pianificazioni utilizzando varie strategie. In linea di principio, e spesso anche in pratica, può in alcuni contesti sfruttare i propri simulatori interni, i propri teatri virtuali, per ricostruire le rappresentazioni dei movimenti degli oggetti, quindi per "seguire l'evoluzione dei fenomeni" in modo molto diretto. In questo caso esegue sostanzialmente delle "simulazioni a basso (o nullo) livello di astrazione". Alcune di queste possono essere affidate interamente ai simulatori, anche se in genere per tempi brevi, e possono quindi essere generate sfruttando poche regole operazionali. Queste ultime consistono nell'applicazione di quelle leggi della fisica e della geometria che permettono appunto di simulare, per tempi brevi, l'evoluzione del particolare fenomeno in oggetto. Possiamo fare riferimento all'esempio di un simulatore 3D che può essere in grado di seguire e anticipare i movimenti di oggetti, simulandone per alcuni tratti le traiettorie.

In taluni casi lo stesso simulatore può essere usato per compiere inferenze, ad esempio per stabilire se una certa azione si può fare o se una certa congettura può essere o non essere vera (ad esempio per stabilire se una particolare forma geometrica si può incastrare con altre).
Nonostante queste possibilità, sospetto che il nostro cervello utilizzi fortemente, anche per le simulazioni di basso livello, regole di tipo associativo. Penso siano regole di questo tipo a gestire e a "sorvegliare" le regole operazionali che si possono usare nei teatri virtuali.
In effetti molti dei fenomeni reali, anche quando sono simulati a basso livello, non sono prevedibili nei dettagli utilizzando solo regole operazionali, se non per tempi relativamente brevi e in casi particolari. Ad esempio, se l'oggetto che si muove è un corpo solido, e se non ci sono ostacoli lungo la sua traiettoria, allora una simulazione che sfrutta la codifica del suo stato di moto potrà funzionare correttamente fino a quando l'oggetto non viene a contatto con una altro. Se l'urto è "semplice", come ad esempio quello di una palla su un muro, allora si possono, in linea di principio, usare regole operazionali anche per fare previsioni su questo evento. Ma in genere, non appena la collisione è un po' complessa, non esiste alcuna possibilità pratica di simulare realmente l'evento usando le "leggi fondamentali": le piccole inevitabili imprecisioni nella conoscenza delle condizioni iniziali comportano l'effetto di rendere comunque imprevedibile l'evoluzione reale dei fenomeni già dopo pochi passaggi. Se si vuole avere qualche possibilità di generare delle previsioni di qualche utilità è necessario fare intervenire regole di tipo associativo. Queste regole non saranno in grado di prevedere gli eventi in tutti i loro dettagli con precisione quantitativa, ma saranno invece in grado di fare previsioni in un certo senso "più qualitative", che funzionano su **astrazioni strutturali** degli eventi simulati. Nonostante queste limitazioni, previsioni di questo genere sono comunque di grandissima utilità. Sono fondamentali nelle nostre azioni quotidiane dove, ad esempio, abbiamo continuamente a che fare con oggetti che non hanno una forma rigida (si pensi a vestiti, coperte, cavi, lacci, liquidi, fluidi di vari consistenza, sostanze in forma granulare, ecc). Nelle nostre azioni non siamo in grado di prevedere esattamente i movimenti di questi oggetti, istante per istante, nello loro singole parti componenti. Ma siamo in grado in moltissimi casi di prevedere appunto quale sarà l'evoluzione "qualitativa" e il risultato finale del fenomeno in atto perché conosciamo, in modo associativo, molte regole specifiche sul loro comportamento. Se buttiamo per aria un mucchio di foglie secche, ci è impossibile prevedere la traiettoria di ogni una di esse. Ma sappiamo benissimo prevedere quale sarà l'effetto globale del fenomeno e soprattutto quale sarà l'effetto finale (anche se non nei singoli dettagli). Si noti che noi siamo in grado di generare previsioni di questo tipo, ma un moderno simulatore 3D, come un videogioco, non può che procedere simulando lo spostamento di ogni singola foglia e processando i dati "inventandosi" degli spostamenti plausibili, usando in genere algoritmi che utilizzano generatori di numeri casuali. L'effetto globale della simulazione al calcolatore sarà "qualitativamente" simile a quello da noi previsto, ma in nessuno dei due casi sarà esatto nei dettagli.

Mi pare abbastanza evidente che le previsioni che noi siamo in grado di fare si basano sulle esperienze passate. Sfruttiamo il fatto di aver osservato in passato situazioni simili, e ci aspettiamo che il nuovo evento mostrerà alcuni aspetti globali simili a quelli già sperimentati. Credo che queste similitudini, queste corrispondenze, riguardino una serie di proprietà e di relazioni strutturali comuni nelle varie ripetizioni dei fenomeni in oggetto, che possono essere identificate e quindi riconosciute proprio grazie ad operazioni di analisi strutturale. Si tratta già di rappresentazioni un po' astratte, anche se non sono necessariamente astrazioni molto spinte.

Le esperienze del passato e quanto si esperimenta al momento sicuramente non coincideranno nei dettagli, ma spesso saranno presenti una serie di proprietà comuni, che consisteranno in configurazioni simili di "soggetti cognitivi strutturali". Nell'esempio di prima avremo sempre che il mucchio di foglie si "separa in aria", "aumentando di volume", "raggiungendo una certa altezza" "non troppo elevata", "le foglie ondeggeranno", " e ricadranno al suolo", "sparpagliandosi per un area più grande". Tutti questi aspetti del fenomeno in oggetto sono legittimi soggetti cognitivi di tipo strutturale, che sono presenti in tutti i lanci di mucchi di foglie secche. Un sistema cognitivo deve essere in grado di riconoscere la loro presenza in maniera indipendente dai dettagli specifici della forma esatta del mucchio di foglie!

Usando quindi regole associative che connettono particolari insiemi di soggetti cognitivi, riconoscibili dall'analisi strutturale, siamo in grado di eseguire utilissime **previsioni "qualitative"** sull'evoluzione degli eventi. Non dobbiamo farci ingannare dall'uso del termine qualitativo. Nonostante non siamo in grado di prevedere tutti i dettagli, le nostre previsioni possono essere molto accurate per molti dei soggetti cognitivi che saranno presenti. Se lanciamo in aria un oggetto in uno spazio aperto, potremo non essere in grado di prevedere esattamente dove ricadrà, ma sapremo prevedere con pressoché assoluta sicurezza che esso dopo un po' tornerà a terra. Questo tipo di fatti, di risultati finali delle nostre azioni, possono essere previsti con ottima sicurezza e possono essere sfruttati. Queste previsioni "qualitative" sono davvero fondamentali nella nostra esistenza: le utilizziamo continuamente per fare previsioni e per poter usare le "nostre ricette comportamentali" per agire sul mondo. Se stendiamo con un ampio gesto la tovaglia sul tavolo, non siamo in grado di prevedere esattamente come questa ondeggerà prima di posarsi su di esso, ma siamo in grado di prevedere la presenza di alcuni aspetti comuni a tutti gli ondeggiamenti, e siamo in grado di prevedere che ricadrà sulla superficie e che con pochi gesti saremo in grado di sistemarla nel modo voluto. Se versiamo della pasta in un piatto, non siamo in grado di prevedere che forma esatta assumerà il "mucchio", ma sappiamo prevedere che non uscirà da questo se evitiamo di "versarne troppa".

Si consideri che molte delle "entità" che si muovono seguono traiettorie che non sono affatto balistiche, e modificano il proprio movimento secondo logiche che non sono simulabili usando semplicemente le leggi fisiche. Tuttavia il risultato di una parte importante dei loro movimenti possono essere anticipati con una certa accuratezza utilizzando regole associative e regole di tipo misto. In effetti una parte importante delle "cose interessanti che si muovono" sono animali, persone e macchine! Gli effetti dei loro movimenti e delle loro azioni, sono spesso ben prevedibili, anche se non nei dettagli.

Penso sia inoltre spesso possibile integrare regole associative ed operazionali. Molti movimenti di animali, persone e cose, sono costituiti da una successione di "fasi balistiche", separate dall'applicazione delle forze che servono per correggere le traiettorie o per intervenire in maniera netta sul cambiamento dello stato di moto. È probabile che molte regole associative possano essere usate per decidere quali regole operazionali utilizzare per prevedere come si svolgerà un movimento durante le sue fasi balistiche. Alcune regole operazionali sono anche applicabili per valutare le accelerazioni. Penso quindi che le nostre capacità di prevedere come si svolgono i movimenti consistano nella composizione di fasi dove le traiettorie sono ben prevedibili, e fasi dove invece non si è in grado di proporre previsioni accurate ma solo "qualitative", come già illustrato più indietro.

Queste idee portano quindi a pensare che sia possibile generare questo tipo di previsioni, che sono tutto sommato ancora di "basso livello di astrazione", facendo lavorare in profonda sinergia regole associative e operazionali.

Le regole associative possono servire sia per decidere, di volta in volta, quali regole operazionali applicare per la simulazione di parte dei movimenti, sia per fornire comunque delle previsioni (seppure di tipo qualitativo) per le fasi dei movimenti o delle evoluzioni dei fenomeni che non si possono simulare con precisione.

È probabile che le regole associative richiedano un hardware diverso da quello necessario per implementare al meglio molte regole operazionali. Credo sia da esplorare l'ipotesi che nel sistema nervoso questi ruoli siano affidati a parti diverse del cervello: alla corteccia potrebbe essere affidato principalmente, anche se non esclusivamente, il compito di implementare regole associative, mentre al cervelletto quello di occuparsi di mettere a punto, e implementare poi di fino, alcune tipologie particolari (probabilmente non tutte) di regole operazionali. Ripeto: si tratta solo di una ipotesi di lavoro che potrebbe risultare errata.

Quanto fin qui descritto vale per regole di basso livello di astrazione. In un sistema cognitivo sono però estremamente importanti le regole che coinvolgono invece rappresentazioni più astratte, quelle che utilizzano principalmente soggetti cognitivi di alto livello.

Penso che quando si sale con il livello di astrazione prevalgono nettamente regole associative. Si tratterà di regole che connettono particolari codifiche logiche, anche con metodi fuzzy, di insiemi strutturati di soggetti cognitivi. Alcune di queste codifiche svolgeranno il ruolo di "cause", altre il ruolo di "effetti".

Credo che una parte importante delle regole associative che utilizziamo per generare previsioni sia ricavata sostanzialmente dall'osservazione di quanto accade. Come visto, credo che un sistema cognitivo possa riuscire, analizzando in modo opportuno le informazioni sensoriali, a ricostruire delle rappresentazioni delle cose del mondo e dei suoi fenomeni, e possa anche individuare in queste una serie di regolarità, molte delle quali potranno essere poi implementate nelle reti del sistema sotto forma di regole utili sia a generare previsioni che per altre tipologie di inferenze. È probabile che una parte importante di queste regole possano essere ricavate dalle registrazioni delle varie rappresentazioni, sovrapposte gerarchicamente, dello svolgersi degli eventi. Altre regole invece dovranno essere in un certo senso "inventate dal sistema", che dovrà procedere per tentativi.

### 9.14 Reti di memorie da vagliare in continuazione per l'implementazione di regole

Ritengo importante insistere sul concetto che in un sistema cognitivo i dispositivi di memoria non servono solo a immagazzinare ricordi, ma sono fondamentali per l'implementazione di buona parte delle regole che costituiscono il "motore" dell'attività cognitiva. Nel modello che propongo le principali reti di un sistema devono incorporare miriadi di memorie attive, in grado di analizzare gli input che ricevono, di confrontarli con il proprio contenuto e di valutare quando è il caso di proporre il proprio output. Queste reti sono sempre attive, vagliano di continuo le informazioni che ricevono e sono suddivisibili in sottosistemi che lavorano spesso contemporaneamente sui medesimi input.

I concetti di **memoria da vagliare di continuo**, insieme a quello di "indirizzamento speciale" (che, nel caso dei semplici ricordi, può essere per "contenuto parziale") sono molto importanti. Se è possibile costruire un sistema cognitivo che funziona secondo le modalità della computazione classica, allora è necessario ricorrere a **memorie a vaglio continuo**, e a modalità di "**indirizzamento per contenuto** ".

Come visto, in generale i dispositivi che si trovano ai nodi di queste reti non sono delle semplici memorie, ma degli oggetti che implementano delle funzioni: verificano se ai loro input sono o meno presenti una serie di condizioni, e in caso positivo generano delle risposte. A questo punto potrebbe sembrare inopportuno, o almeno incompleto, continuare a chiamare questi dispositivi "memorie". Sicuramente non sono semplici memorie, poiché devono svolgere delle operazioni attive. Se riflettiamo, non è difficile convenire che ogni dispositivo di memoria deve eseguire operazioni di verifica di una serie di condizioni; anche in quelle utilizzate nei calcolatori tradizionali i singoli dispositivi ricevono degli input (l'indirizzo) e, se questi soddisfano certe condizioni (se l'indirizzo rientra in quelli che il dispositivo gestisce), generano il proprio output.

Quindi, quando diciamo che si tratta di dispositivi di memoria, intendiamo spesso dire che una delle funzioni principali è quella di memorizzare, cioè di incamerare dei contenuti e di riproporli quando opportuno.

Sospetto che molte delle regole associative debbano essere implementate proprio in questo modo. Una parte significativa delle regole associative consistono nell'associazione diretta tra "condizioni iniziali" e "risultati". Quindi per la loro implementazione servono un insieme di dispositivi in grado di memorizzare "le condizioni iniziali" e di confrontarle con le informazioni che ricevono in input. Quando il confronto dà esito positivo, altri dispositivi di memoria, direttamente connessi ai primi, dovranno proporre in output il proprio contenuto.

Anche le reti di analisi mostrano in parte queste caratteristiche. Abbiamo visto che in esse devono essere presenti dei dispositivi in grado di "riconoscere" le caratteristiche strutturali e di generare, come output, una singola informazione elementare che costituisce l'esplicitazione dell'avvenuto riconoscimento. Anche all'interno di reti di analisi saranno identificabili delle porzioni che si occupano di valutare parallelamente i medesimi input.

Altro concetto importante su cui insistere è che durante l'attività la maggior parte di queste memorie rimane **silente**. Come illustrato, questo silenzio è però solo apparente, poiché in realtà questi dispositivi devono essere sempre attivi, dovendo continuamente computare i dati che ricevono in input per valutare se proporre o meno il proprio output.

Le condizioni in input possono essere computate in vario modo: possono essere scomposte in varie "porzioni", che a loro volta possono costituire condizioni sufficienti ma non necessarie (in OR), oppure condizioni necessarie (in AND) ma a volte da sole non sufficienti, o delle vie di mezzo tra le due. In taluni casi si devono eseguire dei calcoli particolari sulle informazioni in input (ma anche le semplici operazioni di AND e di OR richiedono dei calcoli). Spesso inoltre deve essere presente una gerarchia di operazioni, ad esempio gruppi (o porzioni separate), di informazioni che sono valutate prima in OR (o OR like), poi in AND (o AND like)… eccetera.

Le varie "porzioni" da valutare in input possono presentarsi in vari formati: dalle semplici singole informazioni esplicite, ai pattern di informazioni, a rappresentazioni strutturali estese, ecc..

L'output può consistere, secondo i casi, nella semplice esplicitazione del riconoscimento di un soggetto cognitivo, nella rievocazione di un ricordo, in una rappresentazione compatta che costituisce una previsione su quanto può accadere, nell'attivazione di un obbiettivo o di un pericolo, oppure, come vedremo tra poco, nel suggerimento di un comportamento da tenere (anche complesso).

Le reti e i sistemi che generano previsioni devono ricevere input direttamente dai teatri virtuali e dalle reti di analisi che al loro interno contengono la rappresentazioni della situazione del presente. In taluni casi possono ricevere input anche da rappresentazioni che non si riferiscono al presente attuale, ma che ritraggono situazioni ipotetiche, generate durante attività di "ragionamento interno".

Buona parte delle reti per la generazione di previsioni possono dunque essere costituite da "**memorie associative**". Possono essere composte da moltissimi dispositivi che "osservano" a gruppi lo stato delle reti che rappresentano la situazione del presente e, quando trovano le condizioni corrette (che significa anche che la previsione potenziale contiene "soggetti importanti"), generano il proprio output in modo esplicito, attivando una serie di nodi delle reti opportune, destinate a contenere le "proiezioni per il futuro". Queste ultime reti saranno quindi dedicate a contenere le previsioni esplicite e, dovendo rappresentare "situazioni previste della realtà", saranno in pratica "parzialmente parallele" a quelle dedicate alla situazione del presente, nel senso che condivideranno per buona parte gli stessi soggetti cognitivi.

Come illustrato nel paragrafo precedente, una parte delle previsioni possono essere implementate utilizzando i teatri virtuali; ma, come detto, in pratica dovrebbero sempre essere delle regole di tipo associativo a decidere se generare queste previsioni e quali regole specifiche utilizzare. Ne consegue che anche la gestione dei simulatori deve essere affidata a reti che implementano principalmente regole associative.

Per comprendere più nel dettaglio come possono essere implementati e gestiti questi dispositivi, credo sia necessario ragionare ponendosi il problema di come si apprendono queste regole. Per il momento posso anticipare che un aspetto interessante consiste nel fatto che molte regole associative, sotto alcune condizioni, possono essere semplicemente "osservate" nella registrazione temporale dei risultati delle analisi strutturali (e funzionali) applicate al flusso delle informazioni sensoriali. Il concetto sottostante è che con buone capacità di astrazione è possibile identificare insiemi di fatti correlati che, ripetendosi nel tempo con regolarità, costituiscono a tutti gli effetti delle regole associative valide.

### 9.15 Rappresentazioni delle azioni e dei comportamenti

Una delle facoltà più importanti per un sistema cognitivo consiste nel sapere come agire sul mondo, che significa in primo luogo essere in grado di pianificare azioni e comportamenti in modo efficiente.

Quando vogliamo intervenire sul mondo esterno, fare in modo che le cose cambino secondo un determinato progetto, possiamo agire a più livelli. Possiamo semplicemente muovere i nostri arti per attuare un'azione fisica usando il nostro corpo, ma possiamo anche produrre dei suoni speciali, o scrivere comandi su una tastiera che, interpretati come simboli linguistici, diventano "ordini ed istruzioni" impartiti ad altre persone o a sistemi artificiali.

Uno degli obbiettivi più importanti dell'apprendimento consiste proprio nell'acquisizione della capacità di coordinare le azioni e di pianificare i comportamenti in modo utile per la risoluzione dei vari problemi che si devono affrontare.

Per partire dobbiamo per prima cosa capire come si possono rappresentare le azioni e i comportamenti in un sistema cognitivo.

In un certo senso le azioni sono **operazioni** effettuate sulla realtà circostante. Non dovrebbe sorprendere quindi che anche per esse si ripresenta il problema incontrato per la rappresentazione delle operazioni elementari di computo strutturale. Anche in questo caso non è possibile rappresentare le cose tramite solo strutture "statiche" di prima specie, e si deve quindi ricorrere all'utilizzo della funzione simbolica. Nel capitolo 3 ho proposto la definizione del concetto di **schema,** inteso come una struttura nella quale ad alcune delle parti che la compongono sono associate (anche passando per la funzione simbolica) delle operazioni elementari o degli altri schemi. In modo strettamente analogo possiamo ora definire il concetto "**schema comportamentale**".

In uno **schema comportamentale** possiamo utilizzare una serie di simboli da associare allo "svolgimento potenziale" di singole azioni standard, ma anche di comportamenti complessi.

Alla base si può iniziare con azioni elementari, quali singoli movimenti. In un sistema artificiale gestito da un calcolatore digitale, l'attuazione o meno di un movimento da parte di un attuatore può essere "gestito", nel caso minimale, attraverso un singolo bit inserito entro un apposito registro di memoria. La convenzione può essere che se il bit è 0 l'attuatore deve rimanere inattivo, mentre se il bit è 1 l'attuatore deve compiere la sua azione. In genere si utilizzano molti più bit, in quanto si desidera fornire uno o più valori numerici che corrispondono ad alcuni parametri che descrivono come l'azione deve essere compiuta. Si possono ad esempio utilizzare dei valori per dire di quanto ci si deve spostare, a che velocità deve essere effettuato lo spostamento, ecc…

Non è difficile vedere che questi valori devono avere delle corrispondenze con "le strutture" dello schema dell'azione che è eseguita (possono corrispondere al percorso spaziale del movimento, alle forze utilizzate, o ad altre informazioni di questo genere).

Accanto però ai bit "finali" che fanno passare veramente all'azione, è spesso utile costruire delle loro "rappresentazioni potenziali". Queste consistono in "copie" del comando, che non sono spedite all'organo attuatore, ma che servono invece per rappresentare all'interno del sistema cognitivo le singole azioni elementari in maniera parzialmente simbolica.

Come detto, nelle rappresentazioni di base si dovrà sempre ricorrere a dei simboli da associare a una serie completa di azioni elementari. Mettendo assieme più simboli si otterrà sempre un oggetto complesso che avrà qualche corrispondenza strutturale con le azioni che sono compiute o simulate. Questo oggetto complesso è a tutti gli effetti uno **schema** (come definito nel capitolo 3).

Partendo dal basso, si possono costruire schemi che corrispondono a sequenze di movimenti basilari. Alcune porzioni di questi schemi corrispondono a valori che determinano l'entità di un movimento, o altri parametri importanti. In modo semplice si può allora passare da questi schemi di base ad altri di livello appena superiore, ma più potenti dal punto di vista della loro utilizzabilità, che sostituiscono i detti valori con delle variabili.

Vedremo fra poco che la strategia generale per riuscire a gestire comportamenti complessi consiste nel demandare l'attuazione dei dettagli a processi già collaudati. Questi processi possono essere pensati come delle entità che richiedono in input dell'informazione compatta che consiste "nell'ordine di attuare il proprio programma" e, appunto, alcune variabili che dipendono dalla situazione specifica e che definiscono meglio la struttura, quindi lo schema, dell'azione. Una volta forniti questi dati, ci devono essere dei processi che si incaricano di portare a termine nel dettaglio le operazioni. Procedendo in questo modo, man mano che si sale di livello non diviene più necessario rappresentare l'intero schema, ma solo una sua versione molto più compatta.

Quando ordiniamo o chiediamo ad altri di fare una certa cosa, non ci preoccupiamo di descrivere esattamente, nel dettaglio, ogni azione che le persone dovranno compiere: ci affidiamo al fatto che i singoli individui sono in grado di affrontare di volta in volta i problemi specifici, e possiamo quindi sintetizzare il nostro ordine in poche istruzioni. Queste istruzioni in genere sono composte da alcuni simboli verbali di alto livello e contengono solo la specificazione degli elementi strutturali essenziali, che di volta in volta definiscono il compito specifico.

In modo simile si può procedere anche all'interno di un sistema cognitivo per la gestione e la rappresentazione di comportamenti complessi. In funzione del contesto, un'azione specifica può essere descritta in maniera più o meno compatta e a diverso livello di astrazione. Una rappresentazione compatta verrà affidata a reti ed apparati che la convertono in rappresentazioni via via più dettagliate.

Come sarà illustrato, è importante tenere presente che in molte attività i dettagli specifici di come portare a termine una certa operazione non possono essere stabiliti a priori, ma devono essere affrontati al momento. Non possiamo decidere prima di iniziare un viaggio precisamente con quali sequenze dovremo azionare i pedali della nostra automobile, di quanto dovremmo sterzare il volante e in quale sequenza ecc.. sono dettagli che non possono essere predetti a priori, ma che vanno affrontati al momento in un processo continuo di adattamento dell'azione ai nuovi dati sensoriali.

L'idea generale è che il tutto possa essere gestito attraverso un ordinamento gerarchizzato sia degli schemi procedurali che delle rappresentazioni dello stato del mondo e delle sue possibili evoluzioni, tra le quali quindi, anche le varie possibili rappresentazioni delle "**situazioni di partenza**" e delle "**situazioni obbiettivo**".

Secondo questa visione, la conoscenza di "**saper come fare le cose**" è costituita da una miriade di abilità complesse, e di strategie per la loro gestione, che sono organizzate per livelli crescenti di astrazione. Alla base stanno abilità semplici e ben verificate. Con esse risulta possibile affrontare obbiettivi relativamente semplici e definibili a basso livello. L'idea è che l'esistenza di questo substrato permette di definire attività più complesse utilizzando la strategia di **demandare** l'attuazione del dettaglio, e soprattutto la verifica della fattibilità, alle abilità sottostanti.

### 9.16 La ricerca di soluzioni ai problemi e la "conoscenza del fare"

Nel modello che propongo sono particolarmente importanti le **reti suggeritrici**. L'attività di queste reti consiste nel suggerire le soluzioni ai problemi che il sistema deve affrontare, quindi nel suggerire comportamenti che possono essere, secondo i casi, molto concreti oppure più astratti. Le reti suggeritrici dovranno quindi fornire come output delle rappresentazioni che vanno dalle singole azioni fisiche fino a rappresentazioni astratte di sequenze di operazioni formali (nei sistemi più evoluti).

Queste reti in genere ricevono contemporaneamente due fonti d'input: la rappresentazione dell'obbiettivo da raggiungere (o del pericolo da evitare) e la rappresentazione della situazione iniziale, che spesso consiste nella situazione attuale (anche se nei sistemi evoluti si possono formulare situazioni iniziali ipotetiche). Queste due rappresentazioni costituiscono la **definizione del problema in oggetto**.

Le reti suggeritrici dovranno reagire a questi input cercando al proprio interno, attraverso un vaglio silente di tutti i dispositivi che le compongono, l'eventuale presenza di "**soluzioni già testate**" per quel particolare problema. Se questo avviene, le soluzioni dovranno essere proposte in uscita in funzione della loro "**probabilità**" di funzionare.

In un sistema cognitivo maturo, le reti suggeritrici devono essere organizzate secondo una gerarchia. Ci saranno reti di alto livello che proporranno "soluzioni astratte" costituite in realtà da una serie di sottoproblemi da affrontare e di obbiettivi intermedi da raggiungere. Sotto di queste ci saranno reti di livello inferiore che si occupano di affrontare i vari sottoproblemi e che a loro volta potranno fare riferimento a reti di livello ancora più basso. Si proseguirà in questo modo scendendo lungo la gerarchia delle astrazioni fino ad arrivare a quelle reti che si occupano di pianificare le azioni concrete e di sorvegliarne (in collaborazione con altre) l'esecuzione.

Credo che la conoscenza semantica di come si affrontano i problemi, quindi la "**conoscenza del fare**", debba essere organizzata in modo naturale per stratificazioni e gerarchie. Questa conoscenza può essere costruita gradualmente durante opportune fasi di apprendimento, partendo dal basso, dalle abilità più basilari, per passare gradualmente a capacità di ordine più elevato. Propongo quindi che l'apprendimento possa avvenire per fasi durante le quali si mettono appunto i singoli "strati" di abilità. Questi strati dovrebbero essere costituiti da una collezione di "soluzioni pronte", quindi già testate, per l'insieme dei problemi affrontabili che si presentano con alta frequenza nella vita pratica.

Con la graduale messa a punto di un buon corredo di abilità si può passare da uno strato a quello successivo. In questo passaggio si possono utilizzare delle rappresentazioni più astratte per descrivere gli stessi problemi e le loro possibili soluzioni. Si possono classificare i problemi appartenenti a un certo strato in "risolvibili" e "non risolvibili", quindi in demandabili o non demandabili ad abilità già acquisite. In questo modo diventa possibile pianificare comportamenti più complessi e a più ampio orizzonte temporale.

La graduale costruzione della **conoscenza del fare** permette anche di passare da **astrazioni puramente strutturali** ad **astrazioni funzionali**. Ad esempio gli oggetti che sono degli strumenti, possono essere classificati in funzione degli scopi pratici per i quali possono essere usati: possiamo classificare come "cavatappi" tutti quegli arnesi che permettono di togliere il tappo da una bottiglia, indipendentemente dalla loro forma specifica, quindi anche se non si assomigliano strutturalmente. Sarà sempre, ovviamente, la loro struttura che consentirà di riconoscerli, ma sarà la loro funzione che permette di classificarli come appartenenti a una categoria comune.

Le reti suggeritrici possono essere distribuite in varie parti di un sistema cognitivo ed essere usate ogniqualvolta si tratta di cercare, negli (sotto) spazi delle operazioni possibili (generalizzate), una soluzione a un problema dato.

### 9.17 Primi accenni al ruolo dell'attenzione selettiva

Riassumo alcuni punti. Ho proposto che l'informazione all'interno di un sistema cognitivo sia organizzata in soggetti cognitivi e ho affermato che questi corrispondono alle suddivisioni dell'informazione in unità che ha senso considerare come entità a sé stanti. Si tratta sempre d'informazioni strutturali, anche quando si eseguono astrazioni. Ho anche affermato che il senso di queste informazioni da esplicitare è quello di rendere possibile l'implementazione di regole utili. Ogni soggetto cognitivo ha senso se è agganciato ad almeno una regola utile.

Abbiamo visto che una delle attività cognitive essenziali consiste nella rappresentazione delle situazioni della realtà, e che spesso le stesse situazioni possono essere rappresentate contemporaneamente a diversi livelli di astrazioni emergenti. Questa possibilità è una delle caratteristiche salienti della cognizione.

Va ora considerato che, a priori, tutte le informazioni che provengono dai sensi possono essere **potenzialmente importanti**. Quindi tutte devono essere opportunamente analizzate al fine di identificare i soggetti cognitivi in esse presenti, e tutti i soggetti cognitivi identificati devono essere vagliati dalle altre reti del sistema cognitivo. Procedendo con le elaborazioni, accadrà che non tutto ciò che è stato identificato costituirà un'informazione che è anche **realmente importante.** Tuttavia quest'importanza non può essere stabilita a priori, ma solo a posteriori.

Il fatto che, di volta in volta, i soggetti cognitivi realmente importanti siano in realtà pochi, fa sì che spesso è possibile rappresentare le varie situazioni in modo molto **compatto**. Si può anzi mostrare che è molto spesso conveniente, e talvolta necessario, **selezionare le informazioni**. In effetti, nelle attività di "ragionamento interno" (quando parliamo, generiamo previsioni o proponiamo soluzioni di problemi), utilizziamo sistematicamente rappresentazioni delle situazioni composte di **pochi soggetti importanti**.

È utile implementare dei meccanismi di **attenzione selettiva** per selezionare e "mettere in evidenza" le informazioni importanti.

L'attenzione selettiva, evidenziando i soggetti protagonisti, permette di cogliere con maggiore efficacia i rapporti di causa ed effetto nei fenomeni rappresentati, e permette inoltre di organizzare le informazioni in maniera che siano più facili da usufruire. In effetti, per essere in grado di pianificare azioni e comportamenti raffinati, è fondamentale organizzare le informazioni relative ai singoli "soggetti protagonisti" in modo che le loro proprietà e le loro implicazioni importanti siano rese subito in modo esplicito. In questo modo i processi di ricerca delle informazioni saranno assai semplificati. In un certo senso l'informazione va organizzata secondo modalità simili a quelli di un grande "database".

Credo si possano distinguere due tipologie di conoscenze semantiche: quelle che si riferiscono a verità "stabili nel tempo" e quindi a regole di ampia validità, e quelle che invece dipendono da fatti contingenti.

Talvolta le conoscenze semantiche contingenti possono essere in contrasto con le regole consolidate e per questo motivo è utile implementare dei meccanismi che diano le opportune priorità alle conoscenze contingenti.

### 9.18 Alcuni punti sull'apprendimento

In questo modello di sistema cognitivo le attività di apprendimento consistono principalmente nella progressiva messa a punto e ottimizzazione dei processi di analisi per la codifica dei soggetti cognitivi legittimi, e per l'identificazione, la codifica e l'implementazione delle regole. Questi due aspetti sono strettamente legati. I soggetti cognitivi servono sostanzialmente per permettere di definire le regole. A loro volta le regole devono essere testate, e questo può essere fatto solo utilizzandole, quindi provando attraverso di esse a generare previsioni che si dimostrano corrette o pianificando azioni che consentono di arrivare alle situazioni obbiettivo prefissate. Se si riesce a verificare che una certa regola funziona, allora significa che anche i soggetti cognitivi che sono serviti per definirla sono corretti (almeno in parte). Ma per chiudere questo cerchio possono servire davvero molti passaggi: si devono analizzare le informazioni sensoriali estraendo molte *feature* di vario livello, che devono essere usate per cercare di riconoscere i singoli oggetti e per ricostruirne le geometrie; a loro volta queste ultime devono essere nuovamente analizzate per identificare tutti i possibili soggetti cognitivi strutturali importanti, in particolar modo quelli che permettono di discriminare le microsituazioni che partecipano all'implementazione di singole regole. In generale, durante questi passaggi, è necessario rendere esplicite molte, talvolta moltissime, caratteristiche strutturali degli oggetti e delle loro relazioni reciproche (spaziali, temporali, funzionali, ecc). Molte di queste consistono in dettagli molto particolari.

Spesso per portare a termine un compito apparentemente semplice, come riuscire ad afferrare un oggetto, è necessario utilizzare delle vere e proprie strategie per pianificare la sequenza dei movimenti. Ogni singola fase di un gesto apparentemente semplice richiede la valutazione di molti particolari strutturali e l'utilizzo di molte regole specifiche, ognuna sensibile a un particolare sottoinsieme di questi "dettagli strutturali".

La "conoscenza del fare" richiede un'organizzazione gerarchica delle competenze. Anche azioni molto semplici richiedono di selezionare tra strategie che devono a loro volta generare una serie di "sottoproblemi" da demandare a regole specializzate di movimentazione, quindi di attivazione delle singole fasce muscolari.

Di fronte alla complessità di questi compiti è facile capire che riuscire a chiudere il ciclo che permette di verificare che le regole messe a punto effettivamente funzionino, è un problema intrinsecamente complesso.

Credo sia possibile identificare varie strategie di apprendimento che permettono di gestire la complessità di questi problemi. Abbiamo già visto che conviene suddividere l'apprendimento in fasi successive nelle quali ci si concentra su obbiettivi limitati e quindi più semplici.

Un'altra possibile strategia di apprendimento, consiste nell'usare, quando possibile, il parallelismo interno del sistema per testare contemporaneamente molte possibili soluzioni e selezionare quelle che danno i risultati migliori.

Come visto, spesso è necessario usare i risultati ottenuti "a valle" per mandare dei segnali di "feedback", di "conferma di validità", ai processi sottostanti, che hanno contribuito a generare i risultati stessi. In taluni casi è possibile utilizzare metodi che propagano all'indietro "l'errore", vale a dire una valutazione della distanza tra quanto ottenuto e l'obbiettivo da raggiungere. Quando ciò è possibile, questa "quantificazione" della distanza dalla soluzione ottimale può essere usata per far convergere più velocemente i processi "a monte" imponendo, per l'iterazione successiva, variazioni minori a quelli che più si sono avvicinati alla soluzione, e maggiori agli altri.

Inoltre è molto probabile, per non dire evidente, che veniamo al mondo con un patrimonio di abilità già precostituite, almeno in forma di abbozzo; è evidente che gli organi di senso sono stati ottimizzati per eseguire solo certe particolari elaborazioni tra le molte potenzialmente possibili. È probabile che anche entro le nostre reti nervose vi siano delle "precablature innate" che indirizzano già dall'inizio i vari processi nelle giuste direzioni. È molto probabile che la natura sia riuscita, durante l'evoluzione, a selezionare dei "percorsi formativi" particolarmente efficaci. Veniamo al mondo dotati di una serie di ricette comportamentali innate che ci spingono, fin dalla più tenera età, a eseguire una serie di esperimenti motori e percettivi particolari. È probabile che questi esperimenti generino una serie di "situazioni tipiche" dove si manifestano regole, sia operazionali sia associative, che riusciamo a cogliere e a inglobare nelle nostre reti nervose.

Non è facile valutare il peso, nel problema complessivo dell'apprendimento, di queste "vie già tracciate" che sono state selezionate dall'evoluzione. Come già anticipato nell'introduzione, credo che se riusciamo a comprendere le finalità delle singole fasi dei vari processi cognitivi, dovrebbe allora essere molto più semplice comprendere come "assistere" eventuali sistemi artificiali a seguire i percorsi ottimali e comprendere quali sono le risorse hardware necessarie per ognuno di questi.

Un altro aspetto importante per l'apprendimento è che esistono famiglie di regole associative di "medio/alto livello", che possono essere relativamente semplici da identificare qualora il sistema abbia già acquisito capacità sufficienti di astrazione. Quando un sistema cognitivo è in grado di costruire delle rappresentazioni astratte delle situazioni contingenti può cercare, entro il flusso delle informazioni provenienti dai sensi, l'eventuale presenza di regolarità che si dimostrano affidabili. Può sfruttare molte di queste come regole utili per generare previsioni, per pianificare azioni e per produrre inferenze di altro tipo.

In un certo senso, una parte significativa delle regole che un sistema cognitivo deve imparare ad utilizzare può essere semplicemente "estratta" dall'esperienza diretta. Per alcune di queste regole, la parte più onerosa del lavoro da fare consiste nel riuscire a codificare i soggetti cognitivi astratti corretti e nell'identificare, attraverso processi di analisi statistica, le loro correlazioni spaziali e temporali. Sono in particolar modo importanti quelle correlazioni che si manifestano in modo tale da costituire dei rapporti tipo "causa ed effetto".

Uno dei problemi è riuscire a selezionare i soggetti cognitivi giusti tra i moltissimi che un sistema cognitivo deve essere in grado di riconoscere quando osserva il mondo esterno. In questo compito possono dare un contributo importante i processi di attenzione selettiva. Credo che anche l'attenzione sia un meccanismo che richiede apprendimento, ma penso anche che, nella sua forma iniziale, possa essere innata, quindi "preprogrammata". In effetti, spesso gli oggetti interessanti, che partecipano da "protagonisti" alla codifica di regole valide, sono quelli che manifestano dei "cambiamenti". L'attenzione è istintivamente attratta da ciò che si muove, dagli spostamenti e dai rumori improvvisi. È inoltre una buona strategia provare in primo luogo a correlare gli oggetti che si trovano vicini spazialmente, o che vengono a contatto durante i movimenti. Non credo sia troppo difficile, utilizzando tecniche di analisi strutturale, dotare un sistema cognitivo della capacità di "discriminare" e rappresentare i movimenti e i cambiamenti.

Un sistema cognitivo deve però anche utilizzare molte regole, sia associative che operazionali, che non sono direttamente estraibili dalle registrazioni di ciò che accade, ma che devono essere sostanzialmente "**congetturate**", e che possono essere scoperte solo tramite dei processi di "**prove casuali**" e di **selezione a posteriori** di quelle che funziono. La scoperta di regole di questo tipo è un processo difficile, che richiede molto tempo e, ancora una volta, l'adozione di strategie appropriate.

Gli esseri umani sono dotati di facoltà intellettive superiori; per esse credo che il linguaggio svolga un ruolo essenziale. Penso che il linguaggio, oltre a mettere a disposizione dell'individuo le scoperte di altri (che contribuiscono a costituire il "patrimonio culturale" di una comunità), forzi l'acquisizione di una serie di astrazioni che altrimenti non si svilupperebbero naturalmente.

### 9.19 Graduale costruzione della conoscenza semantica

La costruzione della conoscenza semantica dovrebbe procedere gradualmente e con lo sviluppo sinergico dei vari sistemi di reti. Poiché il tutto è intrecciato, non è semplice descrivere separatamente le varie componenti e le varie fasi. Ad ogni modo tra le prime cose da fare ci sono le implementazioni delle conoscenze degli oggetti concreti, in particolar modo quelli che conservano la loro forma nel tempo e che appartengono alla classe degli oggetti abituali, con i quali si ha modo di interagire molte volte. Questi oggetti hanno delle geometrie proprie che spesso sono la composizione di alcune forme standard che corrispondono a soggetti cognitivi di medio livello. Questi soggetti dovrebbero essere in grado di fornire in feedback gruppi di parametri ai teatri virtuali. Questi gruppi di parametri sono delle descrizioni strutturali compatte delle forme standard, in grado di rigenerarle. Molti singoli oggetti concreti sono caratterizzati da specifiche giustapposizioni di forme standard, quindi essi possiedono (ed ereditano) i loro parametri caratterizzanti.

Non credo sia difficile comprendere come un sistema cognitivo può costruirsi una specie di "database" degli oggetti concreti dei quali ha esperienza. Osservandoli e interagendo con essi, può ricostruire e memorizzare le loro forme. Queste conoscenze devono essere implementate in nodi e moduli in grado di riconoscere le configurazioni di caratteristiche strutturali che permettono il riconoscimento affidabile dei singoli oggetti. Questi riconoscimenti devono essere associati ai parametri in grado di rigenerarne le forme estese (o quasi estese) entro i teatri virtuali. L'implementazione delle associazioni corrette può avvenire durante le fasi di apprendimento dove il sistema prova, entro il ciclo percezione-azione (di retroazione primaria), se le ricostruzioni tentate funzionano correttamente; vale a dire se i parametri sono in grado di generare delle rappresentazioni 3D che consentono di generare previsioni accurate entro gli stessi teatri virtuali (anche se valide solo a breve termine).

Sia negli oggetti concreti (che hanno spesso forme invarianti), sia nelle loro disposizioni "accidentali", le forme standard possono stare in una serie di relazioni spaziali. Un sistema cognitivo deve essere in grado di riconoscere e classificare un set sufficientemente completo di queste relazioni. Le singole relazioni possono essere identificate da algoritmi di analisi strutturale, da combinare, probabilmente, con processi di attenzione selettiva. Si possono usare varie tecniche per identificare queste relazioni. Ad esempio, semplificando, la relazione "stare sopra" può essere, in linea di principio, identificata verificando che tutte le parti di un oggetto A abbiano coordinate verticali maggiori di un oggetto B. Probabilmente un sistema cognitivo deve utilizzare tecniche più sofisticate e più flessibili. È probabile che convenga eseguire molte classificazioni locali delle relazioni tra le "mini porzioni" degli oggetti e procedere, entro nodi appartenenti a strati di livello più alto (lungo la stratificazione delle reti di analisi), a classificare le possibili combinazioni globali di queste. Ad esempio, se localmente le porzioni in contatto degli oggetti A e B sono sempre in relazione tale che le parti di A hanno altezza maggiore delle relative di B, allora è anche vero che globalmente l'oggetto A è sopra l'oggetto B. Chiaramente si possono avere molte situazioni ibride, e alcune possono essere ambigue (ad esempio se abbiamo due asciugamani di colore diverso aggrovigliati assieme).

In modo analogo sono concepibili metodi per codificare molte altre relazioni spaziali tra oggetti, come quelle di "stare a lato", "essere in contatto", "stare davanti", "stare dietro", "poggiare su un lato", ecc.

I singoli oggetti, in funzione delle caratteristiche delle loro forme, possono assumere "posizioni particolari" come: "stare in piedi", "essere sdraiati in orizzontale", "essere appoggiati a qualcosa", "essere obliqui", ecc, Anche queste posizioni possono essere riconosciute da algoritmi, ad esempio identificando la lunghezza maggiore dell'oggetto, tracciando una linea lungo questa e classificando le sue possibili angolazioni. Ovviamente queste "posizioni particolari" si possono identificare in oggetti che hanno almeno una delle dimensioni che prevale sulle altre. Una scopa può stare dritta, sdraiata per terra, obliqua e appoggiata al muro… Per un pallone queste classificazioni non hanno senso.

Tutte queste proprietà e relazioni spaziali sono molto importanti perché, in funzione di esse, si determinano situazioni alle quali si possono applicare regole comportamentali specifiche. Ad esempio, per fare un panino, la relazione di "stare in mezzo in senso verticale" è ben diversa da quella di "stare in mezzo in senso orizzontale". Ogni bimbo apprende che è importante che la superficie del pane sia sufficientemente orizzontale per riuscire a metterci sopra qualcosa. Inoltre apprende che mentre lo si muove è importante non piegarlo troppo in quanto il contenuto potrebbe cadere. Ci sono moltissime regole di questo tipo che dipendono dalla posizione degli oggetti e dalle loro relazioni spaziali reciproche. Molte regole sono anche sensibili agli stati di movimento e alle relazioni temporali tra gli eventi. Un sistema cognitivo deve apprendere a classificare tutte le possibili varianti anche di questo tipo di relazioni. Un "evento", dal punto di vista di un sistema cognitivo, corrisponde a una variazione nella struttura degli oggetti e delle situazioni da essi composte.

Credo che in un sistema cognitivo sia necessario abbondare con le esplicitazioni delle varie proprietà e relazioni strutturali. Quelle che il sistema deve essere in grado di identificare sono molte di più rispetto a quelle che possiamo descrivere verbalmente. Molte proprietà e relazioni possono essere alquanto peculiari e possono riguardare dettagli molto specifici. Con esse si devono tentare, in parallelo, molte codifiche di proprietà strutturali e relazioni più complesse; saranno poi selezionate quelle che contribuiscono all'implementazione di qualche regola valida. Il tutto dovrebbe essere gestito in modo automatico. La quantità e la tipologia di queste caratteristiche, ma anche delle stesse regole, potrebbero essere tali da risultare estremamente complesse da seguire e descrivere analiticamente. Probabilmente un sistema cognitivo deve puntare "sull'abbondanza" e sul parallelismo. Deve testare, entro le sue reti, una grande congerie di caratteristiche strutturali, selezionando, per livelli gerarchici, quelle combinazioni e codifiche che riescono a partecipare a qualche regola valida.

Mi pare abbastanza evidente che il riconoscimento degli oggetti (con la ricostruzione delle loro strutture), nonché il riconoscimento e la classificazione delle reciproche relazioni spaziali, temporali e di movimento, costituiscono una base importante per la classificazione della varie situazioni. Relazioni di questo tipo sono di base perché sono "interne" alle rappresentazioni del flusso temporale degli eventi. Esse sono contenute entro la simulazione tridimensionale del mondo osservato. Sono contenute nel "film in 3D" di ciò che avviene. Queste relazioni non dipendono dall'uso che si decide di fare di una certa cosa, o dall'obbiettivo che si vuole raggiungere. Proprio per questa ragione sono più basilari di altre.

Queste relazioni possono essere rappresentate a diversi livelli, e in funzione di questi, in modo più particolareggiato e dettagliato, oppure in modo più astratto. Per passare alle astrazioni è utile eliminare qualche informazione di dettaglio, quindi passare a dei morfismi. Queste operazioni si possono fare classificando entro le medesime categorie sia oggetti diversi, sia loro varie relazioni specifiche (spaziali, temporali, di movimento…). Ad esempio la relazione che nel linguaggio esprimiamo con il generico "essere sopra", è spesso già un'astrazione rispetto ai moltissimi modi nei quali un oggetto particolare può stare effettivamente sopra ad un altro. Questo vale anche per i termini linguistici che esprimono relazioni quali: l'essere di lato, l'essere il vicino a, ecc… Ci possono essere moltissimi modi in cui un oggetto è vicino ad un altro. Siamo in grado di classificarli tutti nella medesima categoria probabilmente anche proprio per essere in grado di utilizzare il linguaggio. Il linguaggio, con forte probabilità, forza la costruzione di classificazioni comuni. Quando diciamo a una persona: "troverai l'oggetto X sopra Y e vicino a Z"; non stiamo descrivendo nel dettaglio la simulazione 3D della situazione specifica di X,Y,Z , ma stiamo usando un astrazione. Quest'astrazione si poggia sulla capacità del sistema cognitivo che riceve il messaggio linguistico di gestire per livelli stratificati la pianificazione dei propri comportamenti, ad esempio decidendo che la prima cosa da fare è quella di recarsi nella stanza dove si trovano Y e Z; e sulla sua capacità di identificare, quando osserverà concretamente la scena, le varie relazioni specifiche e le loro varie classificazioni, che permettono di riconoscere quanto descritto in astratto.

Passando dalle classificazioni più specifiche a quelle più astratte, un sistema cognitivo potrà associare alle varie situazioni e a singoli soggetti in esse presenti i vari "ruoli funzionali" che riuscirà a codificare durante le esperienze dirette. Come detto, questi ruoli non appartengono alle "**proprietà interne**" delle "simulazioni 3D", ma sono esterni al loro contenuto informativo. Nelle fasi iniziali dell'apprendimento essi dipenderanno dagli obbiettivi che il sistema persegue.

### 9.20 Alcune idee sull'apprendimento e la gestione degli obbiettivi

Per illustrare un possibile modello di apprendimento della capacità di gestire obbiettivi gradualmente più astratti e generali, si deve partire dal basso; penso sia necessario fare riferimento a un "modulo direttivo" che deve essere innato, quindi precostituito nel sistema fin prima che questo cominci la sua avventura di esplorazione del mondo.

Questo modulo direttivo dovrebbe funzionare un po' come il ciclo primario di un sistema operativo: eseguire una serie task fondamentali, gestendone la scansione temporale. Dovrebbe inoltre avere una struttura minimale che consente di gestire le priorità. Cosa deve produrre questo modulo direttivo? Sostanzialmente dei segnali che forzano altre reti ad agire.

Nelle fasi iniziali dell'apprendimento, quando il sistema non ha alcuna forma di cognizione strutturale presente, questo modulo riceve le richieste dal corpo e risponde generando dei segnali che obbligano le reti suggeritrici ed attuatrici ad agire di conseguenza. Chiaramente se nel nostro sistema cognitivo alle primissime armi non ci fosse nessun "programma di comportamento" precostituito, le azioni non potrebbero che essere del tutto casuali, quindi alquanto scomposte e probabilmente pericolose. Un sistema che non sa cosa fare, ma che riceve l'ordine non ignorabile di agire, sarebbe un sistema "molto nervoso"che tende ad agitarsi inutilmente.

La natura ci fornisce di programmi comportamentali innati, attivi prima ancora di iniziare ad apprendere. Essi sono molto semplici. Alcuni sono semplici riflessi neonatali, che consentono di fare cose quali succhiare il latte, provvedere ai bisogni corporali primari e mandare segnali sonori ai genitori affinché intervengano, ecc; nella pratica consentono di sopravvivere nei primi mesi di vita.

Questo modulo deve anche essere in grado, una volta soddisfatti i bisogni primari, di lasciare tempo per l'osservazione del mondo e per le prime attività di apprendimento. Deve anche essere costituito in modo tale da permettere il dirottamento delle pulsioni di base verso una serie di altri "obbiettivi" che il sistema impara a generare. La logica di questo "passaggio di consegne" è, almeno in linea di principio, abbastanza semplice: se una certa attività, un certo comportamento, la presenza di una certa cosa, implica il soddisfacimento degli stimoli primari, questo potrà acquisire un "valore di desiderabilità indotto".

In natura un sistema cognitivo (biologico) deve imparare a cavarsela da solo, a sopravvivere e a darsi da fare per la prosecuzione della specie. Per far questo deve acquisire la capacità di identificare gradualmente **situazioni** e **attività**, in maniera progressivamente più astratta, che facilitano enormemente la possibilità di avere successo nel soddisfacimento delle necessità di base.

Il cibo non è sempre a immediata disposizione nell'ambiente, non si può sempre contare nell'azione "intelligente di altri" che provvedono concretamente ai nostri bisogni (come i genitori che provvedono alla prole). Il cibo va cercato, cacciato, coltivato; se si fa parte di una specie che cerca di sopravvivere in ambienti ostili, in climi freddi e scanditi da stagioni dove non si trova cibo, è necessario imparare ad accumularlo, a costruire tane e abitazioni, a costruire difese dai possibili aggressori.

Il riuscire ad attribuire un valore a sé stante a un obbiettivo astratto (come ad esempio quello di avere un rifugio dove poter riposare, mangiare, ripararsi, far crescere la prole in un ambiente sicuro dai pericoli, ecc) offre enormi vantaggi. Tutte queste cose devono essere tradotte in obbiettivi da raggiungere, in attività, realizzazioni, situazioni da perseguire (anche molto astratte), che costituiscono in se stessi degli obbiettivi dotati di un proprio valore di desiderabilità. Ma per riuscire ad attribuire a queste cose un valore indotto di desiderabilità o di indesiderabilità è necessario acquisire le capacità di rappresentare e riconoscere le **singole "situazioni"** e anche **le singole "attività"**.

In particolare ora mi preme mettere l'accento sul concetto di **attività**. Un'attività è qualcosa che si fa e che dura del tempo. Molti degli obbiettivi più importanti consistono in attività: mangiare, dormire, coltivare il cibo, accudire qualcuno. Chiaramente ci sono attività più semplici da definire e da descrivere, e attività che sono più complesse e spesso anche più astratte. Si noti che molti obbiettivi che consistono in "situazioni astratte" ereditano la loro desiderabilità dal fatto di consentire o facilitare lo svolgimento di alcune attività che sono valutate come positive.

Le domande da porsi a questo punto sono: come si fa a codificare cognitivamente le varie attività? Come può un sistema cognitivo imparare a riconoscerle e a rappresentarle? Come può imparare ad attribuirgli un valore di desiderabilità?

All'inizio le attività sono semplici: mangiare, dormire, soddisfare i bisogni corporei, ecc. E' probabile che un bambino inizi a riconoscerle e a differenziarle in base ad alcuni soggetti importanti in esse presenti, alla descrizione di come si trova il corpo, alla tipologia delle azioni che si compiono.

Si noti che molte "attività" semplici possono all'inizio essere definite, anche se in modo rudimentale, usando le classificazioni delle posizioni che il corpo assume, anche rispetto alle cose circostanti, e usando le varie prime classificazioni delle azioni che si compiono. Le prime attività saranno rappresentate in visione soggettiva, ma con il tempo si potrà cogliere gli elementi comuni sul loro aspetto esterno, e con ciò si potrà riuscire a riconoscerle anche quando sono svolte da altri.

Molte attività richiedono la presenza di specifiche relazioni spaziali e temporali tra oggetti, tra gli oggetti e la descrizione strutturale dei movimenti che si compiono durante le azioni, e tra gli oggetti e le parti del corpo.

Per un bambino, e penso per un sistema cognitivo alle prime armi, un'attività come "mangiare" è caratterizzata dai suoi aspetti esterni ricorrenti: il portare qualche cosa alla bocca, che sparisce in essa. Dormire significa stare in posizione orizzontale e con gli occhi chiusi. Per un adulto mangiare significa anche deglutire e digerire quello che si è inghiottito, mentre il dormire richiede anche il "prendere sonno" e non basta stare sdraiati. Ma nelle fasi iniziali dell'apprendimento, la capacità di analizzare la realtà si limita ai soli aspetti "esteriori" che sono più facilmente discriminabili e riconoscibili.

Molte attività possono essere descritte grazie alla presenza di alcune "microsituazioni tipiche" che le caratterizzano e che un sistema cognitivo deve imparare a riconoscere. Nel modello proposto queste microsituazioni sono riconoscibili da reti di analisi strutturale, capaci di riconoscere i singoli oggetti, le singole cose concrete, di classificarle (per forma, aspetto delle superfici, dimensioni, attributi vari…), e capaci di fare lo stesso per le loro varie relazioni reciproche.
Con l'acquisizione di esperienza le cose concrete potranno dunque essere classificare non solo in funzione della loro struttura, ma anche in funzione di ciò che con esse si può fare, e in funzione degli effetti che la loro presenza può determinare: quindi estendendo le loro proprietà dall'aspetto puramente strutturale a quelle determinate dal complesso delle regole nelle quali partecipano.

Molte situazioni ed attività complesse sono costituite da una serie di situazioni ed attività più semplici. Ad esempio, un bambino comprenderà che l'attività di "cenare" non consiste semplicemente nel mangiare: si deve anche stare seduti a tavola, si devono rispettare una serie di regole di "bon ton", deve essere sera, e varie altre cose. Con il tempo comprenderà anche quali sono i motivi generali che danno un senso a questi comportamenti, anche se all'inizio sono semplicemente imposte dai genitori.
Per quanto riguarda il problema dell'attribuzione del valore di desiderabilità o di indesiderabilità alle varie situazioni e attività, che si impara gradualmente a rappresentare e a riconoscere, penso che questo all'inizio sia attribuito in funzione della loro capacità di soddisfare i bisogni primari, quindi di ottenere delle gratificazioni, degli stati di benessere, o al contrario di generare sensazioni negative.
Ma sviluppando le capacità di inferire, e di associare ad ogni situazione ed attività il complesso delle loro possibili conseguenze (attraverso le varie reti deputate a far ciò), l'attribuzione di valore dovrebbe gradualmente passare da un processo puramente associativo, e/o ereditario tra classificazioni, ad una gestione secondo una chiara logica utilitaristica (ovviamente emergente).
Riassumendo, penso che con il tempo si impari a riconoscere cosa una persona sta facendo in base ad informazioni quali la postura, il vestiario, gli oggetti che ha attorno, ma soprattutto in base alla sequenza delle azioni che compie. Cose analoghe valgono anche per i comportamenti degli animali e dei congegni automatici. Ma a queste cose mancano ancora ingredienti molto importanti, che consistono nelle rappresentazioni "degli stati interni" dei sistemi cognitivi che svolgono le attività. Infatti spesso per capire la logica di quanto viene compiuto da esseri che hanno comunque una qualche forma di attività cognitiva, è molto importante non limitarsi all'aspetto esterno delle azioni che sono compiute, ma è anche importante capire secondo quale "logica interna" sono attuate. È importante saper rappresentare "le intenzionalità" e il "complesso delle credenze" di chi compie le azioni. Per far questo è necessario costruire delle rappresentazioni sugli "obbiettivi" di chi sta agendo, ma anche implementare delle regole che descrivono la capacità di chi agisce di risolvere i problemi. In altre parole è spesso importante avere un modello della cognizione altrui in termini di motivazioni, abilità e abitudini comportamentali. Chiaramente non è solo importante cercare di costruirsi una rappresentazione della cognizione degli altri, ma è fondamentale essere in grado di rappresentare in certa misura anche la propria attività cognitiva.

### 9.21 Alcuni appunti sulla possibilità di costruire delle "meta-rappresentazioni" della stessa attività cognitiva

Una possibilità importante per codificare soggetti cognitivi astratti di alto livello scaturisce dalla possibilità di costruire rappresentazioni della stessa attività cognitiva.
In questo contesto si colloca il problema di trovare il modo per codificare alcuni concetti astratti che sono molto potenti, ma che nel contempo sembrano difficili da imbrigliare e definire con precisione. Si pensi ad esempio al concetto di "impedimento", o al concetto di "condizione necessaria", a quelli di "obbiettivo da raggiungere", di "situazione iniziale", di "problema", ecc. Come possiamo codificare, all'interno di un sistema cognitivo come quello descritto fino ad ora, dei soggetti cognitivi che corrispondono a questi concetti?
Continuando nel solco delle idee presentate fino ad ora, il problema consiste nel trovare il modo per implementare, in qualche rete di alto livello, dei nodi che siano in grado di attivarsi ogniqualvolta è riconosciuto qualcosa che costituisce un impedimento, un obbiettivo, una condizione necessaria, ecc, indipendentemente dalla situazione specifica rappresentata o dalle cose concrete che, di volta in volta, assumono quel ruolo specifico.
Se riusciamo in questa impresa si potranno utilizzare questi soggetti cognitivi attivamente per definire "obbiettivi" molto astratti, come ad esempio quello di "imparare a risolvere i problemi"!
Poiché ci appressiamo a trattare di rappresentazioni che trattano di rappresentazioni, si può utilizzare, come si usa in questi casi, il termine **meta-rappresentazioni.**
Come potrebbe quindi un sistema rappresentare, attraverso opportune operazioni di astrazione, parte della propria attività interna? Idealmente per far questo dovrebbe avere la possibilità di osservare e rappresentare queste attività "dall'alto". Ovviamente non ha senso costruire una rappresentazione di tutto quello che il sistema fa nel dettaglio, ma è necessario cogliere uno schema emergente dell'attività globale. Secondo le idee presentate questo si dovrebbe ottenere se si riesce ad eseguire delle operazioni di quoziente e di morfismo sulla rappresentazione dell'attività globale, che consentano di rappresentare in maniera astratta e molto compatta le dinamiche generali, i passaggi salienti, degli schemi dei vari processi di elaborazione.
Si noti quanto segue.

La medesima rappresentazione di una certa situazione può assumere a seconda dei casi dei ruoli diversi: può essere una rappresentazione che appartiene alla situazione del presente, oppure essere una previsione per il futuro, oppure un obbiettivo da raggiungere, e altro ancora.

Nel modello che ho proposto fino ad ora, questi "ruoli" dipendono essenzialmente **da "dove" la rappresentazione si trova**: quindi se essa si trova in una rete di rappresentazione della "situazione del presente", o in una rete di proiezione delle previsioni, o in una rete degli obbiettivi, ecc. Non è difficile comprendere che questi ruoli costituiscono potenzialmente degli importanti soggetti cognitivi, che potrebbero essere utilizzati per costruire appunto delle rappresentazioni astratte dell'attività globale del sistema cognitivo. Ma per ora nel sistema descritto di fatto **manca una loro codifica esplicita**. Le varie dinamiche sono implementate dagli automatismi di gestione delle varie reti, ma non c'è nulla che codifichi e renda esplicita l'informazione sul "ruolo" che la singola rappresentazione assume nella gestione dell'attività cognitiva. Per **rendere espliciti** questi come soggetti cognitivi, sempre in accordo con il principio di convergenza delle verifiche, si devono produrre delle informazioni elementari con almeno due stati possibili, da associare in maniera univoca al loro riconoscimento. Nel nostro modello dovremmo quindi predisporre una rete speciale i cui nodi sono deputati ad attivarsi quando riconoscono che una certa rappresentazione è, appunto, una "situazione del presente" o "una previsione" o "un obbiettivo da raggiungere".

Chiaramente però, nel modello fin qui presentato, questa codifica è in certo senso banale e inutile perché appunto dipenderebbe semplicemente dalla rete dove le rappresentazioni si trovano. Sarebbe come implementare dei bit che si accendono non appena vi è qualcosa di attivo nelle reti rispettive. Poiché nella normale attività queste reti dovrebbero essere quasi sempre in funzione, le attività di questi bit non sarebbero particolarmente significative.

Perché la cosa abbia senso, serve implementare qualcos'altro; serve, come minimo, una specie di "nuovo teatro interno" in grado di contenere le varie rappresentazioni delle quali vogliamo codificare i ruoli, in modo indipendente dalle reti menzionate.

Questo si può fare, ci possono essere anzi più modi, e più opportunità, per implementare questi "teatri speciali". Tra le varie possibilità possiamo usare la memoria di quanto abbiamo fatto in passato, possiamo usare i "buffer" che sono necessari per contenere le rappresentazioni linguistiche, o meglio ancora possiamo usare i buffer che sono necessari per contenere le pianificazioni e i progetti di quello che ci proponiamo di fare. In effetti in tutti questi casi abbiamo bisogno di memorizzare "sequenze di rappresentazioni" di alto livello del "flusso del nostro pensiero" (o di quello di altri).

Se memorizziamo quanto abbiamo fatto in passato allora ripercorrendo queste memorie saremmo in grado di attivare le varie rappresentazioni senza che queste siano messe nelle reti rispettive (della situazione del presente, delle previsioni, degli obbiettivi, ecc).

Non è difficile vedere che, per essere in grado di comprendere il linguaggio, è necessaria la presenza di un "buffer" capace di contenere le rappresentazioni evocate da quanto ascoltiamo da altri. Anche in questo caso il buffer funziona in modo molto simile alla memoria di quanto abbiamo percepito, pensato e deciso. Si noti anche, che il riuscire a comprendere il linguaggio implica saper distinguere se le varie rappresentazioni evocate da quanto ascoltiamo sono delle richieste di comportamento (quindi qualcosa da inserire nelle nostre reti degli obbiettivi), o la descrizione di qualcosa che accade ma che non vediamo (quindi informazioni che devono far parte della "situazione del presente", o il racconto di qualcosa di accaduto in passato, o una previsione per il futuro fatta da altri, ecc).

Mi sembra anche abbastanza evidente che per essere in grado di pianificare comportamenti sofisticati serve un buffer dove implementare i vari progetti di comportamento, prima che questi diventino "obbiettivi in soddisfacimento". Probabilmente è quest'ultimo il candidato ideale a fare da supporto per la sequenza dei "pensieri interni". Serve infatti quella che viene spessa chiamata **memoria di lavoro**.

Chiaramente le rappresentazioni che possono essere contenute in questi buffer non possono comprendere l'attività globale delle varie reti, ma devono essere rappresentazioni più compatte. Quindi si tratta, per la maggior parte, di rappresentazioni di alto livello, anche se non possiamo escludere che all'occorrenza esse possono avere anche la capacità di evocarne altre di livello più basso.

Quindi, ritornando al problema iniziale, nelle varie memorie di lavoro che contengono le rievocazioni dell'attività cognitiva passata, quelle pianificate, o quelle indote dal racconto di altri, saranno presenti rappresentazioni che potranno assumere diversi ruoli. Un sistema cognitivo evoluto deve essere in grado di distinguerli e classificarli opportunamente. All'inizio dell'apprendimento questi ruoli saranno quelli fondamentali e saranno riconoscibili in funzione delle reti da dove le rappresentazioni provengono o alle quali sono destinate. Ma in fasi di apprendimento più avanzate, sarà possibile codificarne degli altri. Questo è un punto particolarmente importante.

*L'idea generale è che partendo dai ruoli di base sia possibile gradualmente, analizzando le varie "situazioni cognitive", codificarne altri ruoli particolarmente importanti, come quelli di essere una causa, un impedimento, un effetto, una condizione, necessaria, e vari altri.*

Se il sistema ha la possibilità di "etichettare" i vari soggetti in attenzione in funzione del ruolo di base da questi assunto, risulta allora possibile decodificarne altri, come ad esempio quello di "essere un ostacolo". Un "ostacolo", in questo sistema, è quel soggetto X (o quell'insieme strutturato di soggetti attenzionabili), la cui presenza impedisce (di regola) di ottenere l'obbiettivo Y a partire dalle condizioni iniziali Z. Deve essere quindi vera la regola: "Se X non è presente allora si può passare da Z a Y secondo una strategia già sperimentata; diversamente se X è presente allora non ci si riesce". Se si verifica questa condizione allora X assume il ruolo di "ostacolo" all'ottenimento dell'obbiettivo Y a partire dalla condizione iniziale Z.

Un ostacolo può essere una porta sbarrata che impedisce di accedere in una stanza, ma anche una persona che ha il potere di opporsi affinché si faccia una certa cosa, oppure un regolamento, un infortunio, la mancanza dello strumento necessario per fare qualcosa, e molto altro.

In generale è importante riuscire a codificare tutti questi ruoli (che sono a tutti gli effetti legittimi soggetti cognitivi emergenti) in maniera astratta, quindi indipendente da cosa sono concretamente X, Y, Z.

I ruoli di base possono dunque essere utilizzati, come visto nell'esempio appena illustrato, per codificarne altri che questa volta non hanno come referente una rete a loro interamente dedicata!

La codifica di questi ruoli genererà quindi un insieme di soggetti cognitivi molto importanti. Con questi soggetti cognitivi si potranno costruire delle meta-rappresentazioni astratte, costituite dalle sole sequenze dei ruoli cognitivi, che le varie rappresentazioni sottostanti assumo. Queste meta-rappresentazioni saranno soggette a delle regolarità e da esse potranno essere estratte delle utili regole. In accordo con la seconda congettura di riferimento, queste regolarità consisteranno, ancora una volta, in coincidenze strutturali. Coincidenze che questa volta riguardano direttamente la "struttura dei pensieri". Una parte di queste regole saranno quelle della logica!

In realtà procedendo in questo modo sulle rappresentazioni gestite dalla mente umana, emergeranno anche molte regole psicologiche e di comportamento, che indicano come gli esseri umani tendono a pensare e a comportarsi quando devono affrontare certi problemi, quando perseguono particolari obbiettivi partendo da particolari condizioni al contorno, in funzione degli istinti, delle convenzioni sociali, delle pulsioni di fondo, ecc.

È quindi probabile che all'inizio queste saranno un misto di regole a base psicologica, istintuale, di opportunità, ecc (in parte si tratta di ciò che alcuni chiamano "logica naturale"), e regole che hanno invece un senso puramente razionale. L'osservazione sperimentale di quanto avviene nell'uomo sembra indicare che serve molto tempo per riuscire a distinguere nei comportamenti propri e degli altri (si deve considerare che noi apprendiamo da altri molte delle nostre conoscenze) ciò che è una conseguenza necessaria di certe premesse, per ragioni di principi primi, di leggi naturali o di "logica pura", dalle regole comportamentali che dipendono da pulsioni istintuali, da motivazioni di varia natura psicologica e da altro. Anzi l'osservazione sperimentale indica che anche molte persone adulte faticano a distinguere il razionale da ciò che non lo è.

**Alcuni Riferimenti bibliografici.**

Devis Pantano.
Proposta di nuovi strumenti per comprendere come funziona la cognizione.
Preprint submitted to ArXiv March 2014